# Project

# Integrated monitoring of ice in selected Swiss lakes

# Final Report

**ETH Zürich (ETHZ)**

**University of Bern (UniBe)**

**Swiss Federal Institute of Aquatic Science and Technology (EAWAG)**

**REPORT**

**Edited by:** Emmanuel Baltsavias (ETHZ), project coordinator

**Authored by:**
ETHZ: Manu Tom, Mathias Rothermel, Emmanuel Baltsavias, (manu.tom, mathias.rothermel, manos)@geod.baug.ethz.ch
UniBe: Melanie Suetterlin, Stefan Wunderle, (melanie.suetterlin, stefan.wunderle)@giub.unibe.ch
EAWAG: Damien Bouffard, damien.bouffard@eawag.ch

**Other contributions in the project:**
ETHZ: Charis Lanaras, Mikhail Usvyatsov
UniBe: Helga Weber
EAWAG: Ulrike Kobler, Love Vinna Raman, Luca Cortese, Sebastiano Piccolroaz, Martin Schmid, Marco Toffolon (University Trento)

**Responsibility for individual Sections:**
ETHZ: 1, 2, 3, 4, 7, Appendices 1, 4, 5
UniBe: 5
EAWAG: 6, Appendix 2
All, with coordination of ETHZ: Executive summary, 8, 9, 10 (References), Appendices 3, 6

The financial support of the Swiss Federal Government, via the Federal Office of Meteorology and Climatology MeteoSwiss (Swiss GCOS Office) is highly appreciated. MeteoSwiss personnel that participated in the administration of the project and/or provided scientific remarks include:
Fabio Fontana, Michelle Stalder, Ulrich Hamann, Anke Duguay-Tetzlaff, Manuela Bizzozzero.

**This report should be cited as:**

Tom, M., Suetterlin, M., Bouffard, D., Rothermel, M., Wunderle, S., Baltsavias, E., 2019. Integrated monitoring of ice in selected Swiss lakes. Final project report. Available at https://prs.igp.ethz.ch/research/current_projects/integrated-monitoring-of-ice-swiss-lakes.html.

Some data and the report are available at this persistent link https://doi.org/10.25678/0001HC.

## EXECUTIVE SUMMARY

Various lake observables, including lake ice, are related to climate and climate change and provide a good opportunity for long-term monitoring. Lakes (and as part of them lake ice) is therefore considered an Essential Climate Variable (ECV) (WMO, 2018) of the Global Climate Observing System (GCOS). In Switzerland, the implementation of GCOS is coordinated by the Swiss GCOS Office at the Federal Office for Meteorology and Climatology MeteoSwiss. In 2007 (updated in 2018), MeteoSwiss published the first national inventory of the most important climate observations in Switzerland (MeteoSwiss, 2018). For each ECV, including observations of lake ice subsumed under the ECV "Lakes", and international centre, the inventory report identified possible gaps regarding the legal basis, definition of responsibilities, and availability of financial resources for the continuation of observations and operation, respectively. In 2008, considering the findings of this report, the Federal Council approved a financial contribution to secure the continuation of time series and international centres at risk of discontinuation, as a long-term contribution to GCOS. Concerning the ECV "Lakes", in the global National Snow and Ice Data Center (NSIDC), Boulder, Colorado database on lake ice-on/off dates, currently only Lake St. Moritz is partly included (till 2012). Further, existing observations and data from local authorities and publications are not systematic and come from different, uncoordinated and not secured sources. Traditionally, on-shore observers collected the information of lake ice recording the visible frozen-edge. Within the past two decades due to lack of budget and/or human resources, the number of field observations declined and mostly totally stopped. At the same time, the potential of different remote sensing sensors covering varying time periods and spatial coverage to measure the occurrence of lake ice was demonstrated by several investigations. Thus, following the need for an integrated multi-temporal monitoring of lake ice in Switzerland, MeteoSwiss in the framework of GCOS Switzerland supported this 2-year project to explore not only the use of satellite images but also the possibilities of Webcams and in-situ measurements.

The aim of this project is to monitor some target lakes and detect the extent of ice and especially the ice-on/off dates, with focus on the integration of various input data and processing methods. Regarding ice-on/off dates, the GCOS requirements are daily observations and an accuracy of +/- 2 days. The target lakes were: St. Moritz, Silvaplana, Sils, Sihl, Greifen and Aegeri, whereby only the first four were mainly frozen during the observation period and thus processed. These lakes have variable area (very small to middle-sized), altitude (low to high) and surrounding topography (flat/hilly to mountainous) and freeze generally often, covering thus difficult to medium difficulty cases, regarding area, altitude and topography. For all image data, only cloud-free pixels lying entirely within the lake were processed. The observation period was mainly the winter 2016-17. During the project, various approaches were developed, implemented, tested and compared. Firstly, low spatial resolution (250 - 1000 m) but high temporal resolution (1 day) satellite images from the optical sensors MODIS and VIIRS were used. Secondly, and as a pilot project, the use of existing public Webcams was investigated for (a) validation of results from satellite data, and (b) independent estimation of lake ice, especially for small lakes like St. Moritz, that could not be possibly monitored in the satellite images. Thirdly, in-situ measurements were made in order to characterize the development of the temperature profiles and partly pressure before freezing and under the ice-cover until melting. Besides the validation of the results from other data, this in-situ data is used to calibrate a one-dimensional physical lake model so that the criteria for freezing of different Swiss lakes can be derived as a function of meteorological and morphometric conditions. It is expected that the developed methods and software can be used, possibly with some modifications, also for other data acquired for the same lakes or other ones in Switzerland and abroad. The project is a feasibility study, which should lead to a comparison and analysis of the above three techniques and recommendations to MeteoSwiss for further actions. This report presents the results of the project work.

Using MODIS and VIIRS satellite data, ETHZ proposed a processing chain for lake ice monitoring. We tackled lake ice detection as a pixel-wise, two-class (frozen/non-frozen) semantic segmentation problem with Support Vector Machine (SVM) classification. Four different lakes were analysed: Sihl, Sils, Silvaplana and St. Moritz using both MODIS and VIIRS data. While we have concentrated on lakes in Switzerland, the ETHZ methodology is generic and the results could hopefully be directly applied to other lakes with similar conditions, in Switzerland and abroad, and possibly to similar sensors. To assess the performance, the data from both MODIS and VIIRS were processed from one full winter (2016/17) including the relatively short but challenging freezing and melting periods, where frozen and non-frozen pixels co-exist on the same lake. Using MODIS data, we also processed dates from 2011/12 and we demonstrate that the ETHZ approach gives consistent results over multiple winters, and that it generalizes fairly well from one winter to another. For both MODIS and VIIRS, we have also shown that the model generalizes well across lakes.

ETHZ dealt with the feasibility of lake ice monitoring using Webcam images as a supplement or alternative to other monitoring methods. Therefore, publically available Webcam images capturing six Swiss lakes were downloaded from the Internet in the periods of the winters 2016/2017 and 2017/2018. Our evaluation was based on two cameras (high- and low-resolution) monitoring the lake of St. Moritz. To predict the lake ice coverage of the observed water body, fully Connected Neuronal Networks (part of Deep Learning and Artificial Intelligence methods) were utilized. Given an input image, such networks are designed to predict pixel-wise class probabilities. For the problem at hand, the target classes were: snow, ice, water and clutter. Several tests were conducted to identify important parameters for the intensive training of such networks. Moreover, generalization capabilities with respect to differing cameras and to data recorded in different winters were investigated. In contrast to optical satellite data, Webcams typically record multiple images per day. A median-based strategy to fuse such results to derive daily predictions was implemented. Moreover, possibilities of further exploiting temporal redundancy to improve predictions were explored by using two additional network architectures.

The Remote Sensing Research Group, Institute of Geography, University of Bern focuses on the development of a physical approach to monitor lake ice based on data of the satellite sensor VIIRS (Visible Infrared Imaging Radiometer Suite) on-board of NOAA satellites Suomi-NPP. The advantage of the VIIRS data is the daily temporal resolution and improved spatial resolution of 375m considering the high-resolution channels (I-channels). A comprehensive pre-processing chain has been developed and implemented to efficiently obtain projected data of the region of interests (target lakes in Switzerland) from multiple scans. Not only the scientific data records of VIIRS data (e.g., radiometric data, geolocation information) were processed but also environmental data records (e.g., VIIRS M-band surface temperature), and intermediate products (e.g., cloud masks), which were needed for product retrieval. The developed approach considers surface temperature and reflectance values in the visible and near-infrared spectra. The method relies on the assumption that a frozen water surface has a temperature below 0°C and an ice layer increases the reflectance in the visible/near-infrared wavelengths. Retrieval of surface temperature using only one channel in the atmospheric window requires a procedure to correct for atmospheric effects (attenuation). Therefore, a single channel PMW (Physical Mono Window) model has been adapted to the thermal I-band data of the VIIRS sensor (I05) considering atmospheric data from the European Center for Medium Weather Forecast (ECMWF), which were required for the radiative transfer modelling to minimize atmospheric attenuation. The accuracy of the retrieved VIIRS I-band PMW Lake Surface Water Temperature (LSWT) has been assessed with cross-satellite comparison and temperature based validation for Lake Geneva (station Buchillon) and the target lakes Lake Greifen and Lake Sils.

The assessment of thermal infrared-derived surface temperatures indicates an overall good performance of the physical mono window LSWT retrieval method for VIIRS I-band data. Therefore, lake ice detection, based on LSWT, the normalized difference snow index (NDSI) and additional thresholds have been accomplished for the two target lakes for October 2016 to April 2017. Considering both the VIIRS I-band reflectance and thermal infrared-derived LSWT were a satisfying approach for lake ice detection, even for small lakes (e.g. Lake Sils). The implemented two-step-approach used these results to determine, without user-interaction, the thresholds to retrieve ice phenology. Finally, this results in an automatic determination of duration of ice cover (phenology) due to the identification of first/last day with ice cover, even for small lakes where only four-six 375m-pixels being not affected by the shoreline. Hence, the developments made for this feasibility study were successful and the pre-operational concept of our modular procedure can be part of an operational service.

In-situ measurements and processing were performed by EAWAG. There is no direct method to measure ice coverage in lakes. Time series of lake temperatures are often used to assess freezing and melting conditions. Yet, the 0°C boundary condition for ice cover in lakes is limited to a very thin layer at the surface (0 cm) immediately under the ice and is practically nearly impossible to measure with moorings. In this project, we tested different approaches to monitor the ice cover period by analysing the changes in lake dynamics during ice-free and ice covered period. Namely, we observed that (i) the correlation in the temperature time series of two closely vertically located sensors changed when the external forcing are modified by the ice. Similar results can be observed through (ii) wavelet and Fourier analysis of the temporal evolution of a single temperature logger with a noticeable drop in the energy (e.g. temperature fluctuations) during the ice covered period. Lastly, (iii), we evaluated the potential of using a high frequency pressure sensor to track the ice-on/off periods. Finally, the ice phenology was investigated with two different numerical models. A fully deterministic hydrodynamic model (simstrat.eawag.ch) provided information related to the sensitivity of the ice coverage to the meteorological forcing (e.g. mostly wind and air temperature) and lake characteristics (e.g. bathymetry). This model together with a second hybrid model were used to evaluate the long term phenology of frozen lakes and will allow estimate of the change in phenology under climate change conditions (with CH2018 dataset).

The results of ice-on/off dates were based on ground-truth, mainly consisting of visual interpretation of Webcam images, which had some deficiencies and cannot be fully trusted. Furthermore, for the satellite data, clouds on and/or close to the ice-on/off dates led in some cases to large errors in the determination of these dates. The comparison of the different monitoring methods was not based on sufficiently extensive and timely co-incident data. In spite of these difficulties, some very valuable conclusions can still be drawn. Small lakes (< 2 $km^2$) are problematic for satellite images with a ground pixel size of about 250-400m and more. Webcams (though in this case with limited tests) and in-situ (temperature-based) measurements showed the best accuracy. Optical satellite data suffer from clouds, cloud mask errors, reflectance variations and other factors, and their accuracy regarding ice-on/off was worse. The two different processing methods (ETHZ and UniBe) of the same VIIRS data had generally similar performance but differences should be investigated. The main problem for all data (apart in-situ) is the separation between transparent/clear ice (usually thin) and water, particularly in the freeze-up (but also less in the break-up) period, because of very similar reflectance. This problem gets worse with reduced spatial resolution of the images. Even careful visual interpretation of image data cannot reliably distinguish between these two cases. Unfortunately, the critical ice-on/off dates involve water and ice (snow usually comes later on ice-on and melts earlier than ice-off) and thus, accurate identification of transparent ice or water is crucial for the accurate identification of ice-on/off dates.

While as known, satellite data is the best operational input for global coverage, but partly also in Switzerland, for Switzerland Webcam and in-situ data are very valuable. Further data, than those used in this project, should include radar satellite data (as highest priority to avoid the cloud problem) and additionally optical high spatial resolution satellite data (especially the free ESA Sentinel-1/2 data), better Webcams (especially with Pan-Tilt-Zoom) and extended in-situ measurements (in combination with local, cheap meteo stations and better Webcams). For in-situ observations and generally, cooperation with Swiss Federal and other agencies and stakeholders would be very beneficial.

Section 8 summarises the main points, without having to read the whole report.

# 1. INTRODUCTION

## 1.1 Motivation

Various lake observables, including lake ice, are related to climate and climate change and provide a good opportunity for long-term monitoring. Lakes (and as part of them lake ice) is therefore considered an Essential Climate Variable (ECV) (WMO, 2018) of the Global Climate Observing System (GCOS). In Switzerland, the implementation of GCOS is coordinated by the Swiss GCOS Office at the Federal Office for Meteorology and Climatology MeteoSwiss. In 2007 (updated in 2018), MeteoSwiss published the first national inventory of the most important climate observations in Switzerland (MeteoSwiss, 2018). For each ECV, including observations of lake ice subsumed under the ECV "Lakes", and international centre, the inventory report identified possible gaps regarding the legal basis, definition of responsibilities, and availability of financial resources for the continuation of observations and operation, respectively. In 2008, considering the findings of this report, the Federal Council approved a financial contribution to secure the continuation of time series and international centres at risk of discontinuation, as a long-term contribution to GCOS. Concerning the ECV "Lakes", in the global National Snow and Ice Data Center (NSIDC), Boulder, Colorado database on lake ice-on/off dates, currently only Lake St. Moritz is partly included (till 2012). Further, existing observations and data from local authorities and publications are not systematic and come from different, uncoordinated and not secured sources. Traditionally, on-shore observers collected the information of lake ice recording the visible frozen-edge. Within the past two decades due to lack of budget and/or human resources, the number of field observations declined and mostly totally stopped. At the same time, the potential of different remote sensing sensors covering varying time periods and spatial coverage to measure the occurrence of lake ice was demonstrated by several investigations. Thus, following the need for an integrated multi-temporal monitoring of lake ice in Switzerland, MeteoSwiss in the framework of GCOS Switzerland supported this 2-year project to explore not only the use of satellite images but also the possibilities of Webcams and in-situ measurements.

## 1.2 Aims

The project aim is to monitor some target lakes and detect the extent of ice, the duration of lake ice and in particular the ice-on/off dates, with focus on the integration of various input data and processing methods. Thereby, various approaches will be implemented, developed, compared and integrated. Firstly, low spatial resolution (250 - 1000 m) but high temporal resolution (1 day) satellite images from various sensors will be used. Several spectral bands will be used, both reflective and emissive (thermal). Secondly, and as a pilot project, the use of existing Webcams will be investigated for (a) validation of results from satellite data, and (b) independent estimation of lake ice, especially for small lakes that cannot be detected in the satellite images. Thirdly, in-situ measurements will be made in order to characterize the development of the temperature profiles before freezing and under the ice-cover until melting. Besides the validation of the results from other data, this data will be used to calibrate a one-dimensional turbulence model so that the criteria for freezing of different Swiss lakes can be derived as a function of meteorological and other conditions in late autumn and winter. It is expected that the developed methods and software can be used, possibly with some modifications, also for data acquired in the future. The project is a feasibility study, which should lead to a comparison and analysis of the above three techniques and recommendations to MeteoSwiss for further actions.

## 1.3 Input data

*Main:*

- Satellite images (multispectral, temporal resolution of 1 day, spatial resolution at best 250m pixel size): MODIS, VIIRS.
- Webcam images: publicly available images.
- In-situ measurements: temperature and pressure measurements.

*As possible ground truth (see Section 2):*

- Manual interpretation of Webcams.
- Satellite images with higher spatial resolution for validation: Sentinel-2A.
- Other validation data: online media.

More details on the MODIS and VIIRS data are listed in Sections 3.1 and 4.1/5.1, respectively. Details on the Webcam data are listed in Appendix 1 and on the in-situ measurements in Appendix 2. Target lakes for which data were acquired and processed are listed in Table 1. Some of these lakes did not freeze during the test period (see Section 1.5) and thus the respective data were partly not acquired and processed.

## 1.4 Target lakes

They include the lakes in decreasing altitude from 1797 to 435 m, altitude being used in this case as the main criterion of freezing or not: Sils, Silvaplana, St. Moritz, Sihl, Aegeri, Greifen. These lakes have variable area (very small to middle-sized), altitude (low (for Switzerland) to high) and surrounding topography (flat/hilly to mountainous) and freeze more or less often, covering thus difficult to medium difficulty cases, regarding area and altitude/topography. Hence, we expect that the results of the project could be applicable to other similar or easier lake characteristics in Switzerland and abroad.

**Other test lake data** (mostly from Wikipedia)**:**

*Lake Sils*
Max length/width 5 x 1.4 km, max depth 71m, residence time 2.2 years, inflows: Inn, Aua da Fedoz, outflows: Inn.

*Lake Silvaplana*
Max length/width 3.1 x 1.4 km, max depth 71m, residence time 250 days, inflows: Inn (named Sela between Lake Sils to Lake Silvaplana), Fexbach, Ova dal Valhun, outflows: Lake Champfèr.

*Lake Sihl*
Max length/width 8.5 x 2.5 km, max depth 23m, inflows: Sihl, Minster, outflows: Sihl.

*Lake St. Moritz*
Max length/width 1.6 x 0.6 km, max depth 44m, inflows: Inn, outflows: Inn, generally frozen Dec.-May.

*Lake Aegeri*
Max length/width 5.4 x 1.4 km, max depth 83m, residence time (average days of molecules/age of water inside lake) 6.8 years, inflows: Hüribach, outflows: Lorze, generally frozen Jan.-Feb.

*Lake Greifen*
Max length/width 6 x 1.6 km, max depth 32m, residence time 408 days, inflows: (Mönchaltorfer) Aa, Aabach, outflows: Glatt.

The lake outlines were digitized from OpenStreetMap (OSM) by UniBe and had an accuracy of ca. 25-50m. To reduce unnecessary tiny details, they were generalized by UniBe and ETHZ using different methods. Although the two generalized results were not compared, a comparison by ETHZ of the raw and generalized outlines showed no important differences, especially considering the rather coarse spatial resolution of MODIS and VIIRS. The lake outlines were backprojected from the ground coordinate system onto the images to guide the estimation of lake ice. These outlines were corrected for the absolute geolocation error of each satellite sensor (as estimated by ETHZ, see Table 5) by both ETHZ and UniBe.

**Table 1. Characteristics of six target lakes (sorted according to altitude).**

| Lake | Altitude (m) | Area (km$^2$) | Average depth (m) | Volume (Mm$^3$) | Altitude (m) | Lake type |
|---|---|---|---|---|---|---|
| Sils | 1797 | 4.1 | 35 | 137 | 1797 | natural |
| Silvaplana | 1791 | 2.7 | 35 | 140 | 1791 | natural |
| St. Moritz | 1768 | 0.78 | 26 | 20 | 1768 | natural |
| Sihl | 889 | 11.3 | max 17 | 96 | 889 | artificial |
| Aegeri | 724 | 7.3 | 49 | 360 | 724 | natural |
| Greifen | 435 | 8.45 | 18 | 148 | 435 | natural |

Data generally from Wikipedia.

## 1.5 Data acquired and processed

Table 2 gives a summary of the dates of data acquired and processed. ETHZ received VIIRS data from UniBe for the winter 2016-2017 in two batches. First, October, December and February, much later November, January and March and even later April data were processed. Thus, the ETHZ investigations listed below started with the first data batch of October, December and February (winter subset) and continued with the full winter.

**Table 2.** Dates of data acquired and processed for the six target lakes and all monitoring methods.

| **St. Moritz** | | | **Silvaplana** | | |
|---|---|---|---|---|---|
| **Method** | **Dates of data downloaded / acquired** | **Dates of data processed** | **Method** | **Dates of data downloaded / acquired** | **Dates of data processed** |
| MODIS | 1.10.11 - 30.4.12<br>1.7.12 - 31.8.12<br>1.10.16 - 30.4.17 | 1.11.11 - 30.11.11,<br>1.1.12 - 29.2.12<br>1.7.12 - 31.8.12<br>1.10.16 - 30.4.17 | MODIS | 1.10.11 - 30.4.12<br>1.7.12- 31.8.12<br>1.10.16 - 30.4.17 | 1.11.11 - 30.11.11<br>1.1.12 -29.2.12<br>1.7.12 -31.8.12<br>1.10.16 - 30.4.17 |
| VIIRS (ETHZ) | See VIIRS (UniBe) | 1.10.16-30.4.17 | VIIRS (ETHZ) | See VIIRS (UniBe) | 1.10.16 - 30.4.17 |
| VIIRS (UniBe) | 10.2.2012 - 31.3.2012<br>1.10.2016 - 30.4.2017 | LSWT: 10.2.2012 - 31.3.2012<br>LSWT and lake ice mask: 1.10.2016 - 30.4.2017 | VIIRS (UniBe) | 10.2.2012 - 31.3.2012<br>1.10.2016 - 30.4.2017 | 1.10.2016 - 30.4.2017 |
| Webcams | 4.12.16 - 12.6.17,<br>1.12.17 - now | 4.12.16 - 12.6.17 (2 cams),<br>29.11.17 - 31.1.18<br>(2 cams) | Webcams | 4.12.16 - 20.6.17,<br>1.12.17 - now | None |
| In-situ | 26.10.16 - 3.7.17 | All | In-situ | 7.1 - 6.7 2016,<br>25.10.16 - 3.7.17 | All |

| **Sils** | | | **Sihl** | | |
|---|---|---|---|---|---|
| Method | **Dates of data downloaded / acquired** | **Dates of data processed** | **Method** | **Dates of data downloaded / acquired** | **Dates of data processed** |
| MODIS | 1.10.11 - 30.4.12, 1.7.12 - 31.8.12, 1.10.16 - 30.4.17 | 1.11.11 - 30.11.11, 1.1.12 - 29.2.12, 1.7.12 - 31.8.12, 1.10.16 - 30.4.17 | MODIS | 1.10.11 - 30.4.12, 1.7.12 -31.8.12, 1.10.16 -30.4.17 | 6.2.12 - 25.2.12, 20.3.12 - 31.3.12, 1.8.12 -31.8.12, 1.10.16 - 30.4.17 |
| VIIRS (ETHZ) | See VIIRS (UniBe) | 1.10.16 - 30.4.17 | VIIRS (ETHZ) | See VIIRS (UniBe) | 1.10.16 -30.4.17 |
| VIIRS (UniBe) | 10.2.2012 - 31.3.2012 1.10.2016 - 31.3.2017 | 1.10.2016 - 30.4.2017 (LSWT & lake ice mask) | VIIRS (UniBe) | 10.2.2012 - 31.3.2012 1.10.2016 - 30.4.2017 | LSWT & lake ice mask: 1.10.2016 - 30.4.2017 |
| Webcams | 4.12.16 - 12.6.17<br><br>1.12.17 - now | None | Webcams | 4.12.16 - 20.6.17 2.12.17 - now | None |
| In-situ | 26.10.16-5.17 | All | In-situ | 19.1-13.4 2016, 16.11.16-18.5.17 | All |

| **Greifen** | | | **Aegeri** | | |
|---|---|---|---|---|---|
| **Method** | **Dates of data downloaded / acquired** | **Dates of data processed** | **Method** | **Dates of data downloaded / acquired** | **Dates of data processed** |
| MODIS | 1.10.11 - 30.4.12 1.7.12 - 31.8.12 1.10.16 - 30.4.17 | None | MODIS | 1.10.11 -30.4.12 1.7.12 -31.8.12 1.10.16 - 30.4.17 | None |
| VIIRS (ETHZ) | See VIIRS (UniBe) | None | VIIRS (ETHZ) | See VIIRS (UniBe) | None |
| VIIRS (UniBe) | 10.2.2012 - 31.3.2012 1.10.2016 - 30.4.2017 | 1.10.2016 - 31.12.2016 (LSWT & lake ice mask) | VIIRS (UniBe) | 10.2.2012 - 31.3.2012 1.10.2016 - 30.4.2017 | None |
| Webcams | 1.12.16 - 12.6.17 2.12.17 - now | None | Webcams | 4.12.16 - 12.6.17 1.12.17 - now | None |
| In-situ | 6.1-18.4.2016, 8.11.16-5.5.17 | All | In-situ | 19.1-20.4 2016, 16.11.16-18.5.17 | All |

## 1.6 Definitions used

**Ice-on** used here as 1st day totally or in great majority frozen, with a similar day after it (same definition as in Franssen and Scherrer (2008), i.e. end of freeze-up).

**Ice-off** used here as the symmetric of ice-on, i.e. the 1st day after having all or almost all lake frozen, when little but clear water appears and in the subsequent days this water area increases. Therefore, we use it here as melting (break-up) start. In most cases, especially in N. America (and in the NSIDC database) ice-off is defined as end of break-up when almost everything is water. We used in this work our own definition as it is symmetric to ice-on, much easier to detect and allows the calculation of the important parameter of the number of ice-covered days.

**Clean pixels** are those that are totally within the lake outline. For Webcams, the lake outline was manually determined. For satellite images, the procedure of finding the lake outlines in the images is described above. ETHZ used strictly this criterion, e.g. even if a pixel was only 1% on land, it was not considered clean.

In all subsequent investigations with image data, cloud-free clean pixels were used. With an exception for St. Moritz and ETHZ processing of VIIRS 2016-17 images, where also mixed pixels were used (see Section 4.3.2).

## 2. REFERENCE DATA

### 2.1 Human observations from lake shore

Initially we thought that we could use such observations from fishermen, local authorities, police etc. as mentioned in Franssen and Scherrer (2008). However, for the time periods of the data used, no such observations existed.

### 2.2 Internet media

A quite extensive search provided information, mainly from e-newspapers, but this information was limited and referred to the "event" which was that the police gave free the frozen lake for people to walk, skate etc. on it, but without ice-on/off dates (see Table 3).

**Table 3.** Ground truth (Information from online media). The table cells are left blank when no information was gathered.

**Lake Greifen**

| Winter | Ice-on | Ice-off | Remarks |
|---|---|---|---|
| 2011/12 | 12.2.12 | 17.2.12 | |
| 2005/06 | 4.2.06 | | |
| 2002/03 | 20.2.03 | | Frozen at least on 23.2.03 |

**Lake Sihl**

| Winter | Ice-on | Ice-off | Remarks |
|---|---|---|---|
| 2011/12 | 12.1.12 | | Frozen at least from 6.2.12 - 25.2.12 |
| 2009/10 | | | Frozen at least on 14.1.10 |

**Lake Aegeri**

| Winter | Ice-on | Ice-off | Remarks |
|---|---|---|---|
| 2005/06 | 4.2.06 | | |

### 2.3 Visual interpretation of Webcams

This proved to be the easiest way of getting ground truth. For each lake, a main Webcam was selected, and others were additionally used. The interpretation was done by minimum one operator. In some cases, one operator evaluated a lake multiple times with a temporal difference of several days to avoid bias. In case of uncertainty, several images per day were evaluated. Due to the problems of the Webcams (as explained in Section 8), this visual evaluation cannot be considered error free. A particular difficulty was the differentiation between water and thin, transparent ice (without snow), especially during the longer freezing period. Apart from that, the Webcams imaged only a part of the lake (see Fig. 80). Thus, there is no safe conclusion on whether the whole lake was frozen, though when a large, central part of the lake was frozen, one could safely conclude that the whole lake was frozen, apart possibly from small areas close to inflows and outflows. For the visual interpretation, we used the following classes and criteria:

Explanation of lake status:
The % below refers to the visible area of the lake, potentially excluding the far part of the lake, which you generally cannot judge, except in case of snow. Clearly, these % are a matter of personal judgement and only approximate.

- ***s*** = snow, when snow is on lake ice, lake frozen to ca. 90-100%
- ***i*** = ice, frozen lake to ca. 90-100%
- ***w*** = water, lake covered by water to ca. 90-100%
- ***ms*** = more snow, ca. 60-90%, but a small part water
- ***mi*** = more ice, ca. 60-90%, but a small part water
- ***mw*** = more water, ca. 60-90%, but a small part frozen

- ***c*** = clouds or fog covering all lake
- ***u*** = unclear, when you cannot judge the lake state
- ***n*** = no Webcam data available

The first two categories are the critical ones to estimate ice-on/off dates. More details are listed in Appendix 4.

A separate visual classification was done for the Webcam images of Lake St. Moritz for the processing explained in Section 7. In this case, the classification was more detailed for each image, using polygons for the classes water, ice, snow and clutter (all rest classes, especially constructions on the frozen Lake St. Moritz for sport events).

## 2.4 Enriched Webcam ground truth using Sentinel-2A

The Webcam ground truth was enriched by using visual interpretation of Sentinel-2A images (see Appendix 5), as explained below. This ground truth was used by ETHZ for the evaluation of MODIS and VIIRS data processing in Sections 3 and 4. The "enrichment" was based on the following procedure.

Firstly, all available dates were assigned one of the above labels (one label per day) based on the visual evaluation of the Webcam data. Then, all available Sentinel-2A data was analysed visually to enrich the Webcam-based ground truth. Days with labels: ***c, u and n*** in between a continuous sequence of fully frozen or fully non-frozen dates have been re-labelled as fully frozen and fully non-frozen, respectively. When there was a contradiction between Webcam and Sentinel-2A, we adopted the visual interpretation of the latter. We assume that the lake freezing and melting are slow processes with the following pattern in transitions ***w*** <-> ***mw*** <-> ***mi/ms*** <-> ***i/s*** and spikes in the ground truth were smoothed out with this assumption. In case of gaps, we performed interpolation of the available ground truth. For lakes Sihl and St. Moritz, no Webcam data is available from October and November 2016. Hence, sparse information from Sentinel-2A is the primary source of ground truth for these months and these two lakes.

**Winter 2016/17.** ETHZ primarily used freely available images from the Webcams pointing at the target lakes to obtain the ground truth. Given the fact that the Webcams do not cover the complete lake surface, and their Ground Sampling Distance (GSD) varies greatly across Webcams, manual interpretation is in some cases uncertain. In addition to the Webcam images, ETHZ used the online media reports, which however provided only scarce information. Additionally, the Webcam-based ground truth is complemented with the information from Sentinel-2A data. Sentinel-2B data were not available for the winter 2016-17. As opposed to the Webcams, Sentinel-2A offers full coverage of the lakes. However, for visual evaluation, the Sentinel-2A images are not usable in some dates due to the presence of clouds. Additionally, the temporal resolution of Sentinel-2A is 10 days at equator. Hence, it cannot be used as the main source of validation, but as a secondary source with Webcams being the primary source. More information on Sentinel-2 can be found at https://eoportal.org/web/eoportal/satellite-missions/c-missions/copernicus-sentinel-2 and https://www.esa.int/Our_Activities/Observing_the_Earth/Copernicus/Sentinel-2.

The S1C (monodate orthorectified image estimated at Top of Atmosphere (TOA) reflectance) product is analysed using the software: *QGIS Desktop* (version: 2.18.2, *Las Palmas*). It is observed that $B_8$ is particularly good for visual evaluation especially on the days with thin clouds. The RGB True Colour Image (TCI, see Fig. 1d) using the bands $B_2$ (blue, see Fig. 1a), $B_3$ (green, see Fig. 1b), and $B_4$ (red, see Fig. 1c) included within the Level 1C product is also very useful. By visually inspecting the Sentinel-2A bands, a frozen lake can be distinguished from a non-frozen lake, thanks to the special texture/colour appearing in some bands due to the presence of ice. Additionally, it is easy to judge visually the presence of snow on ice especially using the TCI. However, it is confusing to judge whether the state of lake is water or a thin layer of ice is present over the water. Consolidating the information from Webcams and Sentinel-2A, one label per day is assigned for the state of each lake. The results of the visual evaluation of Webcams and Sentinel-2A data are listed in Appendices 4 and 5, respectively, while information from online media is in Table 3.

**Winter 2011/12.** ETHZ could not obtain Webcam data for winter 2011/12 (except for Webcam El Paradiso which monitors Lake Silvaplana for which the data is available from January 2012, one image per day). Hence, the scarce ground truth is obtained primarily from the online media (includes e-newspapers and Websites/blogs, see Table 3).

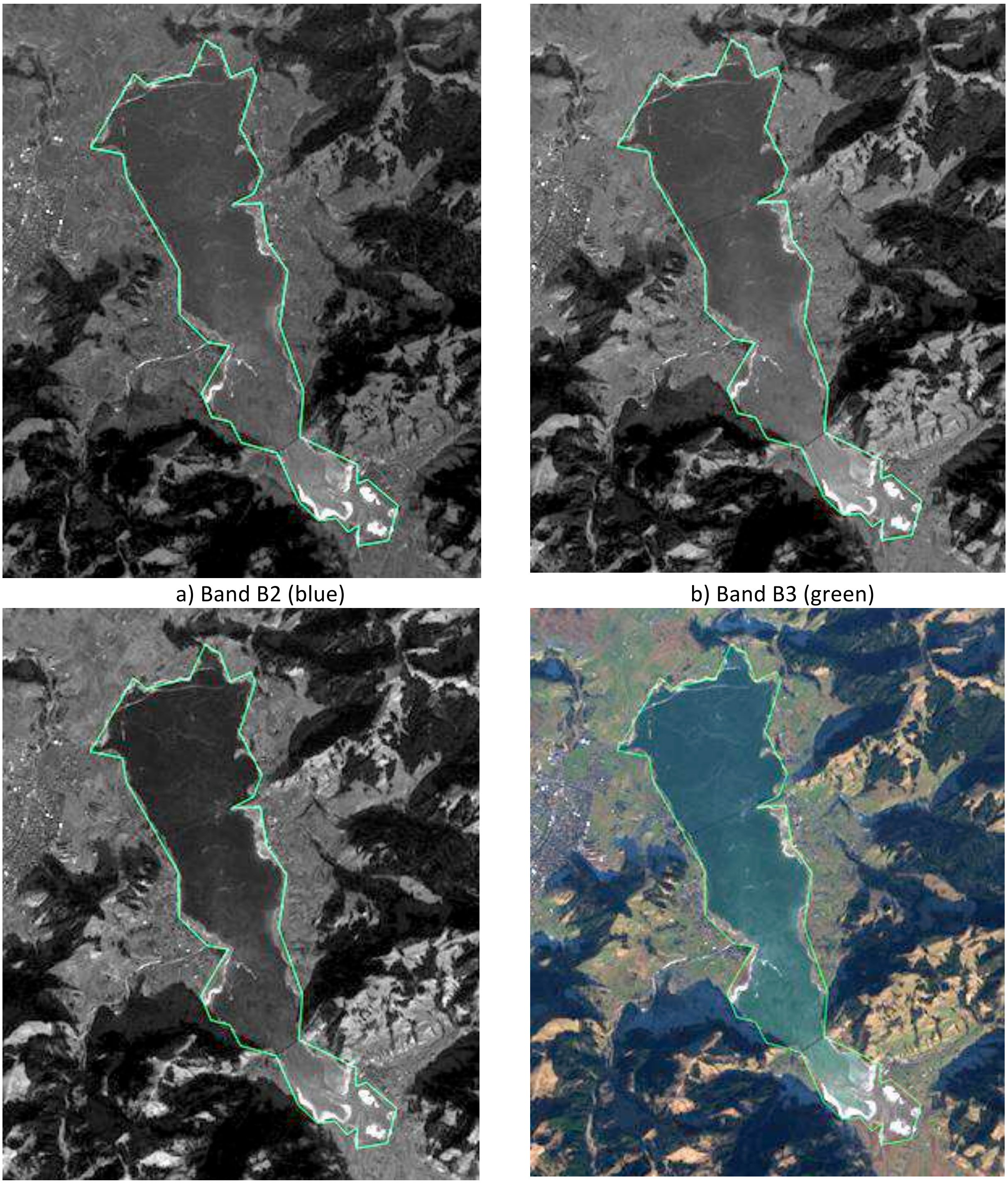

a) Band B2 (blue)

b) Band B3 (green)

c) Band B4 (red)

d) True Colour Image

**Fig. 1.** Selected bands of Sentinel-2A data for Lake Sihl on 31.12.2016 (fully frozen, no snow on ice) and the true colour image, generated from the three bands.

Ice cover changes gradually, so on some days the lake can be only partially frozen. To ensure reliable reference data, only the dates in which the lakes were completely frozen are analysed during this winter. Winter 2011-12 was the coldest among the recent winters. In addition, the larger lakes at the lower altitude were also frozen in January and February. Hence, we assumed that the lakes Sils, Silvaplana and St. Moritz were frozen in January and February 2012. This assumption is substantiated by the very low temperatures during these months and the high lake altitude. However, Lake Sihl was frozen for a lesser period. The freeze information for Sihl is collected from Internet sources (like media reports).

UniBe used for some days of processed VIIRS data Landsat quicklooks for internal visual control.

### 2.5 Conclusions on the ground truth

Clearly, even with visual interpretation, the ground truth cannot be trusted 100%. Especially, for the ice-on date, with difficult image interpretation and separation between water and thin, transparent ice. This was however the best we could do under the given circumstances. A comparison of the results of the different methods can also provide an implicit ground truth. In the sense, that if several methods result in very similar ice-on/off dates and one or two differ a lot, one could rely much more on a weighted average of the first results.

## 3. MODIS PROCESSING

### 3.1 Input data

Two copies of MODIS are in orbit on-board the Terra (Terra Mission, EOS/AM-1) and Aqua satellites. In this work, ETHZ uses only Terra MODIS images, as Aqua MODIS images have inferior quality. The Terra MODIS data is freely available and is downloaded from the *LAADS DAAC* (Level-1 and Atmosphere Archive and Distribution System Distributed Active Archive Center) archive. The Hierarchical Data Format (HDF) files of the relevant MODIS products (*MOD021KM, MOD02HKM, MOD02QKM, MOD03* and *MOD35_L2*) are ordered from the archive based on a time and location based search and further downloaded using File Transfer Protocol (FTP). All MODIS tiles, which cover the rectangular area (either fully or partly) with following geographic co-ordinates, are downloaded:

- Upper left corner of rectangle: 3.837 (longitude) 48.319 (latitude)
- Lower right corner of rectangle: 11.996 (longitude) 44.294 (latitude)

An example MODIS image is shown in Fig. 2.

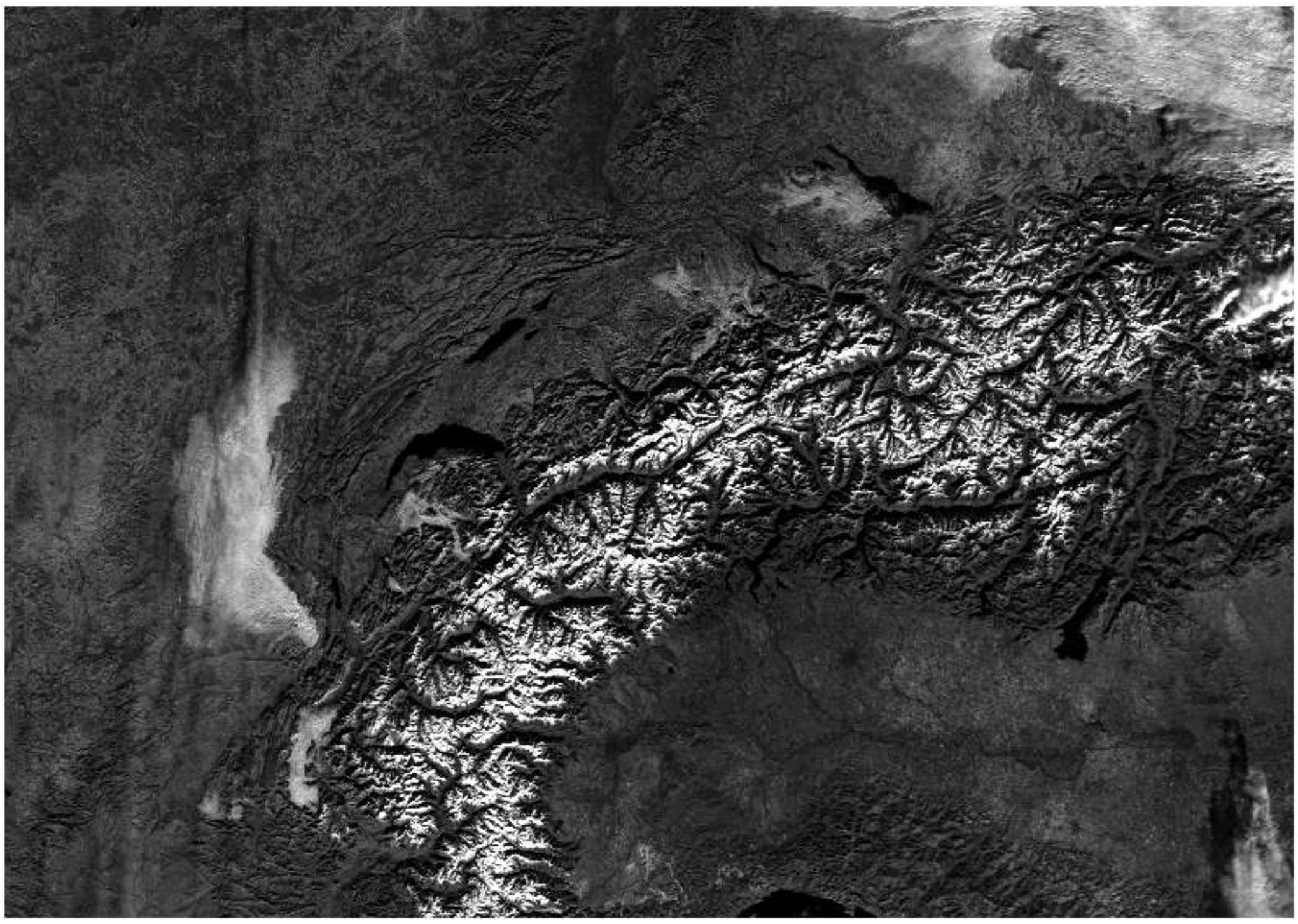

**Fig. 2**. An example of MODIS image (contrast enhanced, Band $B_2$, 250m GSD, 28.12.2016 10:45am) displaying region in and around Switzerland.

#### 3.1.1 Terra MODIS

MODIS is a passive imaging spectroradiometer with 490 detectors, arranged in 36 spectral bands that are sampled across the visible and infrared spectrum. The MODIS sensor captures data at varying spatial resolutions (2 bands with 250m at nadir, 5 bands with 500m at nadir and 29 bands with 1000m at nadir). The multi-spectral data is available with a temporal resolution of ca. 1 day. More details can be found at: https://eoportal.org/web/eoportal/satellite-missions/t/terra#sensors. ETHZ used the following three MODIS products in the analysis: *MOD02*: the level 1B data set with calibrated and geolocated aperture radiances, *MOD03*: the geolocation product containing geodetic coordinates, ground elevation, solar and satellite zenith and azimuth angle, *MOD35*: the 48-bit fractional cloud-mask product. Here, we process only the daytime acquisitions.

### 3.1.2 Useful MODIS bands

Either because of the presence of stripes or saturation, many bands of MODIS are not directly useful. Out of the 36 available bands, twelve are selected as possibly useful, through visual inspection but also non-overlapping spectral bandwidth (with the only exception of two bands, which however had different bandwidth). This set includes eight reflective (R) and four emissive (E) bands with different spatial resolutions (Table 4).

**Table 4.** MODIS, useful bands. Spatial resolution (Res) and spectral bandwidth (BW) of the 12 potentially useful reflective (R) and emissive (E) spectral bands selected after visual inspection.

| **Band** | **$B_1$** | **$B_2$** | **$B_3$** | **$B_4$** | **$B_6$** | **$B_{17}$** | **$B_{18}$** | **$B_{19}$** | **$B_{20}$** | **$B_{22}$** | **$B_{23}$** | **$B_{25}$** |
|---|---|---|---|---|---|---|---|---|---|---|---|---|
| **Type** | R | R | R | R | R | R | R | R | E | E | E | E |
| **Res (m)** | 250 | 250 | 500 | 500 | 500 | 1000 | 1000 | 1000 | 1000 | 1000 | 1000 | 1000 |
| **BW (µm)** | 0.62-0.67 | 0.84-0.88 | 0.46-0.48 | 0.55-0.57 | 1.63-1.65 | 0.89-0.92 | 0.93-0.94 | 0.92-0.97 | 3.66-3.84 | 3.93-3.99 | 4.02-4.08 | 4.48-4.55 |

### 3.1.3 Super-Resolution

Different resolutions of the various MODIS (and VIIRS, and other relevant satellite) bands pose the following challenge: a direct analysis of the data will usually deliver results at the lowest available resolution. A possible way around this problem is to reconstruct the lower resolution bands to the highest available resolution. This procedure is usually termed *multispectral super-resolution* and is a generalization of *pan-sharpening*. The aim of these methods in this case is to exploit the underlying information of all bands and to obtain images with the highest spatial resolution for all spectral bands. This is possible by building on the assumption that the discontinuities observed at the highest spatial resolution should be also present at lower resolutions. Thus, two different methods are used to increase the resolution of the 500m and 1000m bands to 250m; using simple bilinear interpolation and SupReME (Lanaras et al., 2017). SupReME is a method that inverts the linear spectral mixing model, with adaptive, edge-preserving regularisation. It solves the inversion problem for all bands simultaneously, and was applied with the following empirical parameter settings. The subspace dimension is set to p = 7, the spatial regularization as λ = 0.2 and the subspace weights q = [1 1:5 8 15 15 20 20]. A better method developed later within ETHZ and based on deep learning is given in Lanaras et al. (2018) but it could not be tested within this project. An example with the 1000m resolution band $B_{17}$ comparing the above two methods is presented in Fig. 3. As it can be seen, SupReME creates sharper results than the bilinear interpolation. However, SupReME results in a smoother image without visible noise. Since there is no ground truth available to directly evaluate the super-resolution performance, the results can only be judged qualitatively.

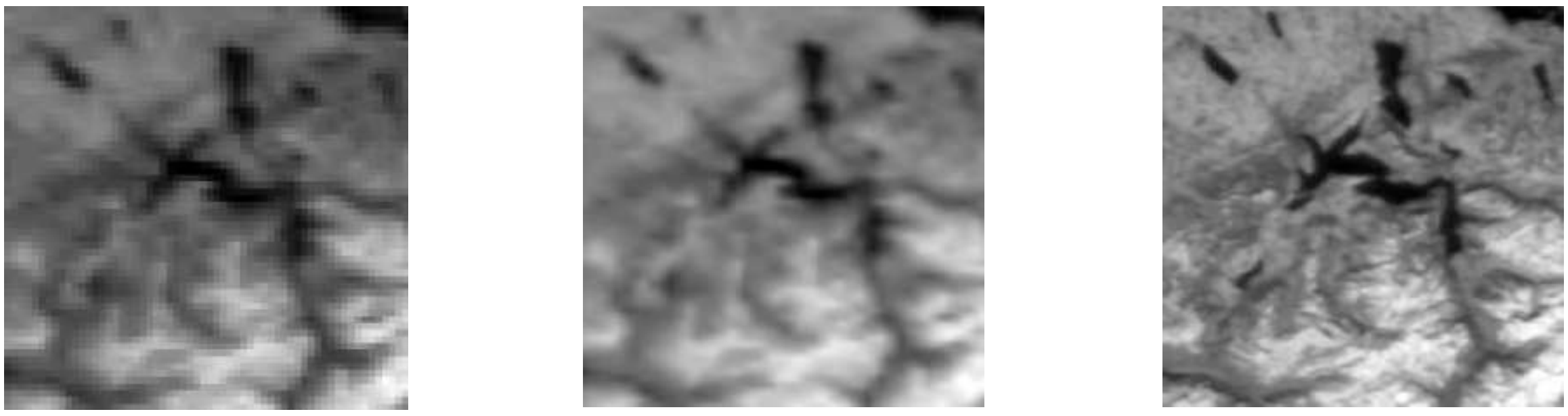

a) b) c)

**Fig. 3**. Region around Lake Lucerne (Vierwaldstättersee). (a) Original band $B_{17}$ with 1000m GSD, (b) Bilinear Interpolation, (c) SupReME. Last two images resampled to 250m.

For SupReME, it is found empirically that better qualitative results are obtained when only the reflective bands are processed. This is likely due to the fact that the emissive discontinuities do not fully coincide with the reflective ones, not fulfilling the assumption mentioned above. Thus, for winter 2011/12, along with bilinear interpolation, SupReME is run in two configurations: including ($SupReME_{RE}$) and excluding ($SupReME_{R}$) the emissive bands in the processing. Although reflective and emissive bands may show different edges (especially in mountainous regions), the lake surface and shores would not exhibit significantly different edges. It can be seen in Section 3.3.1, that the bilinearly interpolated MODIS images are slightly better (quantitative results, winter 2011/12) than the counterparts interpolated with SupReME. This does not indicate that SupReME is an inferior super-resolution strategy compared to bilinear interpolation but that SupReME is not as effective as bilinear for the specific case of lake ice monitoring. Therefore, SupReME super-resolution was performed only on the winter 2011/12 images (initial analysis). However, bilinear interpolation was used with all MODIS images.

## 3.2 Methodology

### 3.2.1 Pre-Processing

**Extraction of HDF**. *MRTSWATH* (MODIS Reprojection Tool Swath, 2017) software is used to extract the 36 different MODIS bands and reproject them to the reference UTM32N co-ordinate system. To achieve this, the respective HDF files along with the geolocation HDF files, downloaded from *LAADS DAAC* archive (see Section 3.1), are fed as input to the *MRTSWATH* software.

**Cloud-mask generation**. A binary cloud-mask is derived from the MODIS 48-bit fractional cloud-mask product by following a conservative approach. The four available classes (*cloudy, uncertain clear, probably clear* and *confident clear*) are combined in the following manner:

- Cloudy (*cloudy, uncertain clear*)
- Non-cloudy (*probably clear, confident clear*)

For more information on MODIS cloud mask, see http://cimss.ssec.wisc.edu/modis/CMUSERSGUIDE.PDF. The *LDOPE* software tool provided by the MODIS land quality assessment group (Roy et al., 2002) is used to perform masking of bits and the *MRTSWATH* software to reproject to the reference UTM32N co-ordinate system. An example image and the derived binary mask are shown in Figs. 4a and 4b, respectively.

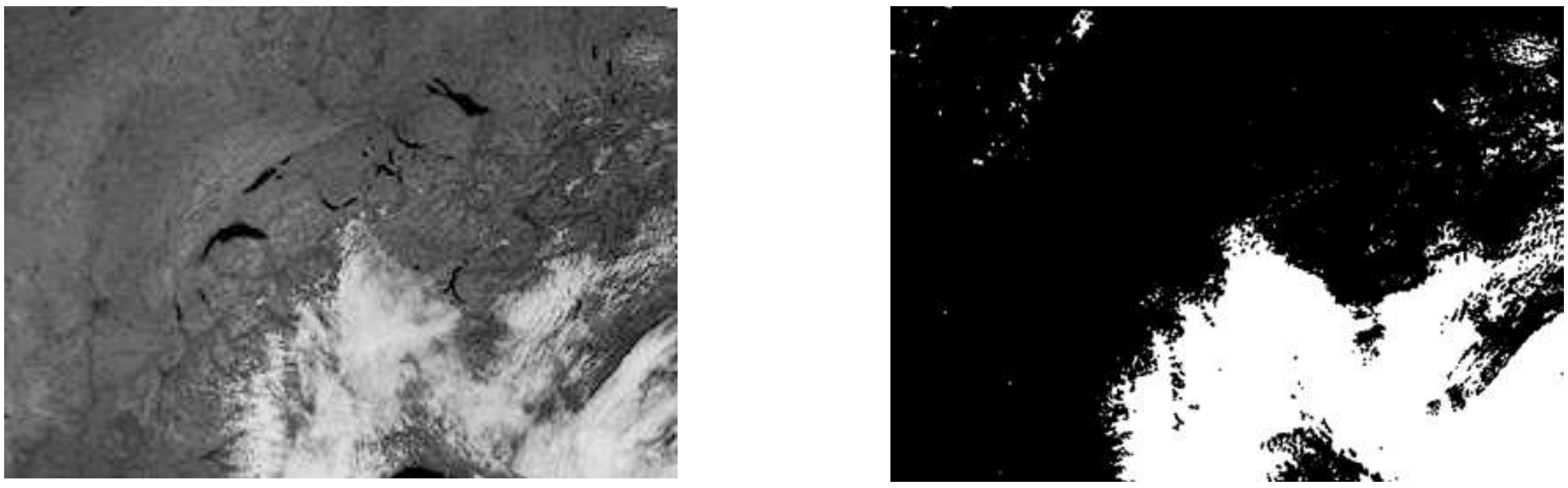

(a) (b)

**Fig. 4.** (a) MODIS band 19 ($B_{19}$) with 1000m GSD (b) binary cloud-mask (*white*-cloudy, *black*-not cloudy) combining the *cloudy* and *uncertain clear* categories.

Each MODIS band has 16-bit grey values of which the invalid pixels have value above 32767. These pixels are also masked out and only the cloud-free and valid pixels are processed. The MODIS cloud-masks are not perfect, still ETHZ rely on them for the moment. Cloud masking is outside the scope of the present project. Our results thus provide a realistic estimate of what can be achieved with the existing erroneous cloud-masks, and a lower bound for a processing chain with improved cloud-masks. The MODIS cloud-mask is available only at a spatial resolution of 1000m. In order to process the data at higher resolution (250m), the mask is upsampled with nearest neighbour interpolation.

An inherent challenge of optical satellite image analysis is missing data due to cloud cover. Images with more than 70% cloudy pixels are not analysed. Cloud detection performance can vary from one sensor to another. Separate cloud-masks are produced for MODIS and VIIRS. Snow/cloud confusion and large view angles can result in considerable over-estimation of clouds and ice (Kraatz et al., 2017). This recent study also reported that there are substantial differences between the MODIS and VIIRS cloud-mask products, especially in the presence of ice. In addition, owing to the fact that the VIIRS and MODIS acquisitions happen at different times of day, the clouds may not be in the same place during the MODIS and VIIRS overpasses on a day, and thus cannot be directly compared.

**Lake outline generation and absolute geolocation correction**. The original outlines of the lakes are downloaded from the Website: http://*overpass-turbo.eu*. These outlines are further generalised by Douglas Peucker Algorithm (Douglas and Peucker, 1973). A comparison of the outlines pre- and post- generalization is shown in Fig. 5. The generalized, digitized lake outlines are projected onto the images to guide the estimation of lake ice. The lake matching approach (Aksakal, 2013) is followed to estimate and correct the absolute geolocation accuracy. Two dates per month are analysed from December 2011 until March 2012. For this, many lakes are used, covering the whole area in and around Switzerland. Fifteen lakes with a minimum area of 500 pixels (in 250m spatial resolution) are incorporated in the analysis. The analysis is performed only if the following two conditions are met: firstly, minimum 40% of each lake area should be cloud-free and secondly, large (partially) cloud-free lakes should exist in at least three out of four corners of the images. The results vary slightly among lakes as well as dates. For each date, the mean translational offsets in both $x$ and $y$ directions are computed by averaging, weighted by the number of cloud-free pixels per lake. In the end, the average final shifts are computed. Final shifts are estimated as -0.75 pixel and -0.85 pixel in $x$ and $y$ directions, respectively. For a comparison, the estimated means shifts of both MODIS and VIIRS sensors are shown in Table 5. For MODIS the correction was rounded to 1 pixel, for VIIRS not. Figs. 6a and 6b shows the backprojected outlines of Lake Sihl on MODIS before (6a) and after (6b) applying the correction. The specified geolocation accuracy of MODIS is 300m (≈ 1.2 pixels at nadir) (Nishihama et al., 1997). For both MODIS and VIIRS, the absolute accuracy error varies slight across lakes. Hence, even after applying a correction based on a global shift value estimated (Table 5), some errors still remain. This can influence the results of smaller lakes such as St. Moritz with very few clean pixels (see Table 6).

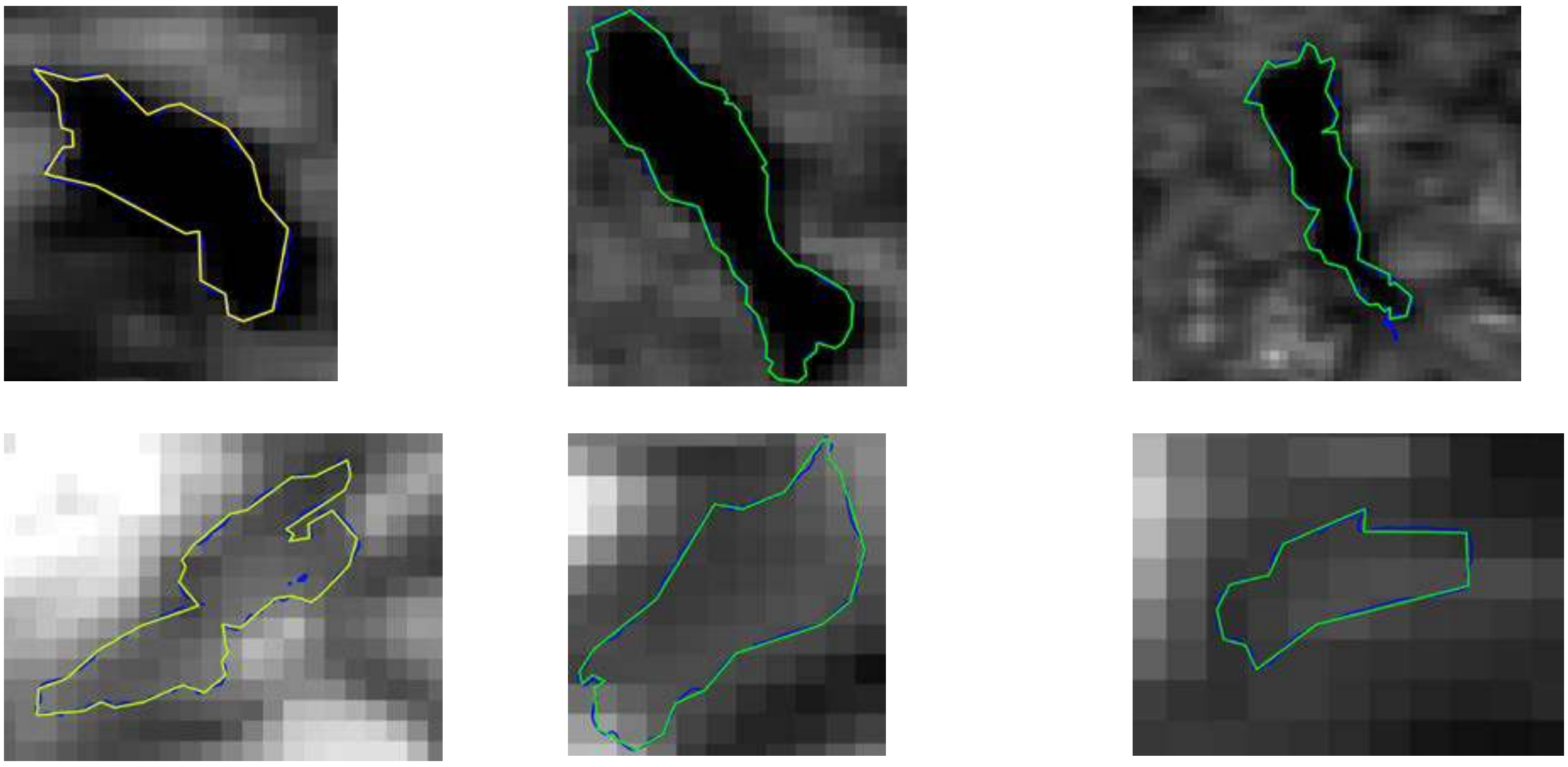

**Fig. 5**. Comparison of original lake outlines (blue) and the generalized outlines (yellow). All images are at 250m resolution but at different zoom levels for better visualization.

**Table 5.** Mean shifts (estimated absolute geolocation errors) for both MODIS and VIIRS. The direction of shifts is shown in Fig. 6c.

| **Sensor** | **Y Shift** (mean, in pixels) | **X Shift** (mean, in pixels) | **Study Period** |
|---|---|---|---|
| MODIS | -0.85 | -0.75 | December 2011 - March 2012 |
| VIIRS | -0.3 | 0.0 | October 2016, December 2016 |

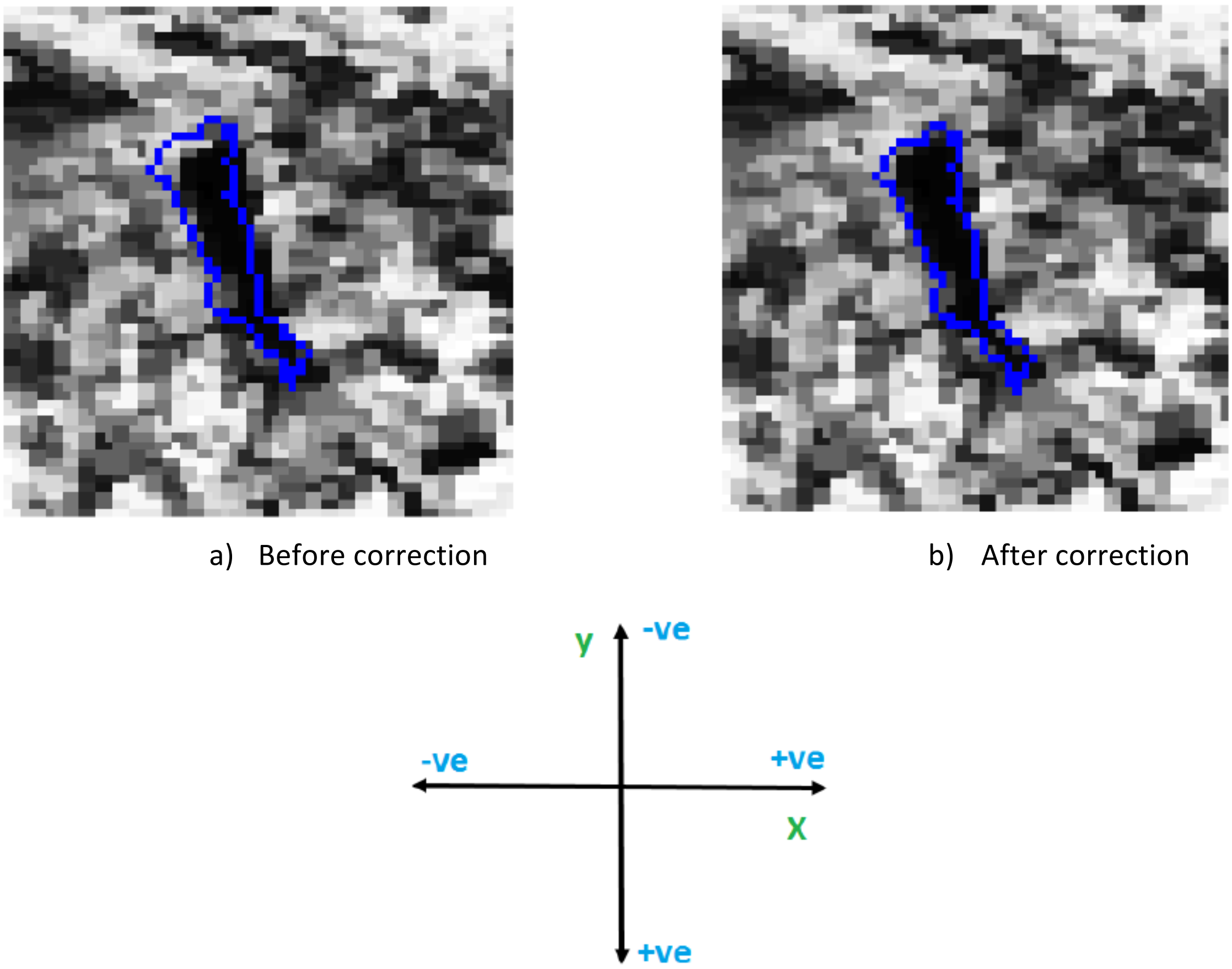

a) Before correction

b) After correction

c) Direction of shifts

**Fig. 6.** Backprojected outline of Lake Sihl (2.8.2012) on MODIS Band 2 ($B_2$, 250m GSD) before absolute geolocation correction (a) and after geolocation correction (b). The direction of shifts are shown in (c).

**Clean pixels vs. mixed pixels**. Mixed pixels lie only partially on the lake as opposed to the clean ones. Only the clean pixels are analysed. To determine the clean pixels, the lake outlines are corrected for absolute geolocation errors, and backprojected on to the MODIS band $B_2$ (250m, see Fig. 2) and the VIIRS band $I_2$ (375m, see Fig. 16). The details of the clean and mixed pixel estimation for all four lakes are shown in Table 6. Since MODIS has two bands with higher spatial resolution compared to VIIRS, it has more clean pixels. Note that the lower the spatial resolution of the images, the higher the proportion of mixed to clean pixels is. This proportion is also increasing the smaller the lake is. Hence, MODIS provides more clean pixels than VIIRS, absolutely and relatively, especially for small lakes like St. Moritz.

### 3.2.2 Band selection

**XGBoost analysis.** XGBoost is a boosting method with shallow decision trees as weak learners. It internally computes variable importance, conditioned on the training labels. The F-score denotes the relative contribution of each band to the model, calculated from the band's contribution to individual decision trees. For more details on XGBoost, see Chen and Guestrin (2016). A higher value implies that the band is more important to make correct predictions. To select the MODIS bands with the best inter-class separability, the cloud-free, clean pixels are fed into XGBoost. The result is shown in Fig. 7. While that method in principle already performs classification, we only use the built-in variable selection. The selected bands are then passed to SVM (Cortes and Vapnik, 1995) for classification (see Section 3.2.3), since it delivered better than XGBoost itself. It can be observed from Fig. 7 that the most significant set of bands to separate frozen dates from non-frozen dates vary across lakes. However, the reflective band $B_1$ and emissive bands $B_{25}$ and $B_{20}$ are significant for all four lakes.

**Table 6.** The number of clean and mixed pixels per acquisition per lake, for MODIS (250m GSD) and VIIRS (375m GSD).

| Lake | MODIS | | VIIRS | |
|---|---|---|---|---|
| | Clean Pixels | Mixed Pixels | Clean Pixels | Mixed Pixels |
| **Sihl** | 115 | 104 | 45 | 63 |
| **Sils** | 33 | 64 | 11 | 37 |
| **Silvaplana** | 21 | 42 | 9 | 24 |
| **St. Moritz** | 4 | 19 | 0 | 11 |

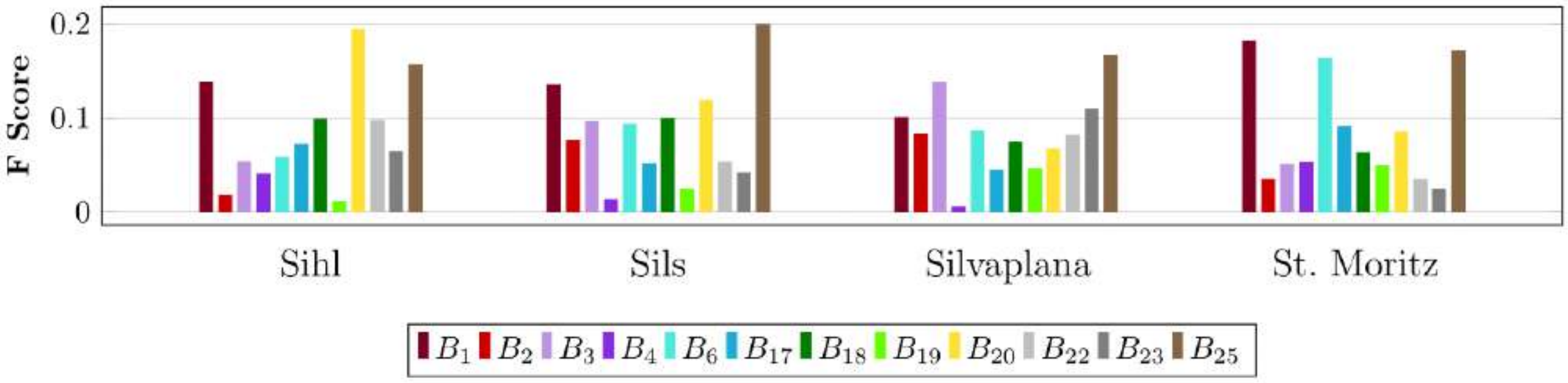

**Fig. 7**. Bar graph showing the significance of each of the 12 selected MODIS bands for frozen vs. non-frozen pixel separation using the *XGBoost* (Chen and Guestrin, 2016) algorithm for all four lakes. All fully frozen and fully non-frozen days during the full winter 2016-17 period (October - April) are included in the analysis.

**An initial experiment.** When a lake is frozen, the water can occur in two different states: firstly, the water freezes and appears as ice and secondly, snow falls and persists on top of the ice. In this experiment, they are considered as two separate sub-classes since the reflective and emissive properties of ice and snow are different. This difference is also reflected in the emissive and reflective spectral responses in different bands of the multi-spectral satellite images. An example for Lake Sils is shown in Fig. 8. For demonstration, band 2 ($B_2$, reflective, 250m resolution) and band 22 ($B_{22}$, emissive, 1000m resolution) are analysed for three different cases: water, snow on ice and snow-free ice. Fig. 8a shows that in $B_2$, the lake pixels appear to be darker when non-frozen (28.8.2012). On the other hand, Fig. 8b illustrates that the pixels are brighter in $B_2$ when frozen and snow exists (6.2.2012). However, it can be observed from Fig. 8c that the pixels are darker in $B_2$ when the lake is frozen, but snow-free (3.1.2012). It can be observed that that in emissive band $B_{22}$, the non-frozen pixels (Fig. 8d) appear brighter as opposed to the frozen pixels (both snow and ice cases in Figs. 8e and 8f, respectively). This example gives two hints. Firstly, snow vs. water and snow vs. ice classification are probably possible with reflective bands. Secondly, the emissive bands might have enough information to distinguish frozen and non-frozen pixels. However, these clues are just based on three selected dates. To confirm these hints, more data needs to be processed.

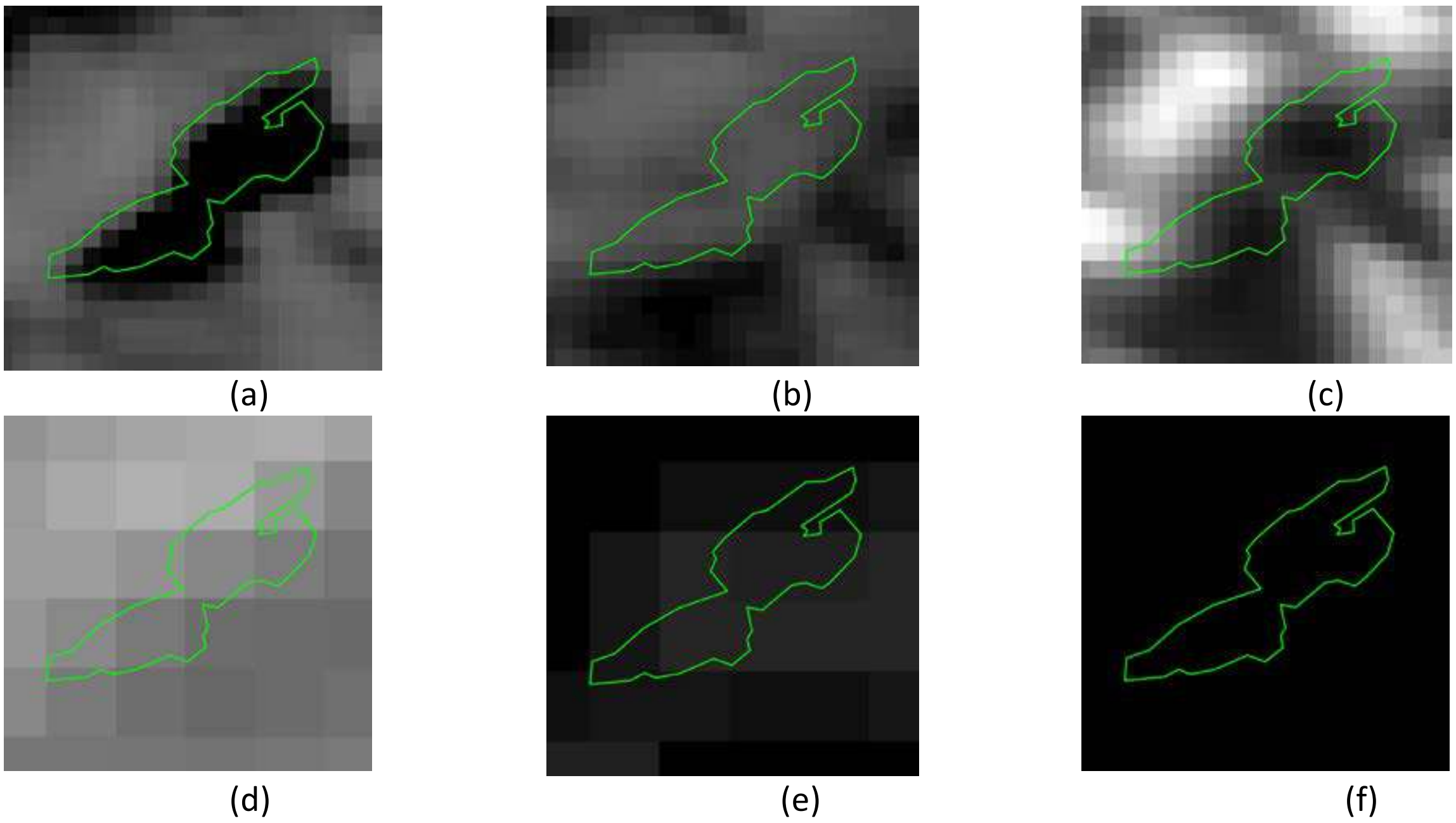

**Fig. 8.** Lake Sils: comparison of lake pixel grey values for non-frozen (left column) frozen and snow covered (middle column) ice, frozen and snow-free ice (right column) in reflective band 2 ($B_2$, 250m GSD, top row) and emissive band 22 ($B_{22}$, 1000m GSD, bottom row). The lake outlines shown are before absolute geolocation correction.

**Grey value histograms.** As a sanity check, we also use the histograms of the target classes in all bands to verify the band selection. For the period of full winter 2016-17 (October - April), the distributions of both frozen and non-frozen pixels in all 12 selected bands of MODIS for Lake Sihl are shown in Fig. 9. It can be inferred from Figs. 7 and 9 that the histograms are in accordance with the band selection by XGBoost. It can be seen that both $B_1$ and $B_2$ have high inter-class separability. Since the two bands are spectrally close (see Table 4) and highly correlated, automatic variable selection chooses only one of them. This applies for other correlated bands too.

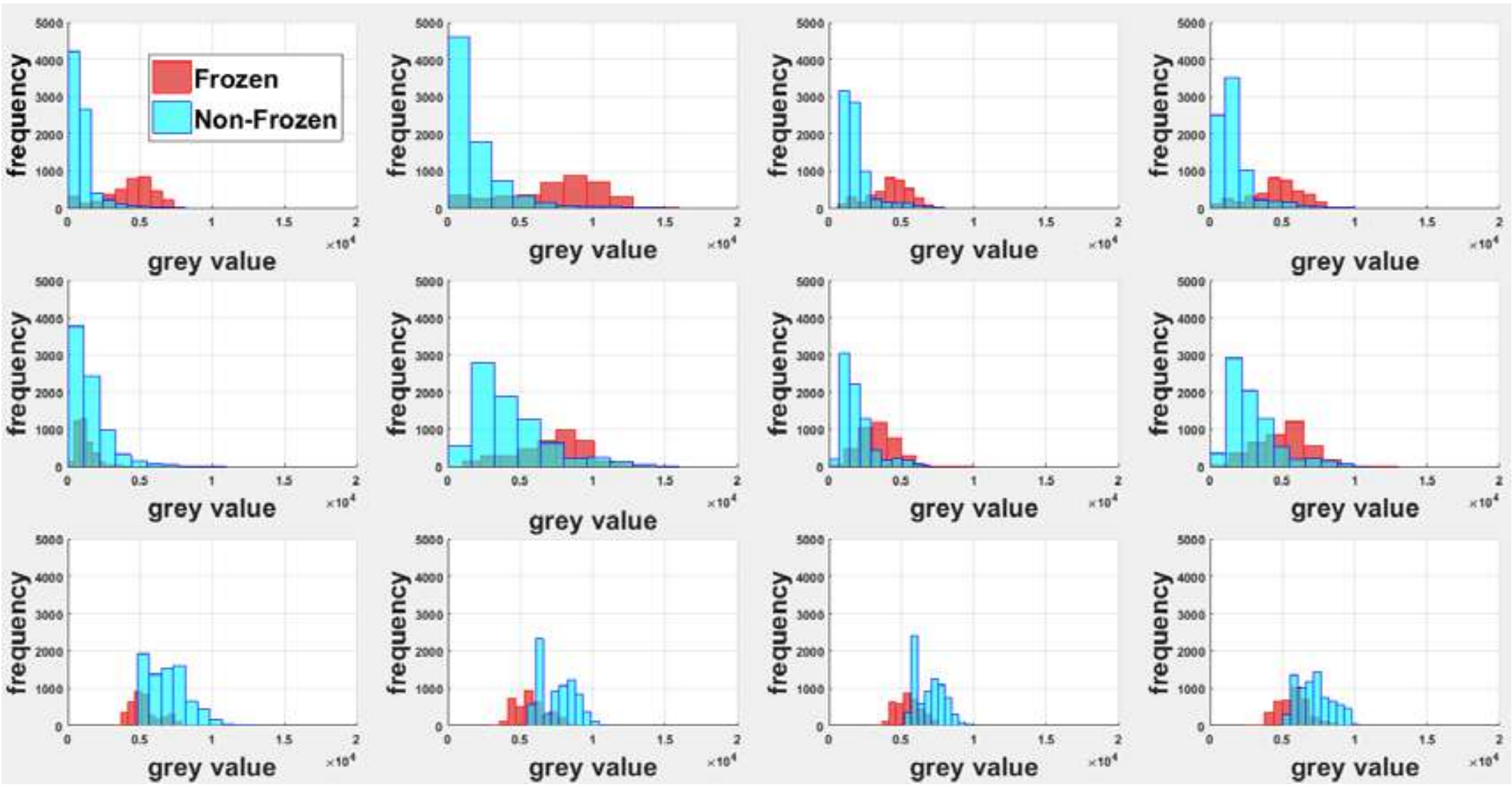

**Fig. 9.** Frozen (red) vs. non-frozen (cyan) statistics of clean cloud-free pixels in 12 selected MODIS bands for Lake Sihl. First row ($B_1$, $B_2$, $B_3$, $B_4$), second row ($B_6$, $B_{17}$, $B_{18}$, $B_{19}$), and third row ($B_{20}$, $B_{22}$, $B_{23}$, $B_{25}$). The band wavelengths are displayed in Table 4. All fully frozen and fully non-frozen days during the winter 2016-17 (1.10.2016-30.4.2017) are included in this analysis.

### 3.2.3 Semantic segmentation methodology

The flow chart of our lake ice detection methodology is shown in Fig. 10. The various image bands are extracted from the raw satellite data after resampling and reprojection to the UTM32N coordinate system. The generalized lake outlines are backprojected on to the image space and corrected for absolute geolocation error. The corrected outlines are then used to identify the clean pixels. Using the binary cloud-mask generated from the cloud-mask product, the cloudy (clean) pixels are discarded.

The lake ice detection pipeline is identical for both MODIS and VIIRS sensors, for MODIS we additionally upsample the low-resolution bands (500m and 1000m GSD) to 250m resolution with bilinear interpolation. Lake ice detection is cast as a two class (*frozen, non-frozen*) semantic segmentation problem and solved using SVM. We experiment with both linear and non-linear Radial Basis Function (RBF) kernels. The RBF kernel non-linearly maps samples into a higher dimensional space. Therefore, unlike the linear kernel, it can handle the case when the relation between class labels and attributes is non-linear. Furthermore, we use the SVM score to estimate the probability for each pixel to be frozen. This is because the frozen and non-frozen areas of a lake can also co-exist in a single pixel. The SVM score is particularly useful in the special case when more than one cloud-free acquisitions are present on the same day. Each acquisition is individually classified using SVM and the scores are averaged to generate one result per pixel per day.

For both MODIS and VIIRS, the images are freely available and can be processed fast, with a processing chain using standard hardware and software tools. The whole pipeline (excluding XGBoost) is implemented and validated using Matlab (version R 2017a). For feature selection using XGBoost, we used the inbuilt python library: *XGBoost.*

**Parameters.** The most important parameters (along with the respective values/choices) are as follows:

- The major sub-parameters of SVM are (this applies to both MODIS and VIIRS processing by ETHZ):
  - *SVM kernel*: This can be set as RBF or linear. In general, RBF gives slightly better results
  - *Kernel scale = 1, kernel offset = 0*
  - *Solver = Sequential Minimal Optimization (SMO)*
  - *Standardize data = 1*
  - *Cost (i,j) = 1 if i≠j, cost (i,j) = 0 if i=j*
  - *Box constraint = 1*
  - *KKT tolerance = 0*
  - *Optimize hyperparameters = none*
- *Cloud percentage*: The minimum percentage of non-cloudy pixels (per acquisition) needed so that the acquisition is processed. We have set this to 30% i.e., acquisitions with more than 70% cloudy lake pixels will not be processed. This applies to both MODIS and VIIRS processing by ETHZ.
- *Absolute geolocation correction*: The lake outlines are corrected (in both X and Y directions) for absolute geolocation error. The respective values used for both MODIS and VIIRS are displayed in Table 5.
- *Bands used*: For MODIS and VIIRS, different combinations bands can be concatenated and fed as feature vector to SVM classifier after standardization.

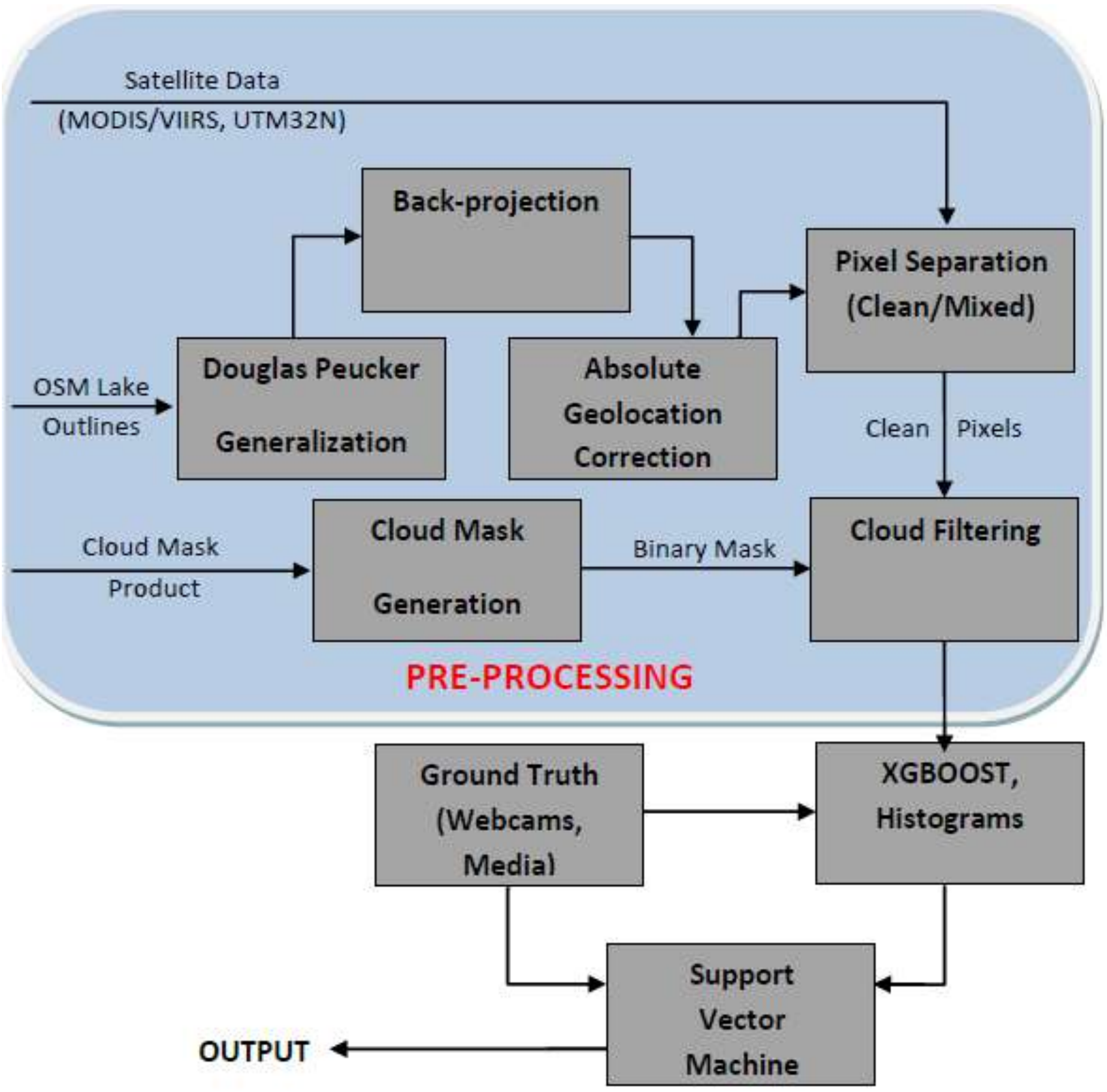

**Fig. 10.** Flowchart of the ETHZ lake ice detection methodology (MODIS and VIIRS).

#### 3.2.4 Post-processing

**Multi-temporal analysis (MTA).** It is very likely that the physical state of a lake (or a portion of it) could be temporally very similar (across adjacent days). Hence, as a post-processing step, for each pixel, a moving average of the SVM scores are computed along the time dimension. The average is calculated for a fixed window length (smoothing parameter) that is determined heuristically. The window, centred at the pixel of interest, slides down along the temporal axis, computing an average over the elements within each window. Since the pixels from each MODIS acquisition are predicted independently by the trained SVM model, MTA is expected to improve the SVM results by leveraging on the temporal relationships. Three different averaging schemes are tried: mean, median and Gaussian. A small drawback of the moving window scheme centred at the current pixel is that, the pixels from future acquisitions are needed to process the pixel of interest. However, for the current application, this is not an issue.

### 3.3 Results and discussion

We report results for the following four lakes: Sihl, Sils, Silvaplana and St. Moritz. In order to assess the performance of a machine learning-based lake ice monitoring system, the data from at least two full winters need to be analysed (at least one full winter for training and the rest for testing). Since the per-day manual interpretation of the Webcam data is a time-consuming task, we were only able to obtain the ground truth for one full winter (2016/17).

Firstly, we performed initial investigations with some selected dates in both winter 2011-12 (see Section 3.3.1) and a subset of winter 2016-17 (see Section 3.3.2). However, the datasets in these investigations had some shortcomings such as limited size of the datasets (for Sections 3.3.1 and 3.3.2) and lack of proper ground truth (for Section 3.3.1). Therefore, these initial investigations were not enough to draw final conclusions. Nevertheless, they provided some vital clues for designing a lake ice monitoring system using data from a full winter (2016/17). Therefore, our primary period of analysis covers one full winter: 01st October 2016 - 30th April 2017 (Section 3.3.4), from which we draw the final conclusions.

**k-fold cross-validation.** In *k-fold* cross-validation setup, the dataset is randomly partitioned (based on date) into *k* equal sized subsets. One of the *k* subsets will be used as test set, the remaining *k-1* sets for training. This process is then repeated *k* times so that each data will be used for testing once. The partitioning is done based on the date, to minimize the temporal correlation between train and test folds.

### 3.3.1. Winter 2011/12 results

We performed the following two experiments using subsets of data from winter 2011/12:

**Experiment 1.** As an initial experiment, ETHZ processed a subset of dates (see Table 7) from winter 2011-12 and summer 2012. The summer 2012 was used to be absolutely sure about the class: *water*. It can be inferred from Table 7, that the dataset used is more representative for winter/summer distinction than for frozen/non-frozen separation and it does not include freeze-up and break-up transition periods.

**Table 7.** Ground truth information on the cloud-free frozen and non-frozen dates (experiment 1, easy dates) processed in year 2012.

| | Lake Sihl | | Lake Sils | | Lake Silvaplana | | Lake St. Moritz | |
|---|---|---|---|---|---|---|---|---|
| | **Period** | **Days** | **Period** | **Days** | **Period** | **Days** | **Period** | **Days** |
| **Frozen** | 06-25 February 2012 | 7 | January, February 2012 | 23 | January, February 2012 | 23 | January, February 2012 | 23 |
| **Non-Frozen** | August 2012 | 13 | July, August 2012 | 26 | July, August 2012 | 21 | July, August 2012 | 22 |

Different combinations of MODIS bands as feature vector are experimented and the results (4-fold cross-validated) are shown in Table 8. For all four target lakes, for this relatively easy dataset, we obtain 100% classification accuracy with SVM and RBF kernel.

**Table 8.** Comparison of 4-fold cross-validated SVM (RBF kernel) results with different feature vectors and super-resolution strategies. $SupReME_{RE}$ and $SupReME_{R}$ indicate the configurations including and excluding the emissive bands. The overall accuracy (defined as an error metric from the very well known confusion matrix) is specified. The dates displayed in Table 7 are used in this analysis. The best results are indicated in green.

| Feature vector | Super resolution | Lake Sihl | Lake Sils | Lake Silvaplana | Lake St. Moritz |
|---|---|---|---|---|---|
| $B_2$ | Bilinear | 98.0 % | 77.0 % | 73.8% | 86.6 % |
| all 12 bands | Bilinear | 94.2 % | 98.6 % | 99.2 % | 96.7 % |
| $B_3$, $B_4$,$B_6$, $B_{17}$, $B_{18}$, $B_{19}$ | Bilinear | 99.6 % | 99.3 % | 99.8 % | 97.8 % |
| $B_3$, $B_4$,$B_6$, $B_{17}$, $B_{18}$, $B_{19}$ | $SupReME_{RE}$ | 94.4 % | 99.8 % | 100.0 % | 98.9 % |
| $B_3$, $B_4$,$B_6$, $B_{17}$, $B_{18}$, $B_{19}$ | $SupReME_{R}$ | 100.0 % | 98.6 % | 99.6 % | 98.3 % |
| $B_{22}$ | Bilinear | 100.0 % | 100.0 % | 100.0 % | 100.0 % |
| $B_2$, $B_{22}$ | Bilinear | 100.0 % | 100.0 % | 100.0 % | 100.0 % |

The emissive channel $B_{22}$ itself is enough for perfect separation of these easy-dates. However, this is contradicting with the statistics of band $B_{22}$ in Fig. 9. (1.10.2016-30.4.2017). This is because Fig. 9 displays the statistics of pixels from a full winter (more challenging dates) as opposed to experiment 1 dataset (Table 7) which is easier. To investigate further into this, the statistics of all 12 channels are re-plotted (shown in Fig. 11) using the dates listed in Table 7. It can be clearly seen from Fig. 11 that the pixels are separable in band $B_{22}$ (and other emissive bands) for the easy dataset. Effectively, being a winter/summer separation, the importance of the emissive channels is overweighted, and the importance of the reflective channels is underweighted.

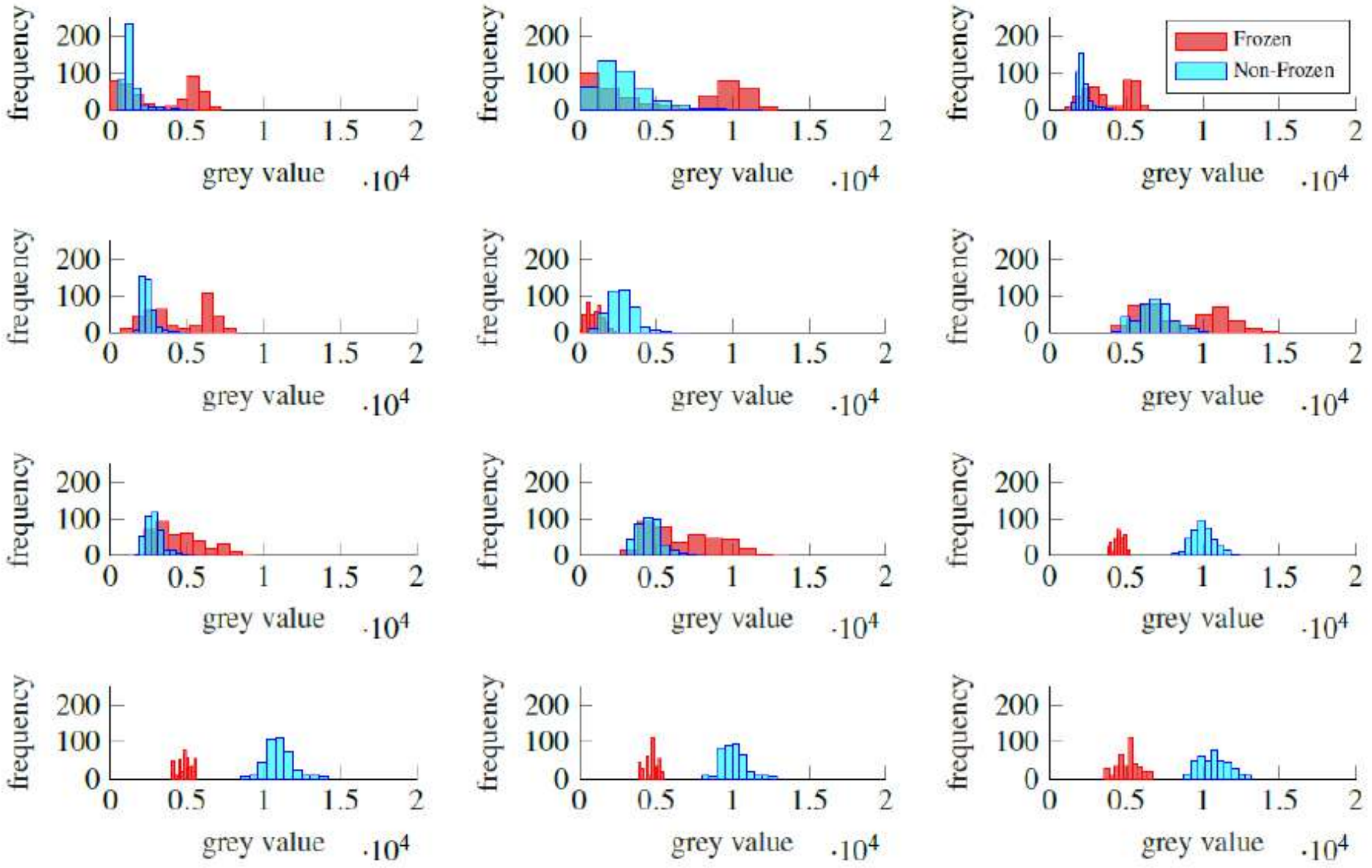

**Fig. 11**. Frozen (Jan, Feb 2012) vs. Non-Frozen (Jul, Aug 2012) statistics of clean cloud-free pixels in 12 selected MODIS channels for Lake Silvaplana. First row ($B_1$, $B_2$, $B_3$), second row ($B_4$, $B_6$, $B_{17}$), third row ($B_{18}$, $B_{19}$, $B_{20}$), and fourth row ($B_{22}$, $B_{23}$, $B_{25}$).

It is to be noted from Table 8, that the performance of all 12 bands combined together is not the best (still >94% for all four lakes) compared to the combination of bands $B_2$ and $B_{22}$ or even $B_{22}$ alone. From Fig. 11, it is clear that not all bands have 100% separability even on the easy dataset (for e.g. bands $B_{17}$, $B_{18}$, $B_{197}$). Ideally, if the dataset is big enough, SVM should automatically learn to weight the features even when all 12 bands are used as feature vector. However, for the current experiment, the dataset size is too small. As a result, the training process has not seen enough samples to assign less weight to the channels with less separability (for e.g. $B_{17}$, $B_{18}$, $B_{19}$). In such cases, it is always good to handpick (based on grey-value statistics or XGBoost analysis) the minimal number of features with maximal separability (in this case any one of the emissive channels). Note also, that in pattern recognition problems, the inclusion of additional parameters beyond a certain point leads to higher probabilities of error (Trunk, 1979).

Table 8 also shows the results (on easy dataset) of SupReME super-resolution approach. For this, the six low-resolution reflective bands ($B_3$, $B_4$, $B_6$, $B_{17}$, $B_{18}$, and $B_{19}$) are used as feature vector after super-resolving them. On this easy dataset, the results of $SupReME_R$ and bilinear interpolation are comparable. However, the results using $SupreME_{RE}$ are slightly worse compared to $SupReME_R$. This is because the reflective and thermal edges are different.

From Fig. 11, we infer that the statistics of all emissive channels are similar and any one of them should be able to separate these frozen (winter) pixels from their non-frozen (summer) counterparts, when fed into the SVM. However, the following two issues exist with experiment 1 setup:

1. Since the dataset involved is very easy, this result cannot be generalized without experimenting with more challenging (representative) non-frozen dates (more closer to the frozen dates).
2. The size of the dataset tested is very small to draw final conclusions on lake ice monitoring.

Experiment 2 below aims at solving the first issue. The second issue still persists with experiment 2. This issue is solved in section 3.3.4, where we process data from one full winter.

**Experiment 2.** A second experiment with more challenging non-frozen dates (which are not far away from the frozen dates compared to the easy dataset, see Tables 7 and 9) is also done, while the frozen dates remain the same. For winter 2011-12, ETHZ collected the non-frozen data from July and August 2012 (Table 7), which are rather less challenging as opposed to the non-frozen dates from March 2012 (Sihl) and November 2011 (Sils, Silvaplana, St. Moritz). These dates for Lake Sihl were selected according to the information from online media (see Table 3). However, for the lakes Sils, Silvaplana and St. Moritz, no information was available from the online media. Hence, we assumed that these three lakes were not frozen in November 2011. This assumption is not fully correct, but not fully wrong (especially for lakes Sils and Silvaplana) too. Hence, this experiment also assesses the performance of the lake ice detector in the presence of ground truth noise. The results of experiment 2 are not conclusive, as they do not include a long time period from water to freeze-up, break-up and again water.

**Table 9.** MODIS (experiment 2) processed in winter 2011-12 at periods with frozen and non-frozen lakes. The table shows the total number of clean, cloud-free pixels (Px) and the total number of acquisitions (Aq), which are at-least 30% cloud-free.

| | **Lake Sihl** | | | **Lake Sils** | | | **Lake Silvaplana** | | | **Lake St. Moritz** | | |
|---|---|---|---|---|---|---|---|---|---|---|---|---|
| | **Period** | **Aq** | **Px** | **Period** | **Aq** | **Px** | **Period** | **Aq** | **Px** | **Period** | **Aq** | **Px** |
| **Frozen** | 06 - 25 Feb. 12 | 7 | 662 | Jan, Feb. 12 | 23 | 656 | Jan, Feb.12 | 23 | 479 | Jan, Feb.12 | 23 | 92 |
| **Non-Frozen** | 20 - 31 Mar. 12 | 6 | 648 | Nov. 11 | 15 | 443 | Nov. 11 | 15 | 300 | Nov. 11 | 14 | 56 |

In experiment 2, the bands 2 ($B_2$, reflective) and 22 ($B_{22}$, emissive) are used to form the feature vectors and classified using SVM (RBF kernel). Bands 2 and 22 originally have 250m and 1000m resolutions, respectively. Band 22 is super-resolved (with different super-resolution methodologies, see Table 10) to 250m resolution prior to analysis. The results of experiment 2 are shown in Tables 10 and 11.

**Table 10.** Comparison of 4-fold cross-validated SVM results (RBF kernel) on dates shown in Table 9 with different super-resolution strategies. $SupReME_{RE}$ indicates the configuration including the emissive bands. The overall accuracy is specified. The best results are indicated in green.

| **Feature vector** | **Super resolution** | **Lake Sihl** | **Lake Sils** | **Lake Silvaplana** | **Lake St. Moritz** |
|---|---|---|---|---|---|
| $B_{22}$ | Bilinear | 98.6 % | 98.1 % | 96.0 % | 92.6 % |
| $B_{22}$ | $SupReME_{RE}$ | 95.7 % | 95.4 % | 90.8 % | 97.3 % |
| $B_2$, $B_{22}$ | Bilinear | 95.1 % | 97.9 % | 99.9 % | 100.0 % |
| $B_2$, $B_{22}$ | $SupReME_{RE}$ | 97.9 % | 96.7 % | 94.7 % | 96.0 % |

It can be inferred from Tables 8 and 10, that the results of SupReME are slightly worse than bilinear interpolation. Though the results using $SupReME_R$ were slightly better than $SupReME_{RE,}$ it does not make sense to upsample the emissive band $B_{22}$ with $SupReME_R$ (excluding the emissive bands). Hence, the results of $SupReME_{RE}$ are compared with that of bilinear interpolation. It can be inferred from Tables 8 and 10, for lake ice monitoring, there is no notable benefit using SupReME when compared to bilinear interpolation. Therefore, SupReME is excluded from the further experiments and the more standard bilinear interpolation is used in all following experiments.

Table 11 shows a comparison of results of experiment 2 with various SVM kernels. It can be seen that, for the experiment 2 dataset, the RBF kernel works better compared to the linear kernel. For Lake Sihl, effectively it is snow vs. water classification. Hence, even the reflective band $B_2$ achieves very good results. Along with snow-covered days, the lakes Sils, Silvaplana and St. Moritz had frozen but snow-free days too. This is the reason why the reflective band $B_2$ alone did not perform well (for both RBF and linear kernels) for these three Lakes. The performance is still relatively good for Lake St. Moritz, which had few days with snow-free ice. In general, adding the emissive band $B_{22}$ to the feature vector solved the problem, and using all twelve bands brought no further improvement. In some cases, the performance of 12 bands decreased slightly, likely due to overfitting to our relatively small training set (as in the case of experiment 1). For Lake Sihl, adding $B_{22}$ causes a significant performance drop, because of two outlier dates (frozen date 24.2.2012 and non-frozen date 20.3.2012). These two dates have similar grey value distribution in the emissive band, such that the data is no longer linearly separable. The RBF kernel alleviates this problem, but does not solve it completely.

**Table 11.** Results from winter 2011-12 for MODIS data on the dates of Table 9. Comparison of 4-fold cross-validated SVM results. The low spatial resolution bands are super-resolved with bilinear interpolation. The best results are indicated in green.

| Feature vector | SVM kernel | Lake Sihl | Lake Sils | Lake Silvaplana | Lake St. Moritz |
|---|---|---|---|---|---|
| $B_2$ | Linear | 99.2 % | 81.1 % | 79.1 % | 96.0 % |
| $B_{22}$ | Linear | 82.4 % | 97.7 % | 93.2 % | 94.6 % |
| $B_2$, $B_{22}$ | Linear | 91.5 % | 98.5 % | 99.9 % | 100.0 % |
| All 12 bands | Linear | 83.1 % | 98.5 % | 100.0 % | 100.0 % |
| $B_2$ | RBF | 99.6 % | 81.3 % | 80.1 % | 96.0 % |
| $B_{22}$ | RBF | 98.6 % | 98.1 % | 96.0 % | 92.6 % |
| $B_2$, $B_{22}$ | RBF | 95.1 % | 97.9 % | 99.9 % | 100.0 % |
| All 12 bands | RBF | 87.5 % | 98.5 % | 99.6 % | 100.0 % |

The above results are also not conclusive since the training set is still small and as a result overfitting is observed. Hence, the superior performance of RBF kernel over linear kernel needs to be investigated further on much bigger and representative datasets (see Section 3.3.4) before we draw final conclusions.

#### 3.3.2. Winter 2016/17 (subset) results

ETHZ also report results on a winter subset (October, first half of December and February) because of the same dates of the initial VIIRS data (see Table 12). These results were used for MODIS winter generalization. In order to avoid confusion, this dataset will be addressed as winter 2016/17 (subset).

As a comparison with the two datasets (experiments 1 and 2) already reported in section 3.3.1, the winter 2016-17 (subset) has more challenging non-frozen dates (colder December 2016 dates as opposed to November, July and August dates in 2012) but less challenging frozen dates (February 2017 dates compared to January and February dates in 2012). This is because, the number of snow-free ice dates were relatively very less (especially for lakes Sils and Silvaplana) in the winter 2016/17 (subset). Overall, the winter 2016/17 (subset) dataset is relatively less complex compared to both the datasets tested in winter 2011/12.

Different band combinations have been tested and the results are displayed in Table 13. For these dates tested, simple linear SVM gave near-perfect results. Hence, experiments with RBF kernel are not conducted. The best results are obtained when all 12 selected bands are fed into the feature vector. It can be seen that, the reflective band $B_2$ itself is enough to obtain a very good classification result. However, band $B_2$ alone was not enough to get good results (especially for lakes Sils and Silvaplana) for winter 2011/12 (see Tables 8 and 11). This is because, as opposed to winter 2011/12, the number of snow-free ice dates were relatively very less (especially for lakes Sils and Silvaplana) in the winter 2016/17 (subset).

**Table 12.** Details of the subset of cloud-free dates processed in winter 2016-17. Frozen data comprises of the acquisitions from February 2017, non-frozen data from October 2016 and December 2016. Only the first half of December 2016 are processed. The table shows the total number of clean, cloud-free pixels (Px) and the total number of acquisitions (Aq), which are at-least 30% cloud-free.

| | Lake Sihl | | | | Lake Sils | | | | Lake Silvaplana | | | | Lake St. Moritz | | | |
|---|---|---|---|---|---|---|---|---|---|---|---|---|---|---|---|---|
| | MODIS | | VIIRS | | MODIS | | VIIRS | | MODIS | | VIIRS | | MODIS | | VIIRS | |
| | Aq | Px | Aq | Px | Aq | Px | Aq | Px | Aq | Px | Aq | Px | Aq | Px | Aq | Px |
| **Frozen** | 7 | 805 | 19 | 774 | 8 | 263 | 15 | 153 | 11 | 227 | 17 | 140 | 9 | 36 | - | - |
| **Non-Frozen** | 17 | 1809 | 29 | 1222 | 15 | 471 | 32 | 328 | 18 | 349 | 32 | 289 | 13 | 52 | - | - |

Additionally, as opposed to Section 3.3.1, the performance of emissive band $B_{22}$ alone is not good. Here, we process additional non-frozen data from the slightly colder December dates (compared to non-frozen dates from November in experiment 2). For the emissive band $B_{22}$, the grey value statistics of the pixels for the dates listed in Table 12, are shown in Fig. 12. Left and right graphs show the statistics including and excluding December 2016 pixels. One can see that the latter shows relatively cold temperatures, which intuitively are more easily confused with cold snow/ice pixels from February 2017 in the emissive bands. This explains the confusion when band $B_{22}$ alone is used as feature vector.

The winter 2016/17 (subset) is a very easy dataset and hence not enough to draw a final conclusion on lake ice monitoring. However, the tests with this dataset showed that the selected bands (feature vectors) for 2011-12 were not sufficient for 2016-17. There can also be year-to-year variability in lake ice formation.

**Table 13**. Results from winter 2016-17 (subset) for MODIS data: Comparison of 4-fold cross-validated SVM (linear kernel) results with different band combinations as feature vector. The overall accuracy is given. The best results are indicated in green.

| Feature vector | Lake Sihl | Lake Sils | Lake Silvaplana | Lake St. Moritz |
|---|---|---|---|---|
| $B_2$ | 97.5 % | 99.3 % | 98.8 % | 100.0 % |
| $B_{22}$ | 89.9 % | 81.1 % | 75.0 % | 72.7 % |
| $B_2$, $B_{22}$ | 99.2 % | 100.0 % | 100.0 % | 100.0 % |
| All 12 bands | 99.9 % | 100.0 % | 100.0 % | 100.0 % |

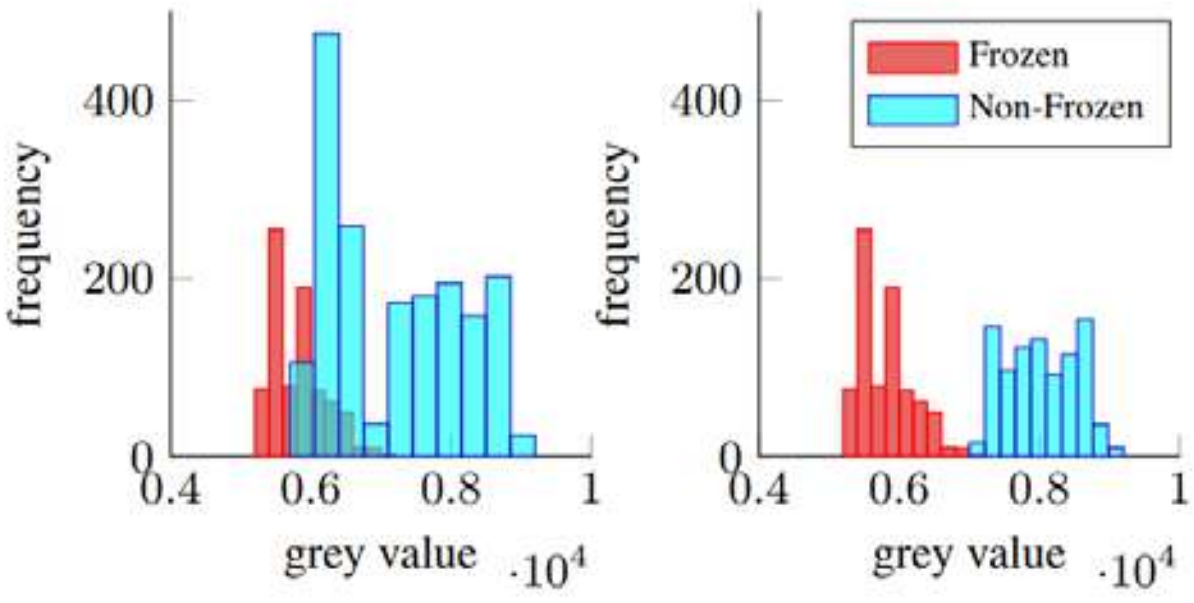

**Fig. 12.** Statistics of pixels in MODIS emissive band $B_{22}$ for Lake Sihl from winter 2016-17 (subset). Bottom left and right parts show the statistics including and excluding pixels from December 2016, respectively.

### 3.3.3. Miscellaneous experiments

Firstly, the MODIS data from both winters (Tables 9 and 12) are combined into one dataset and processed. The 4-fold cross-validated results are shown in Table 14. Overall, the combined dataset clearly improves the correctness of the result. Since both the training and test sets are a mix of images from both winters, we have effectively increased the amount of training data, while the difficulty of the problem remains about the same. In general, the linear kernel performs similar (or even slightly better) than the RBF kernel. This hints that overfitting (compared to winter 2011/12) is reduced (but not fully eliminated, see the performance of 12 bands compared to other feature vector combinations) with more data in the training set. Hence, the experiment provides an indication that the previous datasets (winter 2011-12 and winter 16-17 subset) were still not large enough for the performance to saturate, and the classification can be expected to become more reliable if it receives more training data. What is still missing with the combined dataset are the dates from transition (freezing and melting) periods, where the frozen and non-frozen pixels co-exist in a lake (on the same day).

**Table 14.** MODIS results on the combined data from both winter 2011-12 and winter 2016-17 (subset). The best results are indicated in green.

| Feature vector | SVM kernel | Lake Sihl | Lake Sils | Lake Silvaplana | Lake St. Moritz |
|---|---|---|---|---|---|
| $B_2$ | Linear | 98.2 % | 86.1 % | 86.3 % | 97.5 % |
| $B_{22}$ | Linear | 87.7 % | 92.0 % | 86.1 % | 87.7 % |
| $B_2$, $B_{22}$ | Linear | 95.1 % | 96.5 % | 97.9 % | 97.9 % |
| All 12 bands | Linear | 94.9 % | 99.8 % | 99.7 % | 100.0 % |
| $B_2$ | RBF | 97.7 % | 86.6 % | 85.9 % | 97.6 % |
| $B_{22}$ | RBF | 87.5 % | 91.7 % | 86.7 % | 89.8 % |
| $B_2$, $B_{22}$ | RBF | 96.4 % | 97.3 % | 97.0 % | 97.9 % |
| All 12 bands | RBF | 92.2 % | 99.0 % | 99.9 % | 94.5 % |

The patterns in lake ice formation could vary from one winter to another. As a second experiment, for each lake, an SVM model is trained using all images from winter 2011-12 (Table 9), and then tested on all images from winter 2016-17 (Table 12). This experiment assesses the stability of the classifier over time and how well a classifier trained on one winter transfers to another one. The results are displayed in Table 15. They should be compared with Table 13, since they both share the same test set. While there is a noticeable drop in accuracy (≈5% for lakes Sihl, Sils and Silvaplana, twice as much for St. Moritz) the classification is still correct for 90% or more of the pixels, on all lakes. Owing to the fact that the test sets are different, the results in Table 15 are not directly comparable with Tables 11 and 14.

**Table 15.** Generalization across winters: MODIS results when the SVM is trained with data from winter 2011-12 and tested with data from winter 2016-17 subset.

| Feature vector | SVM kernel | Lake Sihl | Lake Sils | Lake Silvaplana | Lake St. Moritz |
|---|---|---|---|---|---|
| $B_2$, $B_{22}$ | Linear | 96.3 % | 88.4 % | 90.5 % | 86.4 % |
| All 12 bands | Linear | 92.9 % | 97.4 % | 94.1 % | 89.8 % |

The patterns in ice formation could vary across lakes too. In the final experiment, we train the SVM model on the data from three lakes (combined dataset of Tables 9 and 12) and test on the fourth lake. The results are displayed in Table 16. This experiment serves to assess the transferability of the trained model across lakes. It can be seen that our model fairs well in generalizing from one lake to another

**Table 16.** Generalization across lakes: MODIS results when the SVM (linear kernel) is trained on three lakes and tested on the fourth combining data from both winter 2011-12 and winter 2016-17 (subset).

| Feature vector | Lake Sihl | Lake Sils | Lake Silvaplana | Lake St. Moritz |
|---|---|---|---|---|
| $B_2$, $B_{22}$ | 98.1 % | 97.4 % | 97.9 % | 97.0 % |
| All 12 bands | 96.6 % | 98.7 % | 98.2 % | 96.6 % |

It can thus be expected that data for a limited set of lakes, the data from few winters, will be enough to train a classifier which can then be applied to any unseen image. This would clearly be a valuable asset for an operational system.

As we have already discussed, the winter 2016/17 (subset) dataset is relatively less complex than the winter 2011/12 dataset. When both of them are combined, we have effectively increased the amount of training data, while the complexity of the combined dataset remains about the same as that of winter 2011/12 dataset. Additionally, Table 15 reports the results when the pixels of a lake are tested with the model trained using other pixels but from the same lake. On the other hand, Table 16 reports the results when the pixels of a lake are tested using a model generated from the combination of pixels from other three lakes. This explains why the results in Table 15 are better than Table 16.

Two channels produce better results than twelve channels. This is because, the training dataset is not large enough and hence an overfitting issue may exist. As discussed before, feature selection (using XGBoost or histograms) is critical in such situations to discard the channels (features) with less separability.

As specified before, ideally, the data from at least two winters need to be analysed to assess the performance of a machine learning-based lake ice monitoring system. Out of that, at least one full winter must be used in training the SVM model. In addition, the pixels from one full winter must be tested using the trained model. Due to the time-intensive ground truth labelling process, we have restricted our analysis (both testing and training combined) to one full winter (see Section 3.3.4 for MODIS and Section 4.3.2 for VIIRS). This is not ideal, but enough to draw conclusions for a feasibility study where the goal is to prove that the methods work. When more ground truth becomes available, we recommend repeating the analysis with more training data.

### 3.3.4. Full winter 2016/17 (October - April) results

We tested our algorithm on all available dates (at least 30% cloud-free) during the full winter period spanning from 1.20.2017-30.4.2017. On some dates, more than one MODIS acquisitions are available. On the contrary, there exist very few dates with no acquisition at all (for the selected region of interest). Many acquisitions are not processed due to the presence of clouds above the lake. Firstly, all data from the seven months is divided into two categories: *non-transition* dates and *transition* dates. The data from dates in which a lake is fully frozen (at least 90% area approx.) or fully non-frozen (0-10% approx.) belong to the non-transition category, while the partially (10-90% approx.) frozen dates form the transition counterpart. Both *freezing* and *melting* dates belong to the transition category. The ground truth as described in Section 2 is used for this grouping of data. For training the classifier, only the non-transition dates are used. This is to ensure the presence of only the reliable reference data in training set. However, both non-transition as well as transition dates are tested using the trained SVM models. The non-transition dates are further split (manually) into two almost equal halves (*split1* and *split2*, temporally interleaved) based on the following criteria:

- Acquisitions from the same day should remain within the same split.
- Acquisitions are temporally interleaved such that two (day step = 2) adjacent cloud-free acquisitions together belong to the same split. For example, in a set of 12 cloud-free daily acquisitions (d1 to d12), the six acquisitions d1, d2, d5, d6, d9, d10 belong to the split 1 and the rest belong to split 2.

Thus, for each lake, there will be two SVM models trained (*model 1* and *model 2*). All transition dates are tested with both the SVM models. For example, for Lake Sihl, *model 1* is trained using the following data:

- *Split 1* acquisitions of Lake Sihl from winter 2016-17.

- All data from non-transition dates (both splits 1 and 2) of lakes Sils, Silvaplana and St. Moritz from winter 2016-17. This is done to increase the variety in the training set and to induce generalization capability across lakes.
- All non-transition data of Lake Sihl from winter 2011-12. This is done to induce generalization capability across winters. The data from November 2011 was not included in the training set to avoid noisy labels.

*Model 2* for Lake Sihl is trained using the following data:

- *Split 2* acquisitions of Lake Sihl from winter 2016-17.
- All data from non-transition dates of lakes Sils, Silvaplana and St. Moritz from winter 2016-17.
- All non-transition data of Lake Sihl from winter 2011-12. The data from November 2011 was not included in the training set to avoid noisy labels.

Once both models are trained, *split 1* data is tested using *model 2* and *split 2* data using *model 1*. The transition dates are tested using both the models. Choosing the training dataset is very critical and influences the final result. Ideally, the data from at-least one full winter need to be present in the training set to generate a robust model. The more variable the data in the training set, the better the model that will be generated.

For each lake, the five most significant bands selected by XGBoost are shown in Table 17. These five channels are fed as feature vector to the SVM classifier to create the trained models. We also report results on other channel combinations (inspired from experiments in Sections 3.3.1, 3.3.2 and 3.3.3).

**Table 17.** Five most significant MODIS bands selected by XGBoost algorithm for each lake for *frozen* vs. *non-frozen* pixel separation. Data from full winter 1.10.2016-30.4.17 is used in this analysis.

| **Lake** | **Bands** |
|---|---|
| Sihl | $B_{20}$, $B_{25}$, $B_1$, $B_{18}$, $B_{22}$ |
| Sils | $B_{25}$, $B_1$, $B_{20}$, $B_{18}$, $B_3$ |
| Silvaplana | $B_{25}$, $B_3$, $B_{11}$, $B_1$, $B_6$ |
| St. Moritz | $B_1$, $B_{25}$, $B_6$, $B_{17}$, $B_{20}$ |

**Quantitative Results.** Quantitative analysis is done only on the non-transition dates and the corresponding results are shown in Table 18. The overall accuracy is shown for all four lakes. We have experimented with various combinations of bands as feature vector as well as with linear and RBF kernels of SVM. For all lakes, the best results (rows 6 and 7) are obtained when all 12 bands are used as feature vector. This shows that, compared to the experiments in section 3.3.1, 3.3.2 and 3.3.3, the current dataset size is big enough which effectively nullifies overfitting. In addition, the performance of RBF is superior to linear kernel (compare rows 5 and 6, also rows 3 and 4). With RBF kernel, the top five bands selected by XGBoost are enough to get very good results (row 4).

**Table 18.** Quantitative results on MODIS data from full winter 2016-17 for different feature vectors and SVM kernels. The overall accuracy is shown. The best results are indicated in green. The last row shows the results with MTA.

| **Feature vector** | **SVM kernel** | **Lake Sihl** | **Lake Sils** | **Lake Silvaplana** | **Lake St. Moritz** | **With MTA** |
|---|---|---|---|---|---|---|
| $B_2$, $B_{22}$ | RBF | 91.7 % | 84.6 % | 82.7 % | 72.6 % | No |
| Top 5 bands | Linear | 90.0 % | 84.1 % | 81.3 % | 72.1 % | No |
| Top 5 bands | RBF | 91.9 % | 93.2 % | 91.7 % | 86.2 % | No |
| All 12 bands | Linear | 92.3 % | 87.0 % | 81.3 % | 76.4 % | No |
| All 12 bands | RBF | 93.1 % | 94.0% | 92.8 % | 87.9 % | No |
| All 12 bands | RBF | 97.1 % | 97.5 % | 96.1 % | 93.7 % | Yes |

**Qualitative results.** Qualitative analysis is done on all dates (both transition and non-transition), which are at least 30% cloud-free. In case of dates with only one MODIS acquisition, the SVM scores and classification results are used as such. However, in case of more than one acquisition on the same date, these acquisitions are processed separately and then SVM scores are averaged (individually for each pixel). For each pixel, SVM score is a value from 0 to 100 (0:frozen, 100:non-frozen) directly proportional to the probability of being non-frozen. To generate one result per pixel per day, each pixel is then classified as non-frozen if the averaged SVM score is greater than 50 and as frozen if the score is less than or equal to 50. This averaging scheme applies to VIIRS processing too (Section 4.3.2).

Fig. 13 displays the final results on all four lakes for full winter 2016/17 (1.10.2016-30.4.2017). Timelines of lakes Sihl, Sils, Silvaplana and St. Moritz are shown in Figs. 13a, 13b, 13c and 13d, respectively. The SVM classification result (one value per day) showing the percentage of non-cloudy pixels classified as non-frozen (NF) are shown on the y-axis and the corresponding (available) dates which are at-least 30% cloud-free are shown on the x-axis (chronological order, one representative per available date). A date is not displayed if it is more than 70% cloudy or if there is no MODIS data present on that day. Each month is displayed with a distinct colour. The ground truth is plotted in each figure as a continuous cyan line. For better visualization, the ground truth plot shows the following classes:

- class: water (label = w, value = 100)
- class: more water (label = mw value = 75)
- class: more ice/snow (label = mi/ms, value = 25)
- class: ice/snow (labels i/s, value = 0)

The above values (100,75,25,0) for ground truth are used just for better visualization.

For Lake Sihl, it can be seen from Fig. 13a that most of the data from November 2016 and first half of February 2017 is not available for processing. It can also be observed that, during some of the dates in the freezing period (end of December), when the ground truth class is *more water (value = 75)*, the classifier picks the lake as fully non-frozen. The ground truth of Lake Sihl is primarily based on the Webcams, which cover the northern part of the lake, and the sparse information from Sentinel-2 data. This means, there exists some uncertainty in the ground truth. If the ground truth is not fully correct and if a non-frozen (water) pixel is labelled as snow-free thin ice (frozen) or the other way around, this can cause noisy labels in the training set, which can corrupt the SVM model generated. This could end up in wrong classification of thin ice as water and/or vice-versa. This thin ice vs. water confusion could also be due to the inability of satellite images to distinguish between those two classes. This issue can also be noted for Silvaplana (13c) and St. Moritz (13d). During the visual evaluation of Sentinel-2 images for ground truth generation, it was observed that, even for human operators it is sometimes difficult to separate water and thin ice (see Section 2).

For Lake Sihl, no Webcam data is available from 1.10 – 3.12. We assumed that the Lake Sihl was not frozen during this period. This is assumption is supported by the Sentinel-2A observations during this period which confirmed that the lake was fully non-frozen (on the days Sentinel-2A was available and cloud-free). This assumption was needed because Lake Sihl is the biggest of all four lakes and the October and November pixels of Lake Sihl, if left out from the training set, could significantly affect the size of the dataset, which might result in overfitting. Classification results on Lake Sihl (Fig. 13) shows that our assumptions were not wrong.

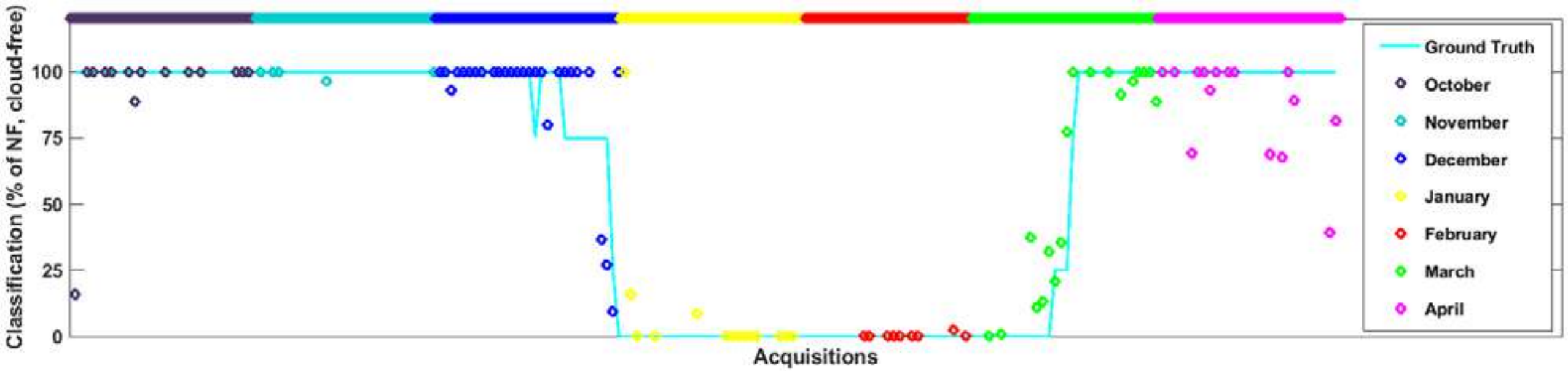

a) Lake Sihl

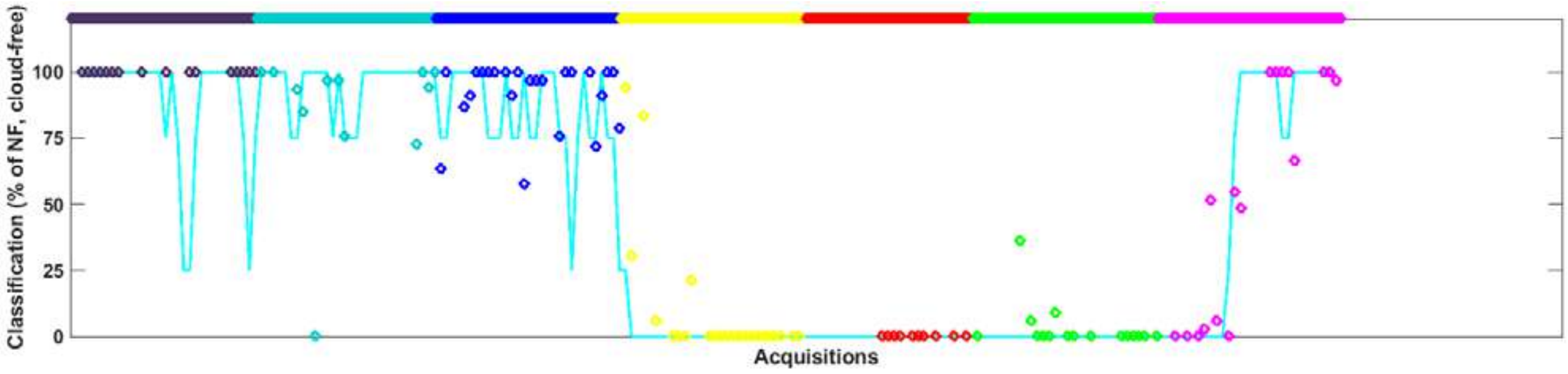

b) Lake Sils

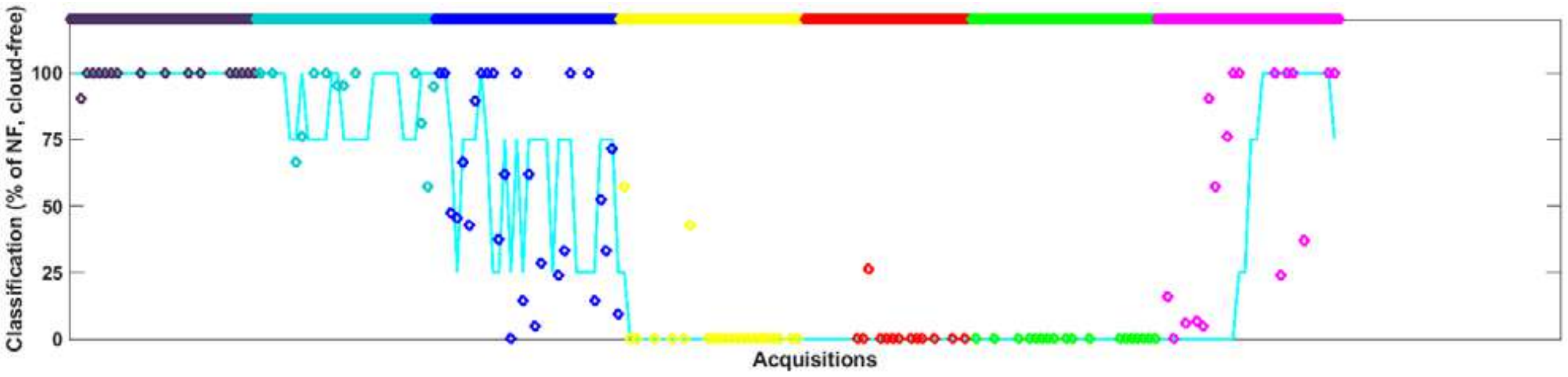

c) Lake Silvaplana

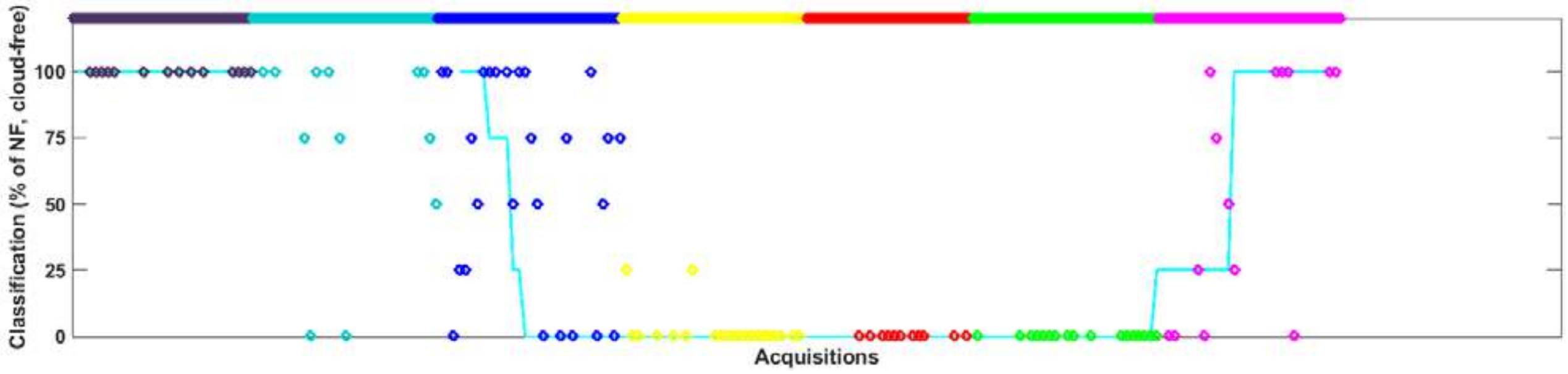

d) Lake St. Moritz

**Fig. 13.** SVM classification results (full winter 2016/17, one value per day, only the result from the best training model (model 1 for Lakes Sihl, Silvaplana, St. Moritz and model 2 for Lake Sils) is shown for the transition dates. For the other dates, either model1 (half data) or model 2 (half data) was used. The ground truth used is shown as a thin cyan line.

Similarly, no Webcam data exist for St Moritz from 1.10 – 3.12.2016. Similar to Lake Sihl, we confirmed from Sentinel-2A that St. Moritz was not frozen in October. However, the lake was partially frozen in November. Hence, the St. Moritz pixels from November till Dec. 3 were excluded from the training set. The SVM results of Lake St. Moritz (Fig. 13d) also validate our assumption (some dates in November were already classified as frozen by the SVM). There also exists a gap in Webcam capture of St. Moritz (see

Appendix 1) during end of March 2017. We do not have any dense (i.e. for every day) ground truth during this period, except for the sparse information from Sentinel-2A and a few images of the lake downloaded from the Internet. The ground truth shown as thin cyan line during this period was interpolated from these sparse sources.

**Multi-temporal analysis (MTA).** In MTA, the temporal moving average of SVM score (after computing one score per day) is computed for each pixel. We used the following three averaging types: mean, median and Gaussian (with a moving window of 3-15 days). Fig. 14 demonstrates the effect of MTA on Lake Sils. The top row displays the SVM classification timeline without MTA (same as MTA with window length one). The second and third rows show the timelines when Gaussian moving average is applied with window lengths 5 and 13 days, respectively. It can be inferred that, for Lake Sils, the higher the window length, the smoother the timeline is. Smoothing is particularly useful to remove outliers, for e.g., a pixel classified as frozen when all (or most) of its temporal neighbours are classified as non-frozen. Smoothing could also wash out the dynamics, e.g., when a part of a lake changes state from completely frozen to partly frozen and back within a short period. This is a trade-off. Generally, short windows of 3-5 days should be preferred.

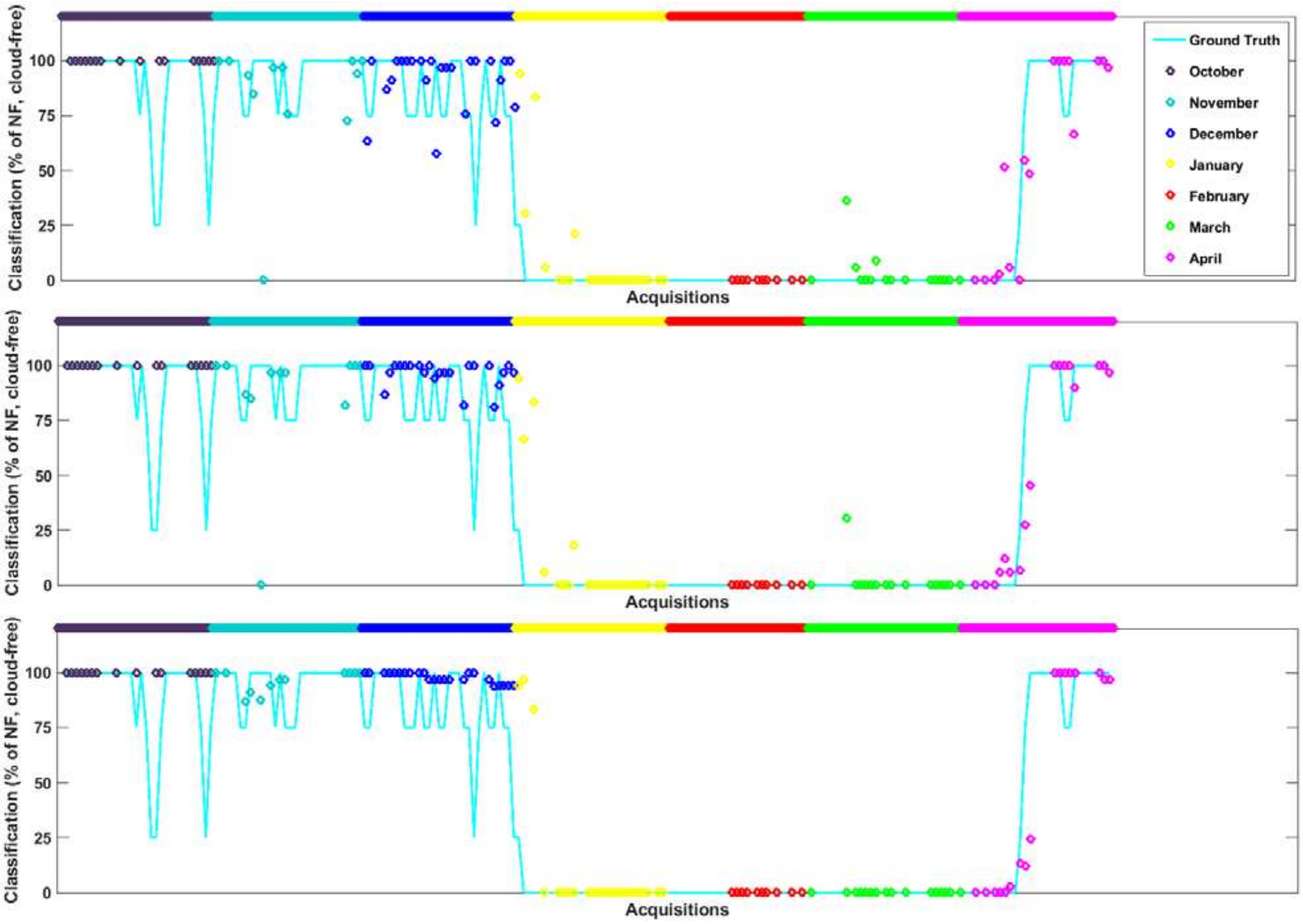

**Fig. 14.** Results of MTA on Lake Sils (full winter 2016/17, one value per day). Top row shows the SVM classification timeline without MTA. Second and third rows show the timelines when Gaussian moving average is applied (window lengths 5 and 13 days, respectively).

MTA was applied only with the configuration, which works the best (all 12 bands as feature vector, RBF kernel). In Table 18, both cases are compared: without (row 6) and with (row 7) MTA and the latter performs better. However, the ideal averaging scheme and window size differs from one lake to another. The best case of all tried averaging schemes and window sizes is shown in the Table 18 (last row). With MTA, there is an improvement in the results.

The effect of each MTA scheme is shown in Fig. 15 for all four target lakes. Figs. 15a, 15b, 15c and 15d show the results of lakes Sihl, Sils, Silvaplana and St. Moritz. In all graphs, the overall accuracy (of non-transition dates) is plotted against the smoothing parameter for all three averaging schemes. Smoothing parameter, i.e. the averaging window in days for mean and median averaging and the width of the Gaussian filter, both in days, with value one is same as the special case when no MTA is performed.

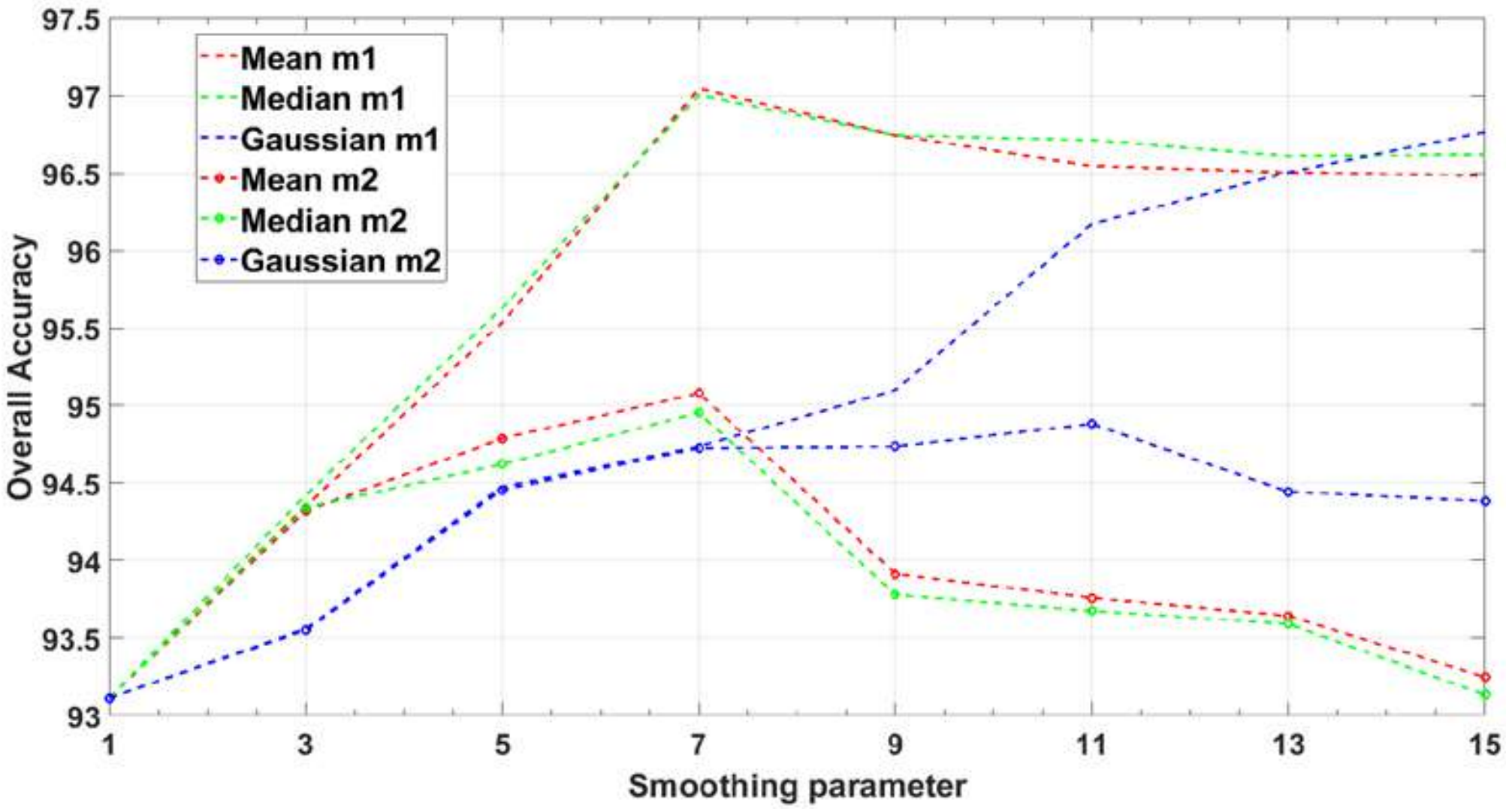

a) Lake Sihl: model1 (m1, trained on split 2) and model2 (m2, trained on split 1)

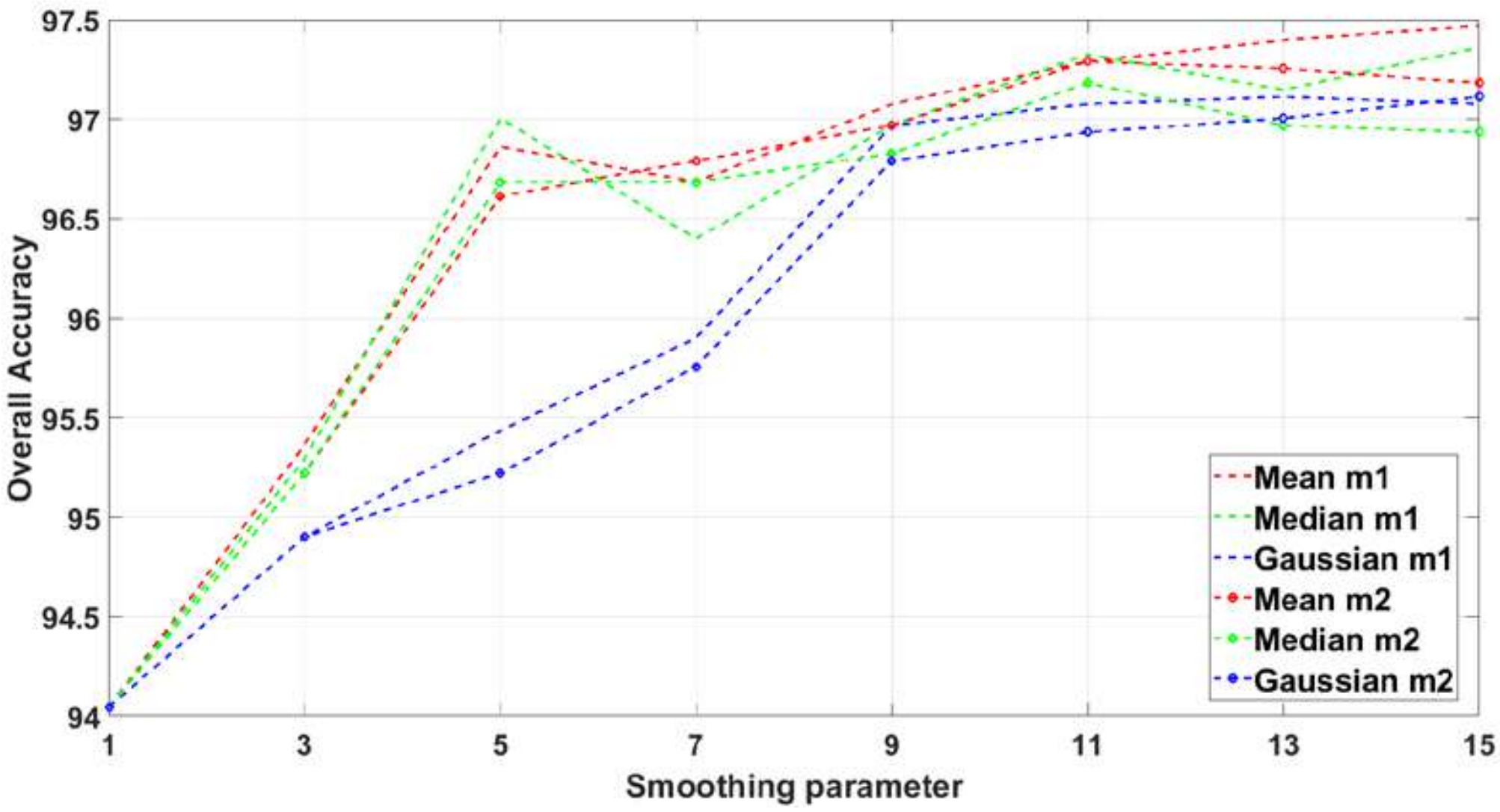

b) Lake Sils: model1 (m1, trained on split 2) and model2 (m2, trained on split 1)

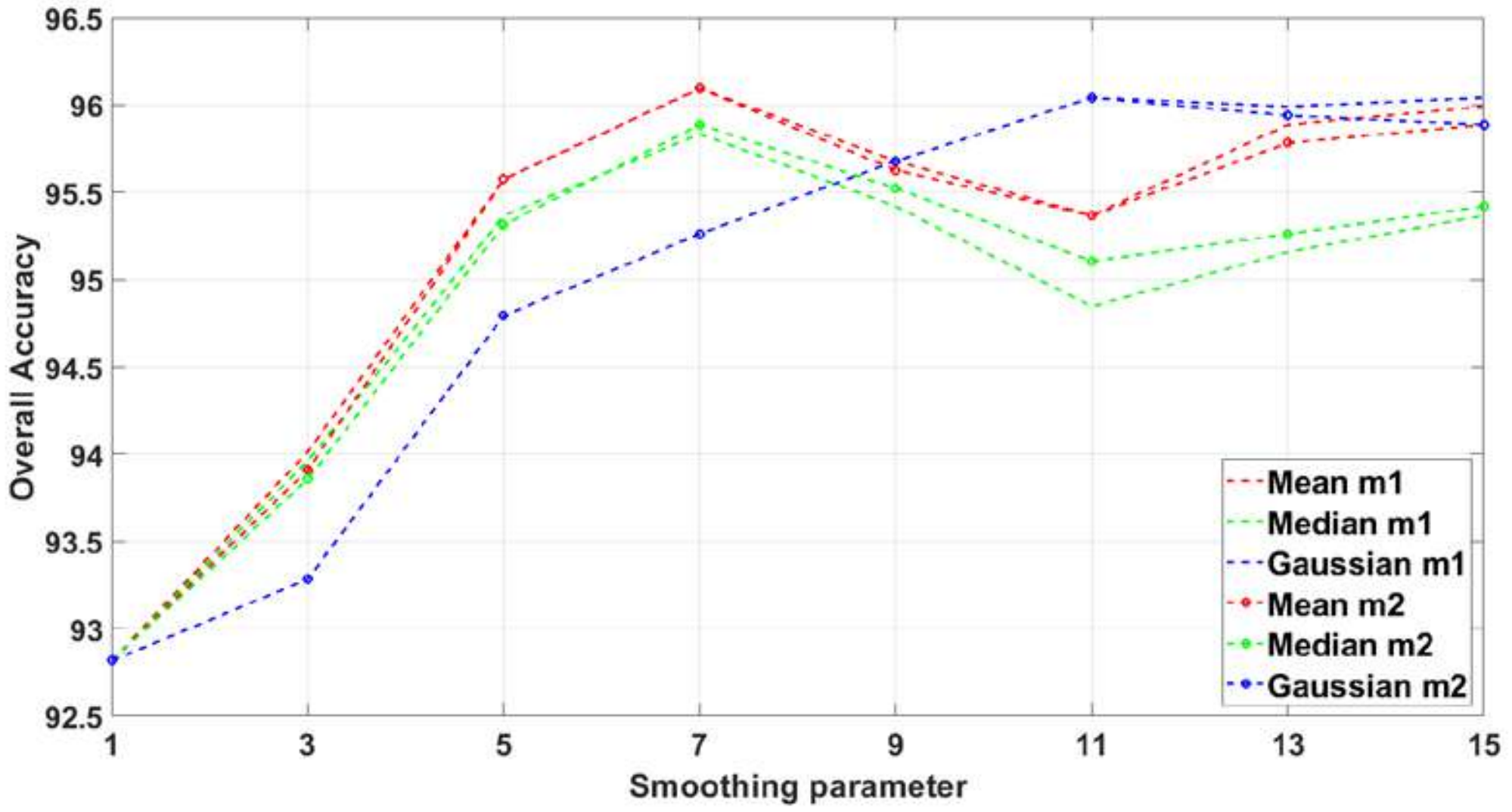

c) Lake Silvaplana: model1 (m1, trained on split 2) and model2 (m2, trained on split 1)

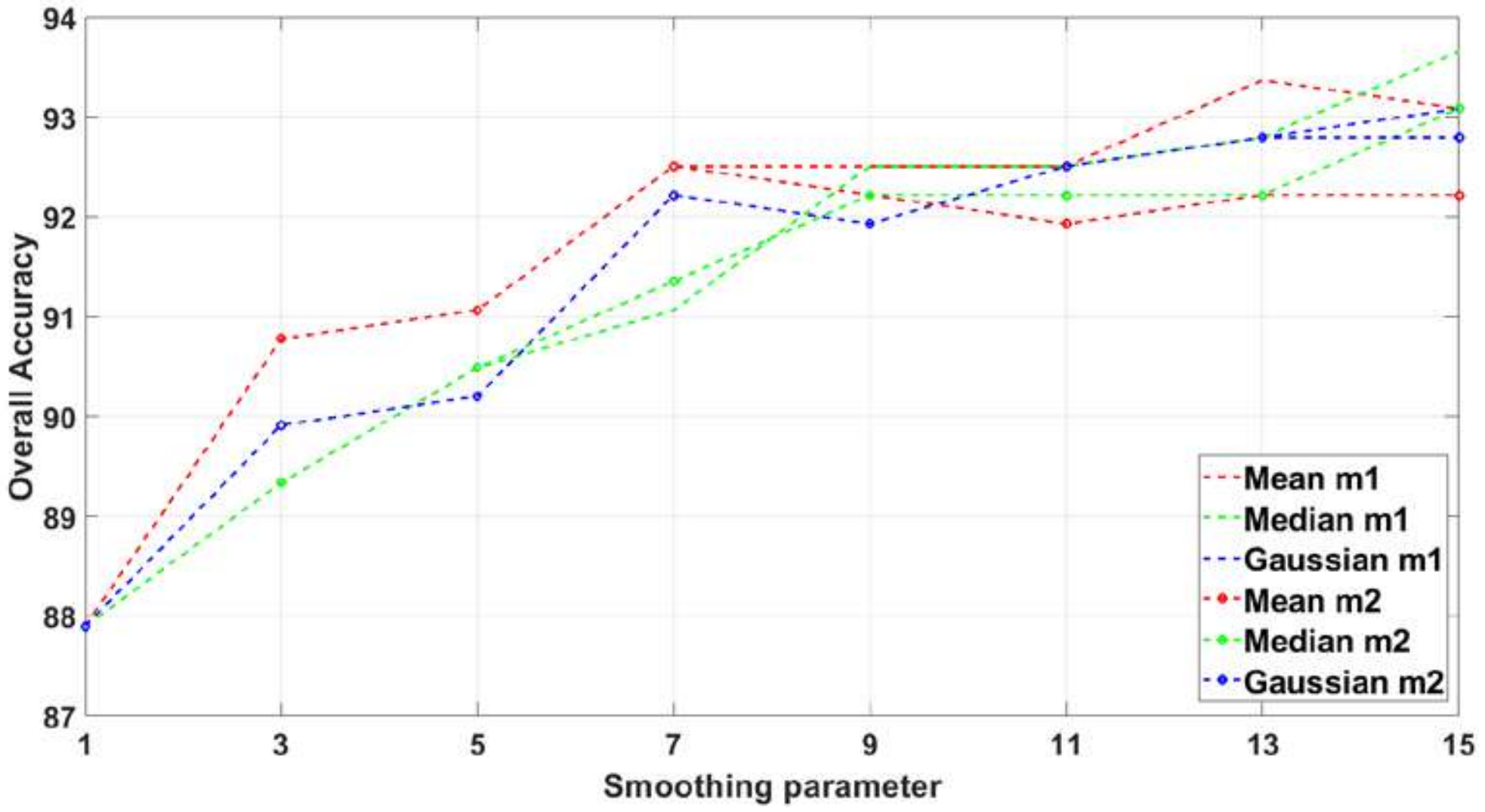

d) Lake St. Moritz: model1 (m1, trained on split 2) and model2 (m2, trained on split 1)

**Fig. 15.** MTA results on MODIS data (full winter 2016/17) for all three averaging schemes on all four target lakes. The vertical axis shows the overall accuracy of all pixels of all non-transition days.

### 3.4. Summary

We proposed a methodology for lake ice detection from MODIS satellite data. While we have concentrated on lakes in Switzerland, the ETHZ methodology is generic and should be applicable elsewhere too. We have shown that a simple, transparent classification approach achieves high accuracy with MODIS data for all tested lakes. We also processed the relatively short but challenging freezing and melting periods, where frozen and non-frozen pixels co-exist on the same lake. We have observed errors in the cloud-masks of MODIS, which are one of the main error sources in the final classification results.

We have demonstrated that the ETHZ approach gives consistent results over multiple winters (with a drop in accuracy of 5 to 10%), and that it generalizes fairly well from one winter to another. We have also shown that the model generalizes well across lakes. We have also tested on one full winter (which includes the challenging transition dates) to evaluate the performance of our lake ice monitoring approach. However, at least two full winters must be analysed to test the performance of a machine learning-based lake ice monitoring approach. This is because the training set should have data from at least one full winter and should include all dynamics from a winter during training. At the moment, we only have one full winter for training and testing combined. This is enough to draw conclusions for a feasibility study. However, it is recommended to assess the performance also using the data from more winters.

**Pros and Cons.** The main advantage of MODIS is the availability of longer time series data. In addition, MODIS has useful bands in various areas of the electromagnetic spectrum. However, MODIS has several disadvantages too. The radiometric quality is not very good and many bands are not useful due to artefacts, saturation etc. thus, only some useful bands were selected, see Table 4. The sensor is very old and the absolute geolocation is less accurate than VIIRS (more important for small lakes). MODIS is to be discontinued and its continuity will be guaranteed by VIIRS.

**Open problems.** A limiting factor is the coarse resolution of the images, which makes it difficult to process very small lakes. This is also indirectly affecting the classification accuracy of small lakes as the number of significant pixels (from the same lake) in the training set is relatively less compared to the bigger lakes. Moreover, it restricts the size of the dataset on which we report our results. Another issue is the uncertainty of the ground truth. Webcams typically do not cover the entire water surface, and it is often difficult to manually interpret the state of the lake. Finally, the standard cloud-masks of MODIS are not perfect and a significant source of error for lake ice detection. In particular, some actual cloud pixels are not detected, which are then classified as snow or ice.

The major issue with our approach is that there is significant confusion between thin ice and water states. Apart from the similar reflectance of these two classes, this could be also due to two factors: either ground truth noise or the inefficiency of the satellite data. A solution to ground truth noise is to use only dates in training for which the ground truth is certain. A practical solution is to avoid in training the dates near the transition periods, for which the ground truth noise occurrence is more probable. However, in that case the generated SVM model will be weak, if it does not have transition date pixels in training. A clear advantage of the Webcam-based approach compared to the satellite image approach is that pixel-wise ground truth labels of the Webcams are available during the transition dates that are used in training the Webcam model. Additionally, in MODIS data, there exist very few dates with snow-free ice, not enough to form a strong training model. Moreover, this thin ice-water confusion exists (even for a human operator) in assigning labels, even using the Sentinel-2 data.

It can also be seen from Fig. 13a that the results for April are not good as they should be. This problem can be reduced by enriching the training data set with acquisitions from May. The light conditions in April are more similar to March and May compared to other non-transition dates in the training set. In addition, there are some days that are classified totally different from its neighbours. This issue of classifying a pixel totally different from its temporal neighbours is reduced with MTA. MTA improved the results, since freezing and melting cycles are relatively smooth over periods of several days. The current SVM model could be strengthened with more training data from September 2016 and May 2017.

From our experiments, for each lake, ETHZ recommends to use all twelve MODIS bands (see Table 18) as the feature vector to be fed to the SVM classifier with RBF kernel. These best results are used in Appendices 3 and 6.

# 4. VIIRS PROCESSING (ETHZ)

VIIRS data has high temporal resolution (at least one image per day), sufficient spectral resolution and spatial resolution (GSD 375m-750m). Here, we process only the daytime acquisitions.

## 4.1. Input data (UniBe)

VIIRS data are obtained from the Comprehensive Large Array Stewardship System (CLASS) archive of NOAA. Almost 25.000 data sets were downloaded (approx. 1.5 TB) and processed for the retrieval of Lake Surface Water Temperature and Lake Ice. In addition to the input data, including the VIIRS Imaging resolution bands scientific data record (SDR), the VIIRS Cloud Masks (IICMO and VICMO), and the related geolocation information (GITCO and GMTCO), the VIIRS Land surface temperature (LST) Environmental Data Record (EDR; VLSTO) is downloaded for validation purposes of LSWT PMW results. As of March 2017, the VIIRS Cloud Mask EDR with the naming convention of VICMO was released, while the preceding cloud masks follow the naming convention of VIIRS Cloud Mask Intermediate Product (IICMO) (Note: the cloud mask is the same product – only the naming convention has changed). An example VIIRS image is shown in Fig. 16. Details of the five VIIRS bands used in our processing are shown in Table 19.

**Table 19.** Details of VIIRS I-bands.

| **Band** | **Spectrum** | **Centre wavelength (μm)** | **Bandwidth (μm)** |
|---|---|---|---|
| $I_1$ | Near infrared | 0.64 | 0.05 |
| $I_2$ | Near infrared | 0.87 | 0.04 |
| $I_3$ | Short-wave infrared | 1.61 | 0.06 |
| $I_4$ | Mid-wave infrared | 3.74 | 0.38 |
| $I_5$ | Thermal infrared | 11.45 | 1.90 |

## 4.2. Methodology

### 4.2.1. Pre-Processing

**Absolute geolocation correction**. Similar to MODIS (see Section 3.2.1), the generalized OSM lake outlines are corrected for absolute geolocation errors prior to backprojection on to the image space. The absolute geolocation error estimation procedure is the same as that of MODIS except for the following: firstly, the band $I_2$ (see Fig. 16) with the best contrast is fed as input to the algorithm. Secondly, the VIIRS shifts are estimated from two months in winter 2016-17 (two cloud-free dates per month). Table 5 displays the estimated final shifts for VIIRS and the analysed months. It can be seen that the absolute geolocation of the VIIRS images is more accurate than that of MODIS.

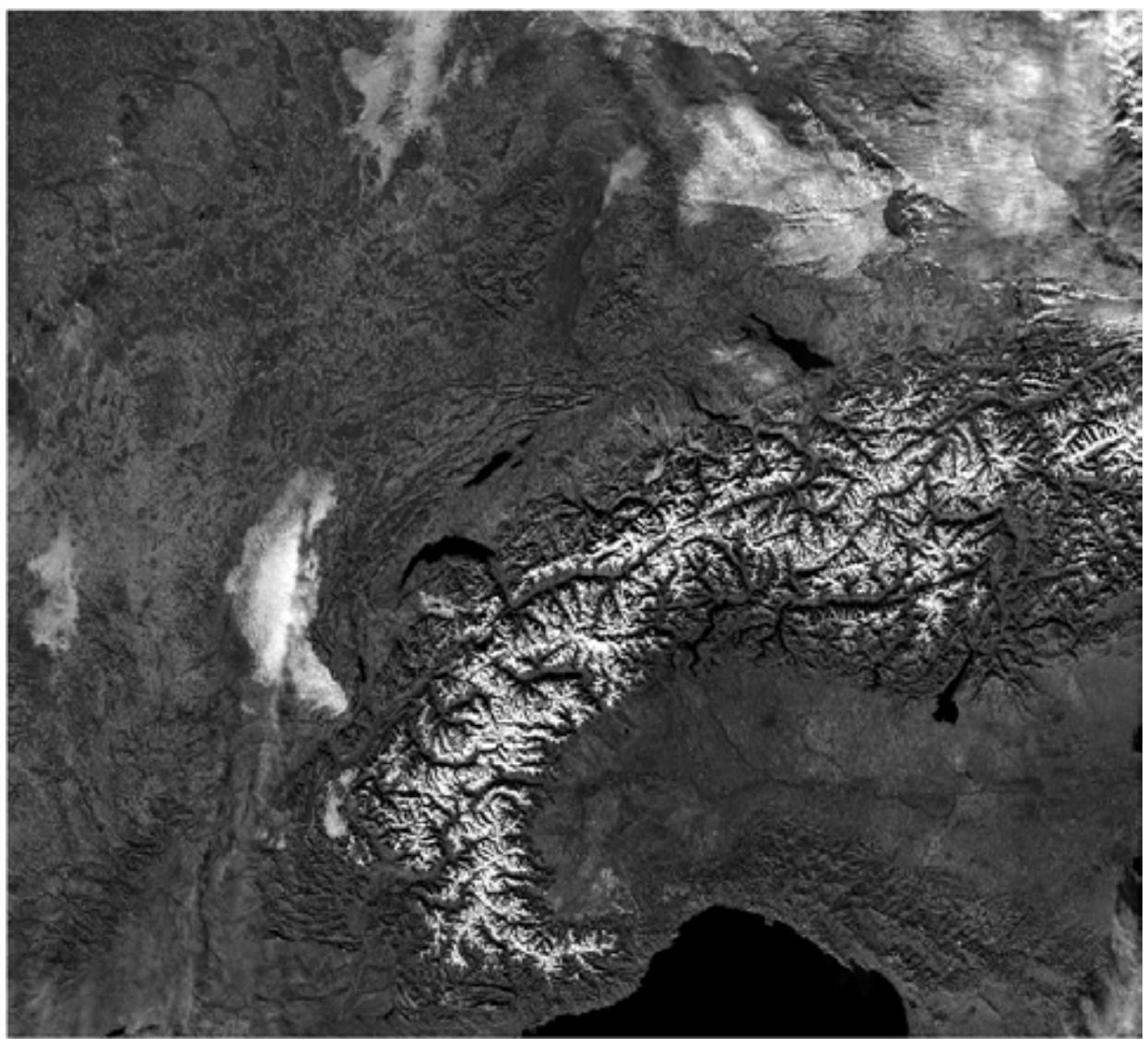

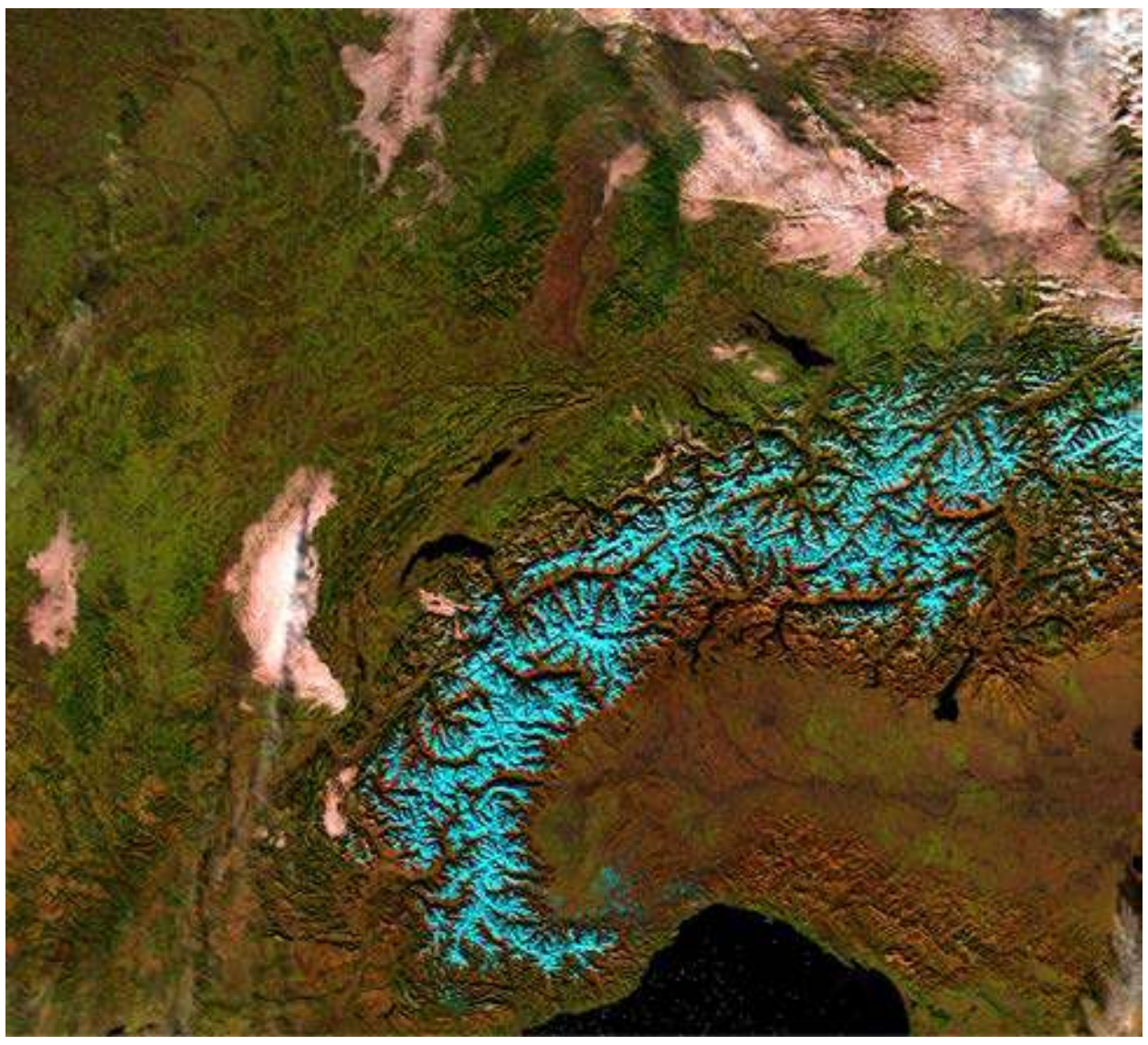

**Fig. 16**. Left: An example of VIIRS band (contrast enhanced version of Band $I_2$, 375m GSD, 28.12.2016, 12: 12 pm) displaying the region in and around Switzerland. Right: True Colour Image (TCI) combining the bands $I_3$, $I_2$, $I_1$ (RGB) from 28.12.2016, 12:12 pm.

**Cloud statistics.** For VIIRS, Fig. 17 illustrates the extent of the issue of data loss due to the presence of clouds. For each acquisition on the x-axis, the corresponding percentage of cloudy pixels is shown on the y-axis, for Lake Sihl. It can be seen that a lot of acquisitions are unusable due to the presence of clouds.

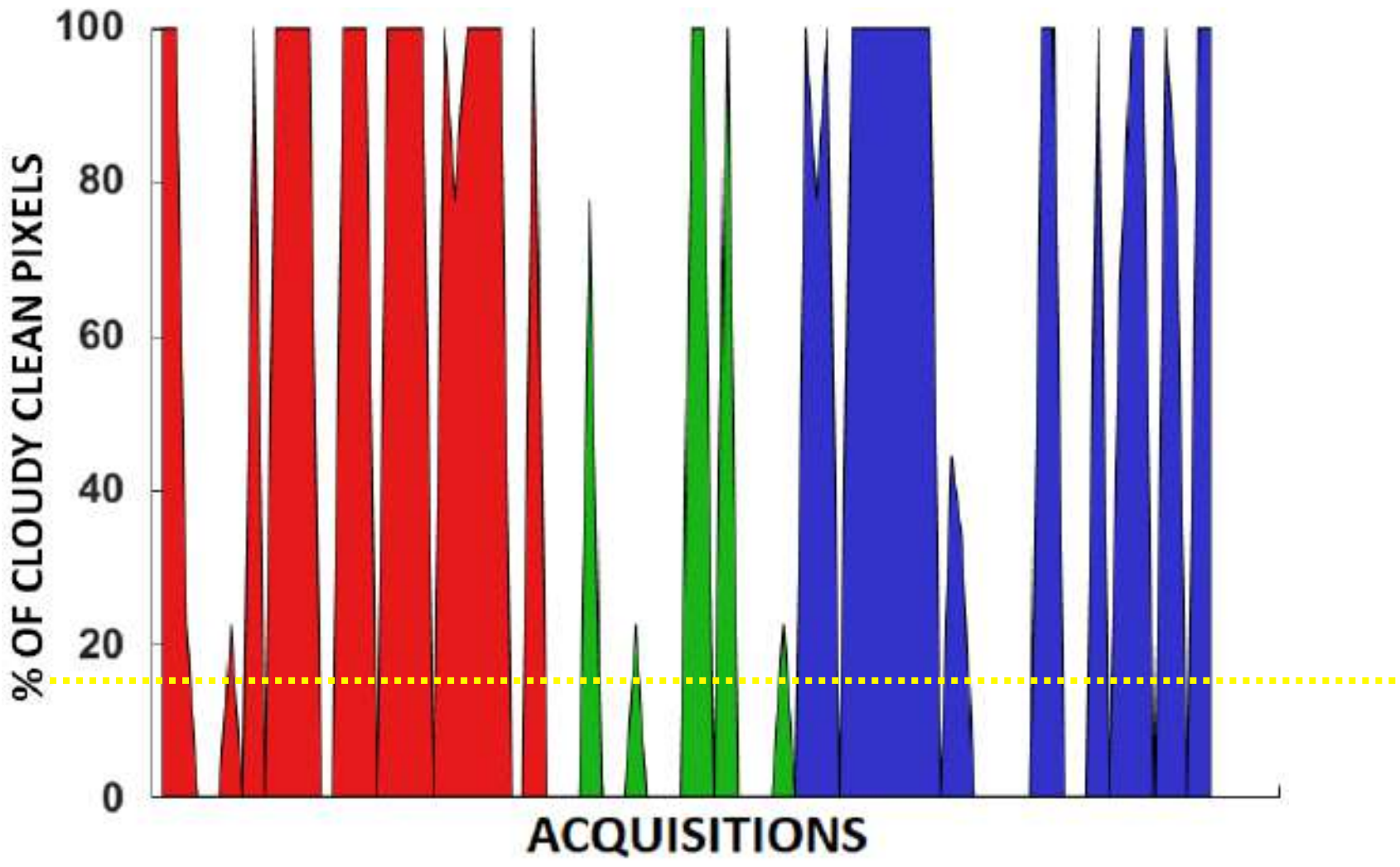

**Fig. 17.** VIIRS cloud-cover for Lake Sihl. Acquisitions from October 2016 are shown in red, December 2016 (first half only) in green and February 2017 in blue in chronological order. The cloud cut-off (30%) is shown with a yellow dotted line.

**Clean pixels vs. mixed pixels**. To determine the clean pixels, the lake outlines are corrected for absolute geolocation errors using the values in Table 5. Fig. 18 shows four lakes with the respective outlines (shown in green) backprojected onto the band $I_2$. The clean pixels are indicated with cyan squares and mixed pixels with red squares.

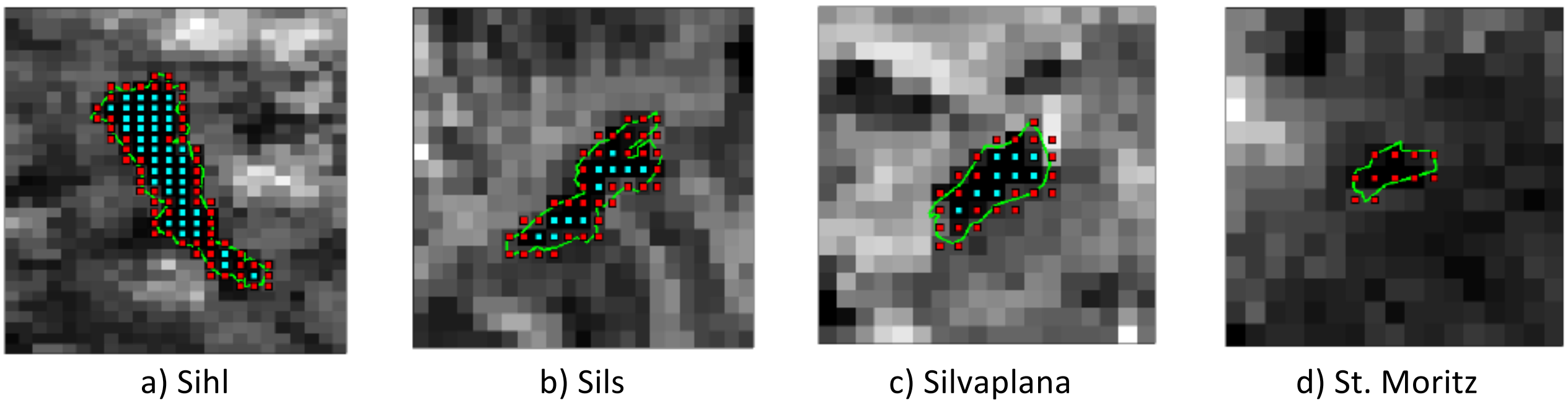

**Fig. 18.** OSM outlines of the four target lakes overlaid on VIIRS band $I_2$ (375m GSD) after correcting errors with absolute geolocation. The clean pixels are indicated with cyan squares and the mixed pixels with red squares. The lake outlines are shown in green. There is no clean pixel for Lake St. Moritz.

#### 4.2.2. Band selection

**XGBoost gradient boosting.** Similar to MODIS, to assess the significance of different bands, we perform a supervised variable importance analysis. We only include the five bands in our analysis. All VIIRS bands are analysed with XGBoost (see Section 3.2.2 for more details). We used the dates mentioned in Table 12 to perform this study. The results of the XGBoost analysis for VIIRS are shown in Fig. 19. It can be inferred that, for the data analysed in winter 2016-17 (subset), the band $I_1$ has the best inter-class separability among all VIIRS bands. Lake St. Moritz was exempted from this analysis due to lack of clean pixels.

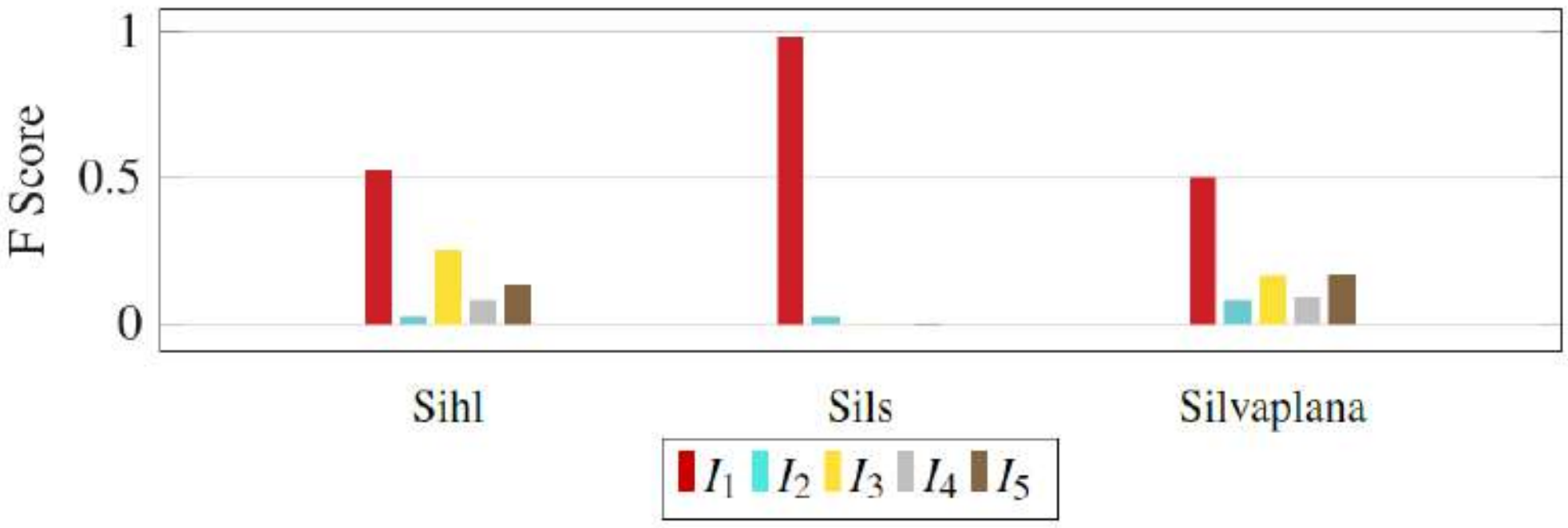

**Fig. 19.** Bar graph showing the significance of each of the five VIIRS bands for *frozen* vs. *non-frozen* pixel separation using the XGBoost (Chen and Guestrin, 2016) algorithm. The dates displayed in Table 12 are used to select the frozen and non-frozen pixels.

**Grey value histograms.** As a second check, we also use the histograms of the target classes in all bands to verify the band selection process. For Lakes Sihl, Sils and Silvaplana, the histograms of both frozen and non-frozen data in all five VIIRS bands are shown in Fig. 20. Both $I_1$ and $I_2$ bands have high inter-class separability. Since the two bands are spectrally close, and highly correlated, automatic variable selection chooses only one of them.

For the final processing all five I-bands were used.

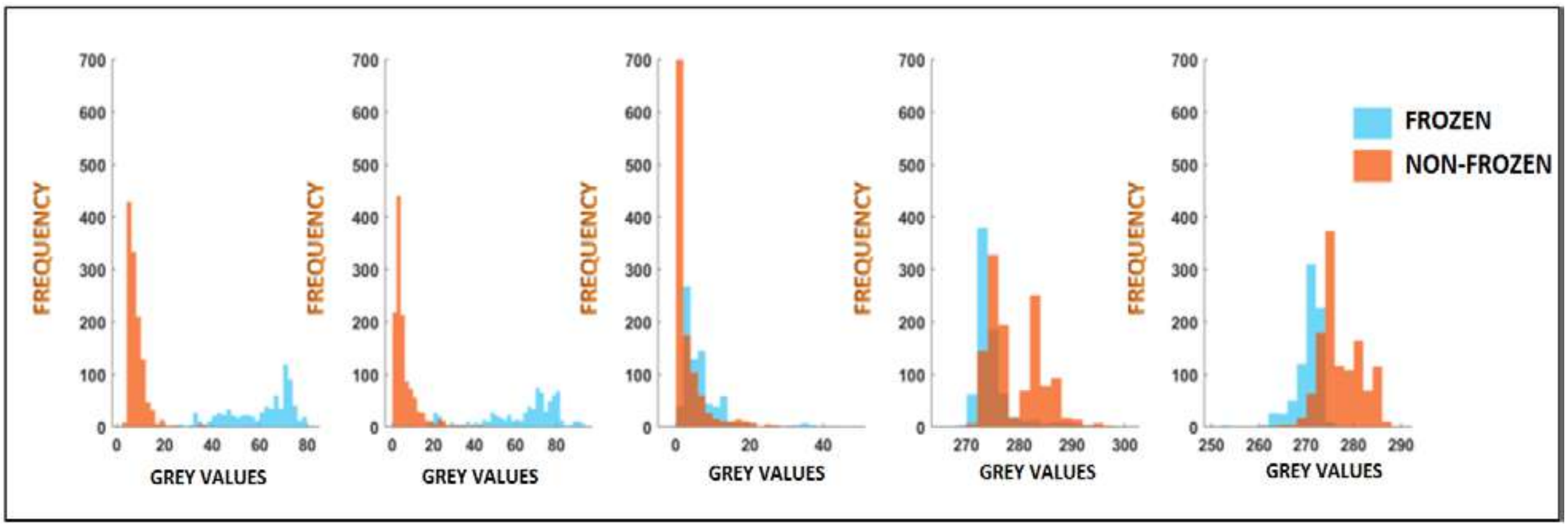

**Fig. 20.** Grey-value histograms of all five VIIRS bands (in each row from left to right: $\mathbf{I_1}$, $\mathbf{I_2}$, $\mathbf{I_3}$, $\mathbf{I_4}$, $\mathbf{I_5}$) for Lake Sihl. Non-frozen data is from images in October 2016 and first half of December 2016. Frozen data refers to images from February 2017.

#### 4.2.3. Semantic segmentation

The semantic segmentation methodology described in Section 3.2.3 is generic and is applicable to both VIIRS and MODIS sensors. The flow chart of the lake ice detection methodology for VIIRS data is shown in Fig. 10. Since we use only the VIIRS bands as feature vector, the super-resolution step shown in the flowchart is applicable only for MODIS.

#### 4.2.4. Post-processing

**Multi-temporal analysis (MTA).** Similar to MODIS processing, for each pixel, a moving average of the SVM scores are computed along the time dimension. The average is calculated for a fixed window length (smoothing parameter) that is determined heuristically. The window, centred at the pixel of interest, slides down along the temporal axis, computing an average over the elements within each window. Like MODIS, three different averaging schemes are tested: mean, median and Gaussian.

### 4.3. Results and discussion

Our primary period of analysis is 1.10.2016-30.4.2017 (Section 4.3.2). We also report results for some selected dates in winter 2016-17 (Section 4.3.1), which will be addressed as winter 2016-17 (subset) in order to minimize confusion.

#### 4.3.1. Winter 2016/17 (subset) results

For winter 2016-17 (subset), the details of the dates processed are shown in Table 12. We report results on the following three lakes: Sihl, Sils and Silvaplana. Lake St. Moritz is not processed, as there exist no clean pixels.

**Qualitative results.** For Lake Sihl, selected qualitative results for VIIRS data are shown in Fig. 21. For Lakes Sils and Silvaplana, the results are shown in Fig. 22. In both figures, the first and second rows display the SVM classification result and score, respectively. In the first row, the pixels that are classified as frozen are shown in blue and the ones classified as non-frozen in red. In both Figs. 21 and 22, the second row shows the pixel-wise SVM scores (confidence value shown as a bar from blue to red, the more blue means more frozen and the more red means more non-frozen). In Fig. 21, the first column shows a successfully classified date (18.2.2017) and the second column displays a failed date (11.10.2016) for Lake Sihl. It can be seen that, for Lake Sihl, on a non-frozen day 11.10.2016 (right column), some pixels are classified as frozen. These mistakes lie near cloudy pixels; we thus assert that the failures are due to the influence of nearby clouds.

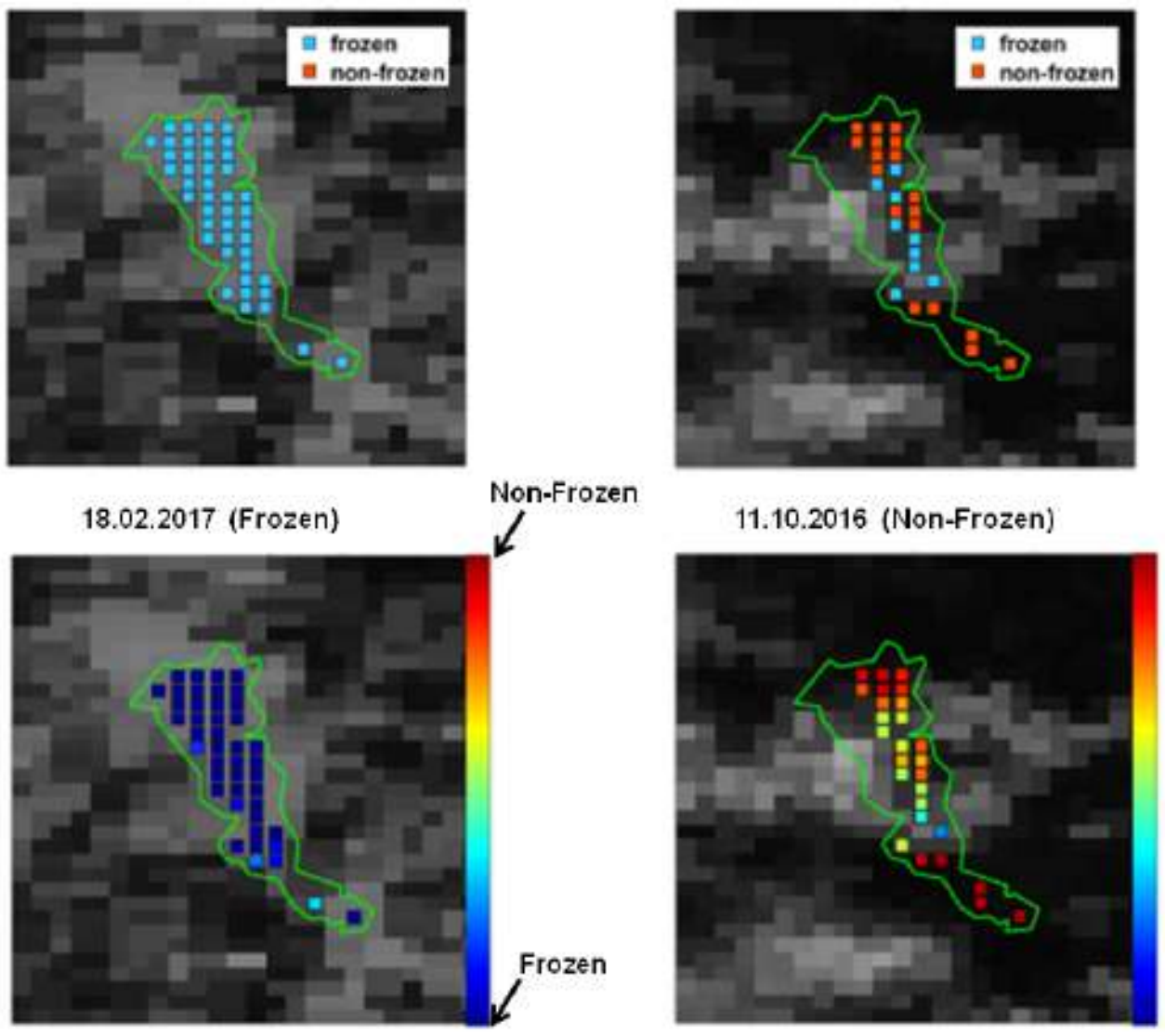

**Fig. 21.** Results from winter 2016-17 (subset) on VIIRS data for Lake Sihl. First and second rows show SVM classification result and score (confidence), respectively. In the second column, more red means more non-frozen and blue means frozen.

For the Lake Sils, all tested dates are successfully classified. A sample result is shown in Fig. 22 (first column) for a non-frozen date (31.10.2016). For the Lake Silvaplana, both failure (second column) and success (third column) cases are displayed. The second and third columns show results for Lake Silvaplana on the same frozen date 11.2.2017 at two different acquisition times. In one of the acquisitions (second column), a pixel is wrongly classified as non-frozen (red), because the pixel appears dark (maybe due to shadows). The lake outlines are overlaid on a reflective band.

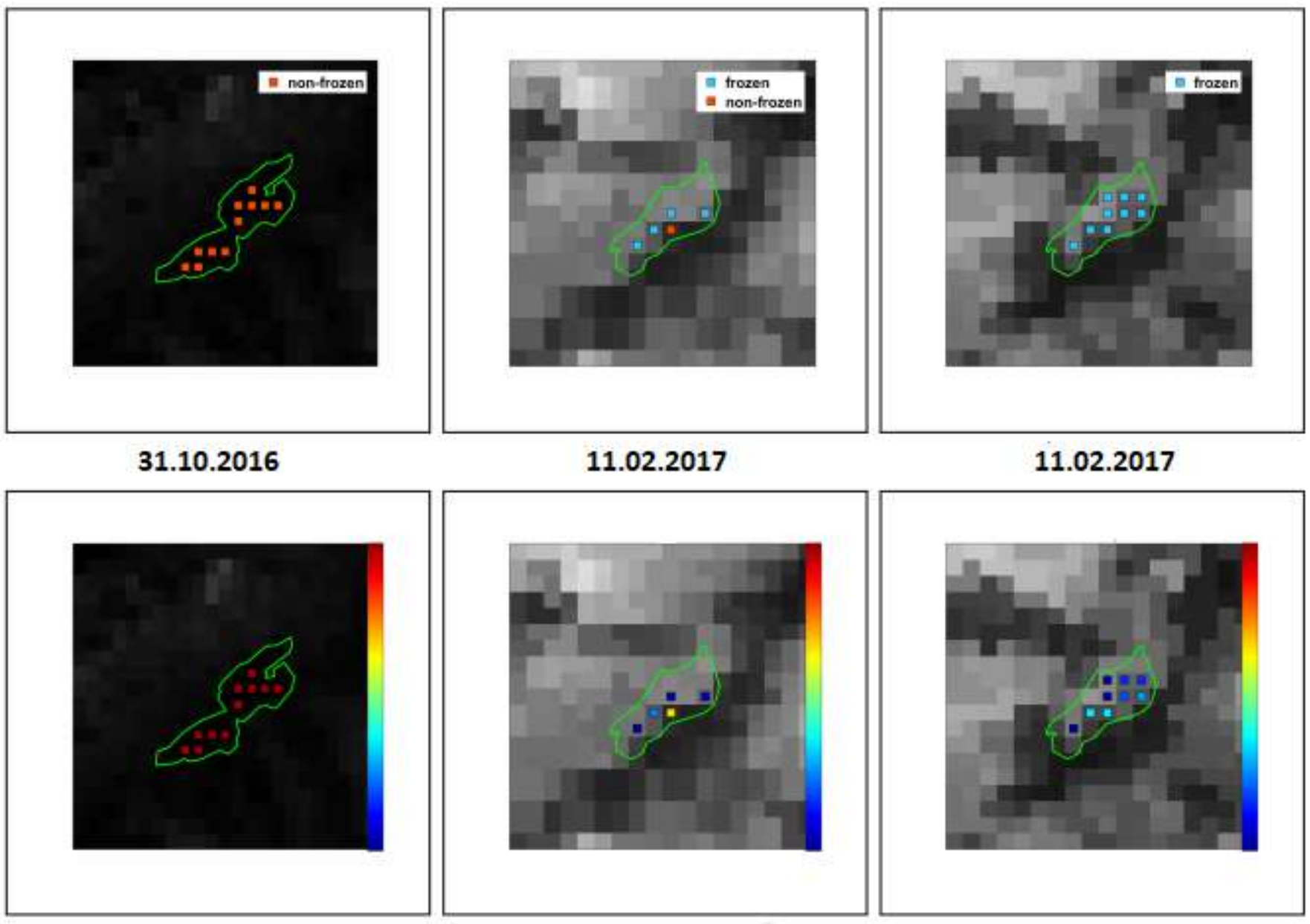

**Fig. 22.** Results from winter 2016-17 (subset) on VIIRS data for Lakes Sils and Silvaplana. First and second rows show classification result and score (confidence), respectively. In the second column, more red means more non-frozen and blue means frozen.

**Quantitative results.** Different combinations of bands are tested as feature vector and the best results are reported in Table 20 after 4-fold cross validation. It can be seen that, even with a linear kernel, SVM delivered near perfect results for the period winter 2016-17 (subset). The best result is obtained when all five bands are used, but, as predicted by XGBoost, even the reflective band $I_1$ alone is enough to distinguish the frozen pixels from non-frozen ones almost perfectly. We also tried the combinations of all reflective bands ($I_1$, $I_2$, $I_3$) and thermal bands ($I_4$, $I_5$). The reflective bands performed markedly better than the thermal ones. It can be seen in Fig. 20 that the separation between snow and water is where the reflective bands $I_1$ and $I_2$ outperform the thermal bands $I_4$ and $I_5$.

**Table 20**. Results (overall accuracy) from 2016-17 (subset) on VIIRS data: Comparison of 4-fold cross-validated SVM (linear kernel) results with different band combinations as feature vector. Best results are shown in green.

| Feature vector | Lake Sihl | Lake Sils | Lake Silvaplana | Lake St. Moritz |
|---|---|---|---|---|
| $I_1$ | 98.8 % | 99.8 % | 98.8 % | - |
| $I_4$, $I_5$ | 91.2 % | 95.2 % | 96.7 % | - |
| $I_1$, $I_2$, $I_3$ | 99.3 % | 100.0 % | 99.1 % | - |
| $I_1$, $I_2$, $I_3$, $I_4$, $I_5$ | 99.3 % | 100.0 % | 99.5 % | - |

The current VIIRS dataset (winter subset) is very small, not enough to draw final conclusions. Hence, we report results on one full winter (Section 4.3.2).

### 4.3.2. Full winter 2016-17 (October - April) results

Similar to MODIS (see Section 3.3.4), we divided the dates into two parts: *transition* dates and *non-transition* dates. For all four lakes, all non-transition dates are 4-fold cross-validated and all transition dates are tested with a SVM model (separate model for each lake) trained using VIIRS data from all non-transition dates. For all three lakes except St. Moritz, pixels (from non-transition dates only) from the same lake are used in generating the SVM model. For Lake St. Moritz, since there exist no clean pixel, the clean cloud-free non-transition pixels from other three lakes are used to generate the SVM model. We report results using both linear and RBF kernels.

**Quantitative Results.** The quantitative results are reported only on the non-transition dates, which are either fully frozen (labels **i (ice), s (snow)**) or fully non-frozen (label **w (water)**) with acquisitions, which are at-least 30% cloud-free (see more details on class labels in Appendix 4). The results are shown in Table 21.

Firstly, following the processing of MODIS, we used all five bands as feature vector and tested using both SVM kernels. It can be seen from Table 21 (row 3) that the linear kernel gives very good results with 5-bands. However, the RBF kernel outperforms the linear kernel for all four lakes (see row 4). We also tested with just the band $I_1$ as feature vector. It does not make sense to run an RBF kernel with one channel feature vector. Therefore, we tried linear kernel with band $I_1$. It is quite impressive that even with a single channel and linear kernel the results are very decent.

The results on Lake St. Moritz are based on mixed pixels. Hence, the absolute values are not 100% trustworthy. Still, the relative results (linear vs. RBF kernel) follow the trend as in other three lakes tested on clean pixels.

**Table 21.** Quantitative results on VIIRS data from full winter 2016-17. The overall accuracy is shown. For the same SVM test configuration, the third and fourth rows shows the results without and with MTA are listed. The best results are indicated in green.

| Feature vector | SVM kernel | Lake Sihl | Lake Sils | Lake Silvaplana | Lake St. Moritz | With MTA |
|---|---|---|---|---|---|---|
| $I_1$ | Linear | 96.1 % | 88.9 % | 91.9 % | 76.6 % | No |
| All 5 I-bands | Linear | 96.4 % | 92.4 % | 92.4 % | 79.0 % | No |
| All 5 I-bands | RBF | 98.2 % | 93.9 % | 96.2 % | 86.6 % | No |
| All 5 I-bands | RBF | 98.6 % | 96.3 % | 98.6 % | 88.1 % | Yes |

**Qualitative results.** Qualitative analysis is done on all dates (transition and non-transition). Results on the lakes Sihl, Sils, Silvaplana and St. Moritz are shown in Figs. 23a, 23b, 23c, and 23d. Similar to MODIS plots, the results (October 2016 until April 2017) are listed in a chronological order on the x-axis and the SVM result is shown on the y-axis as the percentage of cloud-free pixels that are classified as non-frozen (NF). Similar to MODIS processing, the ground truth is shown as a cyan colour line.

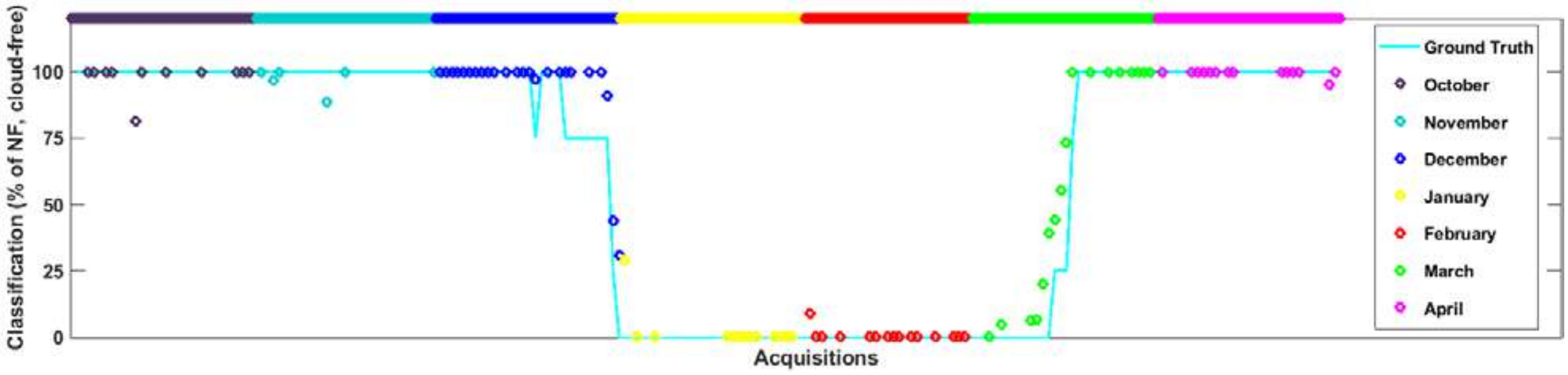

a) Lake Sihl

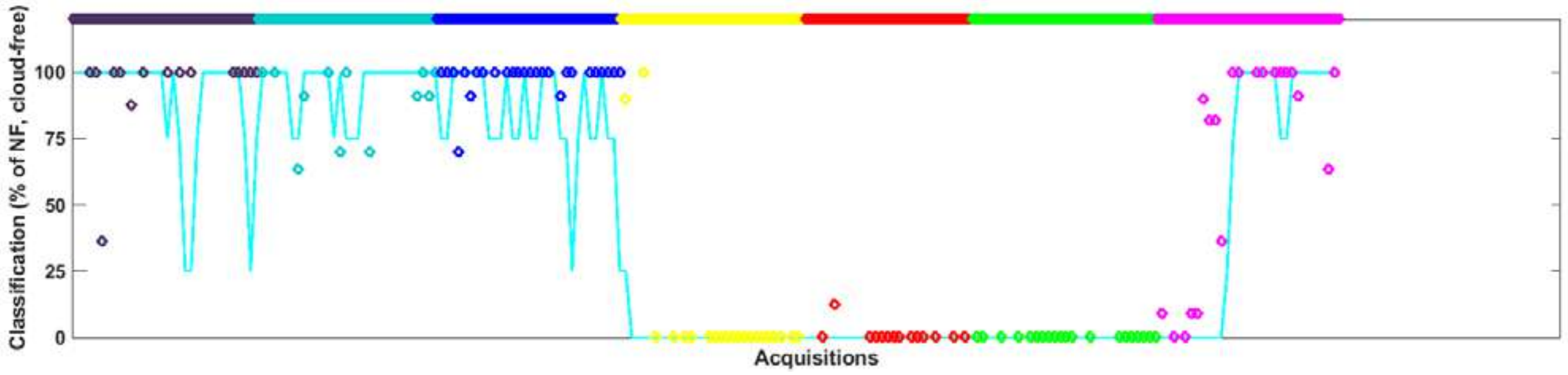

b) Lake Sils

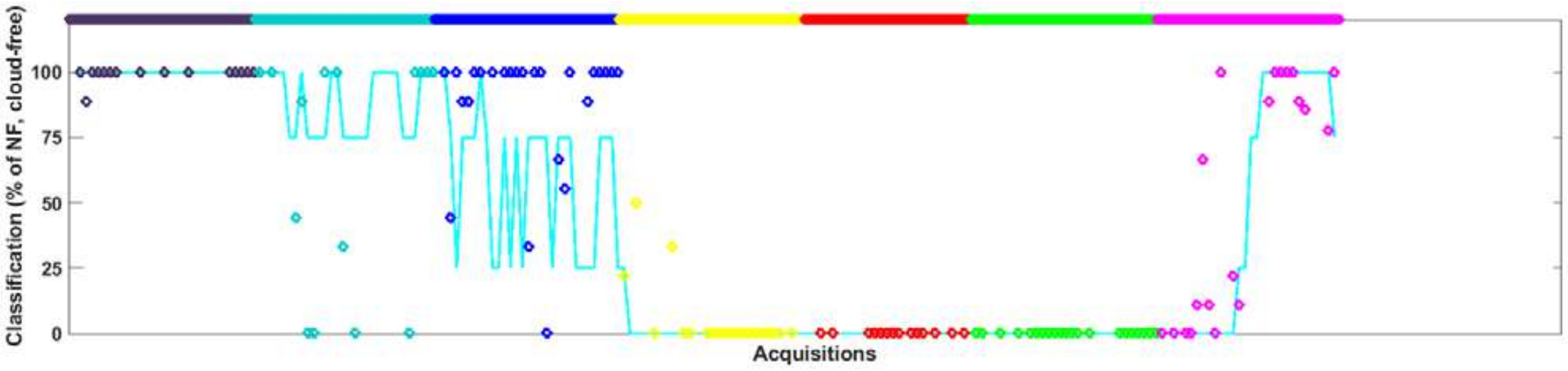

c) Lake Silvaplana

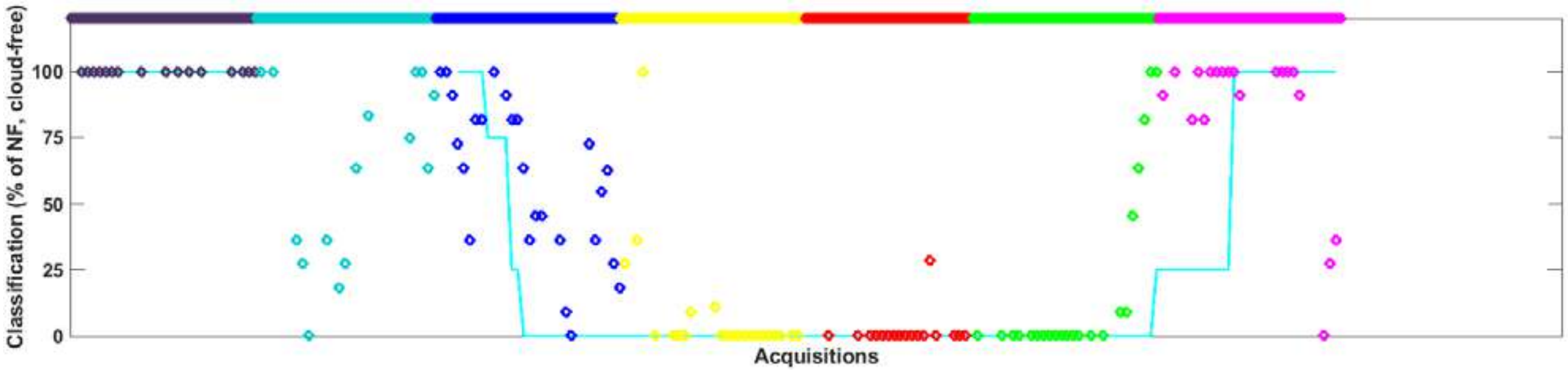

d) Lake St. Moritz

**Fig. 23.** VIIRS qualitative results for full winter (1.10.2016-30.4.2017). All dates that are at least 30% cloud-free are shown in chronological order on the x-axis. The classification results (percentage of non-frozen cloud-free pixels) are plotted on the y-axis.

From Fig. 23a, it can be inferred that there are no major failure cases for the Lake Sihl.

For Lake Sils (23b), it can be observed that, for some transition dates, the classification result is not exactly following the ground truth. This issue exist for Lake Silvaplana (23c) too. This could be due to incorrect interpretation of the Webcam data. If there are noisy labels in the ground truth, it can corrupt the SVM model being trained too. This can eventually result in wrong classifications. However, re-training with less noisy ground truth could solve this issue.

For Lake Silvaplana, it can be seen there that, compared to the class: frozen, there exist few non-transition dates for the class: non-frozen (there are many transition dates near the non-frozen dates). Hence, during the training process, the SVM has seen few non-frozen pixels than frozen counterparts (training samples not balanced for both the classes). This is the reason why the results of class frozen are almost perfect compared to the class non-frozen. Even though the SVM model is generated based on a sparse set of support vectors from both classes, there should be a minimum number of training samples (from each class) to generate a robust model. For Silvaplana, this is missing in the case of the class non-frozen.

Analysis of Lake St. Moritz using the clean pixels is not possible, since it has none. Hence, we trained a SVM model using all clean pixels of the three lakes (Sihl, Sils and Silvaplana). From these lakes, only the acquisitions from the *non-transition* dates were used in generating the SVM model. Using this model, we tested the eleven mixed pixels (see Table 6 and Fig. 18d) of Lake St. Moritz on all available dates (at least 30% cloud-free) from winter 2016-17. Our classification result shows that Lake St. Moritz was partially frozen during some dates in November 2016. This is in accordance with the results of MODIS too (Section 3.3.4)

**Multi-temporal analysis (MTA).** The effect of each MTA scheme is shown in Fig. 24 for all four target lakes. In all graphs, the overall accuracy (of non-transition dates) is plotted against the smoothing parameter for all three schemes. The smoothing parameter (window length) with value one is same as the special case when no MTA is performed. The overall accuracy of Lake Sihl is excellent even without MTA. Hence, the post-processing using MTA is not really needed for Sihl. For the other three lakes, the overall accuracy increases by around 2% with MTA (see Table 21, rows 4 and 5). In general, a value of 3-5 is recommended for the smoothing parameter. Owing to over-smoothing, higher values for smoothing parameter could be counter-productive.

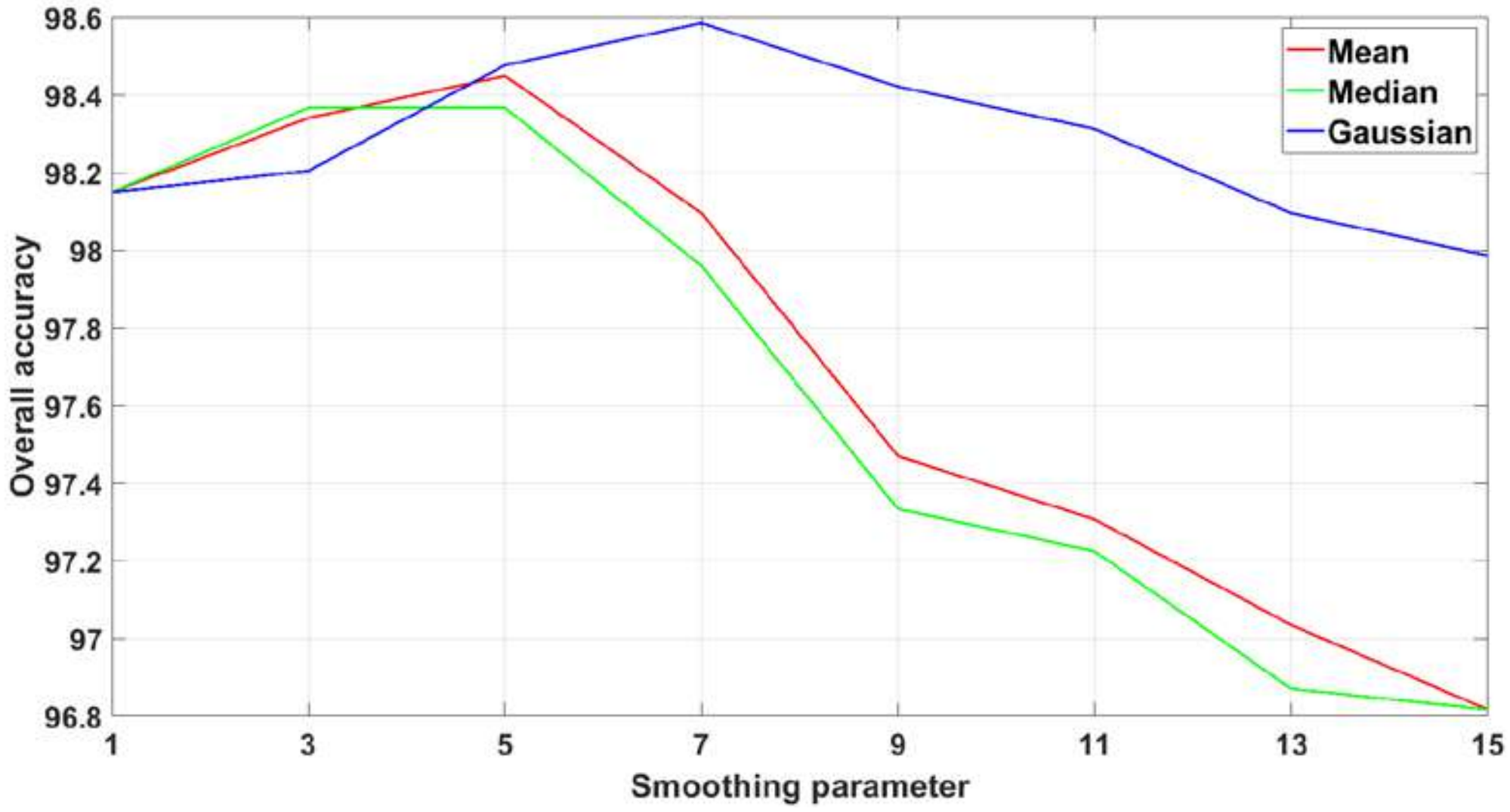

a) Lake Sihl

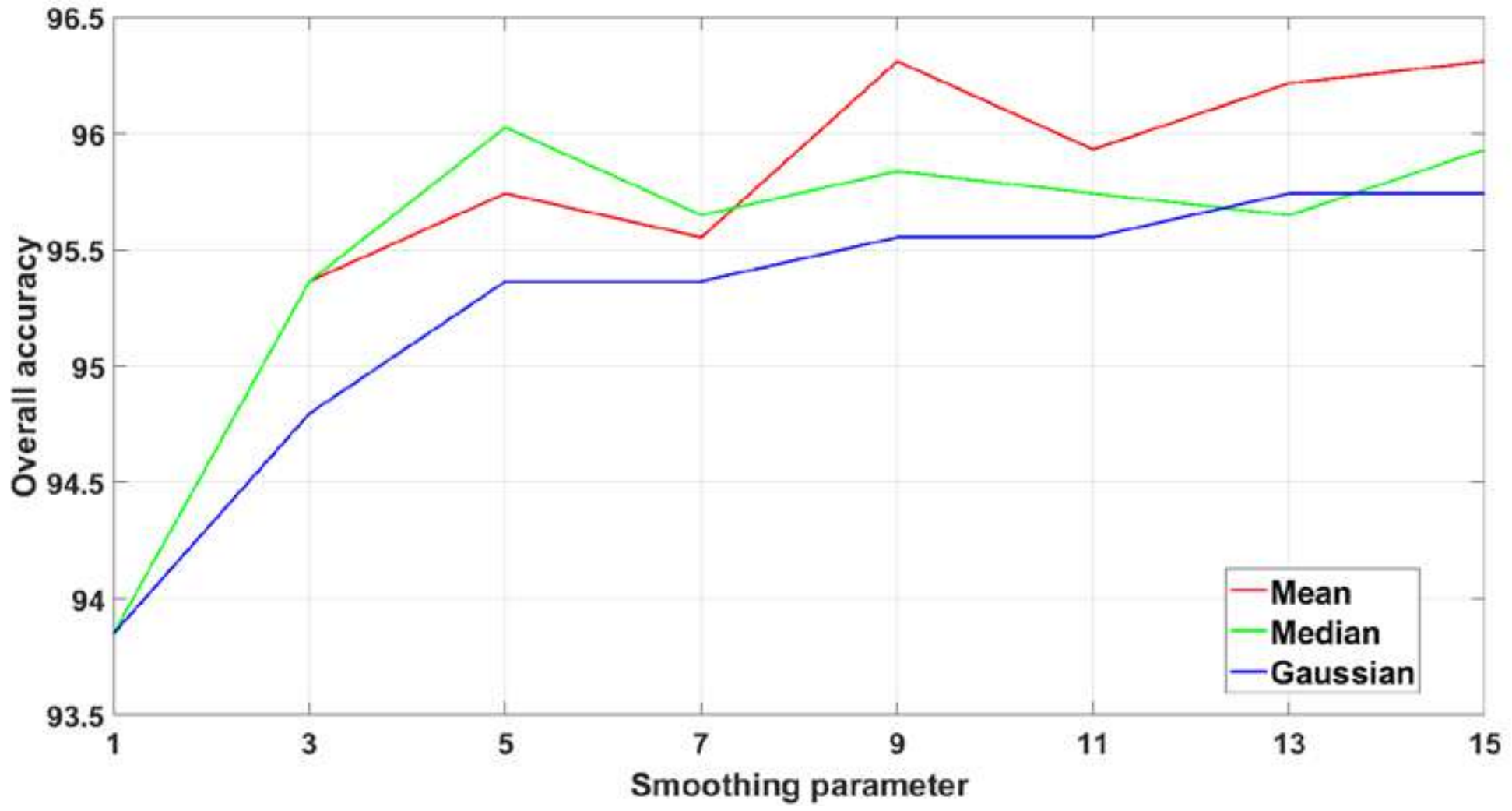

b) Lake Sils

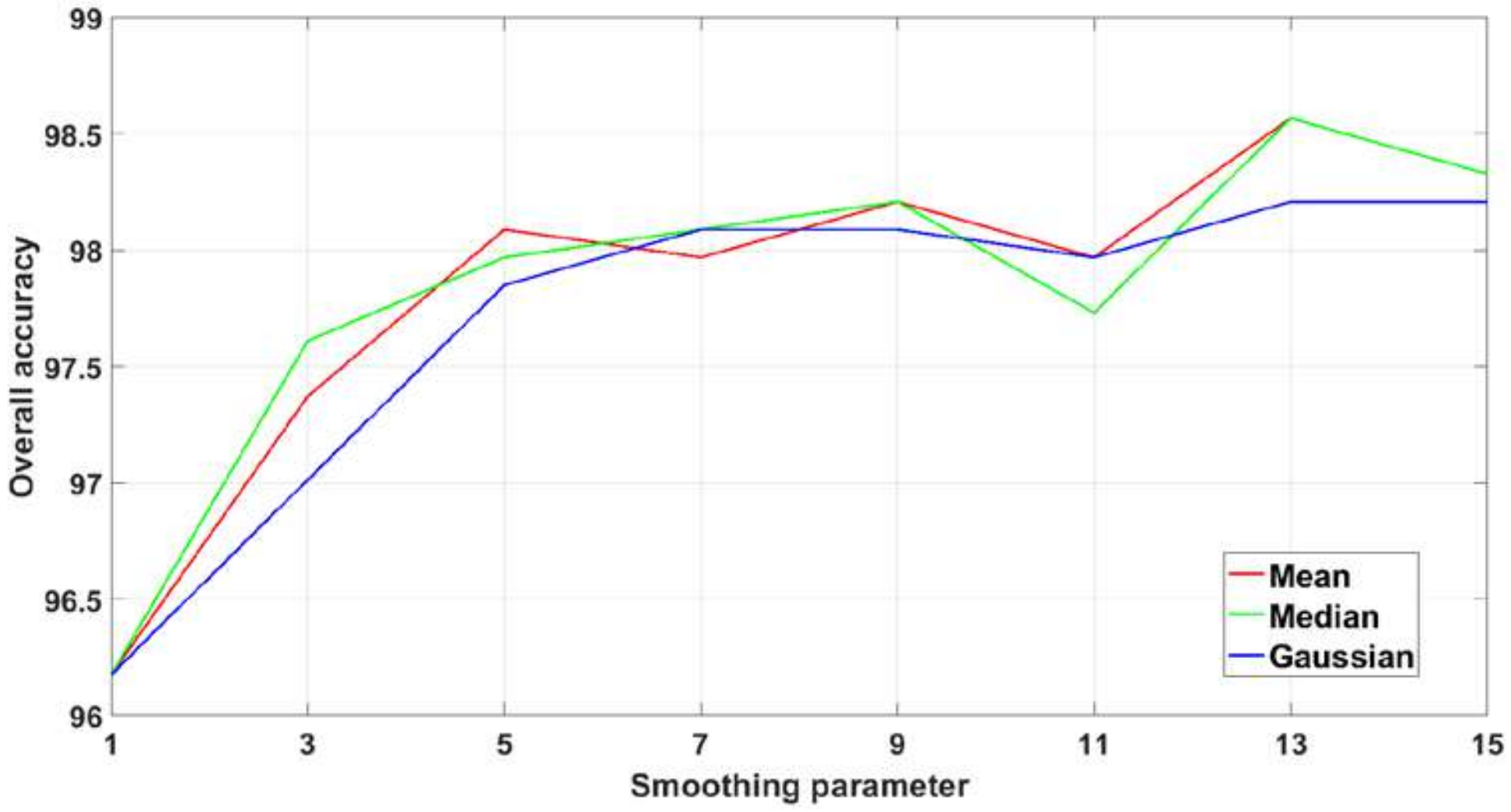

c) Lake Silvaplana

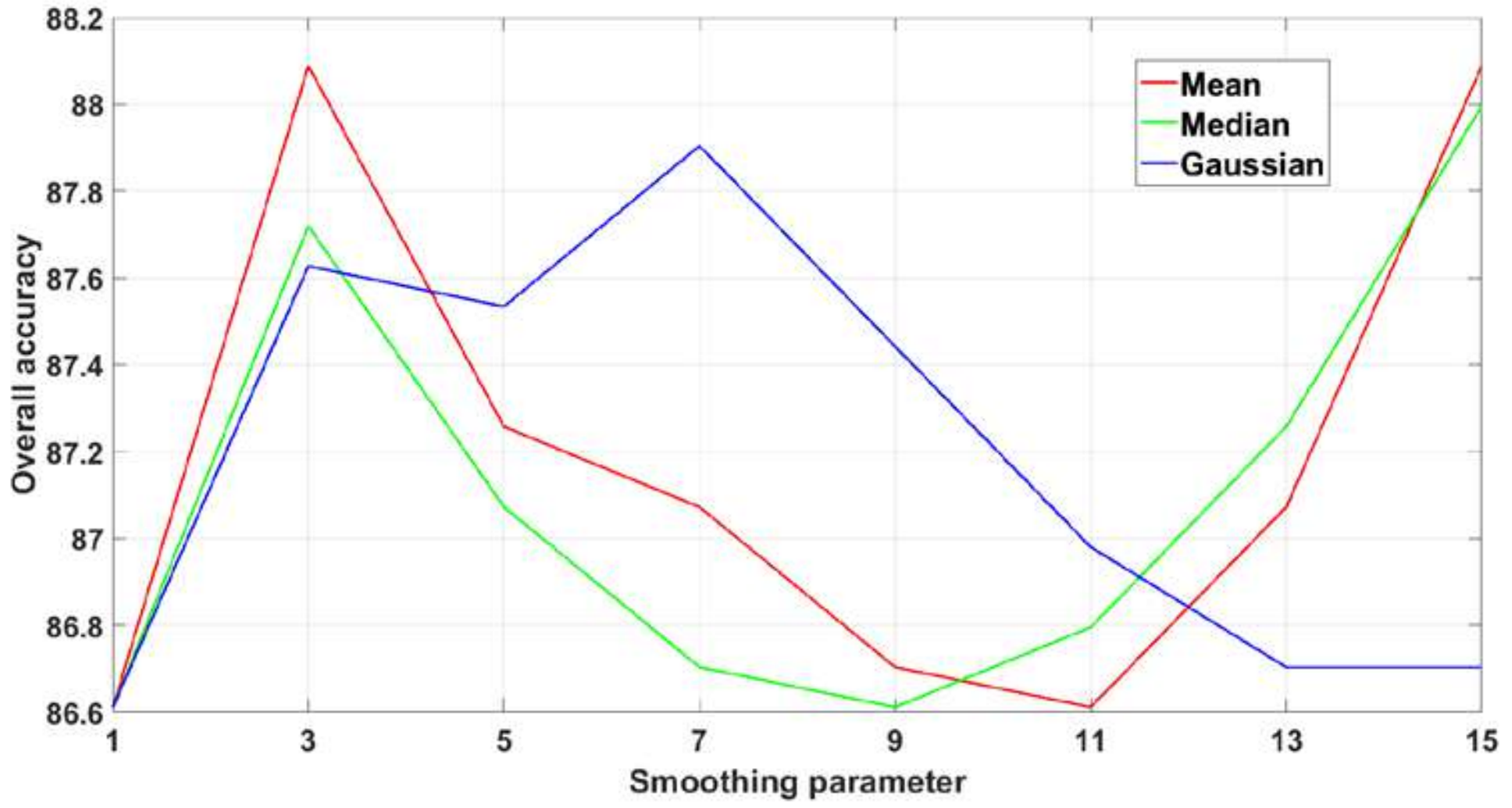

d) Lake St. Moritz

**Fig. 24.** MTA results of VIIRS data (full winter 2016/17, October-April) for all three averaging schemes.

**Generalization across lakes.** In order to assess the transferability of a trained model across lakes, the SVM model trained on pixels of a lake (or a set of lakes) is used to test the pixels from a completely different lake. The results are displayed in Table 21. It can be seen that, very good results are obtained even when the model is trained using the pixels from an entirely different lake. Additionally, the more lakes are used in the training, the better the results. The best results are when the pixels from the same lake are included in the training process. Except for Lake St. Moritz, testing is done only on clean pixels. For St. Moritz, we performed tests on the mixed pixels. Information from Sentinel-2A data was not incorporated in the ground truth when this experiment was performed. Hence, the ground truth uses information only from Webcam images. In addition, the data from April 2017 is also not used since it was not available to ETHZ when this experiment was conducted. This explains why the absolute values of the results differ for the lakes compared to Table 21, without MTA. However, this is not critical since the relative accuracy mainly matters in this experiment.

**Table 22**. Quantitative results from winter 2016-17 (October-March) on VIIRS data. Classification is done using SVM with RBF kernel. All five I-bands are fed as feature vector to the classifier. The overall accuracy is given. Best results are shown in green.

| **Tested on:** | **Trained on:** | **Overall accuracy** |
|---|---|---|
| **Sihl** | **Sihl** | 97.9 % |
| Sihl | Sils | 95.3 % |
| Sihl | Sils, Silvaplana | 96.8 % |
| **Sils** | **Sils** | 99.4 % |
| Sils | Sihl | 97.5 % |
| Sils | Sihl, Silvaplana | 97.4 % |
| **Silvaplana** | **Silvaplana** | 99.2 % |
| Silvaplana | Sils | 94.3 % |
| Silvaplana | Sihl | 96.7 % |
| Silvaplana | Sihl, Sils | 97.0 % |
| St. Moritz (mixed pixels) | Sihl | 84.1 % |
| St. Moritz (mixed pixels) | Sils | 79.8 % |
| **St. Moritz (mixed pixels)** | **Sihl, Sils, Silvaplana** | 85.6 % |

## 4.4. Summary

We have shown that the semantic segmentation methodology we proposed for MODIS data is applicable for VIIRS too. We performed various experiments for the period winter 2016-17 mainly for the following three lakes: Sihl, Sils and Silvaplana. Lake St. Moritz is excluded from the primary analysis, as there are no clean pixels for that lake. However, we report results for Lake St. Moritz by testing on the mixed pixels using a training model developed using the clean pixels of other three lakes. We also have shown that the model trained on one lake generalizes well to another (Table 22). The algorithm gives the best results when the pixels from the lake being tested are used in the training set. Pixels from more lakes if available in the training set, the better. However, even the pixels from a completely different lake can give very good results.

**Open Problems.** We have presented results on one full winter. Like MODIS, one of the primary sources of errors is the erroneous cloud masks. We also suspect some noise in the ground truth labels generated by visual evaluation of Webcams (esp. for lakes Silvaplana and Sils), which could be another source of error.

For all lakes, RBF kernel outperforms the linear kernel. However, the performance of linear kernel is also very good. ETHZ recommends using all five VIIRS bands as feature vector to be fed as input to SVM (RBF kernel) for lake ice detection using VIIRS data. These best results are used in Appendices 3 and 6.

# 5. VIIRS PROCESSING (UNIBE)

## 5.1 Data of operational NASA satellite Suomi-NPP

The Visible Infrared Imaging Radiometer Suite (VIIRS) on-board of the operational satellite systems Suomi-NPP and Joint Polar Satellite System (JPSS-1) has the unique combination of high temporal resolution (daily) with high spatial resolution in the thermal spectra (375m). Furthermore, the sensor will be also on the following missions JPSS-2 and -3 offering continuity until 2030. These features make the system very attractive for a feasibility study to test the capability for lake ice detection, even for small lakes in Switzerland.

Data are obtained from the Comprehensive Large Array Stewardship System (CLASS) archive of NOAA. Almost 25.000 data sets were downloaded (approx. 1.5 TB) and processed for the retrieval of Lake Surface Water Temperature and Lake Ice. In addition to the input data, including the VIIRS Imaging resolution bands scientific data record (SDR), the VIIRS Cloud Masks (IICMO and VICMO), and the related geolocation information (GITCO and GMTCO), the VIIRS Land surface temperature (LST) Environmental Data Record (EDR; VLSTO) is downloaded for validation purposes of Lake Surface Water Temperature (LSWT) Physical Mono Window (PMW) results. It has to be noted that as of March 2017, the VIIRS Cloud Mask EDR with the naming convention of VICMO was released while the preceding cloud masks follow the naming convention of VIIRS Cloud Mask Intermediate Product (IICMO).

### 5.1.1. VIIRS imaging resolution bands SDR

The specifications of the VIIRS sensor not only outperforms the legacy operational AVHRR sensor in spatial, spectral, and radiometric quality but also significantly outperforms both MODIS and AVHRR with the 375 m spatial resolution of the VIIRS thermal imaging bands. Moreover, while thermal infrared images, for example from instruments on-board Landsat-7, have high spatial resolution (e.g., 60 m at nadir), the noise performance requirements are low, on the order of 0.4 K at 300 K, compared to VIIRS thermal imager of better than 0.1 K at 300 K (Cao et al., 2013; Cao et al., 2014). The spatial resolution of the I-band data increases from 375 m at nadir to 790 m at the end of the scan with a maximum scan angle of 56.28°. Because of a unique data aggregation scheme, this pixel growth rate, typical for MODIS, AVHRR, and other instruments, is considerably smaller than those of the heritage sensors with a pixel size increase at a ratio greater than four times (Lee et al., 2006; Wolfe et al., 2013). With an array of 32 detectors the scan width increases from 11.7 km at nadir to 25.8 km at the end of the scan due to the panoramic effect, called “bow-tie” effect, which is still observed after aggregation (Wolfe et al., 2013). This bow-tie effect leads to scan-to-scan overlap (i.e., data redundancy), which is minimized within the on-board pre-processing to save the downlink bandwidth. Across the second image zone (31.72° to 44.86°), the four outermost sample rows of an individual scan (two on each end) are replaced with fill values and in the third image zone (44.86° to 56.8°), the eight outermost sample rows of an individual scan (four on each end) are replaced with fill values in order to minimize the bow-tie effect. This approach is called “bow-tie deletion” and creates a striped pattern across the second and third zone at the edges of the swath in raw images (Fig. 28a (left)), which reflects the samples containing fill values. The stripes appear, if the data is displayed in pixel space, but are removed when the scan is projected onto the Earth’s surface. Fig. 28b (right) displays projected VIIRS band data extracted from the raw data) for the broader region of interest of this project. This satellite data were acquired on October 4, 2015.

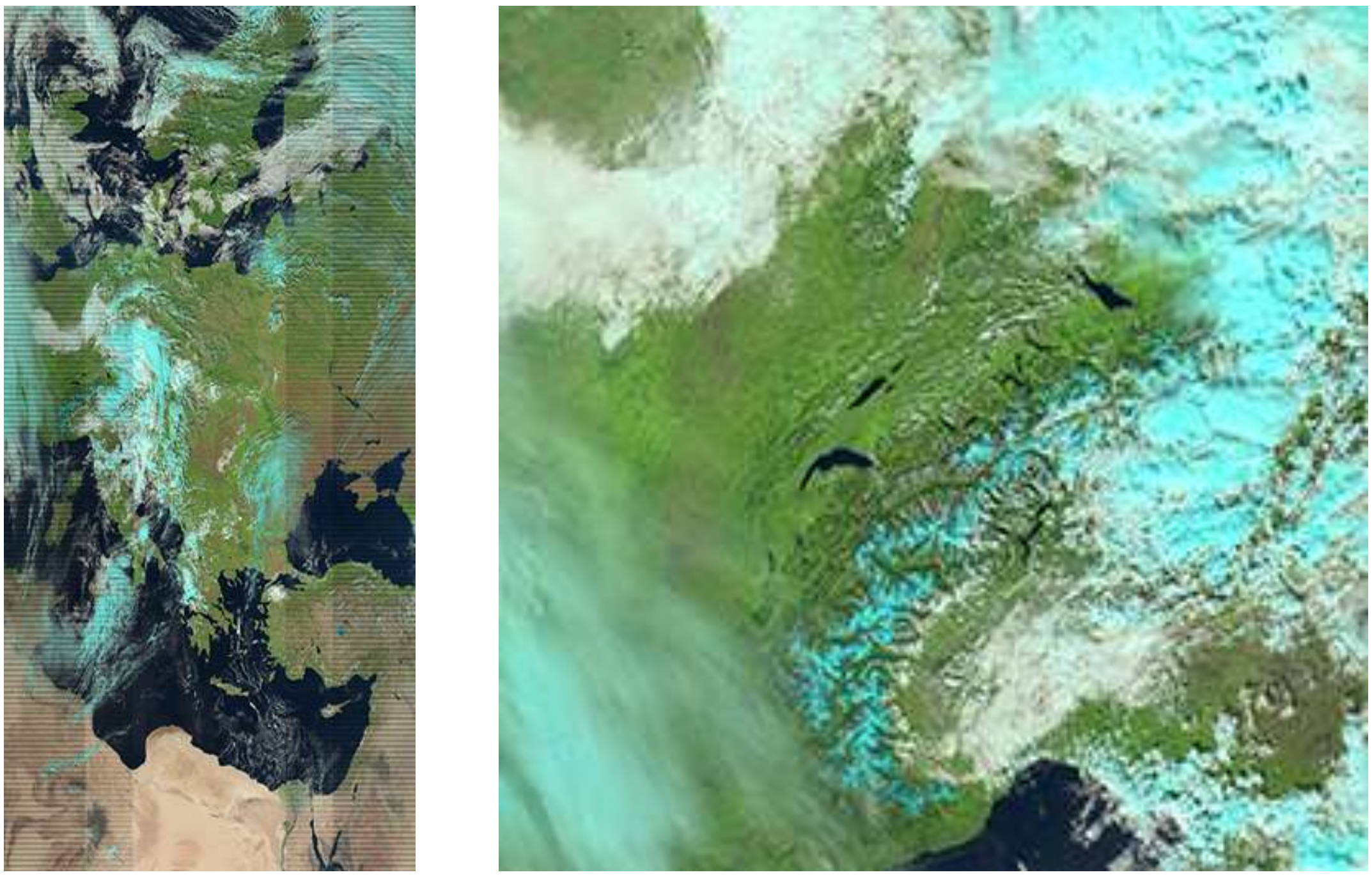

(a) (b)

**Fig. 28.** (a) Eight granules including the bow-tie deletion, (b) subset of interest in UTM zone 32N projection extracted from these granules.

### 5.1.2. Geolocation and cloud information

In order to remove the stripes of fill values resulting from the bow-tie deletion and project the granules onto the Earth's surface, geolocation information is required, especially as we use data from multiple scans. Geolocation data computed with and without terrain correction for each VIIRS pixel in the M-band and I-band are provided. These data sets are distributed as separate SDR geolocation products (HDF5 files) and include pixel geolocation (i.e., latitude and longitude) and geolocation related data, such as terrain height and satellite and solar geometry (i.e., zenith angle and azimuth angle). In this project, both the terrain corrected geolocation information in the I-band (GITCO) and M-band (GMTCO) (Note: Cloud Masks and VIIRS M-Band LST are only provided for M-Bands). Therefore, geolocation for the M-Bands is needed for the projection of these data sets). The pixel geolocation data is mainly utilized for mapping the data while the terrain height and satellite zenith angle data are additionally utilized as input for the pixel-based LSWT processing scheme.

The determination of LWST from space-based TIR measurements, furthermore, requires accurate cloud screening. Cloud contamination is found to significantly impact the surface temperature retrieval and additional cloud screening is strongly recommended in surface temperature applications (Liu et al., 2015). The VIIRS Cloud Mask (VCM) technique incorporates a number of cloud detection tests, including several infrared window threshold and temperature difference techniques and computes for each moderate-resolution VIIRS pixel a confidence that clouds exist (Godin, 2014). Four levels of cloud confidence are provided: confidently cloudy, probably cloudy, probably clear, or confidently clear. The final determination is a combination of the individual confidences of all applied tests. LSWT and lake ice phenology is only retrieved for pixels that are confidently clear.

## 5.2. Methodology

Fig. 29 shows a simplified flowchart of the lake ice detection processing scheme. This section explains the main processing steps, including the pre-processing, the LSWT retrieval, and the ice detection and retrieval of the lake ice phenology, in more detail.

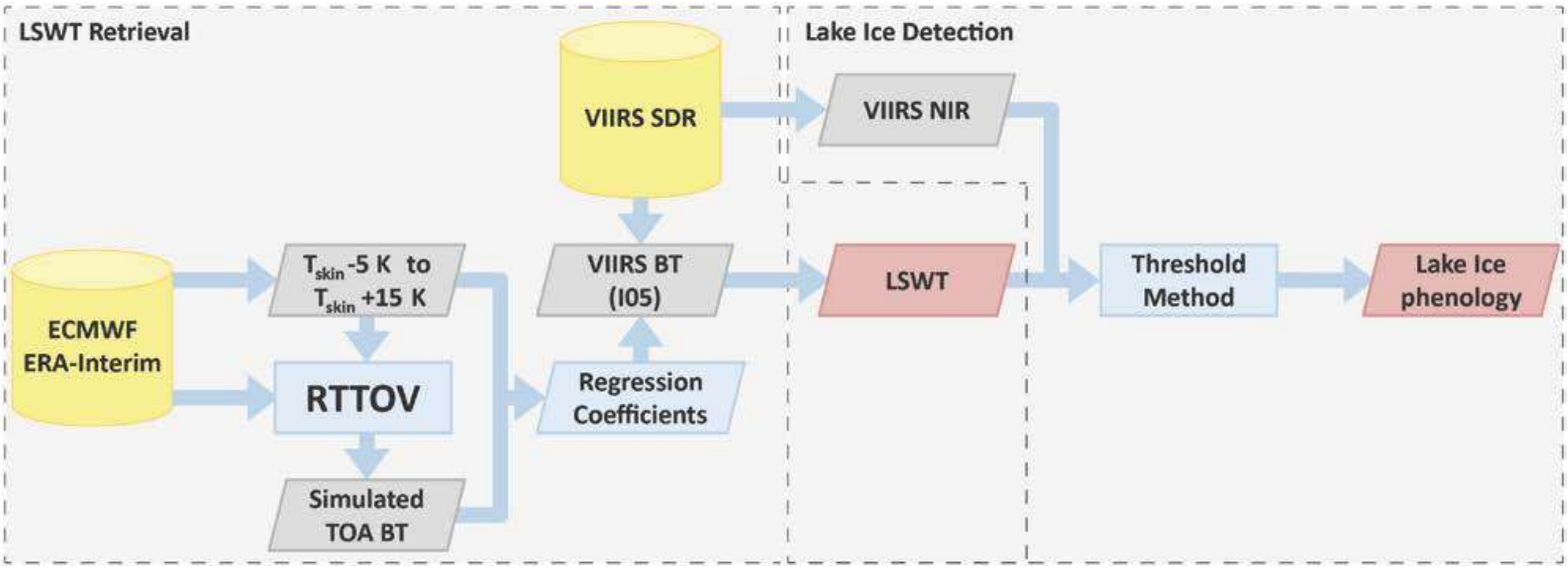

**Fig. 29.** Flowchart of the physical mono-window (PMW) model and lake ice phenology retrieval.

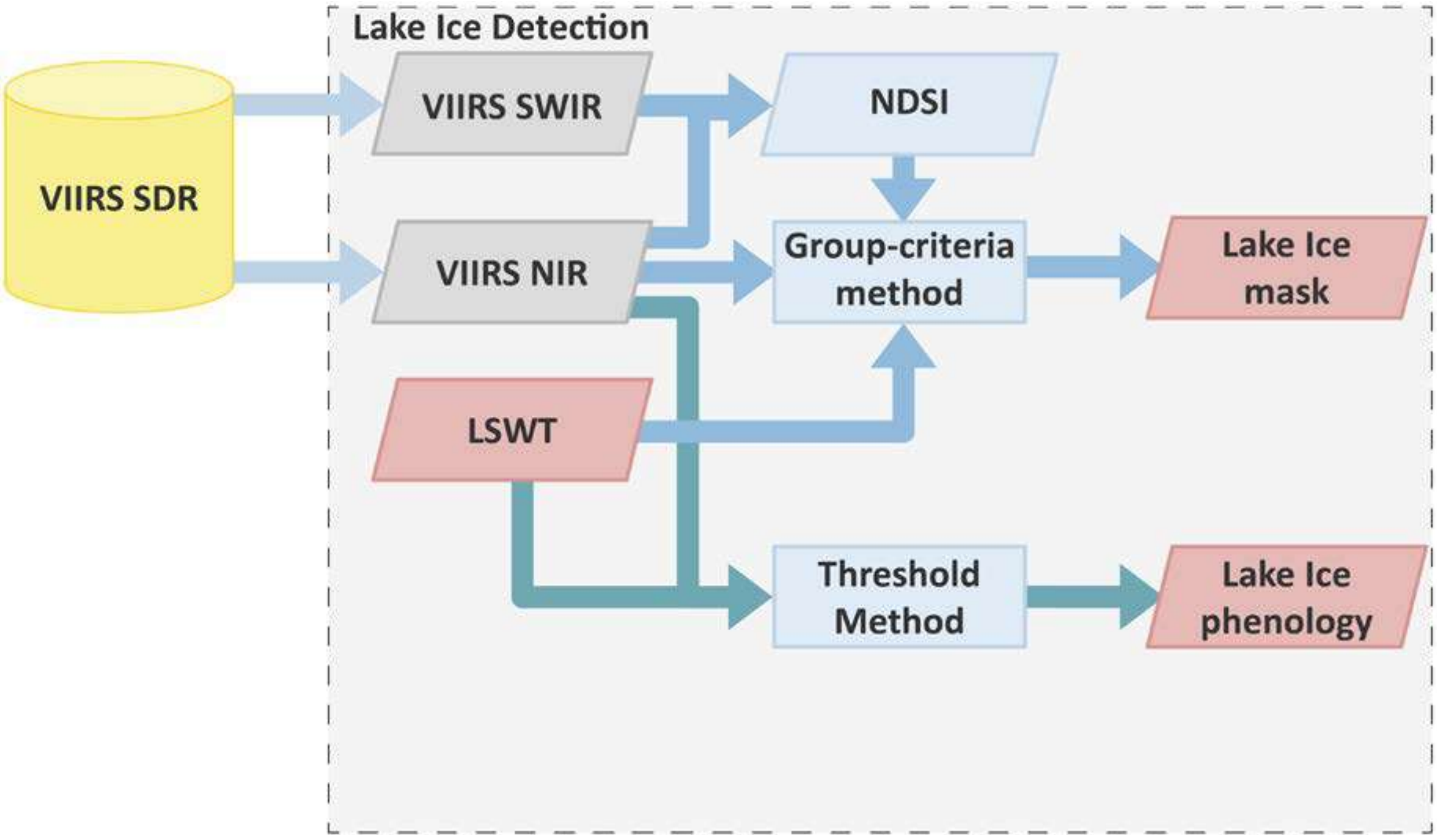

**Fig. 30.** Additional details of lake ice retrieval approaches (lake ice mask based on group-criteria method) are shown in this figure. The procedure considering the Normalized Difference Snow Index (NDSI) and group criteria is used for detecting lake ice, whereas the dynamic approach based on threshold method is shown for extracting information on lake ice phenology.

More details of the different processing steps and both retrieval methods to determine ice-covered pixels are explained in the following sections.

### 5.2.1. Pre-processing

Although this project is primarily a feasibility study, a comprehensive pre-processing chain was developed to process in a short time a huge amount of VIIRS data. The pre-processing includes python modules from Pytroll, which is a free and open source software suited for most of the pre-processing (i.e., reading, resampling, interpretation, and writing) of weather satellite data. The current version of Pytroll provides a direct read-in routine for VIIRS SDR data, but not for intermediate products (IP) and Environmental Data Records (EDR). Details about the Pytroll framework are described in Raspaud et al. (2018). Therefore, a script implementing subroutines of Pytroll was developed for the processing of both the cloud data (IICMO and VICMO) and the VIIRS moderate resolution LST EDR (VLSTO). The following three main procedures are included in the pre-processing chain:

- Loading the image band data and related geolocation information of the granules of interest
- Assembling, mapping, and resampling the data while removing samples with fill values resulting from the bow-tie deletion
- Writing reprojected (UTM Zone 32N) image band data covering the subset of interest (Fig. 28b)

This pre-processing chain is very efficient, and requires a minimum of user interaction.

### 5.2.2. Physical Mono-Window LSWT retrieval

The retrieval of the surface temperature of land or inland water from data recorded in the thermal band of a remote sensing system can be achieved by the solution of the thermal radiance transfer equation (Duguay-Tetzlaff et al., 2015a). The thermal radiance received by the sensor on-board a satellite can generally be divided into the ground radiance ($B(T_s)$), which is attenuated by atmospheric absorption, the upwelling atmospheric path radiance ($L^{\uparrow}$), and the downwelling atmospheric path radiance ($L_{\downarrow}$) reflected by the ground. Considering all possible effects from both the atmosphere and the ground, while approximating the Earth's surface as Lambertian emitter-reflector and neglecting the atmospheric scattering, the top of the atmosphere (TOA) radiance recorded in band c of a sensor can be expressed with a simplified radiative transfer (Eq. 1) (Li et al., 2013).

$$L_c(\theta) = \tau_c(\theta)\big(\varepsilon_c B_c(T_s) + (1-\varepsilon_c)L_c^{\downarrow}\big) + L_c^{\uparrow}(\theta) \qquad \text{(Eq. 1)}$$

where, $L_c(\theta)$ is the radiance recorded under view zenith angle $(\theta)$ and $T_s$ is the surface temperature of land or inland water. $\varepsilon_c$, $\tau_c$, $L_c^{\uparrow}$, and $L_c^{\downarrow}$ are the surface emissivity, atmospheric transmittance, upward atmospheric path radiance, and downward atmospheric path radiance, respectively, for the considered band. The ground emittance in band c ($B_c(T_s)$) is expressed as the Planck function at surface temperature $T_s$. The PMW model, applied in this project, is based on the direct inversion of the simplified radiative transfer equation (Eq. 1) with the Planck function approximated for one band of finite spectral band width in the thermal infrared (Eq. 2) (Chédin et al., 1985; Li et al., 2013; Yu et al., 2014).

$$LSWT \approx \left( \frac{c_2 \nu_c}{\ln\left(\frac{c_1 \nu_c^3 \tau_c(\theta)\varepsilon_c}{L_c(\theta) - L_c^{\uparrow}(\theta) - L_c^{\downarrow}(1-\varepsilon_c)\tau_c(\theta) + 1}\right)} - \beta \right) \Big/ \alpha \qquad \text{(Eq. 2)}$$

where, $c_1$ and $c_2$ are constants from Planck's law, $\alpha$ and $\beta$ are coefficients that depend on the spectral characteristics of the considered band, and $\nu_c$ is the band central wavenumber. The three parameters $\tau_c$, $L_c^{\uparrow}$, and $L_c^{\downarrow}$ can be estimated using information on atmospheric humidity and temperature from atmospheric profiles and the TOA brightness temperatures for a specific atmospheric profile can be simulated using a radiative transfer model. In this study, the Radiative Transfer for the Television Infrared Observation Satellite Operational Vertical Sounder code (RTTOV) is applied and atmospheric profiles from ECMWF ERA-interim are utilized to run RTTOV and simulate TOA brightness temperatures (BT) for temperatures ranging from skin temperature (Tskin) minus 5 K to Tskin plus 15 K (Bento et al., 2017;

Matricardi et al., 2004; Saunders et al., 1999). For each VIIRS observation, RTTOV simulations for model atmospheres with 37 pressure levels are performed considering the elevation and view zenith angle of each pixel and using ERA-Interim profiles interpolated in space and time according to the satellite acquisition. Since this requires several radiative transfer runs during the processing, it is crucial to use a fast radiative transfer code, such as RTTOV (Saunders et al., 2017). The coarse resolution of ERA-interim data in relation to the size of Swiss lakes does not influence the accuracy of the simulated skin temperatures as shown by Lieberherr et al. (2017) applying RTTOV for a validity study of LSWT retrieval. This includes also the varying path length through the atmosphere. More critical is the temporal allocation to consider the dynamic of water vapour in the atmosphere but due to the four analyses per day of ERA-interim the time shift between VIIRS overflight and ERA-interim is minor. Finally, the surface temperatures and the simulated TOA brightness temperatures are related by linear regression and the regression coefficients are used to derive the PMW LSWT from the VIIRS band 5 (11.45 µm) brightness temperature measured at the sensor level.

### 5.2.3. Ice detection and lake ice phenology retrieval

After careful assessment of the PMW LSWT retrievals with temperature based (T-Based) validation using in-situ measurements (Fig. 37), and cross satellite comparison including VIIRS moderate resolution LSWT (VLSTO) and AVHRR LSWT, the PMW LSWT and VIS, NIR, and SWIR reflectance values are utilized for the detection of lake ice and the retrieval of lake ice phenology (Fig. 30). As for sea ice and snow, the albedo of lake ice and snow-covered lake ice is very high at visible wavelengths and low at wavelengths longer than 1.4 µm due to stronger absorption and less backscattering in the shortwave infrared, with higher snow albedo in the VIS (Bolsenga, 1983; Mullen and Warren, 1988). Traditionally, ice and snow are detected with group-criteria methods relying on these spectral properties by applying several thresholds. The well-established Normalized Difference Snow Index (NDSI; see Eq. 3) that takes advantage of these spectral differences to identify snow and ice versus other features in a scene is defined as (Dozier, 1989; Justice et al., 2013):

$$NDSI = (R_1 - R_2)/(R_1 + R_2) \quad \text{(Eq. 3)}$$

where, R1 is the reflectance of the band in the visible or near infrared wavelength range and R2 is the reflectance in the shortwave infrared band. In this project, a first ice detection is performed by using I2 (0.846 – 0.885 µm) and I3 (1.58 – 1.64 µm) for R1 and R2, respectively, and a pixel is identified as ice covered if the NDSI value is larger than 0.45, the reflectance in band I2 is higher than 0.08 (Hall et al., 2002; Riggs et al., 1999), and the PMW LSWT is lower than 275.0 K (Liu et al., 2016).

The NIR and PMW LSWT data are further used to derive the lake ice phenological events following the two-step extraction method described in Weber et al. (2016). In contrast to the pixel-based lake ice detection, for each lake, time series of NIR and LSWT are obtained by averaging the NIR and LSWT values of all lake pixels, respectively (Latifovic and Pouliot, 2007). Instead of applying arbitrary or lake specific thresholds to retrieve the phenology from this time series, with this method the thresholds are derived from the LSWT and NIR time series themselves. In the first step, the NIR data is used, while in the second step, the LSWT and information (i.e., thresholds based on reflectance values as shown in Fig. 32) gained from the data during the first step are utilized. The lake ice phenology events are then extracted as the first and last dates of the frozen and open periods defined by the LSWT and are referred to as start of break-up (BUS), end of break-up (BUE), start of freeze-up (FUS), and end of freeze-up (FUE).

The automatic extraction procedure relies on LSWT and NIR reflectance values and is very sensitive to noise or unrealistic values. A careful noise reduction and smoothing (Fig. 31) has to be applied to avoid biased thresholds. The aim of this pre-processing procedure is to minimize the high variability of the data but to impact the seasonal signal as little as possible. After de-spiking a running 3d median filter is applied with an interpolation step to fill the de-spiked gaps. Based on these data sets a fitting procedure of the lower and upper bound curve is applied using a running window size of 15d followed by smoothing of these curves using a running mean of 15d (7d) for the lower (upper) bound curve. The lower bound curve was calculated to extract open lake, whereas the upper bound curve was used for frozen lake conditions. The lower (upper) bound curve was calculated with the maximum of minimum (minimum of maximum) values (Weber et al., 2016).

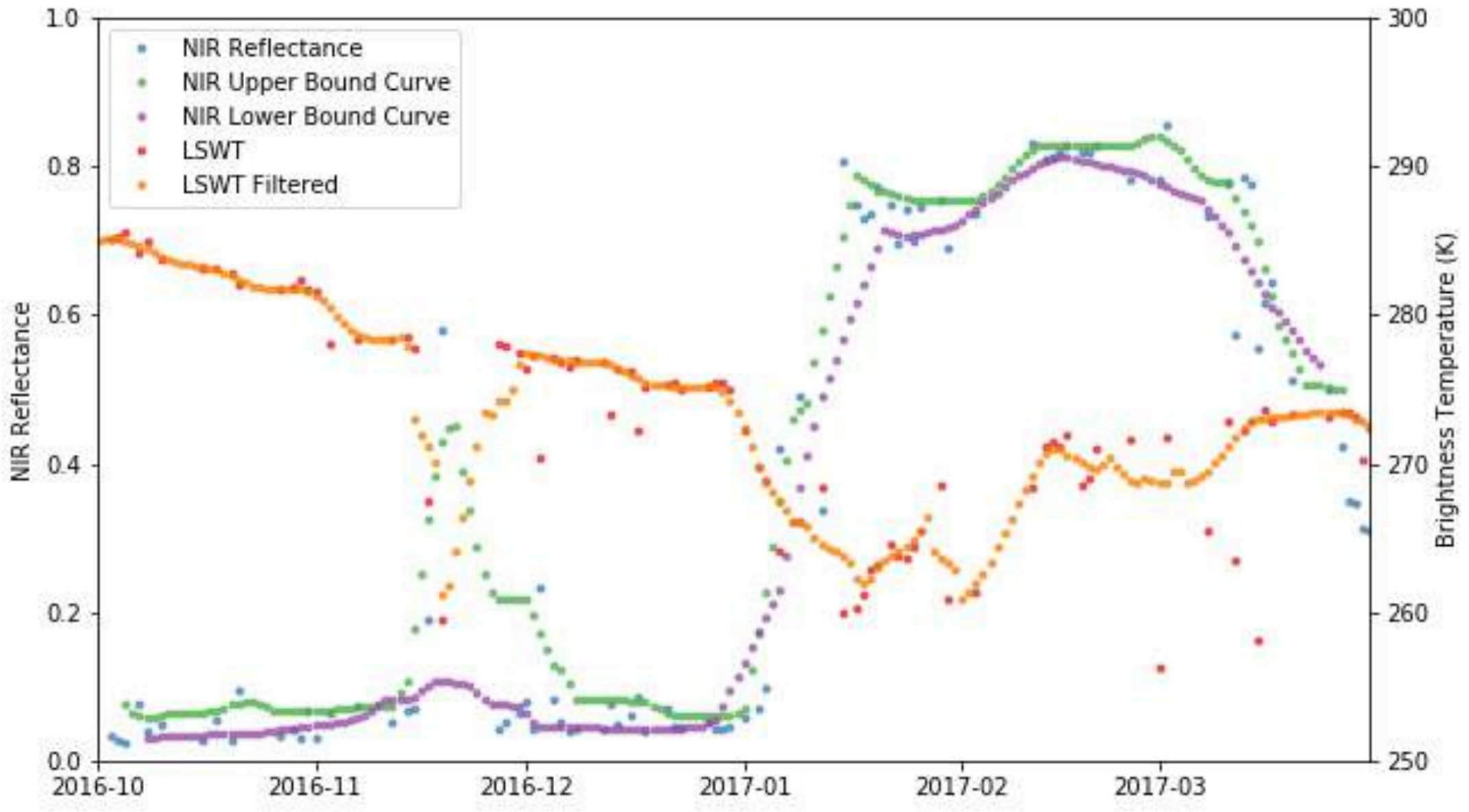

**Fig. 31.** The time series (Oct.2016 – March 2017) show the smoothed NIR values (including the upper (green) and lower (purple) bound curve) and LSWT values (red; orange = filtered).

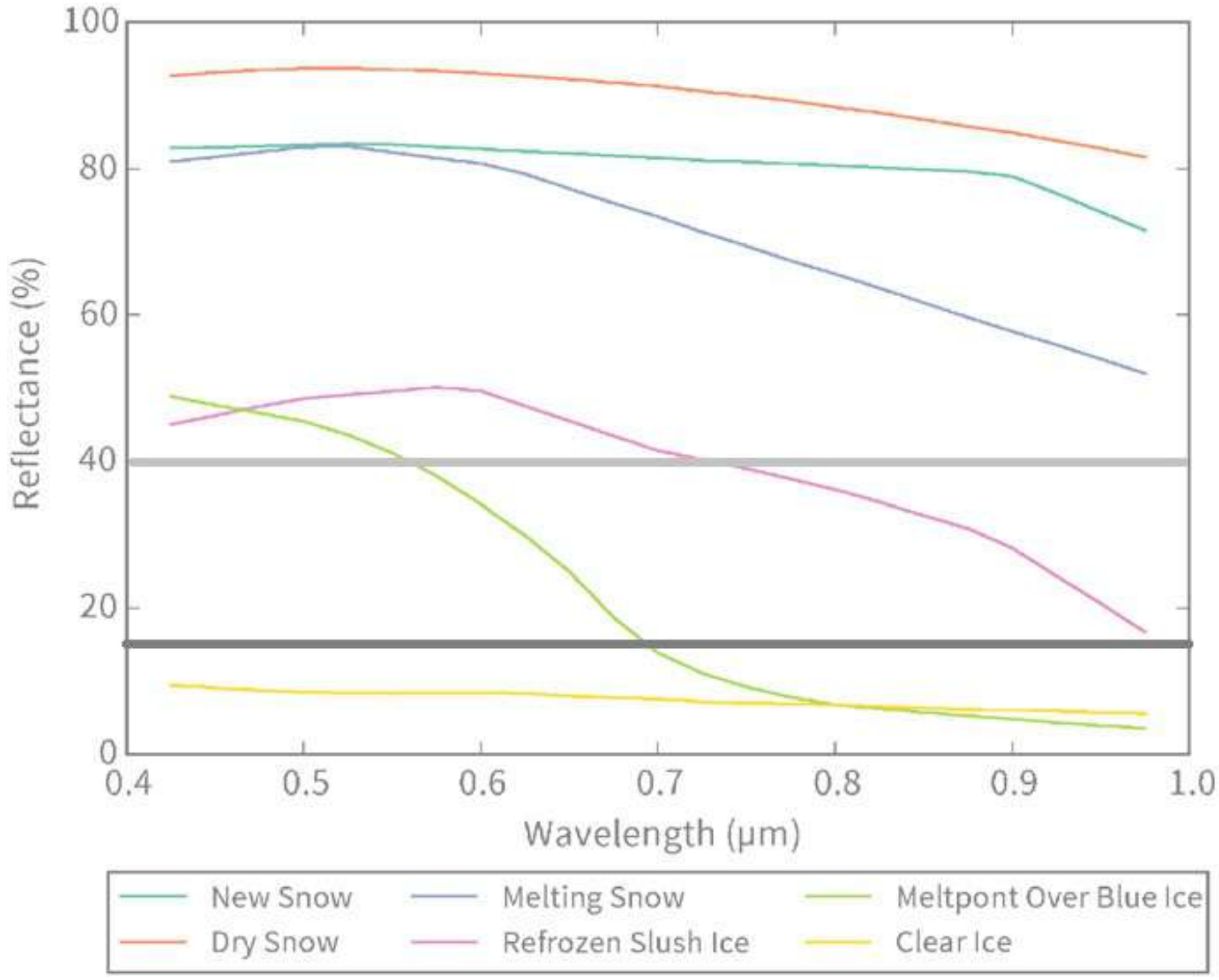

**Fig. 32.** Selected NIR thresholds based on reflectance values of in Weber et al. (2016) NIR reflectance values above 40 % (grey line) are used as threshold for frozen lakes and NIR values below 15% (dark grey) indicate open water.

It is obvious from Fig. 32 that clear ice and melt ponds over blue ice are not detectable using VIS and NIR reflectance values because also open water has very low reflectance in this spectral range.

The original NIR values are smoothed (Fig. 31) and the extracted minimum- and maximum values as lower and upper bound curve are visualized for Lake Sils. The first step of the lake ice phenological approach determines open water if more than 70% of all lake pixels have a reflectance value of open water. In addition, if 70% of NIR reflectance has values above 0.4 the lake is classified as frozen (Fig. 33). This preliminary information about the lake status is used for the next step to determine automatically the threshold considering the histogram of retrieved temperature.

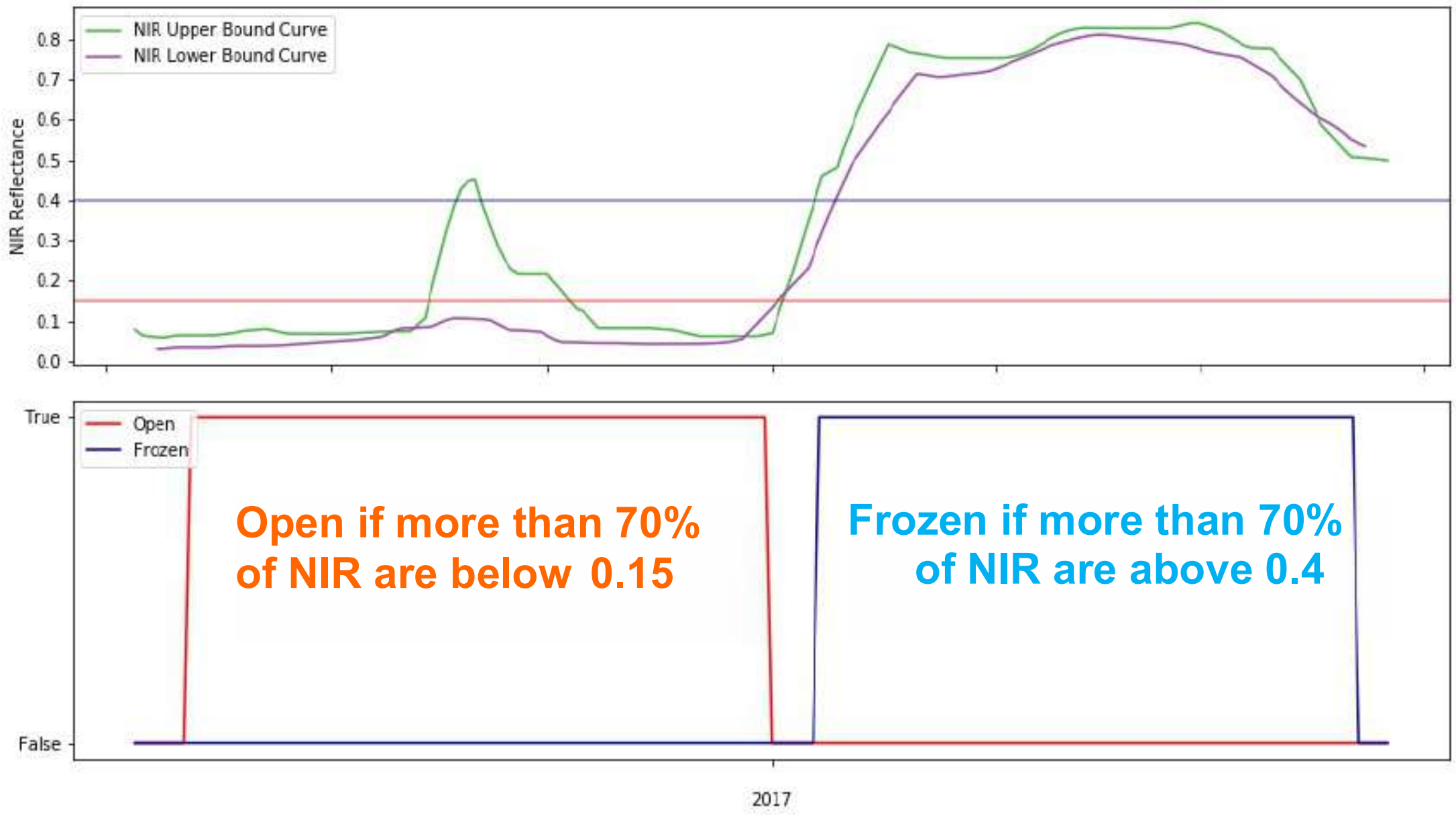

**Fig 33.** Smoothed upper and lower boundary of NIR reflectance with defined thresholds for open water and frozen status. The lower figure part depicts if the test was true or false.

The lake pixels, which are classified as frozen lake or open water with a high confidence considering the thresholds of the NIR values are the basis for the next step to define the thresholds for the TIR values. All LSWT pixels, which are indicated as either open water or ice, are selected for the automatic approach to define the LSWT thresholds. Histograms of LSWT are $90^{th}$ (frozen) and $10^{th}$ (open) percentiles are calculated to define the thresholds for the subsequent steps. Especially the open lake sample needs a longer time series to derive the threshold in the positive range. As shown in Fig. 34 for Lake Sils the length of the time series is not sufficient because the automatically calculated threshold for open water has a negative value. This could be avoided by including LSWT values from spring season.

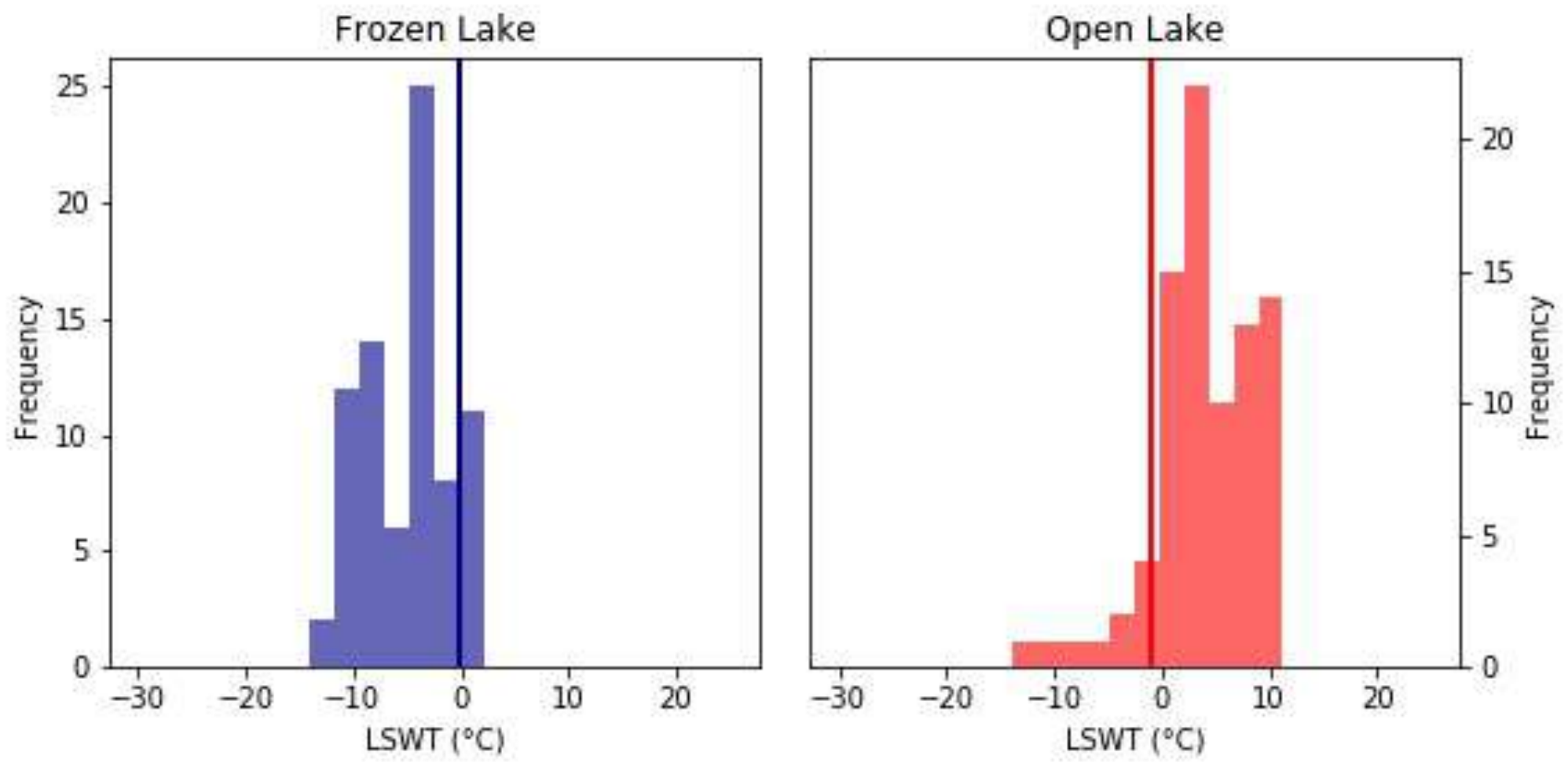

**Fig. 34.** Lake Sils: Histogram of LSWT values for frozen lake (left) and open water (right).

Based on this second step of the approach, the period of frozen lake (start of freezing, end of lake ice cover) are automatically determined. The smoothed (15d) LSWT time series of Lake Sils is shown in Fig. 35 with the determined thresholds for open lake (red line) and frozen lake (blue line). Note: the red line should be above the frozen threshold for proper retrieval of open water. A longer time series including some spring or summer months will result in increased values for the open water threshold as shown in Weber et al. (2016). To avoid that the retrieved dates of start of freezing / end of frozen conditions are influenced by erroneous values the majority of pixels (70%) has to fulfil the criteria of the thresholds. Open water is determined if more than 70% of the pixels LSWT is above (below) the threshold "open water" ("frozen conditions"). The detection of a short period of frozen days is suppressed by a moving window of 15 days. This can be improved with a shorter moving window but it is recommended to have at least 7 days otherwise the LSWT curve is too noisy to extract meaningful phenological dates.

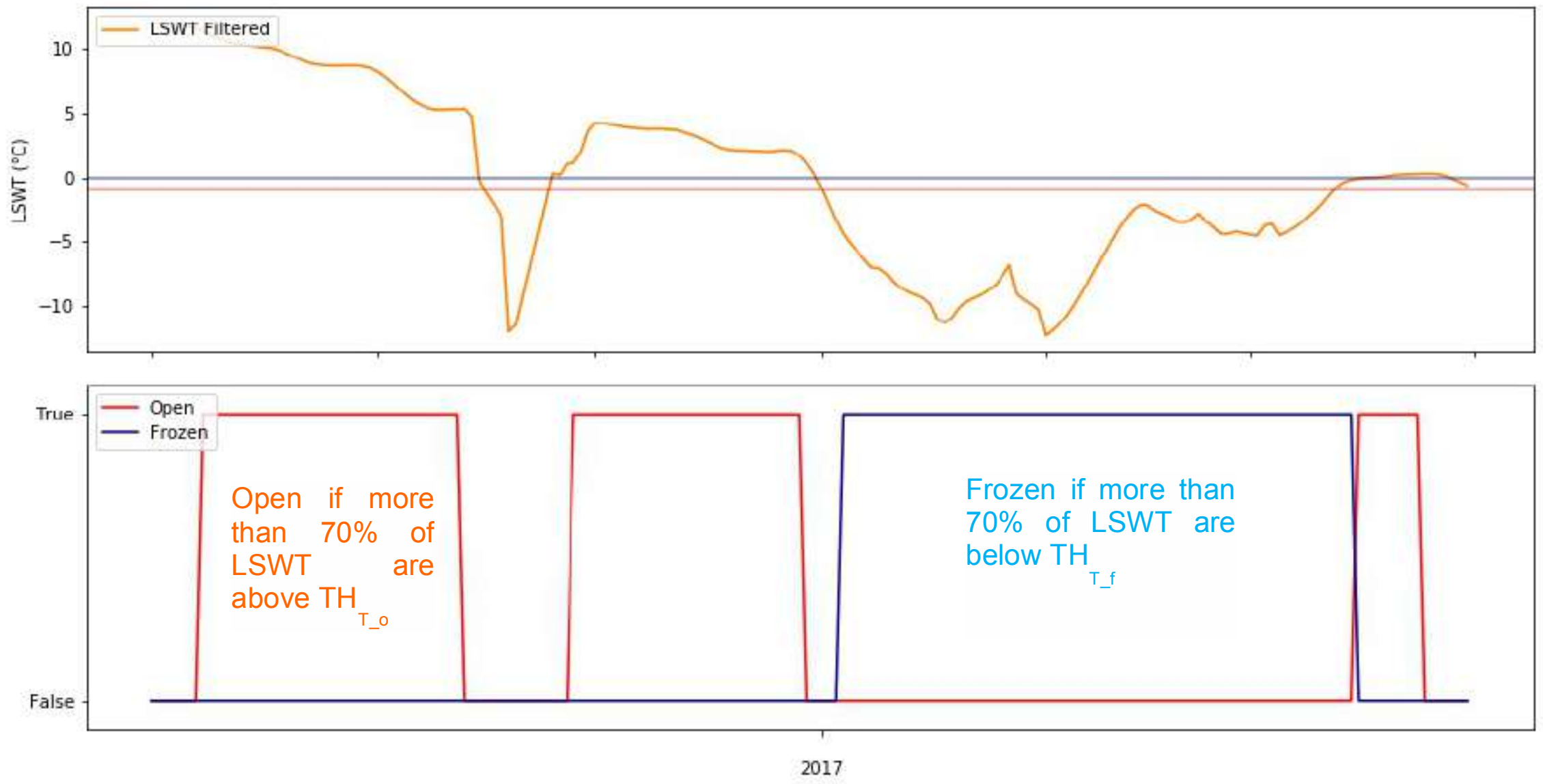

**Fig. 35.** LSWT time series of Lake Sils (upper panel; brown curve) with automatically defined thresholds (red and blue line in the upper panel). The lower figure part shows the periods of open lake (red) and frozen conditions (blue).

## 5.3. Results and discussion

### 5.3.1. Cross-satellite LSWT validation

The cross-satellite (scene-based) comparison is widely used for surface temperature evaluation and involves comparing a new satellite LSWT product with a cross-satellite measurement, such as a heritage surface temperature product or a product derived employing a different approach (Guillevic et al., 2013; Hulley and Hook, 2009; Trigo et al., 2008). The VIIRS LST EDR (VLSTO) that is archived and distributed by NOAA CLASS and has been operationally produced since August 11, 2012, is utilized for the comparison with the VIIRS I-band PMW LSWT. VIIRS LST EDR provides measurement of skin temperature over global land coverage including coastal and inland-water and has reached validated V1 stage maturity defined as "using a limited set of samples, the algorithm output is shown to meet the threshold performance attributes identified in the JPSS level 1 requirements". A single (baseline) split window algorithm utilizing brightness temperatures measured in the moderate resolution bands M15 (T15) and M16 (T16) centred at 10.76 µm and 12.01 µm, respectively, is applied to derive the LST (Yu et al., 2005). The VIIRS EDR LST is retrieved for land surface considering an emissivity of 0.98, which is almost identical of water (0.99) and relies on the M-channels with a spatial resolution of 1km. A comparison between LST and the results based on the implemented PMW approach gives the proof that the atmospheric influence could be modelled with sufficient accuracy independent on the amount of water vapour. The main advantage of the PMW approach is the usability of I-channels with a spatial resolution of 375m.

The cross comparison of the VIIRS I-band PMW LSWT with VIIRS M-band split window (SPW) LSWT is conducted at pixel level for the target lakes. Fig. 36 shows the comparison results for Lake Greifen and Lake Sils for October to December 2016. The bias and coefficients of determination resulting from the comparison are 0.960 K and 0.909 for Lake Greifen and 1.421 K and 0.867 for Lake Sils, respectively. This indicates that VIIRS M-band SPW LST EDR and VIIRS I-band PMW LSWT, although based on different assumptions to describe the atmospheric effect (i.e. physical mono window and split window method), are in a good agreement. Even so, in both the I-band PMW LSWT and M-band SPW LSWT data cloud leakage (i.e., cloudy but identified as clear) caused some very low surface temperatures with values below 265 K for Lake Sils.

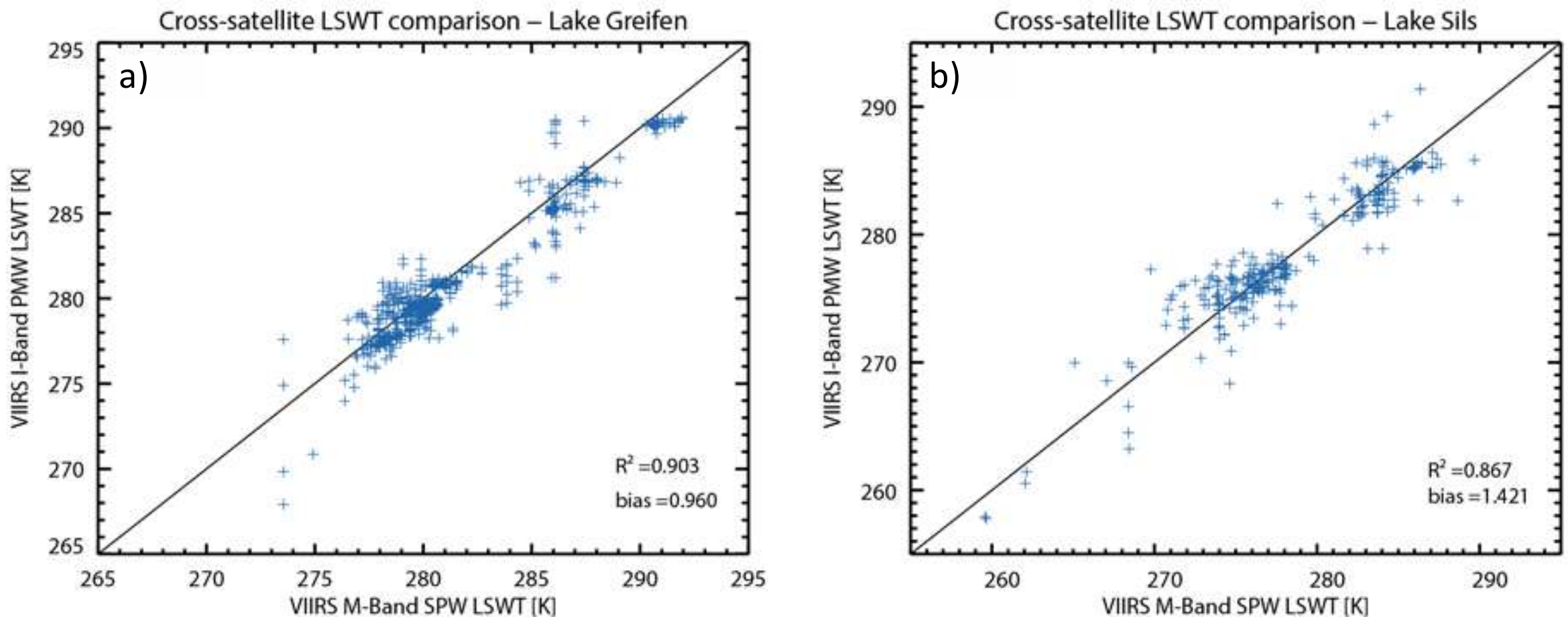

**Fig. 36.** Cross-comparison results between VIIRS I-band PMW LSWT and VIIRS M-band split window LSWT under cloud clear condition for Lake Greifen (a) and Lake Sils (b).

The comparison provides useful quality information on spatial agreement and disagreements between surface temperature products. However, different products based on similar formulations (e.g., split-window) or input data sets can be highly consistent with each other while at the same time significant discrepancies are observed when compared to ground reference measurements (Guillevic et al., 2014). Therefore, the intercomparison of surface temperature products does not represent a comprehensive validation and additional validation efforts with ground-based reference data are required.

### 5.3.2. Temperature based LSWT Validation

The temperature based validation approach involves a direct comparison of the satellite-based estimates with the corresponding ground-based measurements of surface temperature, and has been frequently used to validate surface products derived from several satellite instruments. This validation method is particularly suited for studies over inland water bodies, such as lakes, as they provide large spatially homogenous temperature targets (Coll et al., 2009; Hulley et al., 2011). In contrast, spatially heterogeneous validation sites require a large number of in-situ measurements for precise characterization (Guillevic et al., 2012). In this project, surface temperature measurements from radiometers and temperatures measured just below the surface of the target lakes are used to evaluate the performance of the VIIRS PMW LSWT processing scheme. So far, in-situ skin temperature measured by a radiometer mounted at Buchillon starting November 2016 has been available for validation. The LSWT retrieved from the VIIRS thermal I-band (I5) with the physical mono-band approach is compared with the in-situ skin temperature measured for Lake Geneva and with AVHRR LSWT derived applying the split window method for the period from November 1 until December 31, 2016, in Fig. 37. Temporally, the start time of the granules is matched with the closest ground site observations. The upper plot (Fig. 37a), shows a pixel-based validation and LSWTs derived from cloud-covered pixels are indicated with grey circles. In the bottom plot (Fig. 37b), the cloud clear LSWT is averaged over a 3x3 pixel window and LSWT derived from a window with more than 2 pixels obscured by clouds are indicated as cloud covered.

The comparison shows that VIIRS PMW LSWT and AVHRR SPW LSWT produce consistent measurements in the same range. With a bias ranging from 0.87 K to 1.42 K and a coefficient of determination greater than 0.9, the cross comparison results show that the VIIRS I-band PMW LSWT are in a good agreement with the operational VIIRS M-band split window (SPW) LSWT. Our AVHRR LSWT product was used because it is well validated in previous studies (Riffler et al., 2015; Lieberherr and Wunderle, 2018) and gives a first indication about the stability of VIIRS PMW LSWT product during three winter months. From around November 4 to 23, the in-situ measurements include disturbed and missing data. For the days with valuable in-situ data, the validation demonstrates that the VIIRS PMW LSWT method performs very well. Furthermore, satellite-based surface temperatures, which were lower than in-situ measurements were identified as cloud contaminated indicating an overall good accuracy of the cloud detection (Fig. 37). However, while cloud leakage seems to be less of an issue, from visual comparison, it was found that sometimes snow cover is partly masked as cloud. Consequently, frozen lakes with snow cover might be classified as cloud. Snow, ice, cloud differentiation has been listed as the major issue for further improvement regarding EDRs (Kopp et al., 2014). The accuracy of the VIIRS cloud mask, however, seems to be appropriate for this feasibility study. The quality of the cloud mask was visually checked using RGB-images with an overlay of the cloud mask during the days with first ice cover and during the period of ice-off. During these periods, the quality of cloud masks were reasonable. A more detailed study by Kraatz et al. 2017 did not confirm our very preliminary results because they analysed different cloud masks (MODIS and VIIRS) for winter conditions in the eastern part of USA. They found that VIIRS detect fewer clouds than MODIS (Terra: 12%) even if only using “confidential clear”. A detailed study on cloud cover during winter time in Switzerland is out of the scope of this feasibility study and would require a long time-series of VIIRS data to be compared with ceilometer measurements.

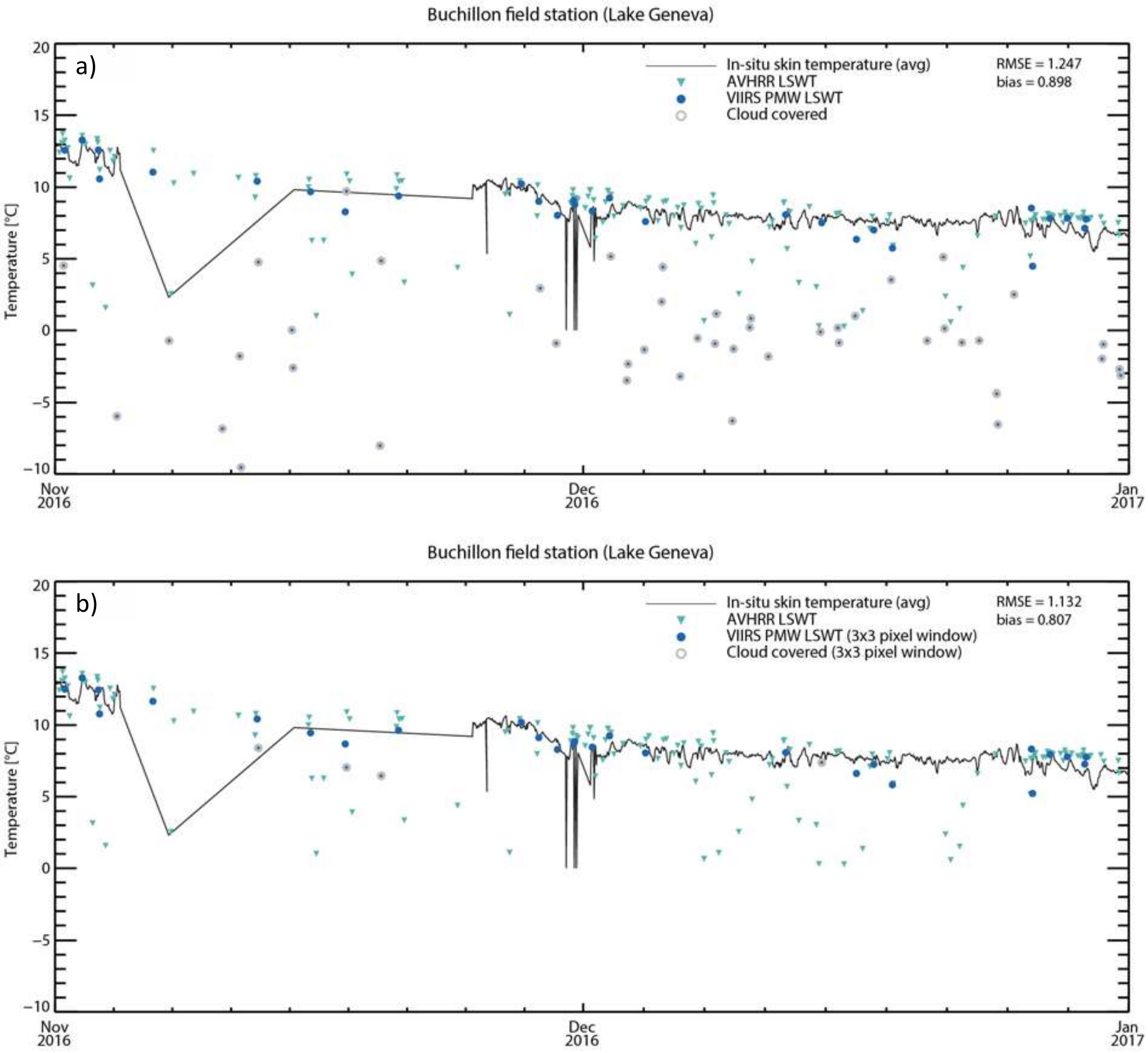

**Fig. 37.** VIIRS PMW LSWT (pixel-based (a) and averaged over a 3x3 pixel window (b)) and AVHRR SPW LSWT plotted alongside skin temperature measured by a radiometer mounted at Buchillon (Lake Geneva) from November 1 until December 31, 2016.

### 5.3.3. Ice detection

As the cross-satellite comparison and the validation with in-situ data indicate a high accuracy of VIIRS PMW LSWT, a first lake ice detection by applying thresholds on the NDSI, the NIR reflectance, and the VIIRS PMW LSWT was conducted for Lake Greifen and Lake Sils for October until December, 2016. While the results for Lake Greifen indicate open water during all three months, for Lake Sils under cloud-free conditions lake ice was detected for a few days in November and several days in December. The daily /twice daily observations of Lake Sils showed for the year 2017 that in the beginning of ice-on conditions the lake was obscured by clouds for 2-3 days resulting in an uncertainty of approximately 2 days to determine the date of ice-on (Fig. 39). A similar situation was observed for Lake Sihl where one to two days with cloud cover prevent the detectability of ice-on (Fig. 40). A comparison with the results of the other datasets (Webcam) indicate that the determined date of ice-on is very likely. This is consistent with a visual assessment of Landsat quicklook data indicating that in winter 2016/2017, Lake Sils was not frozen for longer than a few days up until January. Fig. 38 illustrates an example of the lake ice detection for Lake Sils on December 22, 2016. The bottom right image includes the lake mask that was computed by applying a 200 m buffer. White and purple lake mask pixels denote ice and open water, respectively. In the next steps, the accuracy of the ice detection needs to be assessed. Furthermore, the methodology for deriving the lake ice phenology described in Section 5.2.3 will be applied on the VIIRS data.

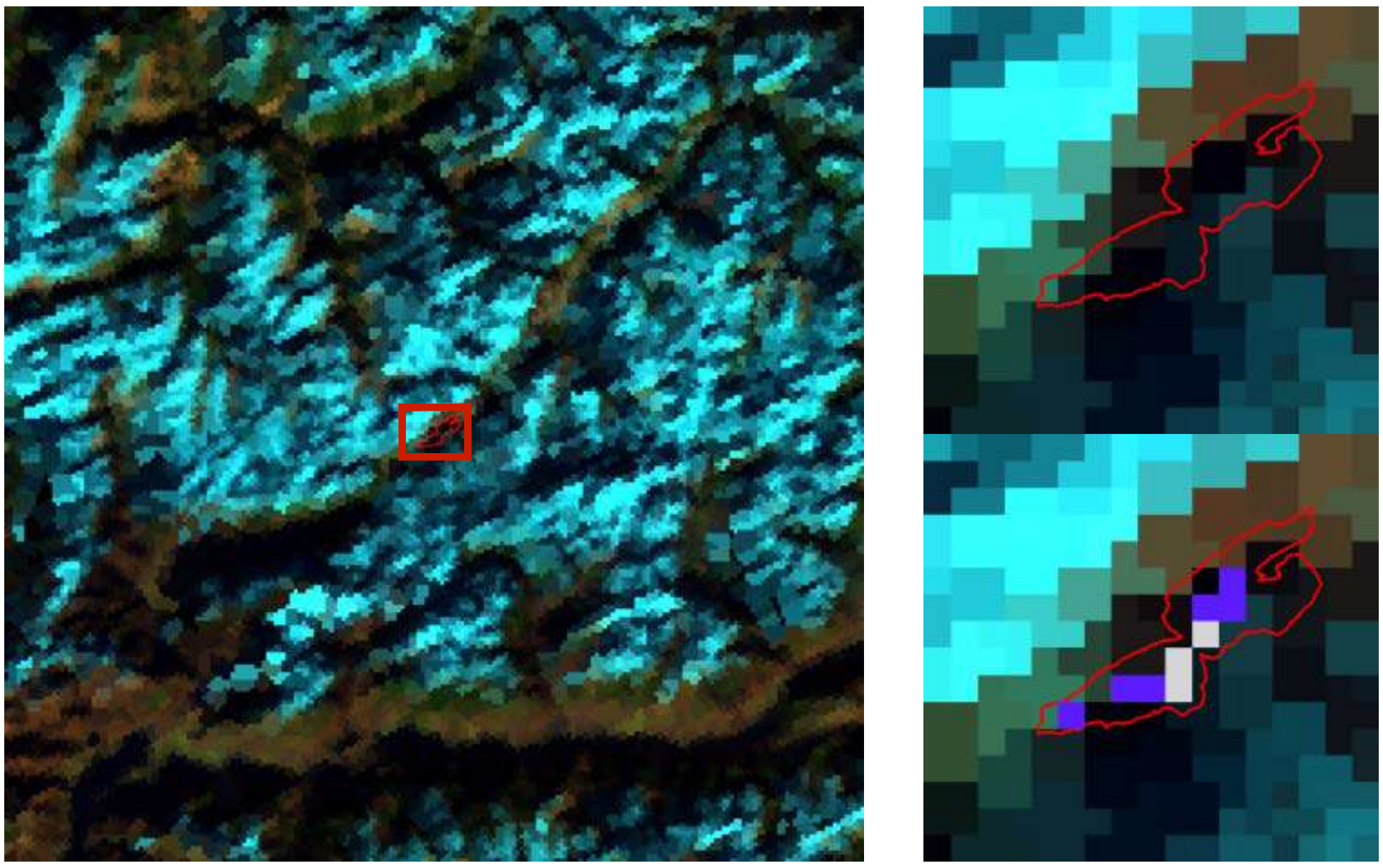

**Fig. 38.** Example of lake ice detection for Lake Sils on December 22, 2016. The natural colour RGB (I03, I02, I01) illustrations display the broad region around Lake Sils (left) and the zoomed-in region of interest (right) with a spatial resolution of 375 m. The bottom image on the right includes the lake ice mask, where white and purple pixels indicate lake ice and open water, respectively. Only pixels totally within the lake were processed. The result was visually controlled.

Fig. 39 shows a short sequence of daily images covering Lake Sils in the beginning of January 2017. These examples were selected to show the influence of cloud cover even for operational satellite systems with daily temporal resolution. During the 3rd and 4th of January some areas of the lake were cloud-free but the central part was cloud-covered (dark grey pixels). Hence, no information about the status (open water, lake ice) for this part of the lake exists. After 1.5 days with overcast conditions, the first unobscured view occurred on January 6 showing that the lake was completely frozen (blue pixels), which was detected by a successful classification process based on NDSI values and additional thresholds. The fixed thresholds are selected from peer-reviewed publications (Dozier, 1989; Hall et al., 2002; Justice et al., 2013) and visual inspection of images with apparent ice cover. Depending on environmental conditions, snow on ice may start to melt resulting in a decrease of NIR reflectances (Fig. 32), which prevent reliable ice detection. In addition, a melting snow cover has a temperature near 0°C, which could prevent detection of ice depending on the selected temperature thresholds. Furthermore, black ice conditions are not detectable by NDSI and NIR-threshold.

The Figs. 40 and 41 show the dynamic of ice cover of Lake Sihl. The first example (Fig. 40) depicts the beginning of freezing in January 2017. In the morning of the 1st January, some pixels with ice cover (light blue) were detected but during noon the ice changed conditions or melted. After a cloudy day (2nd January), the lake was completely covered by ice (3rd of January). The next days a persistent cloud cover hampered the detectability of lake ice and no information could be retrieved.

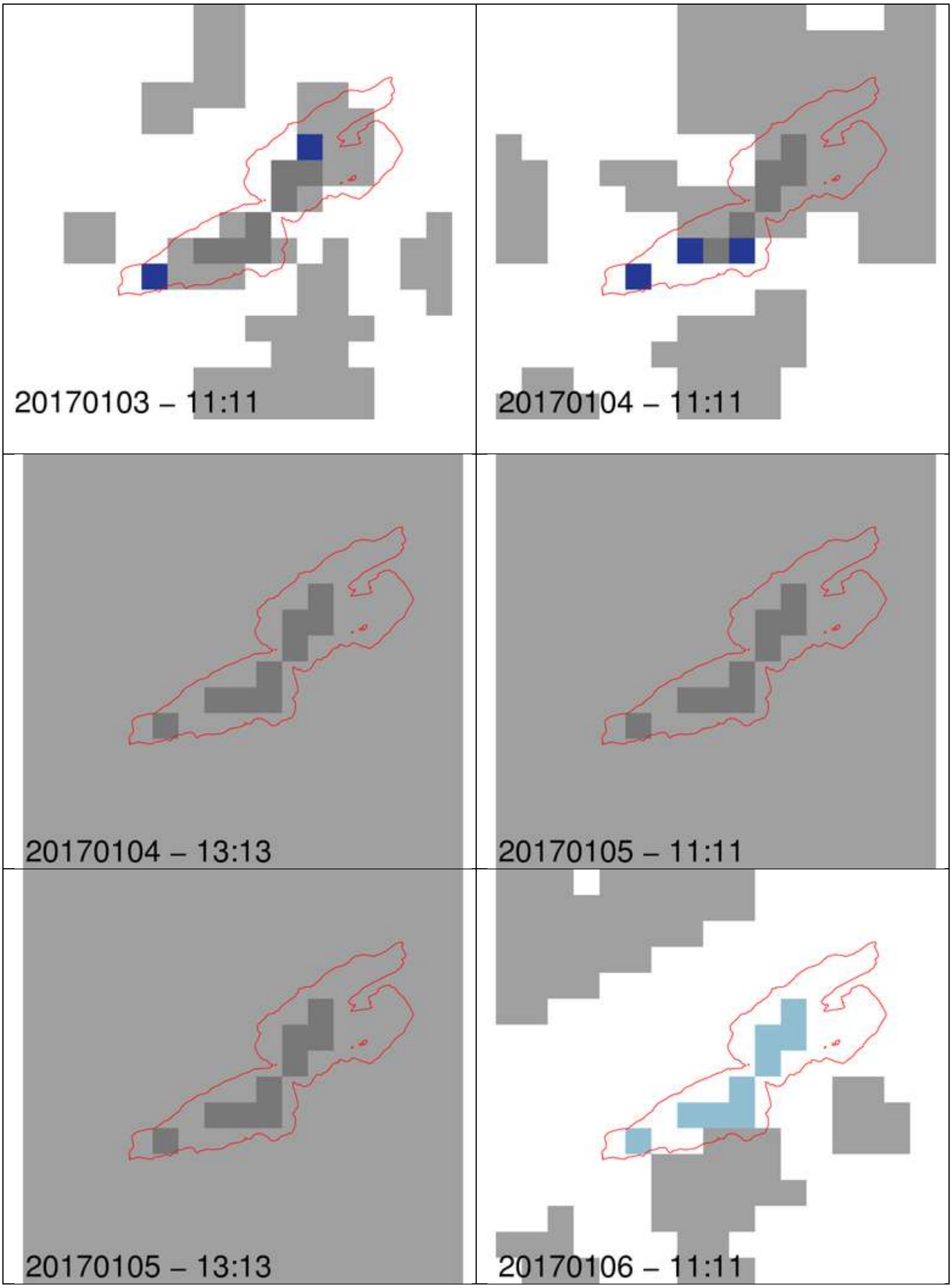

**Fig. 39.** Ice detection for Lake Sils based on VIIRS I-band data. The short time series (3. Jan. – six. Jan. 2017) showing open water (dark blue), clouds (grey), cloud covered lake pixels (dark grey) and frozen lake (light blue), respectively.

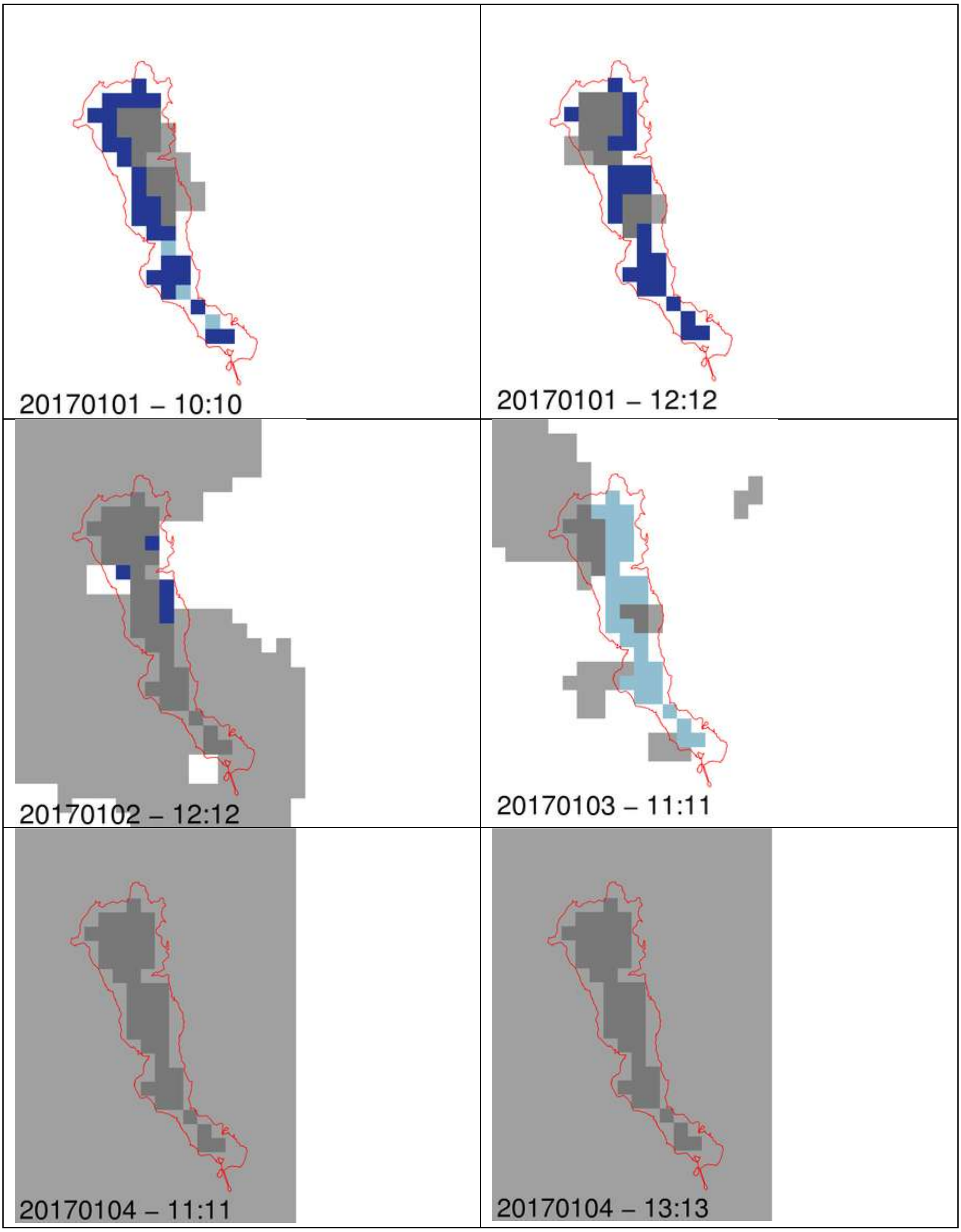

**Fig. 40.** Ice detection for Lake Sihl; Ice-on started on the first of January 2017 with full ice cover on 3rd of January 2017. Afterwards persistent cloud cover hamper ground visibility. Dark blue: open water; light blue: lake ice; grey: clouds.

After the winter season with increased warming of the atmosphere the ice cover of Lake Sihl was melting. During the three days, selected to show the process of melting, only some scattered clouds prevent detectability of the lake surface. On the 13th of March, the southern part of Lake Sihl showed open water (dark blue) but in the northern section all pixels were classified as lake ice. One day later the amount of lake ice decreased along the lakefront (dark blue pixels). Most likely strong melting caused changes in reflectance and temperature of the ice cover resulting in wrong retrieval of open water in the afternoon (14.3.2017) because one day later the ice cover showed almost the same distribution as on the $14^{th}$ at noon. The melting continued and on the $16^{th,}$ only some ice-covered pixels remained at the northern part of the lake. The advantage of VIIRS I-bands with its high spatial resolution is clearly visible to detect the gradual changes of lake ice retreat during the melting period. In addition, with the high temporal resolution (daily – two times per daytime) of the NPP satellites the chance is quite high to detect start of melting / end of ice cover also from small lakes in Switzerland.

## 5.4 Summary

UniBe has developed a semi-automatic processing chain to retrieve lake ice using the I-bands of the VIIRS sensor on NPP satellites. Only hand-over of data sets between the used modules (ECMWF ERA-interim and RTTOV, download of VIIRS and pre-processing, PMW approach and lake ice detection) need manual interaction but this can be automated for an operational application. The advantage of the I-bands is the high spatial resolution of 375m in thermal infrared making the sensor a proven tool to monitor small lakes in Switzerland. As shown in Fig. 39 of Lake Sils, a size of $4km^2$ is the limit of a retrieval but this depends also on the shape of the lake. A more regular shape (e.g. Lago Poschiavo) may support a retrieval even of smaller lakes. Assuming that only nine VIIRS pixels are needed for lake ice detection and a distance of 200m to the coastline is required one may apply the developed method for lakes larger than $2.5km^2$. The lake ice retrieval can be split into three steps: pre-processing of VIIRS data and atmospheric correction based on RTTOV runs and atmospheric data from ECMWF; lake surface water temperature retrieval and calculation of NDSI; and finally the automatic detection of ice phenology. The developed method at UniBe has demonstrated the feasibility to detect lake ice and show the spatial dynamics of freezing and melting during wintertime. However, it has to be noted that even operational satellite systems with a high temporal resolution (daily) are sometimes limited by persistent cloud cover. Hence, the precise start of freezing (or full ice coverage) and ice-off cannot be determined with the needed accuracy. Depending on the cloud condition, the determined ice phenology may vary by ± 5 days. Ice detection based on NDSI and additional thresholds is a robust method but the quality relies on well-defined thresholds, which may vary depending on lake conditions. The first differentiation of open water and ice can be seen as the first step of the UniBe approach. The second step uses the pre-defined open water and ice covered pixels to determine automatically the phenological dates of ice-on and ice-off based on LSWT information. The last step does not need pre-defined thresholds and is independent from the selected lake. A critical procedure of this second step is determination of the filter window to smooth the noisy data (TIR and NIR). A tested 15-day window results in a good smoothing of the original data but suppresses the detectability of ice events of short duration (< 3 days). Hence, a 7-day smoothing window might be the better solution but this depends on the noise in the data set caused mainly by sub-pixel clouds, which were not detected by the NPP cloud mask retrieval.

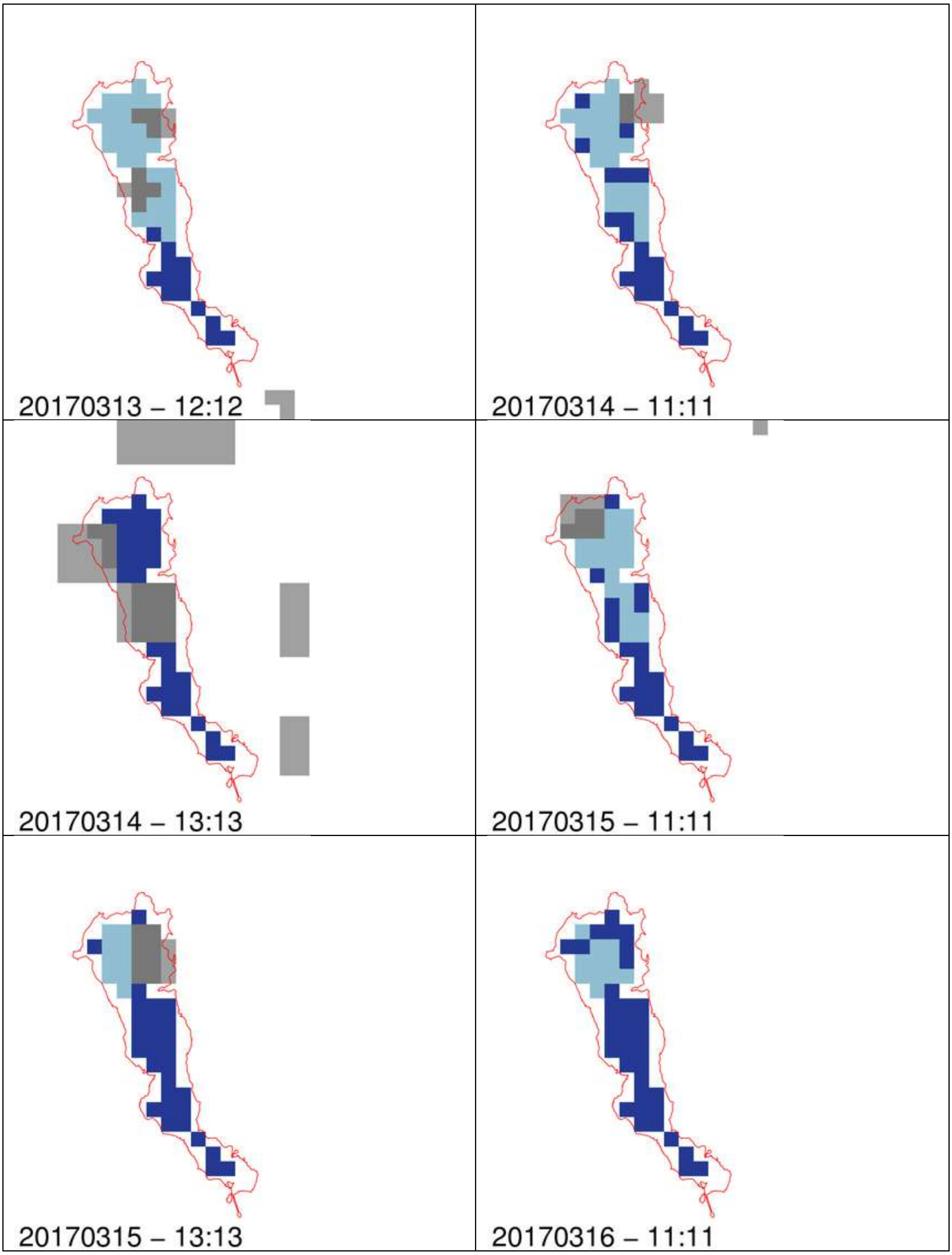

**Fig. 41.** Ice detection at Lake Sihl. Ice-off retrieved in middle of March 2017. The six time steps show the decrease in ice cover to due almost cloud-free conditions. Dark blue: open water; light blue: lake ice; grey: clouds.

## 6. IN-SITU MEASUREMENTS

The objectives of in-situ observations are: (a) to evaluate the potential of in-situ observations for ice-on/off periods on Swiss lakes, (b) to retrieve in-situ temperature data to validate other methods, and (c) to characterize criteria for freezing of Swiss lakes as a function of meteorological data and lake characteristics (such as bathymetry, altitude, etc.).

### 6.1. Sensors and data acquisition

Two winter field campaigns were conducted in 2015/2016 and 2016/2017 in a selection of Swiss lakes at high altitudes. In each lake, a mooring was deployed to measure the temporal evolution of the temperature in the water close to the surface. Each mooring was equipped with at least 10 temperature loggers and one pressure sensor (measurement interval 10 minutes). Details of the moorings can be found in Appendix 2.

For ice-covered lakes, traditional moorings usually only reach 1 - 2 m below the surface for safety reason to avoid surface buoy breaking from the pressure induced by ice and mooring breaking from ice drift. The later constraint is especially relevant for large lakes. Yet, the situation is slightly different in high altitude Swiss lakes were lakes are generally small. Hence, we investigated the possibility to measure temperature up to the surface, and thereby provided a better near surface thermal resolution than traditionally observed.

### 6.2. Data processing

#### 6.2.1. In-situ data and processing

The data processing is divided into three phases:

1. Data quality check.
   The data were first quality-checked by standard procedures including basic statistics and visual data inspections. Note that such standard quality check can be automatized and thereby used for real time monitoring
2. Data interpretation
   In this report, we focus on winter limnology with lake temperature ($T$) below the temperature of maximum density ($T \approx 4$ °C). In this situation, the lake is inversely stratified, that is colder water (T < 4 °C) is lighter and thus floating on top of heavier warmer water (Fig. 42a). Given this stratification, the temperature criteria ($T = 0$ °C) for ice formation is limited to a very thin layer at the surface (0 cm) and requires temperature logger as close as possible to the air-water interface. An illustration of the temporal evolution of the near-surface water temperatures is shown in Fig. 43 for Lake Silvaplana. We notice the inverse thermal stratification (T < 4 °C) and temperatures close to 0°C only in the first cm below the surface.

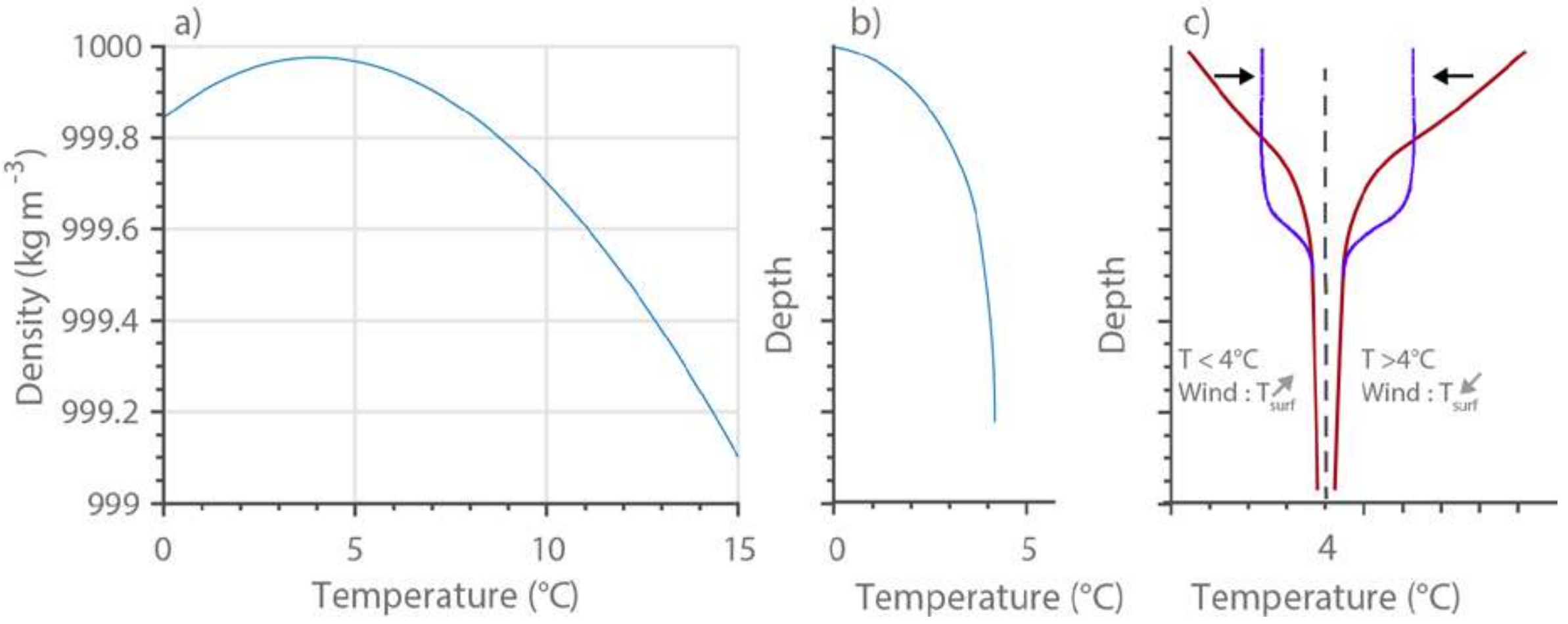

**Fig. 42.** a) Evolution of the density of water as a function of the temperature. b) Typical temperature profile in ice-covered lakes. c) Typical evolution of a temperature profile (red profile changed to blue profile) as a result of a wind event with T > 4° C (right) or T < 4° C (left).

During the ice melting period in late winter, the water temperature is directly warmed by solar radiation penetrating through the ice which leads to a sharp temperature profile between the ice-water boundary at $T$ = 0° C and the radiatively heated layer underneath. Here again, the distance between the temperature logger and the air-ice interface can lead to temperature significantly larger than 0° C (Fig. 42b).

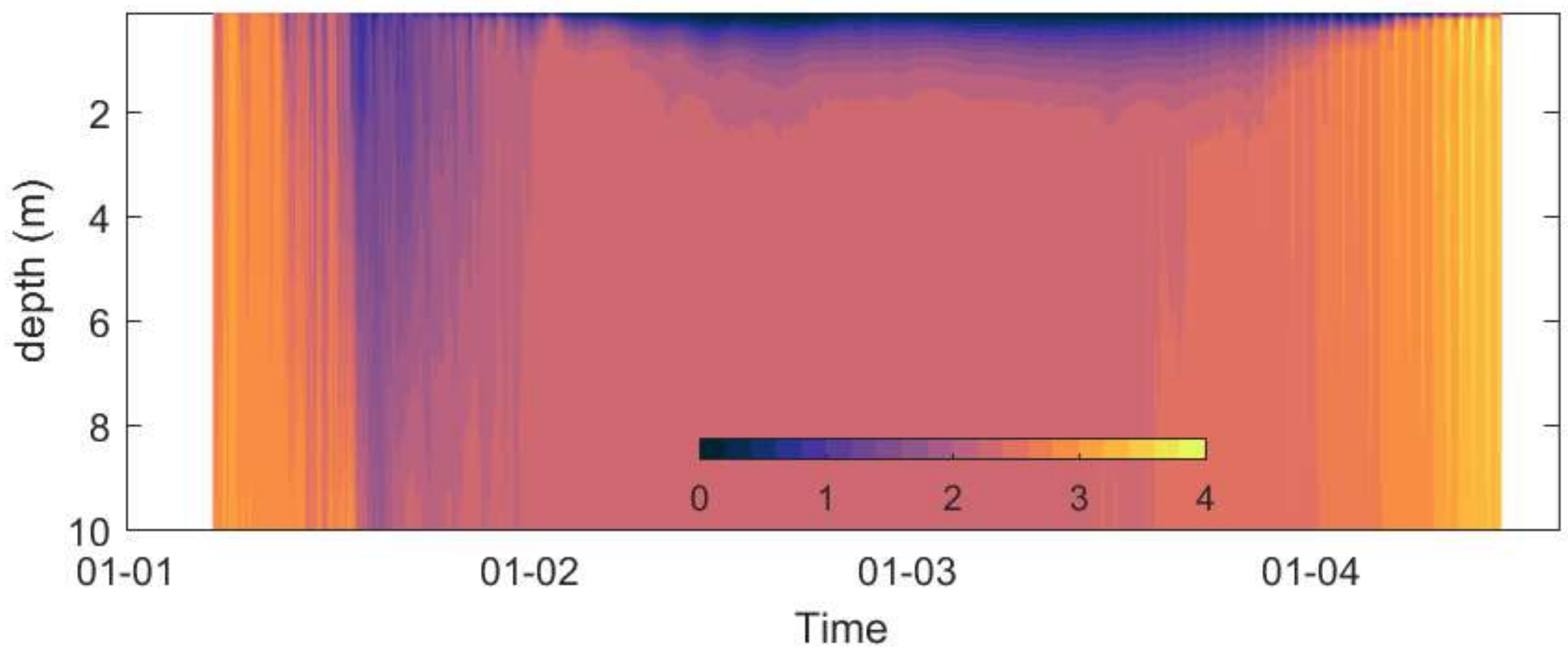

**Fig. 43**. Contour plot of the temporal evolution (dd-mm) of Lake Silvaplana temperature over the first 10 m in winter 2015-2016. The colour bar indicates the temperature (° C).

Despite our attempt to monitor ice-on /off with temperature sensors located as close as possible to the surface, this method was not found reliable as the measured temperature highly depend on the distance from the surface (see Fig.42b). Instead, we analysed temperature time series by investigating the change in the fluctuating response of the time series. For this study, this was done by visual inspection of the time series and manual flagging of the freezing period, yet the method can be automatized. Decision regarding the daily presence or absence of ice cover (as stated in Appendix 6) was taken by specifically looking at the time series at 1 pm local time, that is when the solar radiation is close to its daily maximum. Here, we focused on the sensors located near the surface and compared the relative evolution of the temperatures (typically of the first two or three sensors). The freeze-up time was defined as the time when the temporal evolution between two closely located sensors evolve from a correlated to an uncorrelated pattern. An example is provided in Fig. 44b for the temperature loggers located in Lake Silvaplana at 0.5 and 1.25 m below the surface. Before 20/01/2016, there was a strong correlation between the temperatures measured at the two different depths. On 20/01/2016, the layer at depth 0.5 m starts to respond differently from the layer at depth 1.25 m, which suggested a change in the lake forcing and response. The criteria for break-up were the opposite as for ice formation; the ice-free condition is defined as the time when sensors close to the surface started recording the same temporal pattern and absolute temperatures. An example is provided in Fig. 44c. We observed on 16/03/2016 that the temperatures at 0.5 m and 1.25 m became identical for more than a day. In both cases, the physical argument was that the ice-cover disconnected the lake from the wind and air temperature, thereby strongly reducing the mixing and allowing the formation of vertical temperature gradients in the near surface layer (i.e. change in correlation between near surface temperature sensor).

The dynamical change of the temperature time series was also investigated with wavelet and Fourier transformation analysis to estimate the ice phenology. The physical idea being tested was that the dynamical response of a lake strongly differs between the ice-free and ice-covered periods (Bengtsson, 1996). During ice-free conditions, wind is acting on the lake surface. Briefly, wind will push surface water to the end of the basin (i.e. windward) inducing a horizontal gradient in the water level. Once wind stops, the lake will come back to its equilibrium condition (i.e. horizontal isotherms or surface level) through decaying oscillations around the equilibrium state. This motion is often referred to as surface seiche (Mortimer, 1974). A similar process can be observed in the lake interior in the thermocline (i.e. region of sharp temperature gradient between the warm surface and cold deep layer). By moving water to one end of the basin, wind will push the thermocline down at the windward end (i.e. downwelling) and up at the leeward end (i.e. upwelling). The relaxation leads to oscillations of the thermocline, called basin scale internal waves (or internal seiches; Bouffard and Boegman (2012)). Both periods of surface and internal seiches can be predicted based on the background stratification (i.e. vertical thermal profile) and the mean lake depth and horizontal length (Bouffard and Boegman, 2012). These waves lead to clearly observable oscillations in the temporal evolution of the temperature at a given depth (Fig. 45). A classical way to observe these features is by spectral decomposition of the signal either by Fourier or wavelet analysis. In the case of ice-covered lakes, the water is now disconnected from the main forcing (i.e. wind) and the internal wave response is strongly reduced with nearly no energy embedded into the internal wave field during the ice-covered period compared to the ice-free period (e.g. this lead to a flattening of the internal wave peak in the frequency domain and a strong damping of the energy). The main advantage of this method compared to a direct investigation of the absolute temperature values is that this approach is not sensitive to the location (horizontally and vertically) of the instrument (T- and P- loggers).

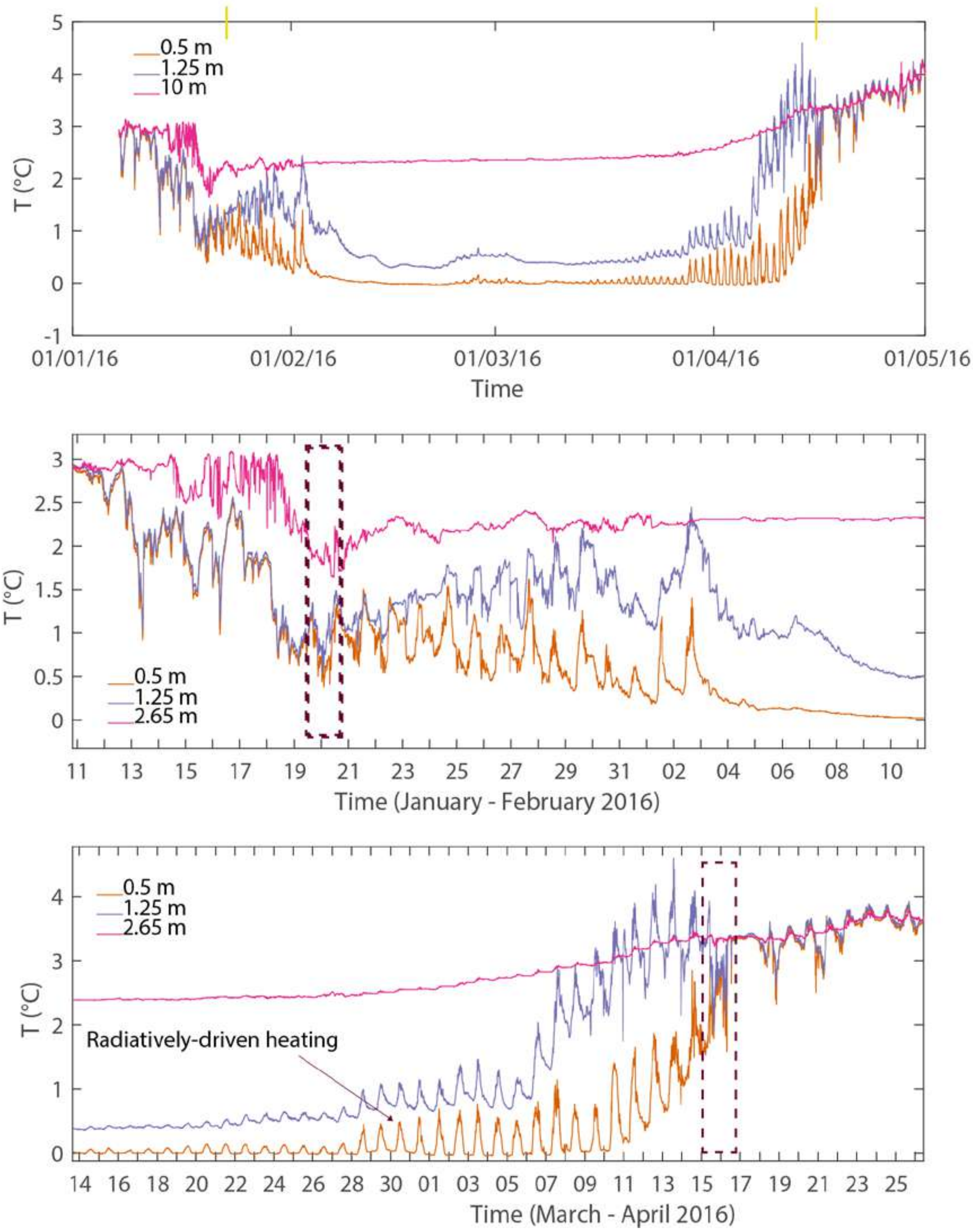

**Fig. 44.** (a) Temporal evolution of the temperature measured at 0.5, 1.25 and 10 m below the surface in Lake Silvaplana (Year 2016). Vertical yellow lines indicate the freeze-up and break-up based on temperature data. b) Temporal evolution of the temperature measured at 0.5, 1.25, and 2.65 m during freeze-up time for Lake Silvaplana. The black dashed rectangle indicates the day of freeze-up. c) Temporal evolution of the temperature measured at 0.5, 1.25 and 2.65 m during break-up time for Lake Silvaplana. The black dashed rectangle indicates the day of break-up.

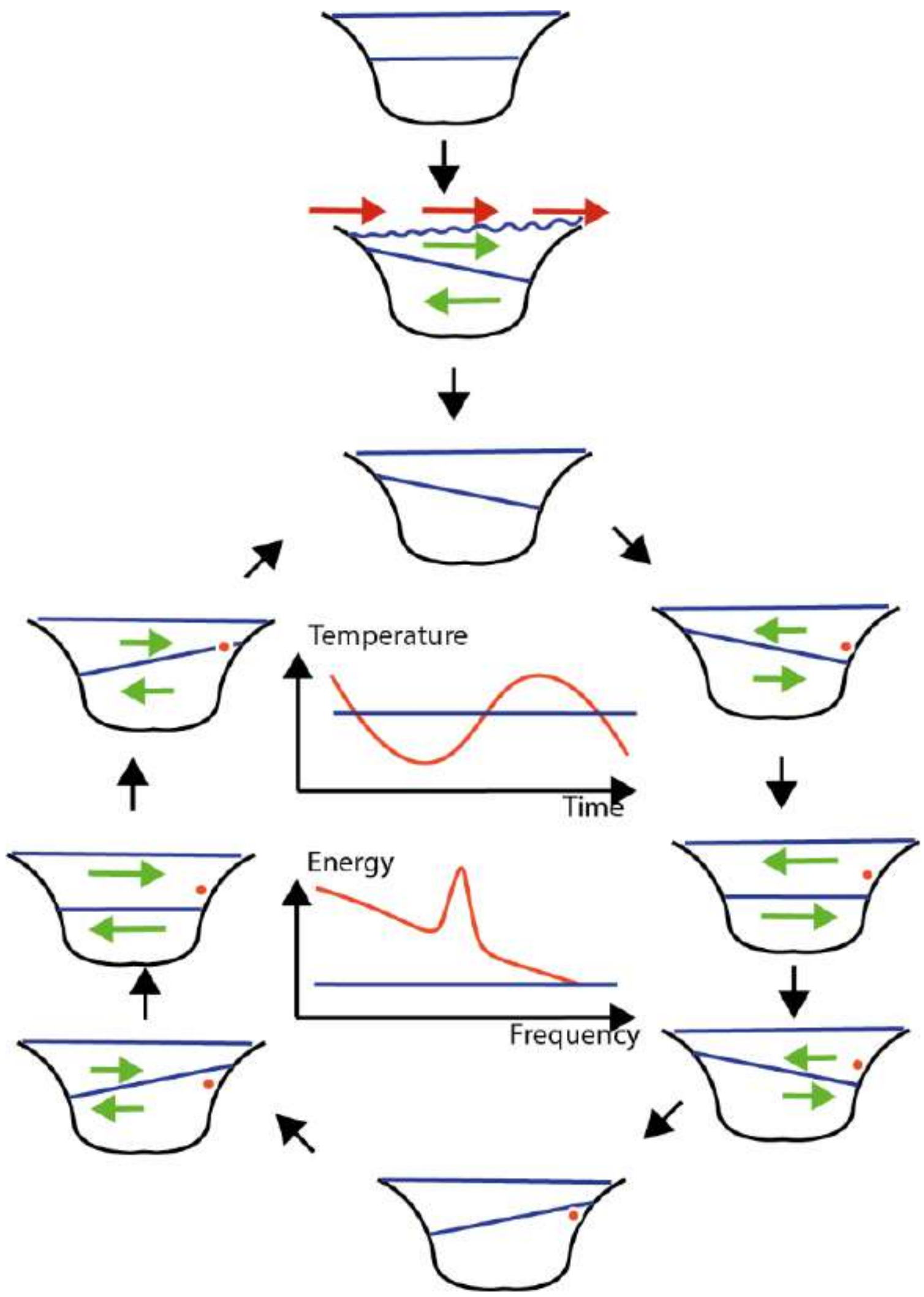

**Fig. 45.** Sketch of the typical lake response of a stratified lake to wind forcing. After calm conditions, the wind (red arrows) is pushing the surface water leeward inducing an upwelling of the air-water interface (up to a few dm) and a downwelling of the thermocline (up to a few m). After the wind ceases, the lake oscillates back to the equilibrium condition (i.e. horizontal isotherms). This process is called internal and surface seiches and lead to clearly visible signals both in the temporal and Fourier (i.e. energy) domains. Under ice-covered conditions, this process is nearly absent and we observe smaller temperature oscillations and weaker energy in the spectral energy domain (blue horizontal line in ice-covered conditions vs. red line in ice-free conditions in the 2$^{nd}$ central panels). The green arrows indicate the main current direction and the black arrows indicate the temporal chronology of the process.

Finally, we used a pressure sensor as another proxy to monitor freeze-up and break-up timing. The physical idea was that the surface buoy will be trapped in the ice, thereby, inducing slight vertical movements of the mooring which should be clearly visible when analysing pressure sensor readings (Fig. 46). The change in the dynamical response of the lake should also hold for the pressure sensor information.

For each method, the ice cover period was manually defined, yet, an automation of the procedure is possible.

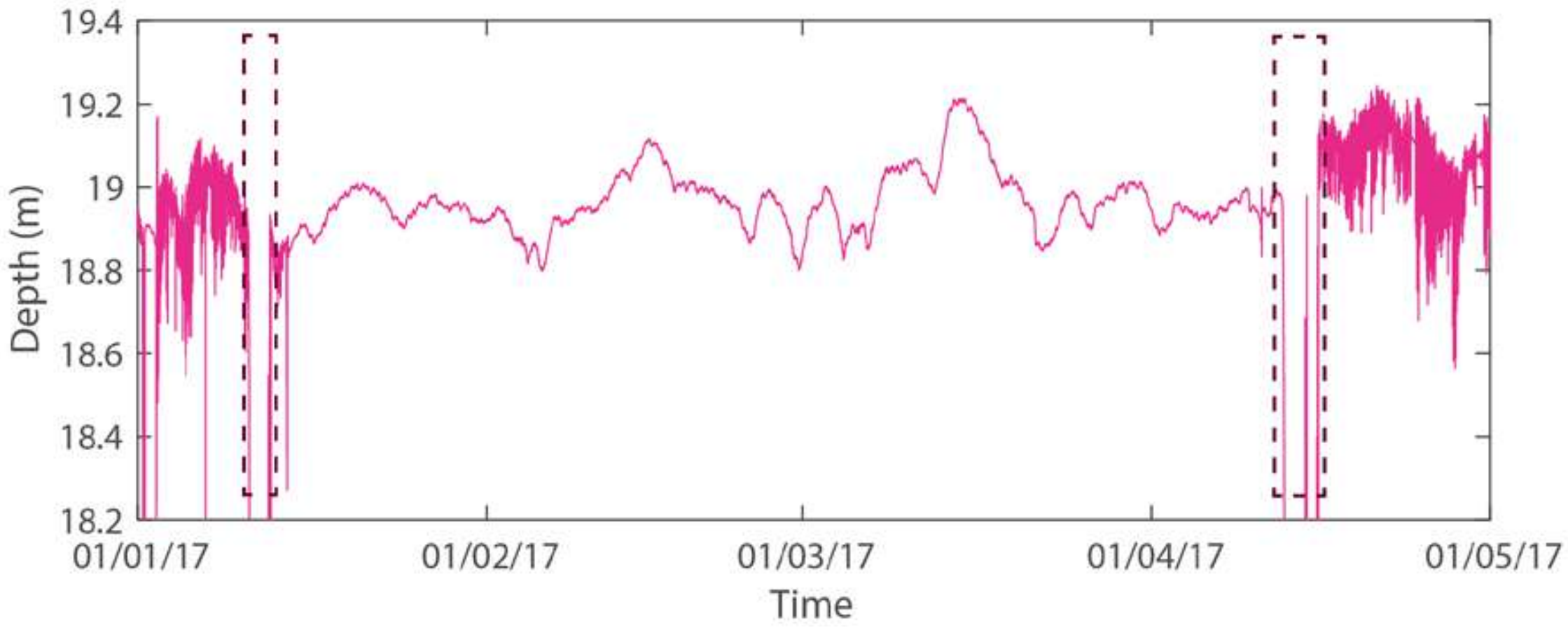

**Fig. 46.** Temporal evolution of a pressure sensor moored in 2017 at Lake Silvaplana at around 19 m depth. The freeze-up and break-up time based on the pressure measurements are shown with the two black dashed rectangles.

### 6.2.2 Ice phenology

Two different modelling approaches were tested in order to evaluate the long-term change of ice phenology and to test the sensitivity of the ice formation and duration on parameters such as meteorological forcing and lake bathymetry.

**Mechanistic model Simstrat.** We used the one-dimensional (1D) hydrodynamic model Simstrat. This model, historically developed at EAWAG, combines a buoyancy-extended k-epsilon model with an internal seiche model to estimate temperature and vertical diffusivities as a function of time and depth (Goudsmit et al., 2002). This model has already been widely used for lakes (Schmid et al., 2014; Schwefel et al., 2016; Gaudard et al., 2017) and requires information on lake bathymetry and external forcing, including meteorological data (i.e. air temperature, short wave radiation, long wave radiation, humidity and wind) and river inflows (discharge and temperature/density). This model was recently updated to simulate ice formation and decay. The model was calibrated with historical and newly collected in-situ temperature data by using a parameter estimation package (PEST, http://www.pesthomepage.org/) (Table 22).

**Table 22.** EAWAG Simstrat Model. For each lake, a calibration period as well as two statistical parameters are indicated.

| Lake | Meteorological data | Calibration period | $R^2$ | RMSE |
|---|---|---|---|---|
| Lake St. Moritz | Samedan SAM | 2006 - 2008 | 0.93 | 0.75 °C |
| Lake Silvaplana | Segl-Maria SIA | 2016 - 2017 | 0.96 | 0.84 °C |
| Lake Sils | Segl-Maria SIA | 2016 - 2017 | 0.92 | 0.98 °C |
| Lake Sihl | Einsiedeln EIN | 2014 - 2017 | 0.95 | 1.1 °C |

**Hybrid model air2water.** While mechanistic models, such as Simstrat, have to resolve the entire heat budget at the air-water interface, thereby requiring many parameters of the local meteorology, a physically-based statistical model (such as Air2Water) is used in order to estimate the surface lake temperature as a function of the air temperature (Piccolroaz et al., 2013). The model expressed as a differential equation indirectly takes other meteorological parameters into account through lake-calibrated parameters. This model was recently extended to take into account ice formation in winter. The main advantage of this model is its simplicity (i.e. relying only on one external forcing parameter) and once calibrated, the model is ideal for rapidly investigating long time series. This part was conducted in collaboration with University of Trento (M. Toffolon).

## 6.3. Results and discussion

### 6.3.1. In-situ temperature data

Here, the timing for ice-on and -off for the selected lakes was defined by visual inspection of the correlation in the temporal evolution of the temperature time series close to the surface. Direct measurement of the surface temperature would be the first obvious method to track ice formation. Yet, this method is not always practical as it is very difficult to keep buoys right at the water surface when freezing. In large lakes where strong wind can be observed (as for instance for lakes Sils, Silvaplana and St. Moritz in the Engadin Inn valley), ice movements may break the mooring and only subsurface moorings are deployed making it impossible to monitor close to surface temperature. In smaller sheltered lakes, the buoys may be slightly lifted up during the freezing process, thus slightly altering the vertical sensor location. Furthermore, the local heating may be modified in the top cm by the buoy (not investigated). Moreover, the inverse thermal stratification makes the temperature at a given depth "disconnected" from the surface temperature. We suggest that instead of the absolute temperature value, a practical way to estimate the timing for freeze-up and break-up of ice is to consider the change in the temporal correlation in the temperature patterns over temperature sensors located at different depth close to the surface (Fig. 44 see the change in correlation between the sensors before 19/01 and after 21/01). Results are provided in Table 23 and an example is provided in Fig. 44. The accuracy of the method was visually estimated to be ~2 days (see for instance dashed black box in Fig. 44), yet, this method is local and represents the lake conditions in the vicinity of the mooring.

Advantages:

- Cheap method requiring 2 to 3 temperature sensors vertically mounted near the surface.
- Accuracy: (~2 day) to be further investigated.

Disadvantages:

- Sensors have to be located in the first meter to the surface. This typically requires a surface buoy.
- Due to the inverse temperature stratification, absolute values of temperature depend on the distance between the surface and the sensor. A precise distance is between sensors and surface is practically difficult to achieve.
- With this method, the information regarding ice coverage remained local (the spatial extent was not investigated.

### 6.3.2. Dynamical analysis of in-situ time series

We then test another process-based approach to use the temperature sensors for lake freeze-up and break-up detection. A common characteristic of temperature sensors is the nearly flat response (i.e. no temperature change) during the ice-covered period compared to the ice-free season (Fig. 44a). We have previously investigated the dynamical change of two closely located temperature sensors. Another method consists in analysing the information embedded into the dynamic of temperature fluctuations of a single logger.

The driving hypothesis is that there is more energy related to internal waves in the ice-free period compared to the ice-covered period. Practically, this hypothesis can be tested by looking at the evolution of the energy of the Fourier transformation or by wavelet analysis. We illustrate this by investigating the dynamic response of the temperature sensor located 10 m below the surface by means of wavelet and Fourier transform analysis (Figs. 47 and 48). From Fig. 47 we notice that there is a peak in energy in the band log (period) of ~1.5 (~30 h) at the beginning and end of the winter and that this frequency contains less energy during the ice-covered period. Similar results can be seen in Fig. 48 with the temporal evolution of the energy embedded at this period. We defined a criterion for freeze-up and break-up time as follows. For freeze-up, we first detect a drop of energy by at least two orders of magnitude. The freeze-up time is defined as the beginning of the drop in energy. The break-up time is defined by taking the mean between the background ice-free and ice-covered conditions. Results are provided in Table 23. Yet, there is still a significant uncertainty regarding the timing for the lake freeze-up and break-up. Further studies will help defining more accurate criteria and automatize the procedure, however, the damping of energy in a lake is not instantaneous and, as a rule-of-thumb, can be expressed as a function of the mean lake depth (damping time scale ~1 day per 20 to 30 m). Yet, this is robust for assessing whether a lake was ice-covered or not during a winter. This can, additionally, be calculated from data of any temperature (or other parameter) sensors moored at any location on a lake. A decisive advantage of this second approach is that this method is not sensitive to the horizontal and spatial location of the sensors.

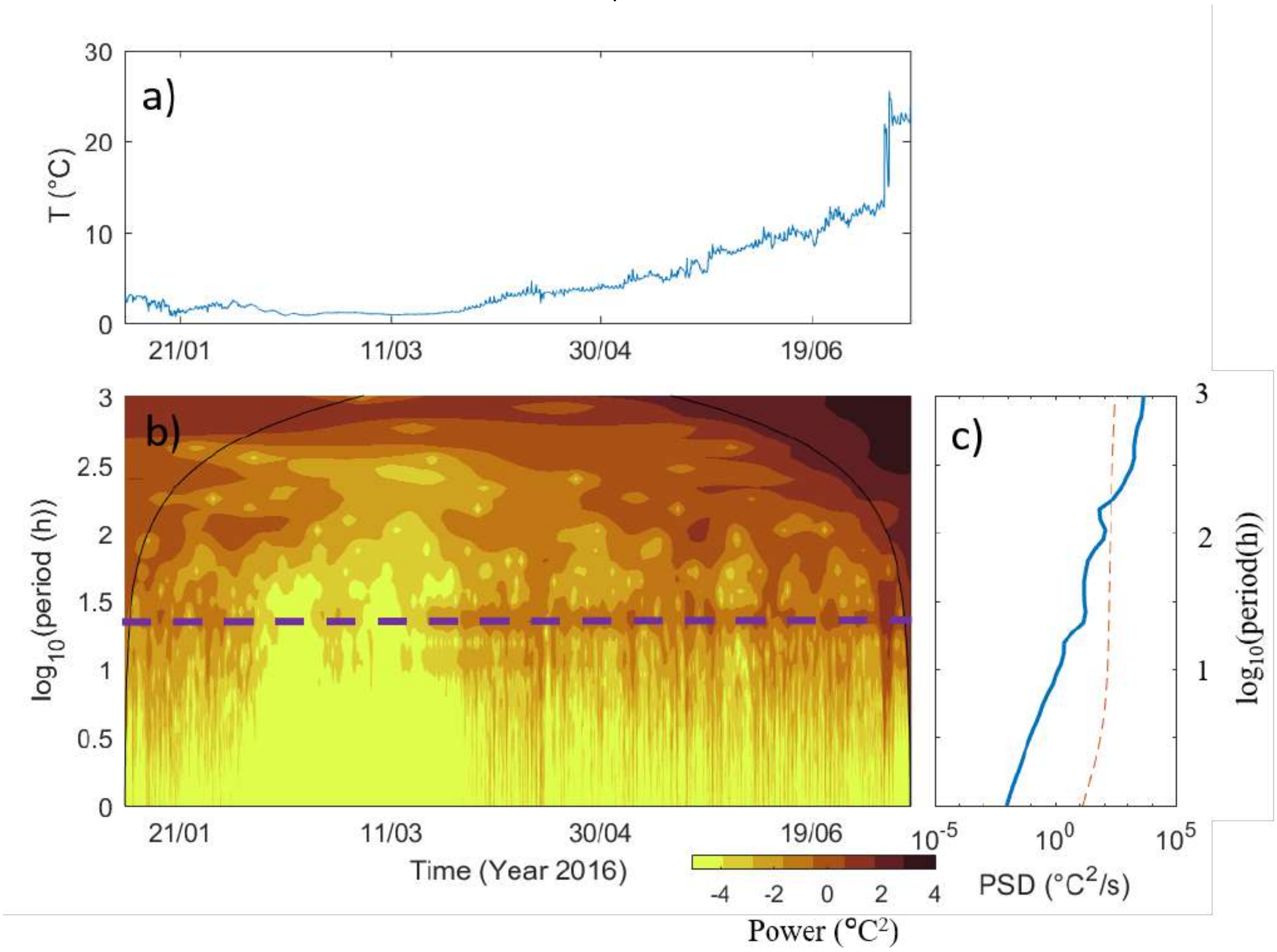

**Fig. 47.** a) Time series of the temperature recorded at 50 cm below the surface of Lake Silvaplana for the first half of 2016. b) Wavelet analysis of the time series presented in a) representing the temporal evolution of the intensity of the energy (colour-coded in log-scale) embedded at a given frequency (or period) with valid values underneath black u-shaped line. c) Power Spectral Density transformation of the time series shown in a) representing the average energy embedded at different periods over the entire season. We notice a peak at log(T) ≈ 1.5 (see dashed line in b) and a small peak in c)). From b) we see that this peak is strong (dark colour) in the ice-free period and nearly absent (light colour) during the ice-covered period.

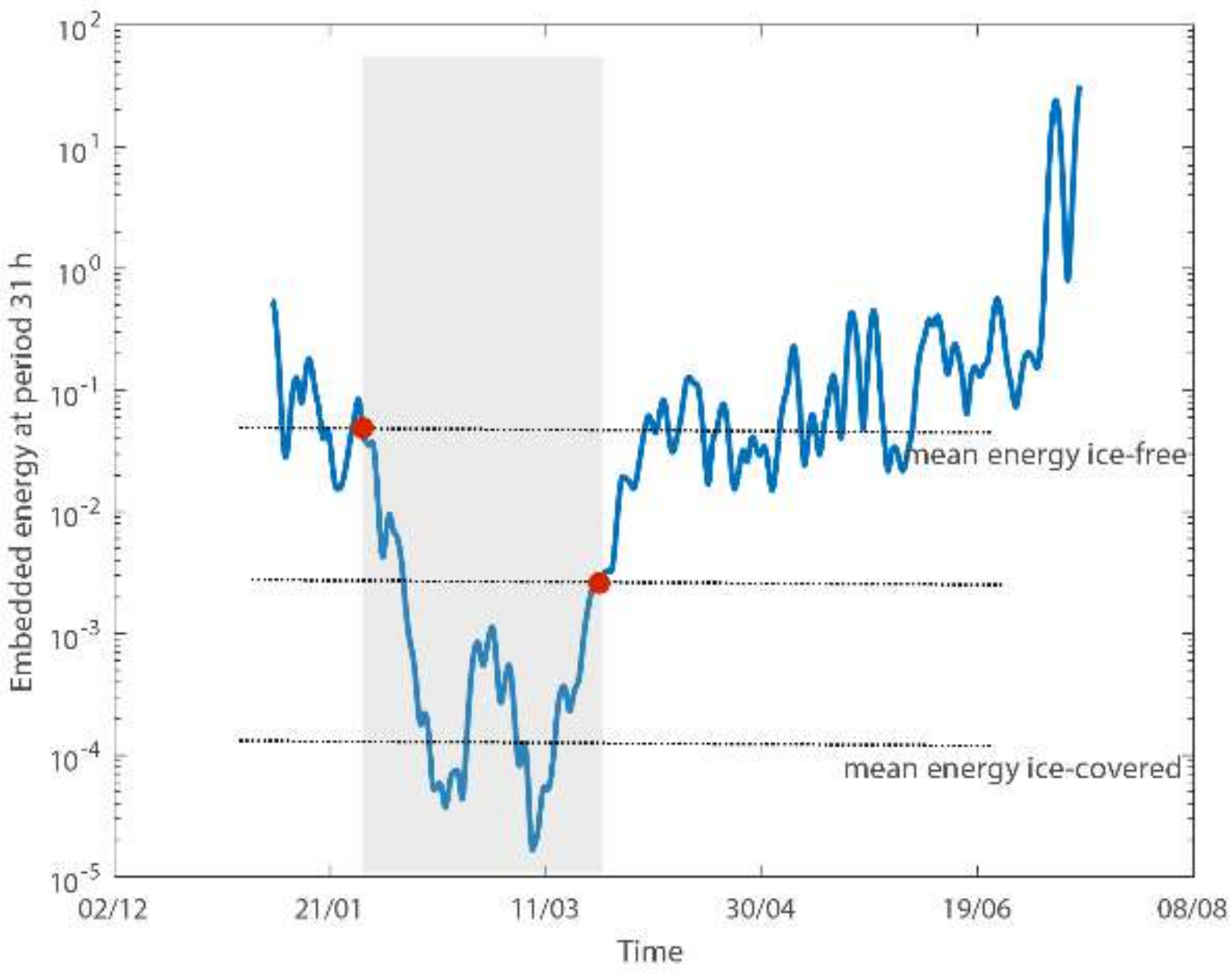

**Fig. 48**. Temporal evolution of the energy embedded at a given period for Lake Silvaplana (2016). The drop in energy suggests the absence of forcing of the internal waves, which is due to the ice-cover at the lake surface preventing the wind to energize the lake water. The grey area represents the estimated ice-covered period (bounded by the red dots).

We finally define a conceptual flowchart for a potential automation of the method in Fig. 49, which is of potential interest for future ice phenology monitoring.

Advantages:

- Basin-scale method. This method is based on the whole lake dynamic and not on the specific sensor location.
- Can be applied to any sensors moored in a lake.

Disadvantages:

- The ratio between the energy in the ice-free and ice-covered period has to be further investigated.
- Energy transfer and damping are not immediate; this method comes with an uncertainty proportional to the mean lake depth.

### 6.3.3. Pressure sensors

We finally tested the potential of using a pressure sensor to estimate lake freeze-up and break-up dates. The physical argument is that with ice formation or breaking, the in-ice trapped surface buoy will slightly move vertically and the resulting abrupt displacement will be recorded by the pressure sensor. An illustration is shown in Fig. 46 and results are presented in Table 23. In most cases, the pressure sensor could very accurately provide information regarding the ice-on and -off timing. Yet, we also observed a few occurrences were no abrupt pressure changes were detected. This method seems to suffer from reliability but is an excellent complimentary source of information.

Advantage:

- Provide a clear signal with an abrupt change from high frequency to low frequency fluctuations.

Disadvantage:

- The results depend on how the mooring was setup (surface or subsurface buoy).

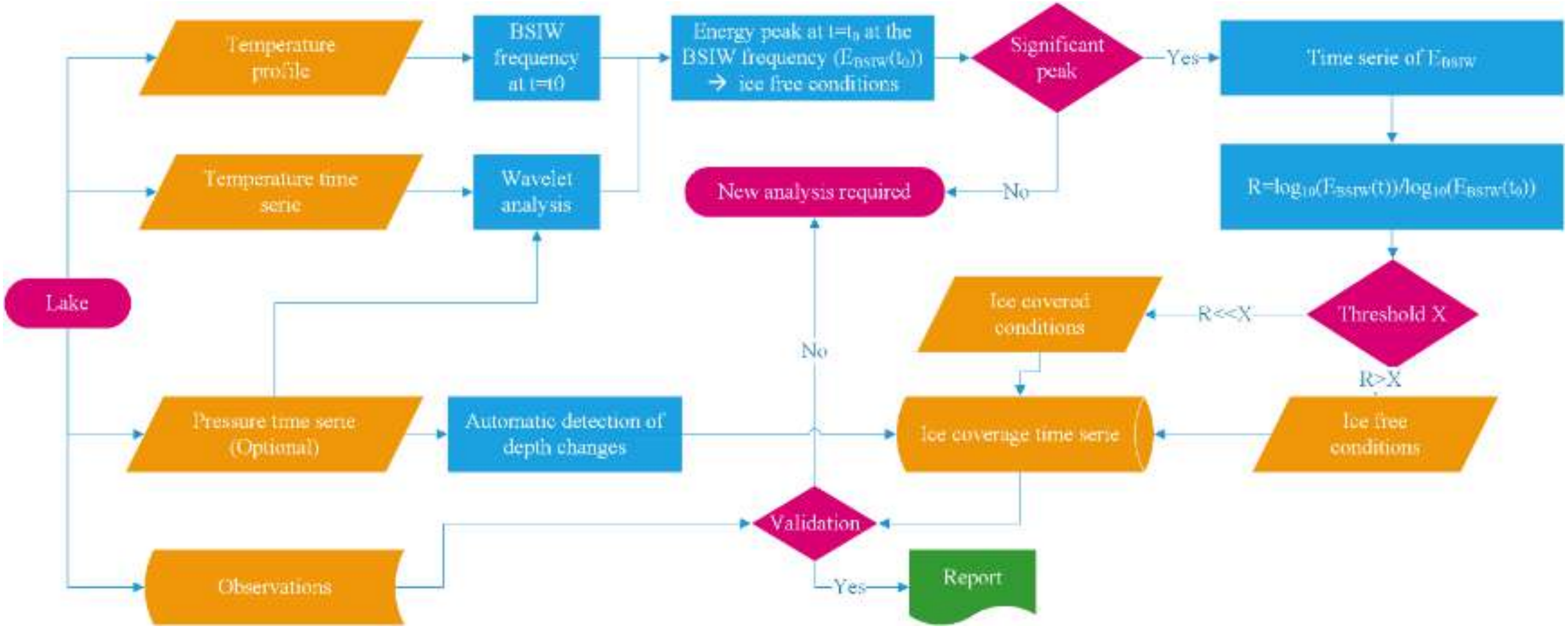

**Fig. 49**. Flowchart for in-situ-based automated lake ice monitoring.

**In-situ method comparison**

We did not find a perfect method allowing for basin scale estimate of the ice cover period with reduced uncertainty (< 2 days). Yet, the combined use of the two first methods (see 6.3.1 and 6.3.2) can provide a way to achieve a global-scale representativeness of the coverage (1 sensor dynamical-based approach, see 6.3.2) and a reduced temporal uncertainty (two-sensors approach, see 6.3.1). Finally, as illustrated by the use of the pressure sensors (see 6.3.3), other sensors can also be used to track the ice-on/off period.

### 6.3.4. Ice phenology

Magnuson et al. (2000) analysed freeze-up and break-up dates for 20 lakes with time series longer than 100 years around the Northern Hemisphere. They showed a significant decay of the ice-covered period with, on average, a later freeze-up (0.58 day per decade) and an earlier break-up (0.65 day per decade). Yet, this trend observed for the period 1845-1995 has increased over the recent decades. Franssen and Scherrer (2008) investigated the ice phenology of 11 Swiss lakes since 1901. They observed a significant reduction of the frequency of the ice-cover events over the last two decades (1986-2006). They also correlated the ice formation to the sum of negative degree days (NDD) (from October to April) and the mean lake depth. Brown and Duguay (2010) reviewed the ice phenology on lakes. They showed that lake depth determines the amount of heat storage in the water, and hence, the time needed for the lake to cool and freeze. The break-up process is found to be more lake independent and mostly controlled by air temperature (Jeffries and Morris 2007, Weyhenmeyer et al., 2004). Martynov et al. (2010) evaluated the sensitivity of lake ice models and concluded that an increase in the ice/snow albedo postpones the break-up date. Magee et al. (2016) conducted a modelling study based on 104 years of ice-cover and temperature observations. They found that ice-cover is only weakly related to the seasonal average wind speed. Nevertheless, the correlation between ice-on date and wind speed is significant, indicating a relationship between decreasing wind speed and earlier ice-on dates. Strong wind events prevent ice formation by disturbing the inverse thermal stratification. Reduced wind speeds allow ice-cover to form slightly earlier. For both ice-on and ice-off, the air temperature has a more significant relationship than wind speed. Finally, Magee et al. (2016) suggested that ice-cover in shallow lakes may be more resilient to climatic change than deep lakes experiencing same forcing.

Ice phenology was investigated through a modelling study. We focused on a sensitivity analysis on the freeze-up by investigating the change in lake depth, air temperature and wind stress. The break-up timing being less dependent on the lake characteristics and more on air temperature and albedo is not investigated here (no information regarding albedo). From a previous study in an ice-covered lake in Russia we observed a change from 0.4 to 0.7 in albedo depending on the day-time (i.e. ice/snow melting) and on the presence/absence of snow (Bouffard et al., 2016).

The cooling period has to be split into two periods. In the first part (i.e. typically September to December), the lake temperature is above $T_{MD}$ and both wind and cooling (i.e. $T_{air} < T_{surf}$) act to reduce $T_{surf}$. For $T_{surf} < T_{MD}$, the cooling will still extract heat from the surface and thereby reduce $T_{surf}$ but the wind will now mix colder surface water with warmer deeper water (i.e. inverse thermal stratification) thereby warming the surface layer. The first period (i.e. until $T_{surf} = T_{MD}$) will increase in duration in a context of global warming. We focus here on the second period.

We define τ as the duration of the second period given by $0 \leq T_{surf} \leq T_{MD}$ (Fig. 50, inset). Simstrat was calibrated based on collected temperature measurements for Lakes Silvaplana, St. Moritz, Sils and Sihl. Details regarding the modelling are provided Appendix 2. For these lakes, the response to changes in air temperature and wind intensity was investigated by increasing and decreasing the mean measured air temperature by ±0.5, ±1, and ±2°C and wind intensity by ±10%, ±30%, and ±50% (Fig. 50). An increase of the mean air temperature or of mean wind speed leads to an increase of τ, i.e. a delay in the freeze-up time. Deep lakes are slightly more sensitive to the increase in air temperature and wind intensity (shallow Lake Sihl compared to deeper Lake Sils and Lake Silvaplana). We finally applied the same meteorological forcing on a hypothetical lake (Lake Silvaplana taken as reference) with changing depth. This numerical study confirms the stronger resilience of ice formation to increased air and wind intensity in shallow lakes.

The modelling study using a mechanistic hydrodynamic model allows assessing the role of internal (i.e. lake) and external (i.e. atmosphere) parameters in ice formation. We show that increased wind or air temperature will delay the time for lake freeze-up. Given the larger stored heat in deep lakes, they are more affected by such meteorological changes. The break-up timing depends on the albedo and the air temperature. The former information being lacking, this timing was not numerically investigated.

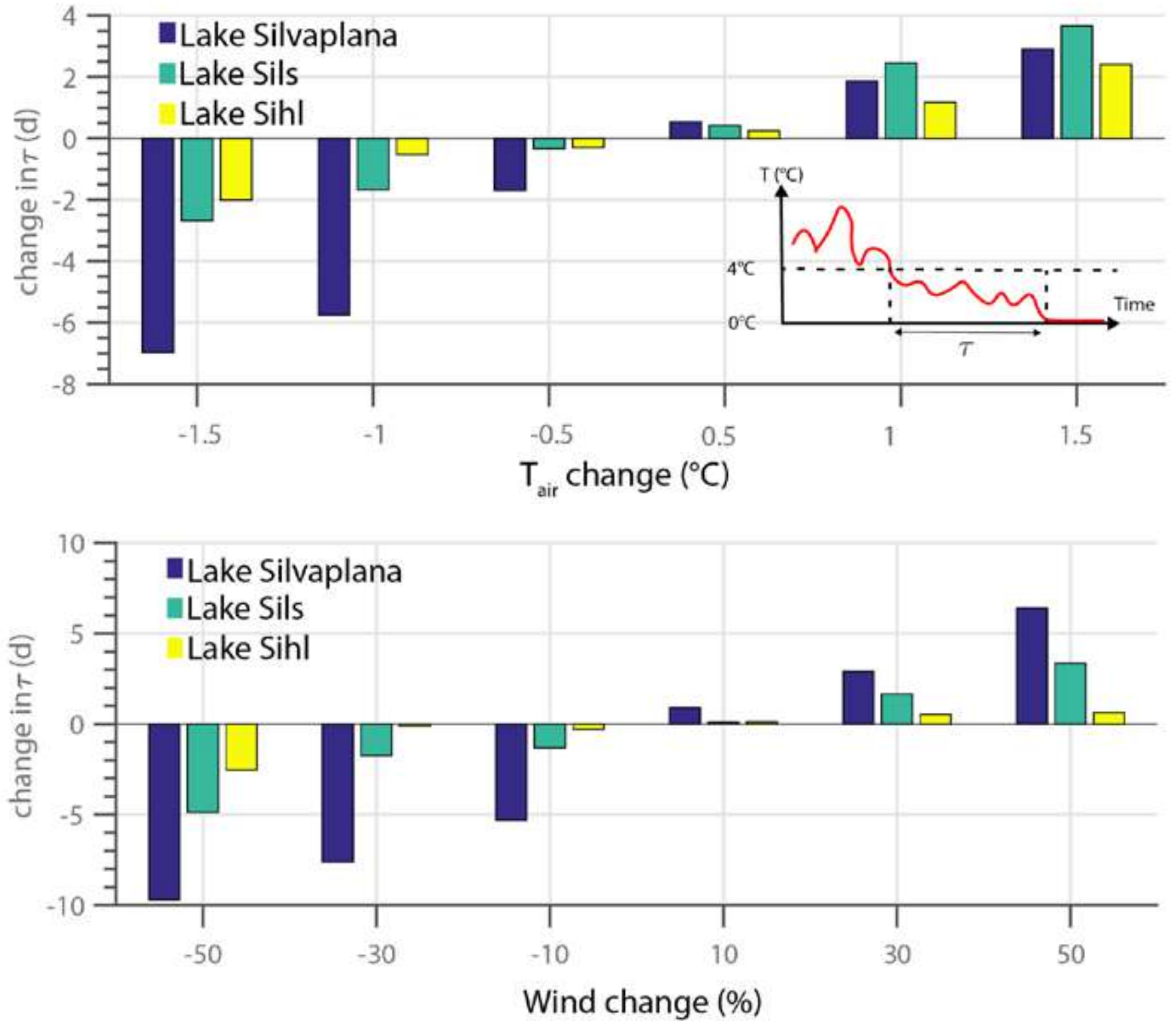

**Fig. 50.** Variations of the freeze-up timing for three lakes due to a change in the mean air temperature (a) or wind intensity (b) for the period 0° C < $T_{surf}$ < 4° C.

Finally, we have only investigated the response in the freeze-up timing based on a change of the mean of air temperature and wind intensity, yet, a similar study should be conducted to assess the role of single storm or cold events during the period of $T_{surf}$ < 4° C.

Advantages:

- Ideal for process-based investigation of lake-ice phenology.
- Hydrodynamic models can be run in operational mode (i.e. short time prediction if coupled to a meteorological forecasting model like COSMO-E or COSMO-1). See also simstrat.eawag.ch.

Disadvantages:

- Requires meteorological data near the lake.
- Information regarding the temporal evolution of the albedo needed.
- Lack of in-situ data to improve the model.

Finally, the global change in the ice phenology is investigated with the hybrid air2water model (including air2ice, collaboration with University Trento, Italy, M. Tofollon, S. Piccolroaz and L. Cortese). The study is conducted on Lake Sihl with a calibration over 3 years (Fig. 51) and a total simulation over 27 years (Fig. 52). The model does not show any significant change in the ice phenology for this lake over the last 27 years (Fig. 53). This result seems in contradiction with literature, yet, agrees with the absence of changes in the timing of ice thickness measurements over the lake. More observational data are needed to fully validate such hybrid models. Yet, the results highlight the potential of this simple model for global climate change studies (i.e. long-term simulations on many lakes using different climate scenarios, such as the CH2018 climate changes scenarios).

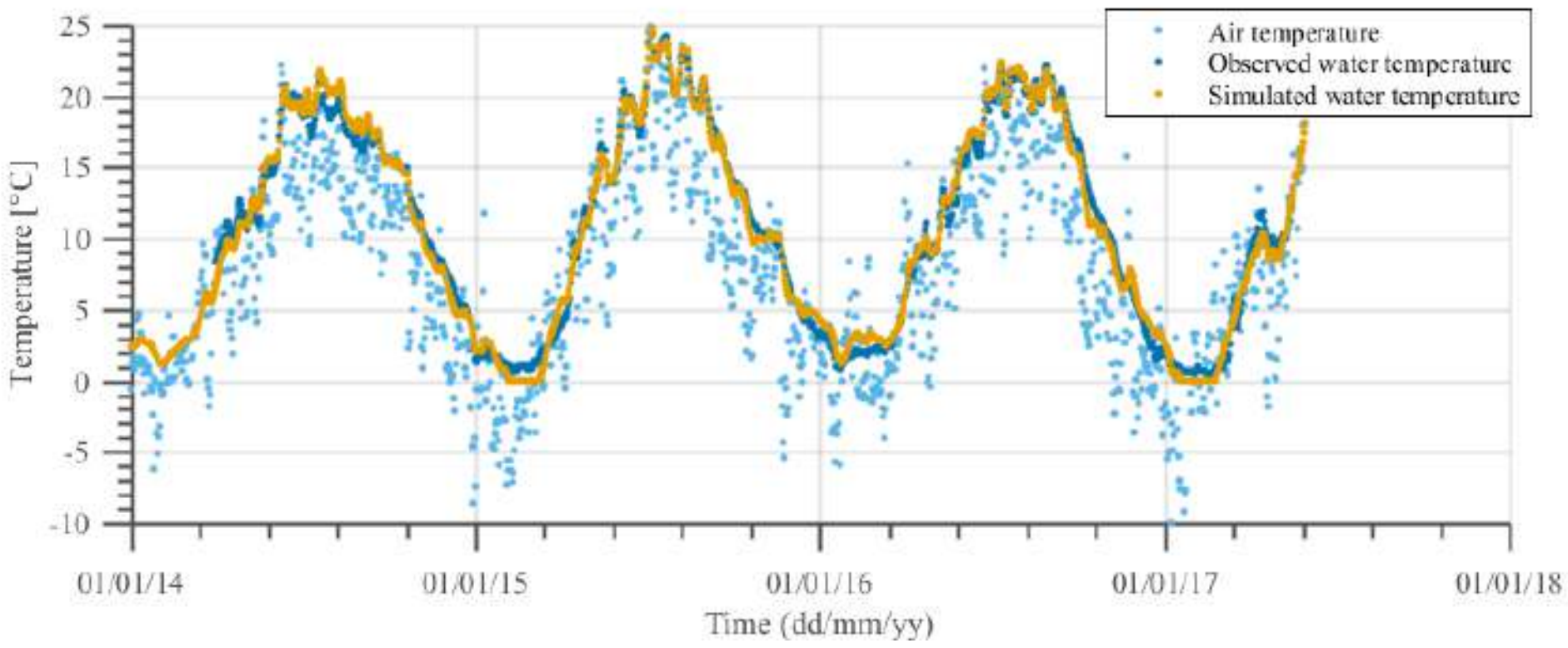

**Fig. 51.** Calibration of air2water (and air2ice) over three years for Lake Sihl. Collaboration with University Trento.

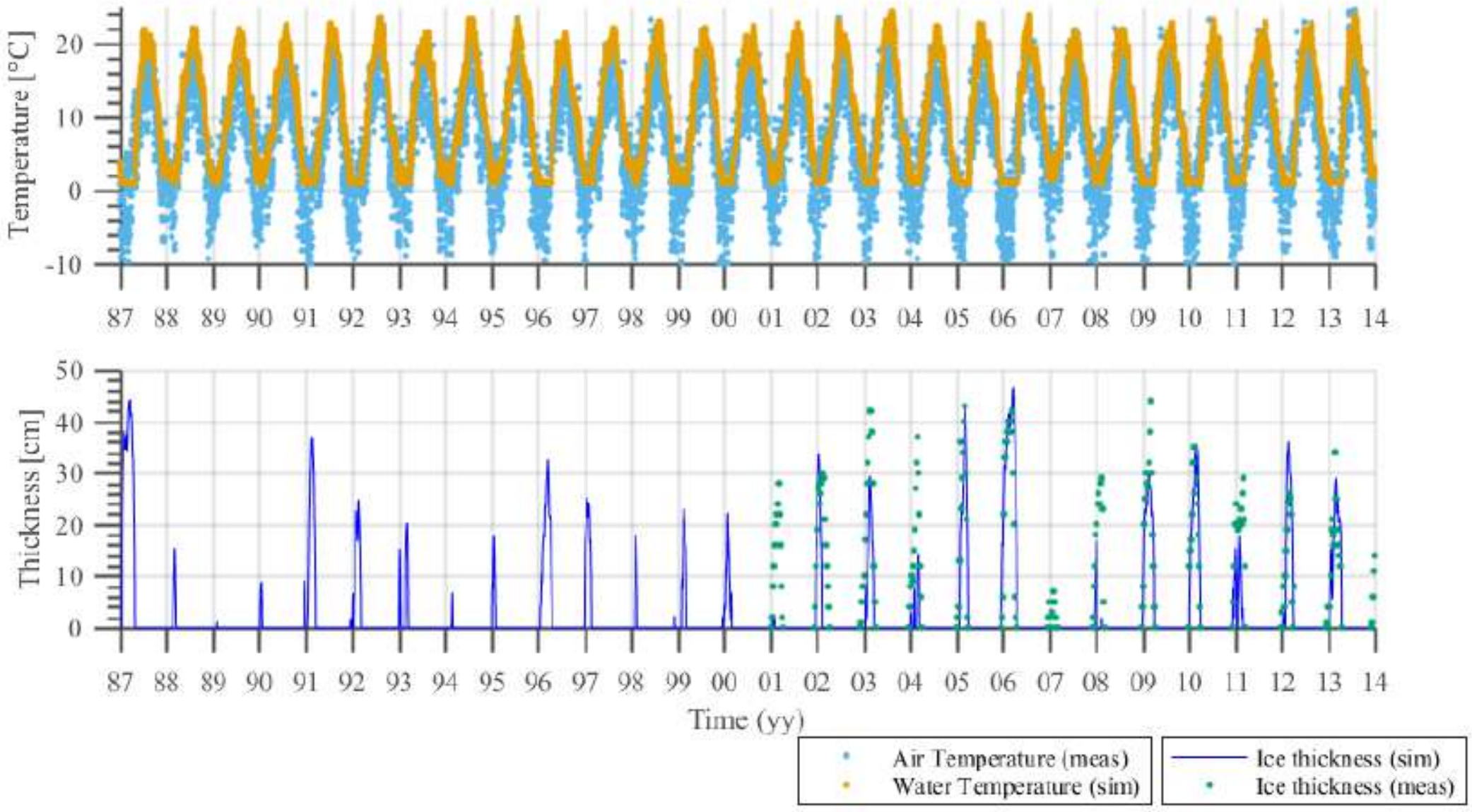

**Fig. 52.** a) Measured air temperature and modelled water temperature for Lake Sihl over 27 years. b) Lake ice-cover phenology projected with air2ice (blue line) and observed by the SBB AG. Collaboration with University Trento (green points).

Advantages:

- Ideal for Swiss, Europe or worldwide-based climate change scenarios.
- Limited input data (e.g. air temperature, widely available)

Disadvantages:

- So far, only preliminary results. Current results have to be taken with caution.
- More in-situ data needed for further validation of the model.

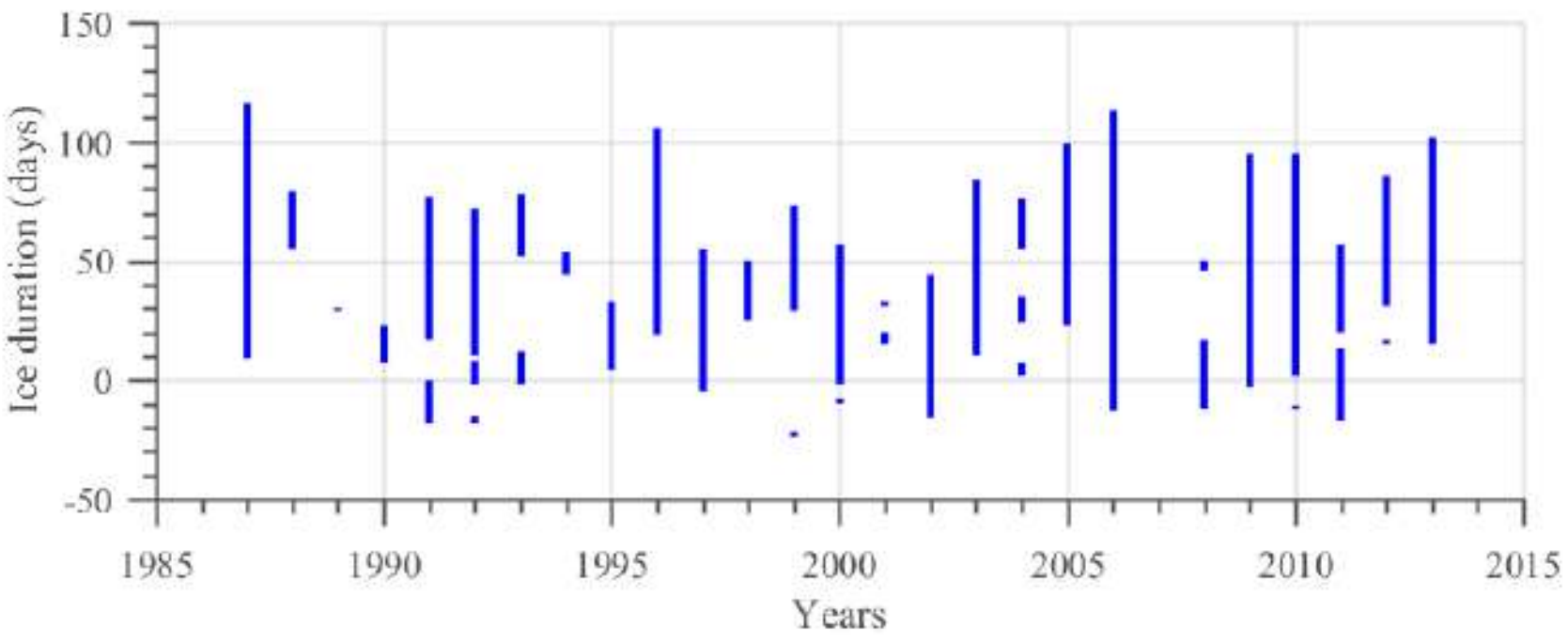

**Fig. 53.** Temporal evolution of the ice duration calculated from the modelling study over 27 years for Lake Sihl.

**Recommendations.** The main problem with in-situ observations of lake freeze-up and break-up is that the 0°C temperature criteria is only reached and maintained in a very thin layer, which is difficult to probe near the surface. Three process-based methods were tested (1) observation of the correlation between different temperature sensors, (2) changes in the energy content in the lake estimated from spectral analysis of temperature sensors and (3) information of the freezing of the surface buoy observed with pressure sensor. While the methods (1) and (3) provide accurate results, they require a certain level of expertise in time series analysis. The method (2) has to be further assessed and is slightly more complicated but can be automatized. Our recommendation is to combine the different methods. The integration of in-situ data into a mechanistic model can also provide short time forecasting capabilities regarding the ice conditions while long time series can be used to validate hybrid models and evaluate the effect of climate change on ice phenology. Yet, it is not realistic to monitor all lakes in-situ and other integrative approaches have to be considered. Further collaborations with the Swiss Federal Office of Environment (FOEN) regarding lake monitoring may offer interesting perspectives. In monitored lakes with high frequency measurements (> 1 measurement per hour) the above-presented methods including in-situ and model-based methods should be further developed and used. For other lakes, simulation with a hydrodynamic model could offer an interesting alternative. In principle, two or three years of data are needed to validate a hydrodynamic model. A calibration of Swiss lakes could be done with a reduced pool of sensors and human resources (one or two lake visits per year for sensor maintenance and data collection) over a couple of years. Satellite and Webcam data can also be used to validate these models. Models could finally be used to interpolate results in between two satellite temperature observations or unclear Webcam images.

**Table 23.** Ice-on/off periods estimated based on the three proposed methods. “X” indicates that the method was not able to provide an estimate of the timing.

| | Temperature-based method (Section 6.3.1) | Dynamic-based method (Section 6.3.2) | Pressure-based method (Section 6.3.3) |
|---|---|---|---|
| Lake Silvaplana 2016 | 20/01 – 15/04 | 19/01 – 16/04 | - |
| Lake Silvaplana 2017 | 14/01 – 14/04 | 12/01 – 13/04 | 14/01 – 15/04 |
| Lake St. Moritz 2017 | 17/12 – 05-08/04 | X - 05/04 | 15/12 – 08/04 |
| Lake Sils 2017 | 31/12 – 10/04 | 28/12 – 08/04 | 31/12 – 10 - 12/04 |
| Lake Sihl 2016 | 09/01 – 16/03 | X | X – 03/02 (potentially false signal) |
| Lake Sihl 2017 | 28-29/12 – 16/03 | 27/12 – 16/03 | X – 16/03 |

## 7. WEBCAM DATA

### 7.1. Data Collection

The Webcam images were downloaded from the following two Websites: https://www.webcam-4insiders.com and https://en.swisswebcams.ch/. For the season 2016/2017, the collection of images capturing all six target lakes was started 4th of December 2016 (except Lake Silvaplana which was started on 7th of December 2016 and the additional Lake Pfäffiker which was started 28th of January 2017). Due to time-limited scripts which required manual restarting after some days and some overlapping with travel and holidays, data was not recorded in the periods: 24th-28th December2016 (5 days), 24th-25th January 2017 (2 days), 22nd-25th February (3 days) and 18th March 2017 to 25th April 2017 (39 days). For the winter season, 2017/2018 data collection started between 29th November and 4th December. As of the time of this report writing, recording is still on-going. Due to improvements of the scripts the images of the complete freezing period were obtained fully automatically without any mayor data gaps. However, for some of the cameras few days are missing due to technical problems on the operators' side. Additional images monitoring Lake Sihl was provided by EAWAG. In 2015/16 EAWAG installed Webcams (Brinno TimeLapse HD Video Camera TLC200, 5fps, 720p) at two locations: Etzelwerk (47.154433, 8.779809) and Wohnhaus (47.146778, 8.757275). Furthermore, in 2016/17 EAWAG installed only one camera at Etzelwerk.

Overall one to six publically accessible Webcam were found for each of the lakes. The lake coverage of the Webcams is crucial for the task at hand. To get an overview, we manually checked each of the Webcams and plotted the approximate lake coverage to orthoimages derived from Google Earth. An example for Lake St. Moritz is shown in Fig. 54.

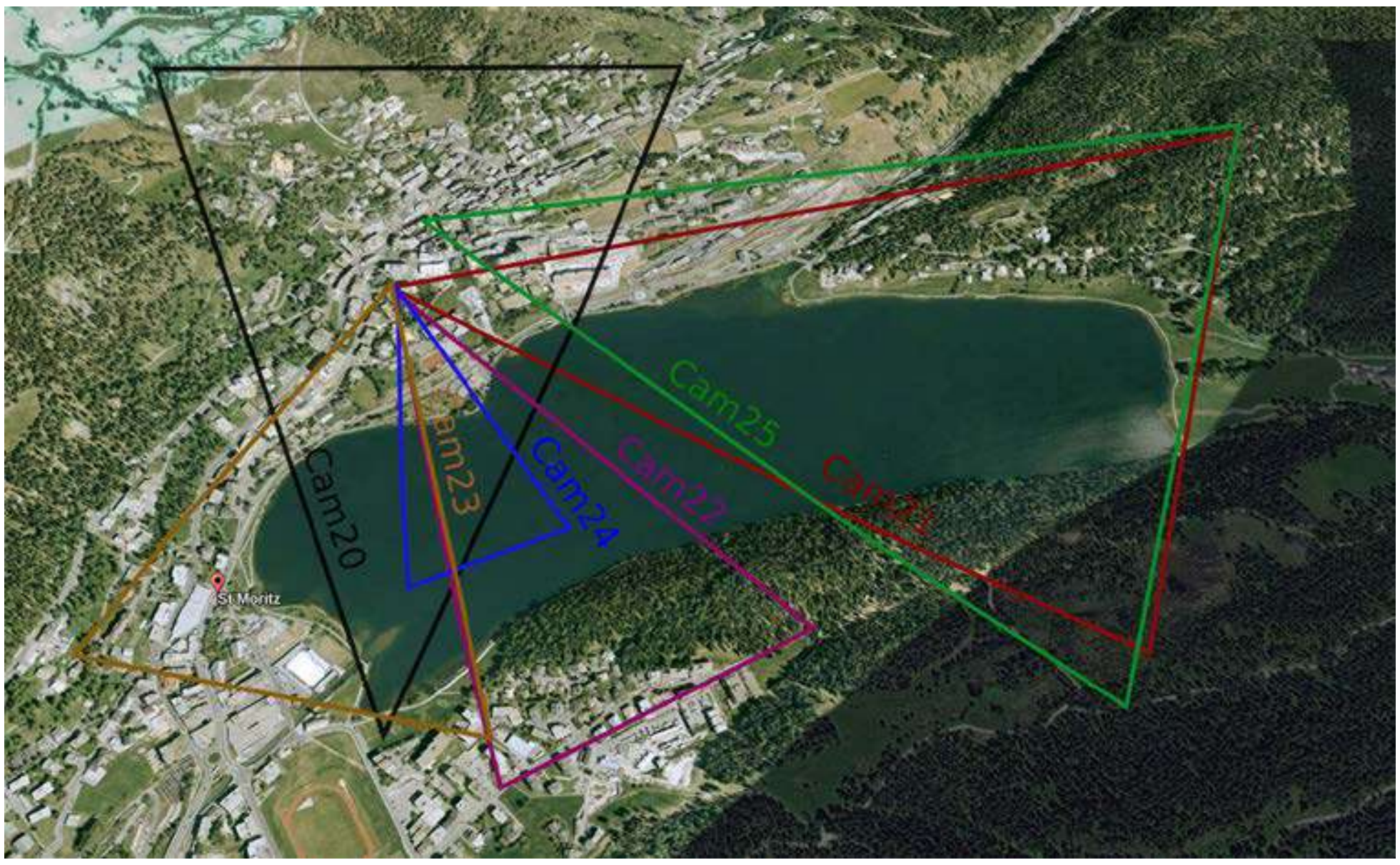

**Fig. 54.** Approximate area covered by each of the Webcams for Lake St. Moritz.

A more detailed review of available Webcam data comprising example images, the number of cameras, location, capturing frequency and type of installation is given in Appendix 1. Furthermore, visualizations of the position and covered lake area by each of the Webcams for all lakes are also shown in Appendix 1.

### 7.2. Webcam sequences for evaluation

Due to time restrictions and rather time consuming training of deep neural networks we limited our evaluations on two Webcams featuring very different GSDs. The GSD is computed based on the assumptions of 1.5 microns sensor pixel size (usually between 1.5-2.5 microns) and 4mm focal length (most common for Webcams with fixed focal length). Also based on a near vertical view, which is never the case for Webcams. The usable GSD depends very much on problems of common Webcams like rather poor radiometric properties, poor optics and many compression artefacts (summarized as poor image quality). However, a major factor is also the intersection angle between camera optical axis and lake surface. The lower this angle is, the larger the area that a pixel images, the greater the pixel distortion (deviation from a square ground pixel) and the less the reliability of the observed reflection will be. Thus, also the lesser this angle, the lesser the distance that a Webcam should observe for usable interpretation results (automated or manual). As an example, camera 5 for Lake Sils, which is positioned a bit higher than the lake surface (51m), has a usable distance of ca. 2 km. While camera 15 for Silvaplana, which lies much higher than the lake (382m), has a usable distance of ca. 4-5 km. Note that these usable distances are given for manual interpretation, for automated processing much shorter distances are needed, e.g. not more than 1 km. Furthermore, the usable distance also depends on the type of lake ice. E.g. snow on ice can be detected at much further distances, but to distinguish ice or even more black ice from water requires much higher GSD and shorter distances. Summarising, the closer the Webcam to the lake and the higher compared to lake surface it is, the better. Probably, the last factor is more important than the first, if an optical zoom lens is used.

Both cameras used image Lake St. Moritz, see Figs. 55 and 56. Images were collected from December 2016 until June 2017 and December 2017 to June 2018. Data gaps for the season 2016/2017 are listed in more detail in Appendix 1. The lake was frozen for a period of approximately four months in winter 2017, starting mid-December. In 2018, the lake started freezing on 3rd of December and melting ended beginning of May. For evaluation purposes we define 3 subsets (for each camera) based on all recorded data. As explained in more detail in section 7.5.1 and 7.5.3 this is to investigate generalization capabilities across winters and test the influence of including large portions of images capturing the lake in an unfrozen state. One of the major differences between the two Webcams is image scale: one camera (Cam21) acquired images with larger GSD, whereas the other one (Cam24) records at higher resolution. Both cameras record at a frequency of one image per hour. The images are stable, especially with respect to wind, such that the maximal temporal movements observed between them are around one pixel (derived by manual exploration of the image sequence). For the experiments, we manually removed images affected by heavy snowfall, fog and bad illumination conditions (early morning, late evening). Methods for automatic detection and elimination of such images have been proposed, e.g. Fedorov et al. (2016), but are not in the scope of this work.

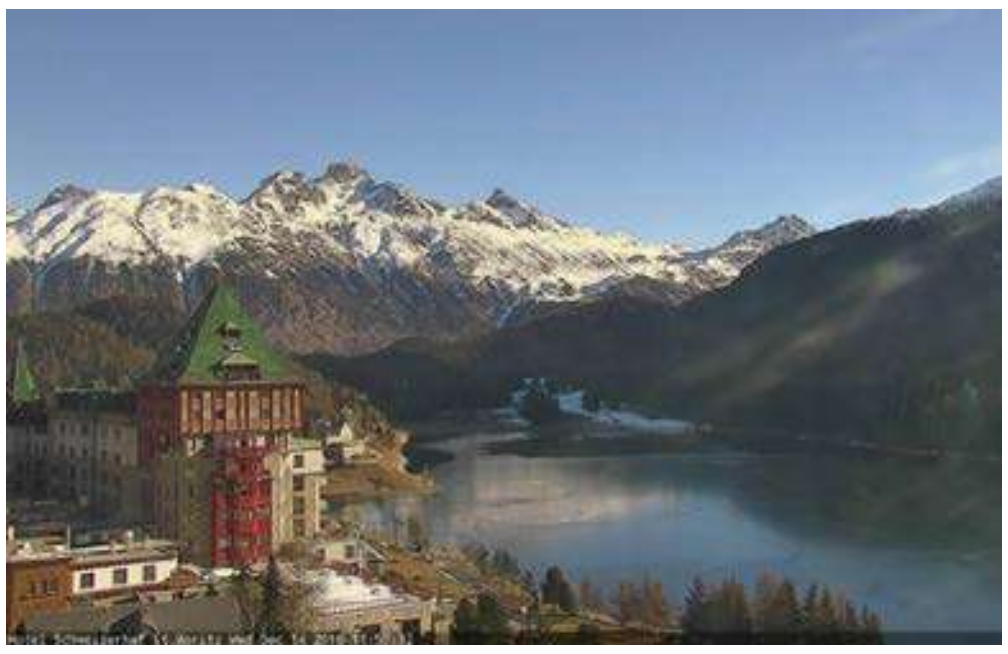

**Fig. 55.** Cam21, low-resolution Webcam.

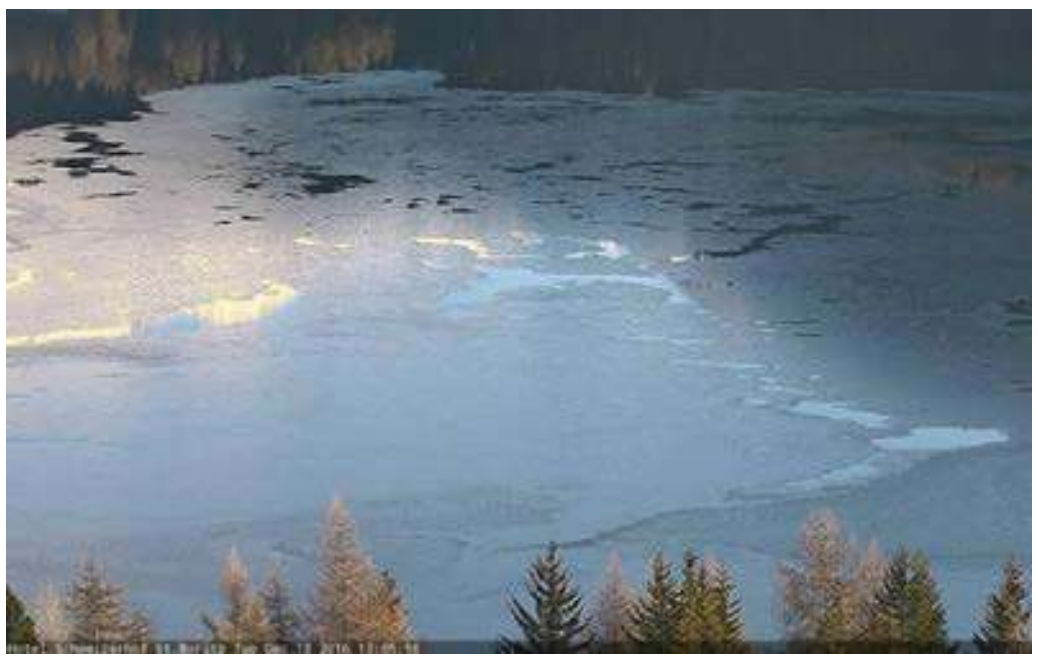

**Fig. 56**. Cam24, high-resolution Webcam.

Interpretation of camera images is challenging because of several reasons. Image resolution is limited and cheap, off-the-shelf Webcams only offer poor radiometric and spectral quality. Beside the limitations of the sensor itself, cameras are mounted with rather horizontal viewing angle such that large parts of the water body can be observed. As a result, large-scale differences (variations of the GSD) within a single image are present. We found that appearance within single classes (water, ice and snow) is rather different throughout the image sequences. This is caused by different ice structures, partly frozen water surfaces, waves, varying illumination conditions, reflections and shadows, etc. Furthermore, appearance across classes can be rather similar which impedes automatic interpretation. In fact, even manual interpretation for some examples was impossible without using additional temporal ques. Examples of patches featuring challenging texture are shown in Fig. 57.

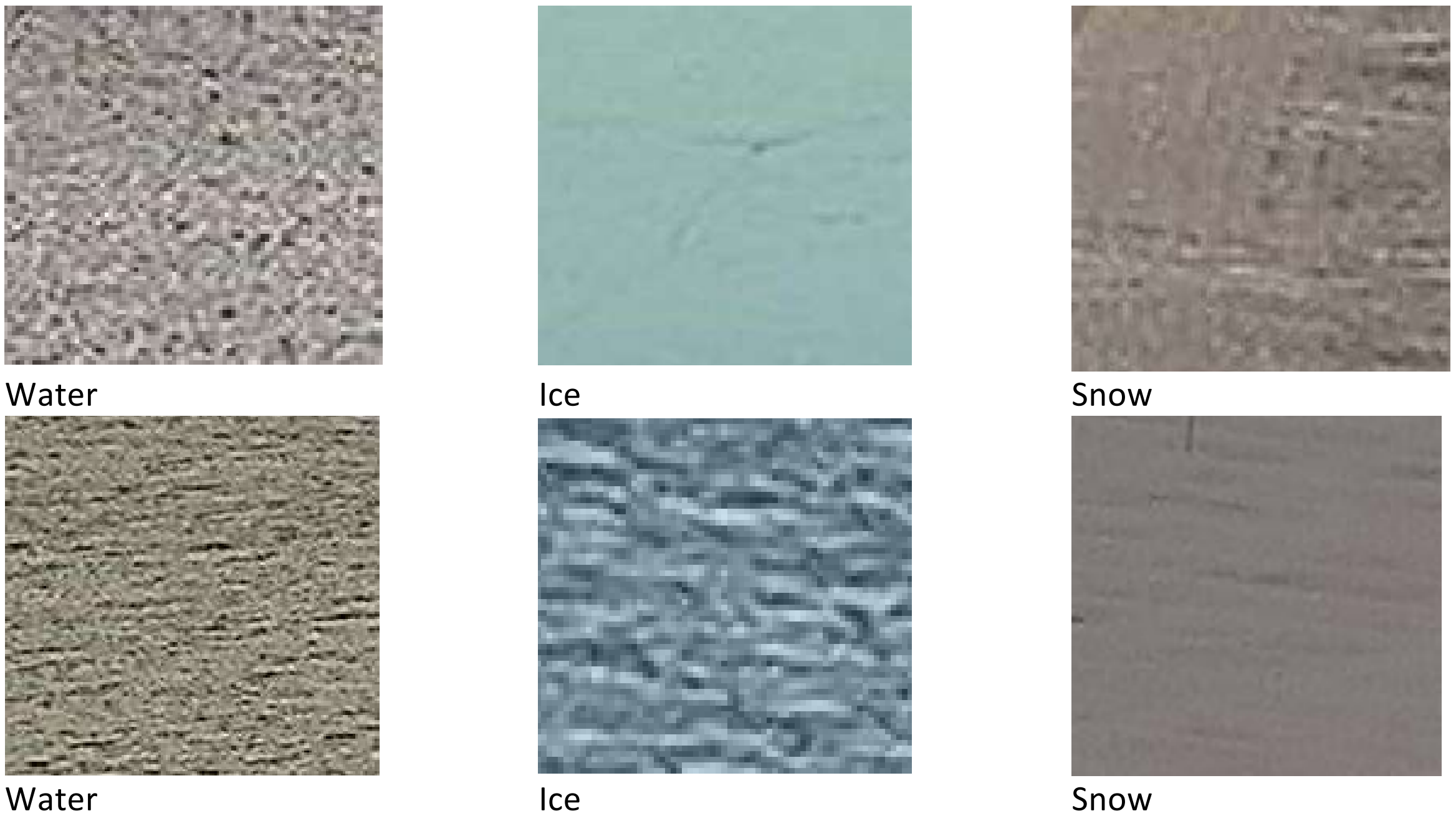

**Fig. 57**. Example patches of challenging texture.

## 7.3. Reference data

To estimate the percentage of the frozen lake area, we aim for a per-pixel classification of Webcam images. For training, testing and validation of our classifier, ground truth data is required. Therefore, ground truth label maps were produced by manually delineating and labelling polygons in the images, with labels water, ice, snow and clutter. Among these, water, ice and snow are the sought attributes of the application, the clutter class was introduced to mark objects other than the three target classes that are sometimes found on the lake, such as boats, or tents which are built up on Lake St. Moritz when hosting horse racing and other sport events. Non-lake pixels were manually annotated as well. For training and testing, we sampled patches only from lake pixels. For the manual labelling task, we used the browser-based tool of (Dutta et al., 2016) enabling annotation of user-defined image regions. Such polygons were then converted to raster label maps with a standard point-in-polygon algorithm (Berg et al. 2008). Overall, 1082 images for Cam21 and 1180 images for Cam24 were labelled for the season 2016/2017. For the season 2017/2018, 645 images were labelled for Cam21 and 635 images were labelled for Cam24.

## 7.4. Methodology

In this study, we investigate the potential of RGB Webcam images to predict accurate, per-pixel lake ice coverage. Technically, this amounts to a semantic segmentation of the image into the classes *water*, *ice*, *snow* and *clutter*, which we implement with a state-of-the-art deep convolutional neural network (CNN). The snow class is necessary to cover the case where snow covers the ice layer, whereas clutter accounts for objects other than the three target classes that may temporally appear on a lake. The key challenge is the limited data quality as discussed before. The core of our system is a variant of the recently successful DenseNet/Tiramisu architecture; however, there is to the best of our knowledge, no published work regarding lake ice monitoring with Webcams.

The rise of deep neural networks for image processing has recently also boosted semantic image segmentation. Based on the seminal Fully Connected Network of (Long et al., 2015), many state-of-the-art segmentation networks follow the encoder-decoder architecture. The encoder is typically derived from some high performance classification network consisting of a series of convolution (followed by non-linear transformations) and downsampling layers, e.g. (He et al., 2016 ; Huang et al., 2017 ; Xie et al., 2017). The subsequent decoder uses transposed convolutions to perform upsampling, normally either reusing higher resolution feature maps (Long et al., 2015 ; Ronneberger et al., 2015 ; Jégou et al., 2017) or storing the pooling patterns of the encoder (Badrinarayanan et al., 2017). In this way, the high-frequency details of the input image can be recovered. The network used in this work builds on the Tiramisu network proposed in (Jégou et al., 2017) and will be shortly reviewed in the following paragraph.

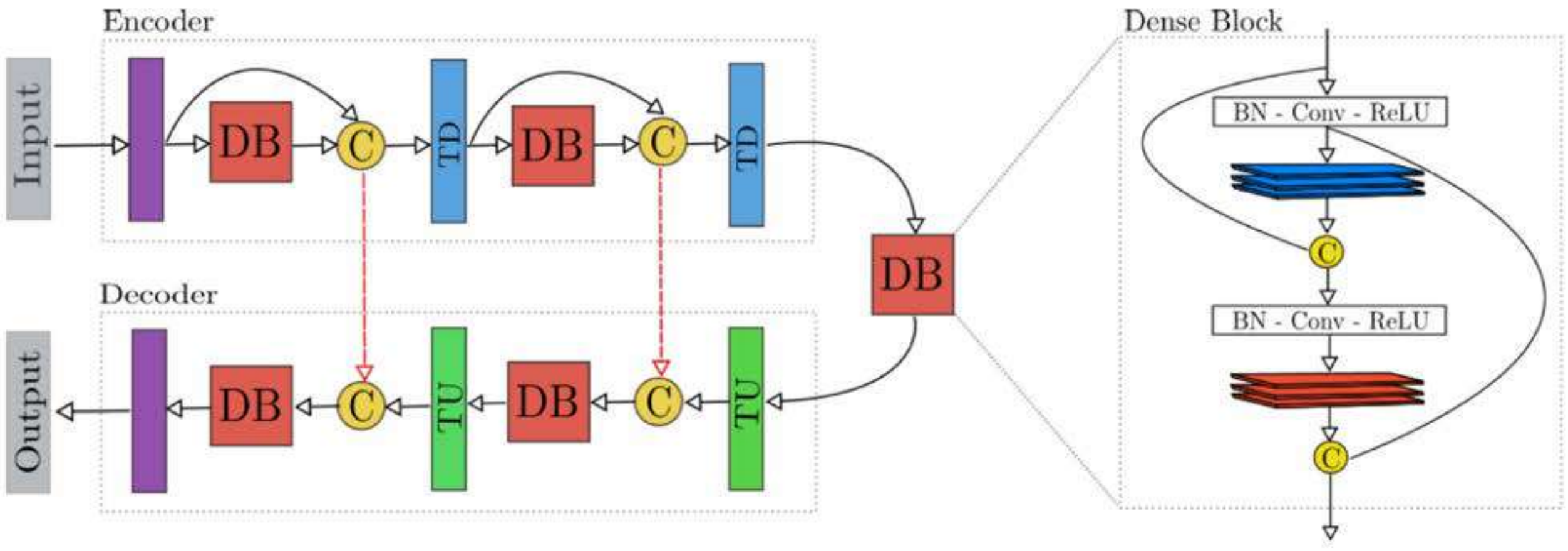

(a) (b)

**Fig. 58.** Schematic illustration of the segmentation framework. (a): The encoder downsamples the image, and thereby increases the field of view. It consists of a sequence of dense blocks (DB) and transition down (TD) blocks. The decoder performs upsampling of feature maps with a sequence of dense blocks (DB) and transition-up (TU) blocks. To recover high-resolution detail, skip connections pass information from intermediate encoder stages to the corresponding decoder stages. (b): Internal structure of a dense block with two layers and growth rate 3.

### 7.4.1. Fully Connected DenseNets

Our segmentation network is based on the One Hundred Layer Tiramisu architecture of Jégou et al. (2017). The network features a classical encoder-decoder architecture (see Fig. 58a). The encoder is based on the classification architecture DenseNet, a sequence of so-called dense blocks (DB) (see Fig. 58b). A dense block contains several layers. Each layer transforms its input by batch normalization (BN, Ioffe and Szegedy, 2015), Rectified Linear Unit (ReLU, Glorot et al., 2011) and convolution. The depth of the convolution layer is called growth rate. The distinguishing characteristic of a dense block is that the result of the transformation is concatenated with the input to form the output that is passed to the next layer, thus propagating lower-level representations up the network. In much the same way, the output of a complete dense block is concatenated with its input and passed through a transition-down (TD) block to reduce the resolution. TD blocks are composed of batch normalization, ReLU, 3×3 convolution and average pooling. To make the model more compact, the 3×3 convolution reduces the depth of the feature maps by a fixed compression rate. The result is then fed into the next dense block. The input feature maps of each transition-down block are also passed to the decoder stage with the appropriate resolution, to better recover fine details during upsampling. The decoder is a sequence of dense blocks and transition-up (TU) blocks. Note that in contrast to the encoder, dense blocks pass only the transformed feature maps, but not their inputs, to the next stage, to control model complexity. Transition-up blocks are composed of transposed convolutions with stride 2, which perform the actual upsampling. Output feature maps from the last dense block are subject to a final reduction in depth, followed by a softmax layer to obtain probabilities for each class at each pixel. The connection between the encoder and the decoder part is a denser block (called “bottleneck”), which has the lowest spatial resolution and at the same time the highest layer depth. It can be interpreted as a sort of abstract “internal representation” shared by the input data and the segmentation map. More detailed network parameters are given in the following two paragraphs.

In practice, the input dimensions are limited by the available GPU memory. To process complete images, we cut them into 224×224 pixel tiles for Cam24 and 56x56 pixel tiles for Cam21. Reduced tile sizes for Cam21 are due to larger image scale and smaller image dimensions. Neighbouring tiles have 50% overlap along the row and column direction, such that each pixel is contained in 4 tiles. Each tile is processed separately, then the four predicted probabilities $p^{c}_{i=0,1,2,3}(\mathbf{x})$ for class $c$ are averaged at every pixel $\boldsymbol{x}$, to obtain $p^{c} = \sum_{i} p^{c}_{i}(\mathbf{x})/4$. The final class is then the one with highest probability (winner-takes-it-all).

The same network architecture is used for both cameras. It features three dense blocks in the encoder (with 4, 6 and 8 layers), and three dense blocks in the decoder (with 8, 6, and 4 layers). The bottleneck, which connects encoder and decoder, has 10 layers. The growth rate is 12. Learning is done with the Nestorov-Adam optimizer (Sutskever et al., 2013). The network is regularized with L2-regularization and dropout (Srivastava et al., 2014) with a rate of 50%. We found empirically that high compression rates of 0.25 to 0.33 were important to ensure good convergence. The network was implemented using Keras (Chollet et al., 2015), with Tensorflow (Abadi et al., 2015) as backend. All experiments were run on Nvidia Titan X and Nvidia GTX1080 graphics cards.

Daily predictions such as freezing and melting dates are of particular interest for climate monitoring. Therefore, we seek to exploit temporal redundancy within image-wise predictions of the same day and estimate the daily percentage of ice and snow coverage for the observed water body. Per image, we sum the pixels of each class to obtain the covered area. We then compute the median coverage per class for each day. Finally, the coverage of the water body by ice, snow and clutter (mainly representing manmade structures erected on the ice) are summed, which gives an estimate of the area covered by ice and snow.

#### 7.4.2. Temporal Extensions of Fully Connected DenseNets

Since working on time series, the question arises if temporal cues of subsequent frames can be exploited to retrieve improved semantic segmentation. The simplest way is to perform a median operation for each pixel in the time dimension. For static scenes, this is expected to improve robustness of the predictions. However, for changing ice cover during the freezing period or floating ice, this strategy might not be optimal since classes per pixels change. This led to the question if temporal features can be learned via a deep CNN framework. Therefore, we tested two CNN architectures: a 3D fully connected (FC) DenseNet and a FC DenseNet embedded in a Long Short-Term Memory (LSTM) framework. Both architectures are described in the following sections.

**3D Fully Connected DenseNet.** The Tiramisu architecture uses 2D convolutions to transform an input image and subsequent feature maps. In contrast to Tiramisu, we use three temporally subsequent images as input of the **3D Fully Connected** architecture. These image blocks and subsequent features are transformed by 3x3x3 convolutions. Thus, temporal as well as spatial information is available for the convolution kernels and in theory enables learning of spatial and temporal features. The idea follows (Cicek et al., 2016), which adapted U-Net (Ronneberger et al., 2015) for the segmentation of 3D data. In our implementation, bands of three input frames are stacked according to their colour spectrum ($R_1R_2R_3G_1G_2G_3B_1B_2B_3$), which forms the input of the network. In the encoder part, convolutions in the dense blocks are adapted to 3x3x3 convolutions. Pooling in the transition-down blocks is performed with a stride of 2x2x1 to implement downsampling in the spatial dimension. The transition up blocks in the decoder features 3x3x3 transposed convolutions and convolutional layers in the dense blocks are adapted to 3D convolutions as in the encoder. The final layer implements feature reduction and softmax layers to produce semantic segmentations of the three input images. As the 3-tuple of input slices overlap in the time dimension, we generate three predictions for each pixel. Thus, for 3 consecutive time steps, 3 redundant predictions are generated. Fusion of such redundant segmentation maps is implemented as follows: per pixel, the maximum of the averaged class probabilities defines the final class prediction. Fig. 59 visualizes the implemented network.

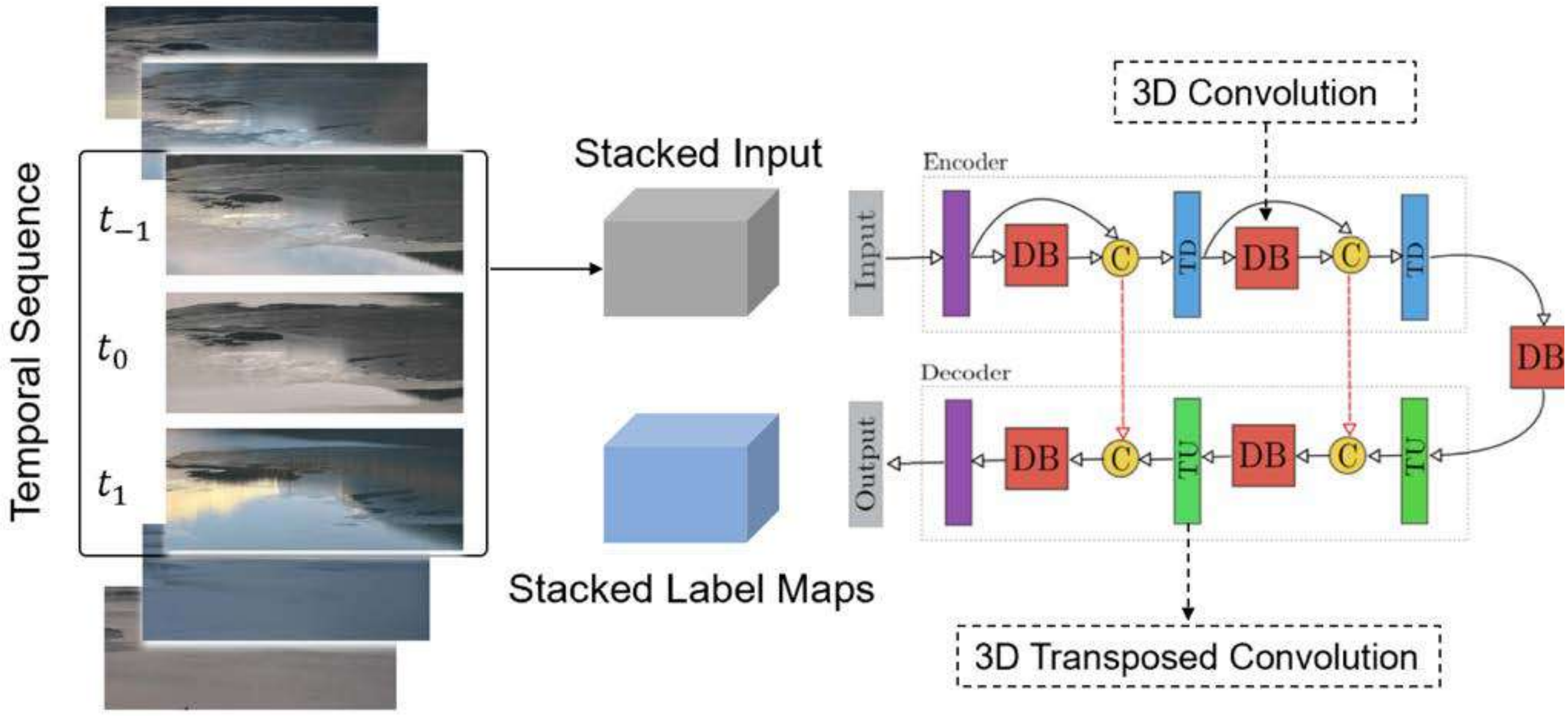

**Fig. 59.** Visualization of the implemented 3D Fully Connected DenseNet.

**LSTM Fully Connected DenseNet.** We implemented a second network, embedding the Tiramisu architecture within a LSTM framework (Hochreiter and Schmidhuber, 1997). LSTM networks are a subgroup of recurrent neural networks. The basic idea of RNNs is to use information of the network's state or output within subsequent inference processes. Thus, they are specifically designed to process data sequences. However, they suffer from the vanishing gradient problem. Training the network over long sequences might be instable or impossible because gradients during backpropagation become very small. In practice LSTM networks are used because they are capable to train networks with long time interactions. A LSTM block takes an input and produces an output, for the problem at hand low level features of the encoder and low level features of the decoder. The block also has a state **s,** which is passed, along with the output **o** to the subsequent time step. For each time step t the state is updated by combining previous state $\mathbf{s}_{t-1}$, previous output $\mathbf{o}_{t-1}$ and current input $\mathbf{i}_t$. A meaningful combination is controlled by so called gates. Fig. 60 shows the implemented LSTM FC DenseNet. The implementation operates on 3 subsequent frames. Network parameters are the same for all encoder strings.

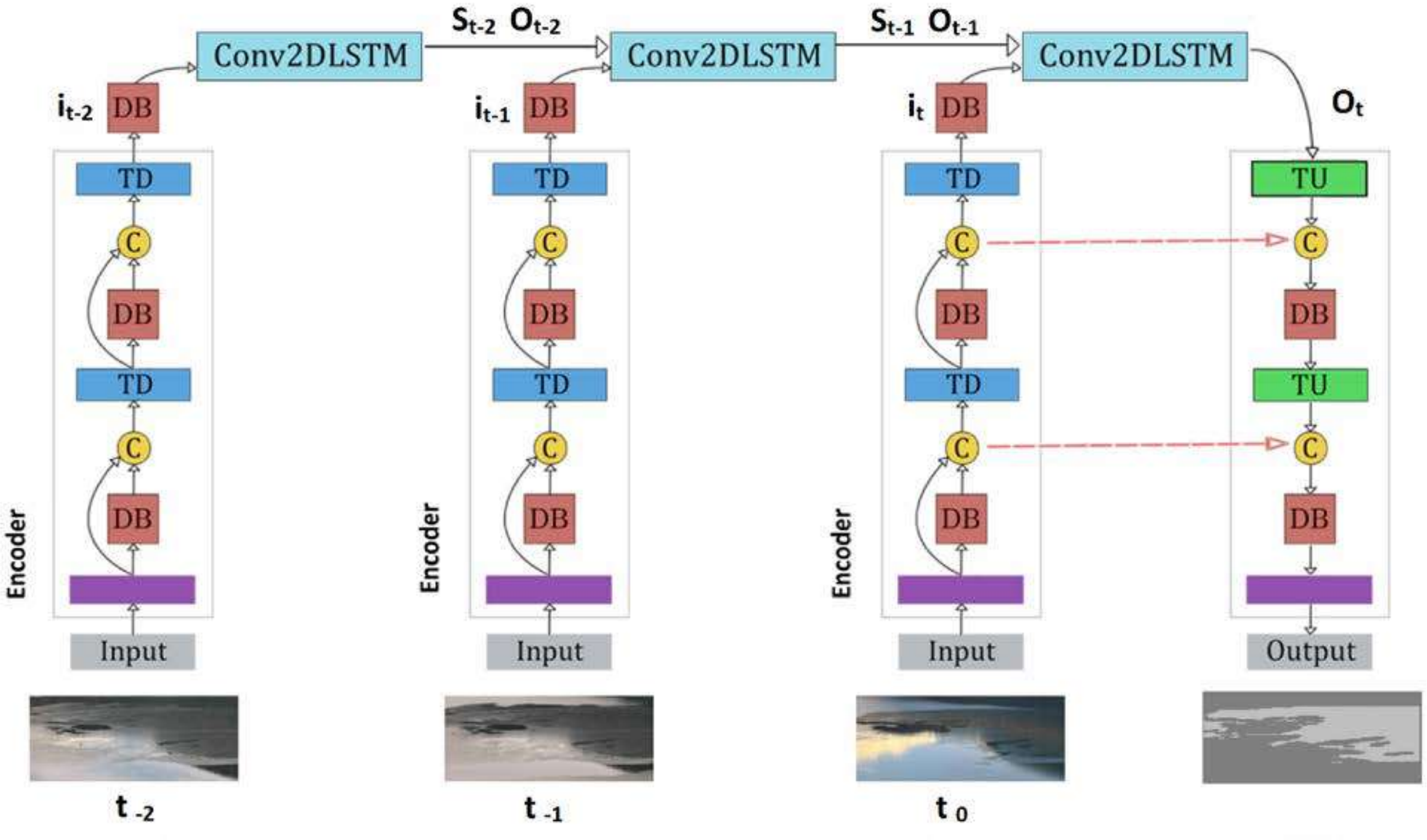

**Fig. 60.** Visualization of the implemented LSTM Fully Connected DenseNet architecture.

## 7.5. Experiments

### 7.5.1. Data Preparation and Evaluation Metrics

The experiments in this section are based on datasets recorded in three different time periods. Table 24 gives an overview of the number of images for each class dataset. Period2 contains all recorded images for the winter 2016/2017. Period1 is a subset of Period2 excluding the large number of images capturing the lake after melting (the actual melting process falls into a data gap). This separation is to test performance of segmentation on data sets not containing images capturing the lake in obviously unfrozen states (see section 7.5.3). Period 3 defines the images in the time span of interest for the winter 2017/2018. Period3 datasets will be used to test generalization capabilities for cross-winter predictions (see section 7.5.6).

For each camera and time period, training, validation and test sets are generated by randomly dividing the whole set of images (60%/15%/25%). Since images cannot be processed in a whole, smaller patches are sampled from those images. For the training set images are oversampled. More precisely, we regularly sample images such that the whole image area is covered. Additionally, we sample random patches since we found that this increases prediction performance (see Section 7.5.4). For test sets, we regularly sample patches with 25%-50% overlap, as explained in Section 7.4.1. After semantic segmentation, we fuse single patches and resulting maps are compared to ground truth segmentations. For our evaluation we use recall, precision, intersection over union (IoU), overall accuracy and mean intersection over union. Let C be the number of classes, TP the number of true positives, FP the number of false positives, TN the number of true negatives and FN the number of false negatives. Then the metrics are define as:

- Recall := $\frac{\text{TP}}{\text{TP+FN}}$ Precision := $\frac{\text{TP}}{\text{TP+FP}}$ IoU := $\frac{\text{TP}}{\text{TP+FP+FN}}$
- Overall accuracy := $\frac{\sum_{c=1}^{C} TP_c}{\text{\# all preddictedd pixels}}$
- Mean IoU := $\frac{\sum_{c=1}^{C} IoU_c}{\text{C}}$

**Table 24.** Number of images used for each class and camera for three time periods.

Period1: 4 December 2016 to 28 January 2017.

| | Water | Mix water-ice | Ice | Mix ice-snow | Snow | Total |
|---|---|---|---|---|---|---|
| Cam21 | 41 | 43 | 75 | 0 | 128 | 287 |
| Cam24 | 46 | 62 | 63 | 0 | 153 | 324 |

Period2: 4 December 2016 to 12 June 2017.

| | Water | Mix water-ice | Ice | Mix ice-snow | Snow | Total |
|---|---|---|---|---|---|---|
| Cam21 | 516 | 43 | 75 | 91 | 357 | 1082 |
| Cam24 | 535 | 62 | 63 | 106 | 414 | 1180 |

Period3: 29 November 2017 to 31 January 2018.

| | Water | Mix water-ice | Ice | Mix ice-snow | Snow | Total |
|---|---|---|---|---|---|---|
| Cam21 | 19 | 56 | 3 | 7 | 235 | 334 |
| Cam24 | 5 | 48 | 0 | 15 | 224 | 308 |

Intersection over Union (or Jaccard Index) is a more popular metric in the computer vision community and the de facto standard to evaluate performance of semantic segmentation methods. It is defined as the ratio of true positives over the sum of true positives, false negatives and false positives, thus combines recall and precision. Mean IuO is defined by the average of IoU values over all classes.

### 7.5.2. Semantic Segmentation for images on winter 2016/17

We train separate networks (i.e., same architecture, but individual network weights) for the Cam21 and Cam24 datasets of the whole season 16/17 (Period 2), so as to adapt the network weights to the specific camera and viewpoint. After training, the network is applied to all test patches of the respective dataset, and the patch-wise predictions are combined to complete per-image segmentation maps with the averaging mechanism explained in Section 7.4.1. A background mask is applied to the images so that only pixels, which correspond to the water body, are evaluated. The resulting pixel-wise class maps per full camera image are the final predictions that we compare to ground truth. Recall, precision and IoU values are displayed in Tables 25 and 26. The segmentation results are promising, reaching an overall accuracy of 94.7% for the Cam21 sequence and 91.2% for the Cam24 sequence. For Cam21, recall and precision of all main classes are in the range of 89.9%–97.1% and 81.5%–98.2%, respectively. For Cam24, recall and precision of the main classes are 87.6%–95.4% and 70.2%–99.8%, respectively. For both Cam21 and Cam24 data sets, the recall and precision of the clutter class are comparably low. This is mostly due to mistakes of labelling on thin structures. We note that the clutter class forms only a tiny portion of the pixels, and would be excluded in post-processing (e.g. temporal smoothing) in most practical applications.

**Table 25.** Cam21, Results of Training/Testing on whole sequence (Period2).

| | Recall | Precision | IoU |
|---|---|---|---|
| Water | 97.1% | 98.2% | 95.4% |
| Ice | 93.4% | 82.0% | 77.5% |
| Snow | 92.7% | 96.7% | 89.9% |
| Clutter | 89.9% | 81.5% | 74.6% |
| Overall Acc. | 94.7% | | |
| Mean IoU | 84.4% | | |

**Table 26.** Cam24: Results of Training/Testing on whole sequence (Period2).

| | Recall | Precision | IoU |
|---|---|---|---|
| Water | 92.2% | 99.8% | 92.1% |
| Ice | 95.4% | 72.4% | 70.0% |
| Snow | 87.6% | 95.2% | 83.9% |
| Clutter | 94.4% | 70.2% | 67.4% |
| Overall Acc. | 91.2% | | |
| Mean IoU | 78.3% | | |

### 7.5.3. Test on reduced input

Freezing and melting periods are of particular interest in climate studies. Semantic segmentation of such images is most challenging because images contain diversified patterns and inhomogeneous class layouts due to drifting ice. Furthermore, only limited training examples are available for those short periods. On the other hand, for Period2 a large number of samples were recorded during the periods for which ice cover is highly unlikely and for which the lake is completely melted. Such samples possibly are not supporting or even degrading the segmentation of samples in the most interesting freezing period. To test this, we train and test a model on a reduced dataset of Period1 capturing the freezing process. Tables 27 and 28 display the results of the test.

For Cam24 the model trained on the freezing period only outperforms the model trained on the dataset including large amount of irrelevant data. For the model trained on all data (Period2) ambiguities might be introduced and the network might be biased towards the predominant water class. Compared to the results of Period2, the mean accuracy of Cam24 increases by 6.5%. For Cam21 the mean accuracy of models trained on Period1 and Period2 are similar. Fig. 61 and Fig. 62 show the daily predictions of lake ice coverage derived from the semantic segmentation maps. Whereas ground truth is reproduced rather accurately for the high-resolution Cam24, some deviations are present in the freezing period for the low-resolution Cam21.

**Table 27.** Cam21, Results of Training/Testing on images of the relevant period.

| | Recall | Precision | IoU |
|---|---|---|---|
| Water | 92.0% | 85.0% | 79.1% |
| Ice | 90.6% | 97.0% | 97.3% |
| Snow | 97.1% | 97.3% | 94.6% |
| Clutter | 91.3% | 81.5% | 71.2% |
| Overall Acc. | 93.7% | | |
| Mean IoU | 84.3% | | |

**Table 28.** Cam24, Results of Training/Testing on images of the relevant period.

| | Recall | Precision | IoU |
|---|---|---|---|
| Water | 95.1% | 99.2% | 94.3% |
| Ice | 98.6% | 97.8% | 96.5% |
| Snow | 98.5% | 98.1% | 96.7% |
| Clutter | 91.2% | 75.1% | 70.0% |
| Overall Acc. | 97.7% | | |
| Mean IoU | 89.4% | | |

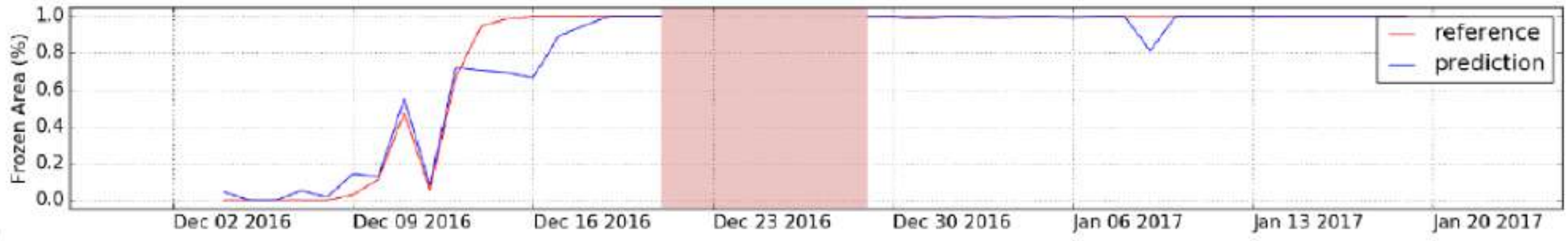

**Fig. 61.** Estimation of frozen area of Cam21 for Period1.

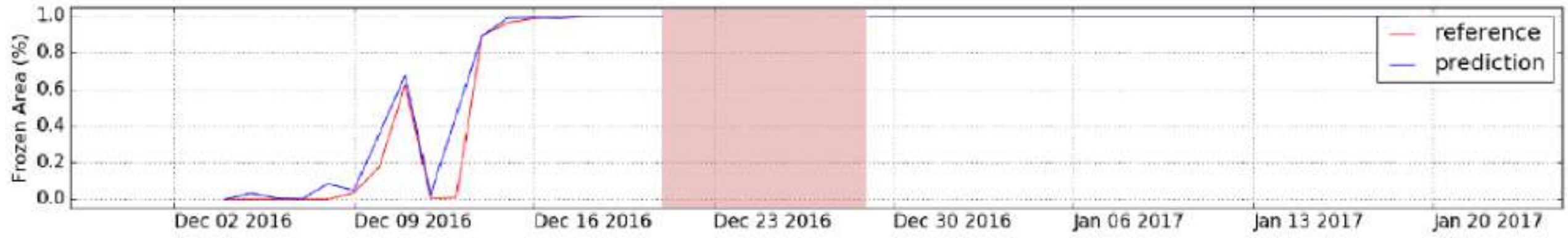

**Fig. 62.** Estimation of frozen area of Cam24 for Period1.

### 7.5.4. Number of sampling patches

As mentioned in 7.4.1, due to memory limitations the images cannot be processed in a whole. Instead, patches are sampled from the images for training and the question of an adequate number of samples arises. Within all tests, we first regularly sample n patches such that each of the images is completely covered. In all experiments so far we additionally sampled 3n patches at random locations to generate a training set of 4n samples. Involving less samples in the training process results in decreased processing time. In this test, we evaluate the influence of lower sampling rates. Therefore, we reduce random sampling to form training sets with only 2n patches. The results in Tables 29 and 30 clarify that performance of models decreases when using reduced number of samples only. More precisely, the IoU for all main classes drops in the range of 5.5 to 9.6 % for Cam21 and for Cam24 IoU decreases in the range of 2.1 to 3.5 % were observed. Test using higher resampling rates of 6n showed no further improvement in accuracy compared to sapling with 4n. Thus, for all further tests we form training sets of 4n patches.

**Table 29.** Cam21: Results for sampling with 2n.

| | Recall | Precision | IoU |
|---|---|---|---|
| Water | 78.5% | 87.7% | 70.7% |
| Ice | 94.3% | 92.7% | 87.8% |
| Snow | 95.1% | 93.4% | 89.1% |
| Clutter | 91.3% | 81.5% | 75.6% |
| Overall Acc. | 91.1% | | |
| Mean IoU | 77.8% | | |

**Table 30.** Cam24: Results for sampling with 2n.

| | Recall | Precision | IoU |
|---|---|---|---|
| Water | 93.3% | 97.1% | 90.8% |
| Ice | 95.8% | 96.8% | 92.9% |
| Snow | 98.5% | 96.0% | 94.6% |
| Clutter | 77.7% | 76.3% | 62.5% |
| Overall Acc. | 96.1% | | |
| Mean IoU | 85.2% | | |

### 7.5.5. Generalization regarding different cameras

In this test, we evaluate generalization capabilities with respect to different cameras. For operational use, good generalization would be highly desirable since a model trained on a specific camera sequence directly could be applied to other camera images. We test the two baseline models of Cam21 and Cam24 trained on Period1 on the test set of the respective other camera. In other words, the model trained using Cam21 data is tested on the Cam24 test set and vice versa. As shown in Tables 31 and 32, performance of cross testing is poor. The overall accuracy amounts to 60.5% for testing the Cam21 model on Cam24 and 52.7% for testing the Cam24 model on Cam21. Also daily predictions do not reproduce the ground truth very well, as displayed in Fig. 63 and Fig. 65. Low performance is expected to be due to large differences in scale, variances in perspective and differing appearance. We note that this is an extreme case, so models trained and tested on more similar image data might generalize better. Interestingly, the model trained on the low-resolution images performs slightly better for the prediction of high-resolution images than vice versa. We assume that this is because the Cam21 model was only exposed to low-resolution patterns, which are present also in the Cam24 data to some degree. In contrast, Cam24 model expects high-frequency information not present in the Cam21 data. Moreover, the Cam21 model could potentially generate low-resolution representations within downsampling at later stages of the encoder. Despite limited generalization capabilities, models trained on both Cam21 and Cam24 perform comparably well. As shown in Tables 33 and 34, an overall accuracy of 85.3% and 91.7% was achieved for Cam21 and Cam24, respectively. Compared to models trained and tested on only one camera this is a drop in overall accuracy of 8.4 % for Cam21 and 6.0 % for Cam24. Despite this drop, the overall accuracy is still larger than 85%, which shows that training a single network on heavily differing image data is possible to some degree (see also Figs. 64 and 66).

**Table 31.** Results of model trained on Cam21 images and tested on Cam24 images.

| | Recall | Precision | IoU |
|---|---|---|---|
| Water | 76.9% | 41.5% | 36.9% |
| Ice | 74.7% | 70.5% | 57.0% |
| Snow | 42.0% | 78.8% | 37.7% |
| Clutter | 49.4% | 38.7% | 27.7% |
| Overall Acc. | | 60.5% | |
| Mean IoU | | 39.8% | |

**Table 32.** Results of model trained on Cam24 images and tested on Cam21 images.

| | Recall | Precision | IoU |
|---|---|---|---|
| Water | 60.3% | 41.0% | 32.3% |
| Ice | 58.6% | 59.1% | 41.7% |
| Snow | 43.7% | 57.7% | 33.1% |
| Clutter | 63.3% | 58.3% | 43.6% |
| Overall Acc. | 52.7% | | |
| Mean IoU | 37.7% | | |

**Table 33.** Results of model trained on Cam21 and Cam24 images and tested on Cam21 images.

| | Recall | Precision | IoU |
|---|---|---|---|
| Water | 77.6% | 83.5% | 67.3% |
| Ice | 75.4% | 90.2% | 69.6% |
| Snow | 96.7% | 84.4% | 82.1% |
| Clutter | 84.6% | 70.6% | 62.6% |
| Overall Acc. | 85.3% | | |
| Mean IoU | 70.4% | | |

**Table 34.** Results of model trained on Cam21 and Cam24 images and tested on Cam24 images.

| | Recall | Precision | IoU |
|---|---|---|---|
| Water | 82.3% | 89.6% | 75.1% |
| Ice | 89.1% | 92.2% | 82.8% |
| Snow | 98.7% | 93.0% | 91.9% |
| Clutter | 80.1% | 74.3% | 62.7% |
| Overall Acc. | 91.7% | | |
| Mean IoU | 78.1% | | |

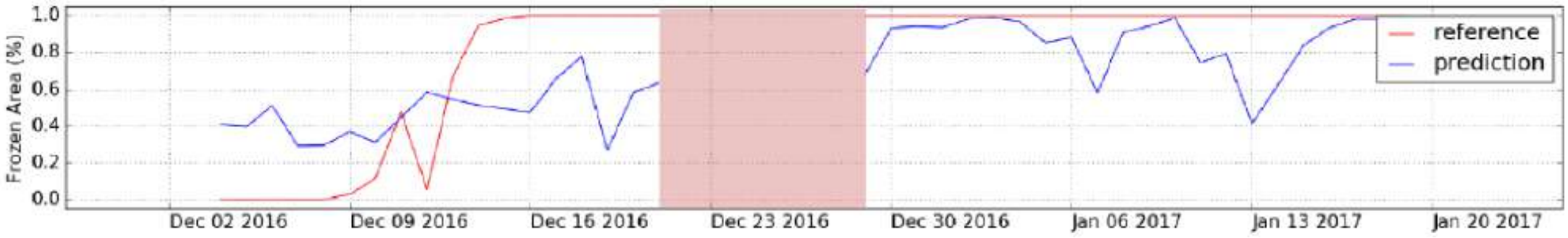

**Fig. 63.** Daily predictions of model trained on Cam24 data and tested on Cam21 data.

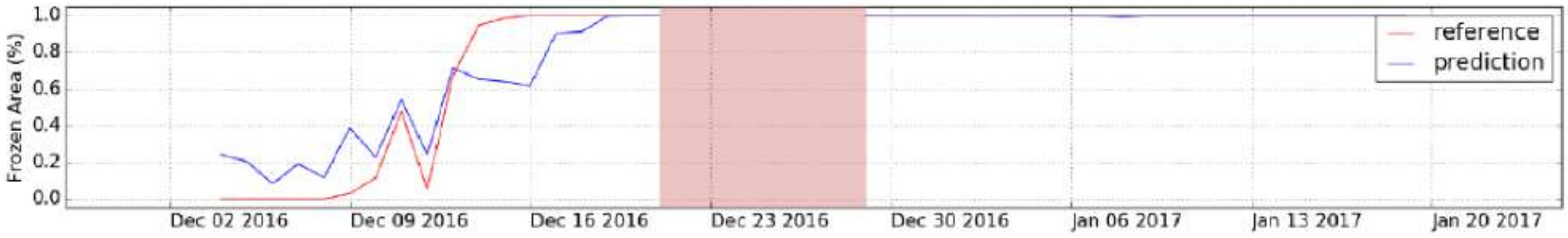

**Fig. 64.** Daily predictions of model trained on Cam21 and Cam24 data and tested on Cam21 data.

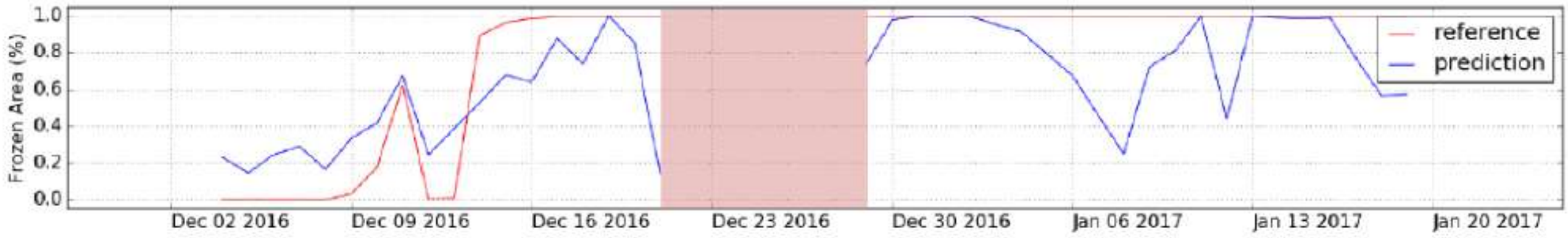

**Fig. 65.** Daily predictions of model trained on Cam21 data and tested on Cam24 data.

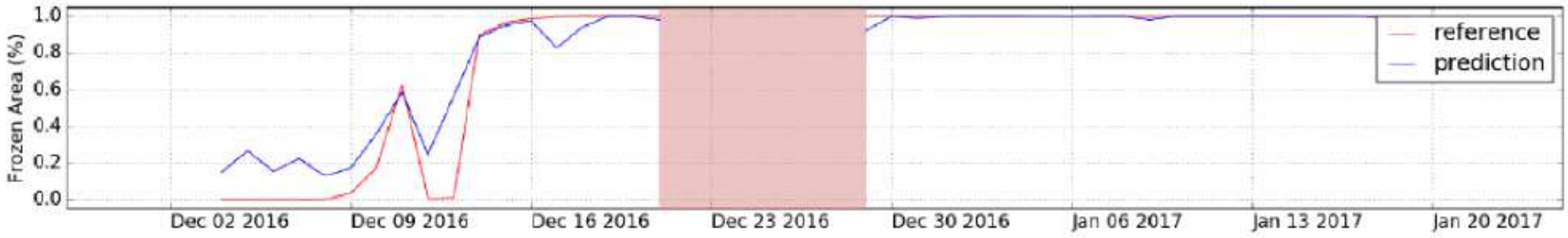

**Fig. 66.** Daily predictions of model trained on Cam21 and Cam24 data and tested on Cam24 data.

### 7.5.6. Generalization with respect to different winters

In this test, the generalization capability of models across winters is evaluated. Therefore, we tested prediction performance of the baseline models trained on data from winter 16/17 (Period1) on samples of the winter 2017/2018 (Period3). As displayed in Tables 35 and 37, a drop in performance can be observed.

Overall accuracy decreases by 13.1 % and 6.4 % for Cam21 and Cam24, respectively. However, still an overall accuracy of 80.6% for Cam21 and 91.9% for Cam24 was achieved. Fig. 67 displays the daily estimation of ice coverage for Cam21. Largest deviations from the ground truth can be observed in the freezing period. For Cam24 (see Fig. 70) ground truth is reproduced rather accurately. We note that this appearance of patches across years is rather different due to different patterns of especially ice and snow. Therefore, generalization capabilities might be better if training data of multiple winters are available.

Furthermore, we show that well performing models can be trained based on data from multiple winters. Therefore, we train and test a network for each camera based on the data of Period1 and Period 3. Tables 36 and 38 show that the overall accuracy amounts to 90.5% for Cam21 and 94.3% for Cam24. This is a slight drop compared to the baseline models trained and tested on Period1 only. More precisely, for Cam21 we observed a drop in overall accuracy of 2.4 % for Cam21 and a drop of 3.4 % for Cam24. However, evaluation of daily predictions (see Figs. 68, 69, 71 and 72) indicates that both models are close to the ground truth. Comparing daily predictions of models trained on data from Period1 in winter 16/17 and models trained on data from both winters, for latter ground truth seems to be reproduced more robustly. We note that for this experiment the capacity of the network was increased by enlarging the growth rate from 12 to 18.

**Table 35.** Results for Cam21, model trained on images from 2016/17 and tested on images from 2017/18.

| | Recall | Precision | IoU |
|---|---|---|---|
| Water | 95.0% | 46.8% | 45.7% |
| Ice | 55.0% | 69.9% | 44.5% |
| Snow | 84.8% | 98.1% | 83.5% |
| Clutter | 76.2% | 46.6% | 40.7% |
| Overall Acc. | 80.6% | | |
| Mean IoU | 53.6% | | |

**Table 36.** Results for Cam21, model trained on images from 2016/17/18 and tested on images from 2016/17/18.

| | Recall | Precision | IoU |
|---|---|---|---|
| Water | 87.5% | 83.0% | 74.2% |
| Ice | 81.2% | 90.5% | 74.8% |
| Snow | 95.5% | 93.2% | 89.3% |
| Clutter | 90.8% | 80.7% | 74.6% |
| Overall Acc. | 90.5% | | |
| Mean IoU | 78.2% | | |

**Table 37.** Results for Cam24, model trained on images from 2016/17 (Period1) and tested on images from 2017/18 (Period3).

| | Recall | Precision | IoU |
|---|---|---|---|
| Water | 85.0% | 90.1% | 77.7% |
| Ice | 68.1% | 78.9% | 57.6% |
| Snow | 96.4% | 94.8% | 91.5% |
| Clutter | 64.3% | 39.9% | 32.7% |
| Overall Acc. | 91.3% | | |
| Mean IoU | 64.9% | | |

**Table 38.** Results for Cam24, model trained on images from 2016/17/18 (Period 1+3) and tested on images from 2016/17/18 (Period1+3).

| | Recall | Precision | IoU |
|---|---|---|---|
| Water | 91.7% | 92.1% | 85.0% |
| Ice | 89.2% | 96.4% | 86.4% |
| Snow | 97.7% | 95.2% | 93.1% |
| Clutter | 80.1% | 65.7% | 56.5% |
| Overall Acc. | 94.3% | | |
| Mean IoU | 80.2% | | |

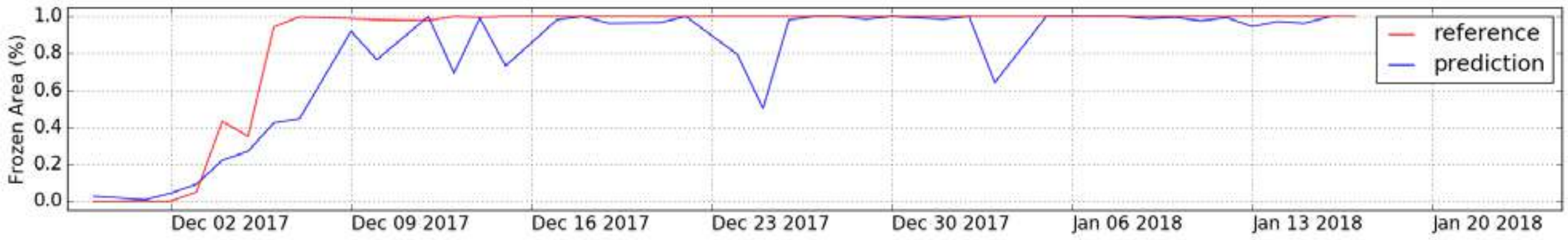

**Fig. 67.** Estimation of frozen area of Cam21, trained on Period1, tested on Period3.

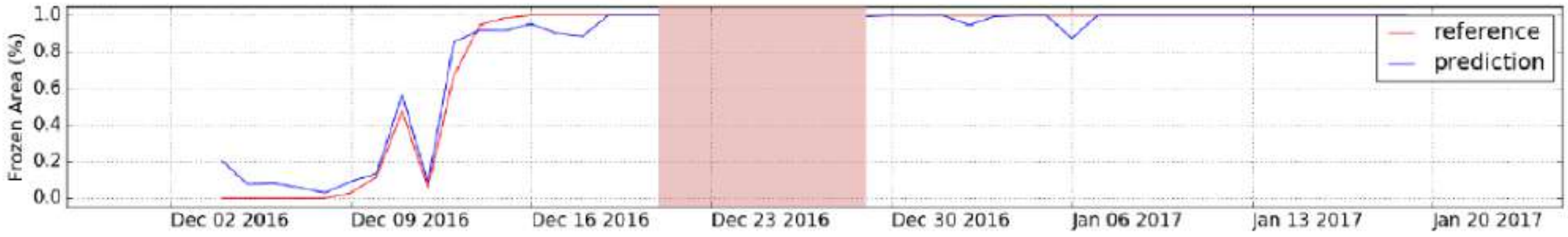

**Fig. 68.** Estimation of frozen area of Cam21, trained on Period1+3, tested on Period1.

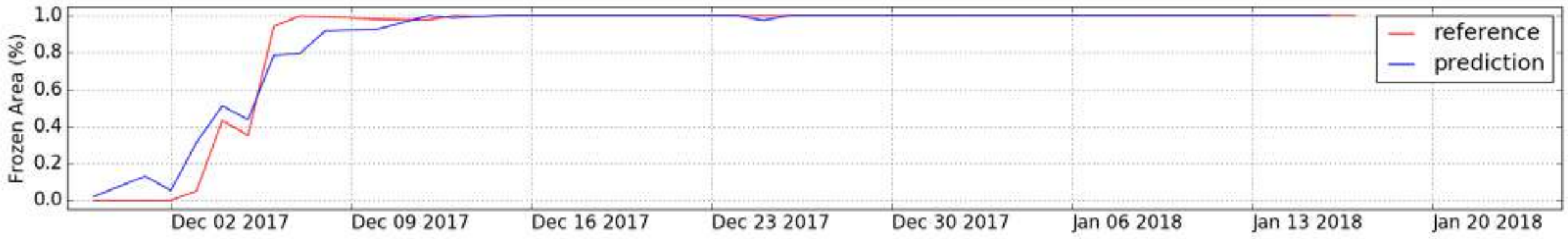

**Fig. 69.** Estimation of frozen area of Cam21, trained on Period1+3, tested on Period3.

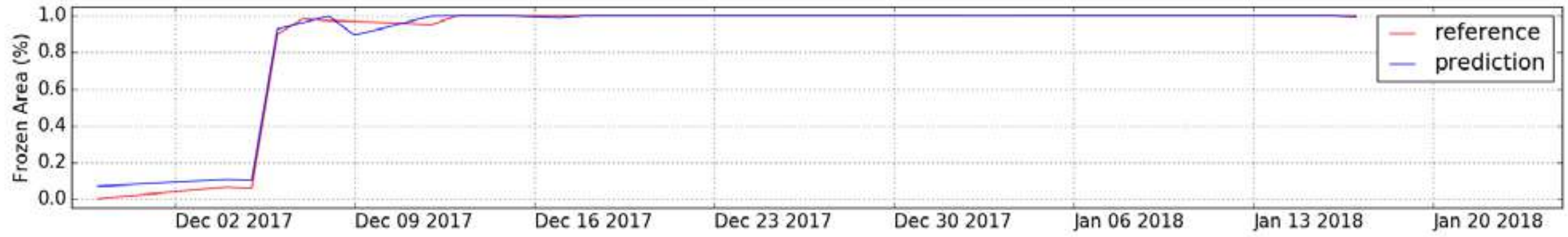

**Fig. 70.** Estimation of frozen area of Cam24, trained on Period1, tested on Period3.

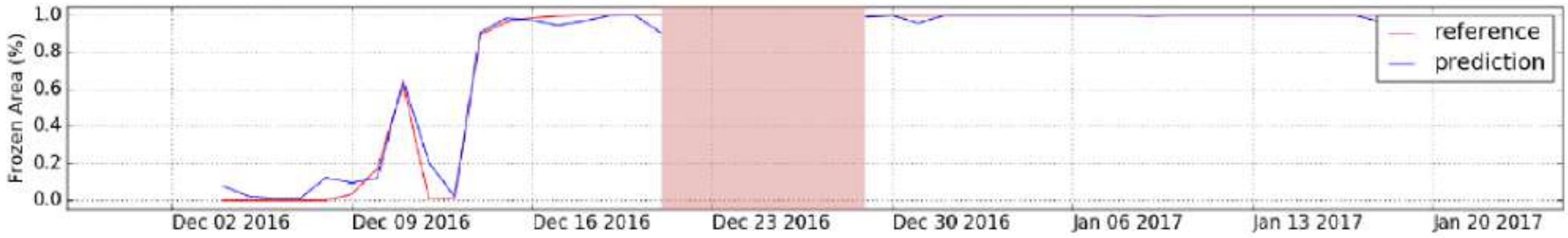

**Fig. 71.** Estimation of frozen area of Cam24, trained on Period1+3, tested on Period1.

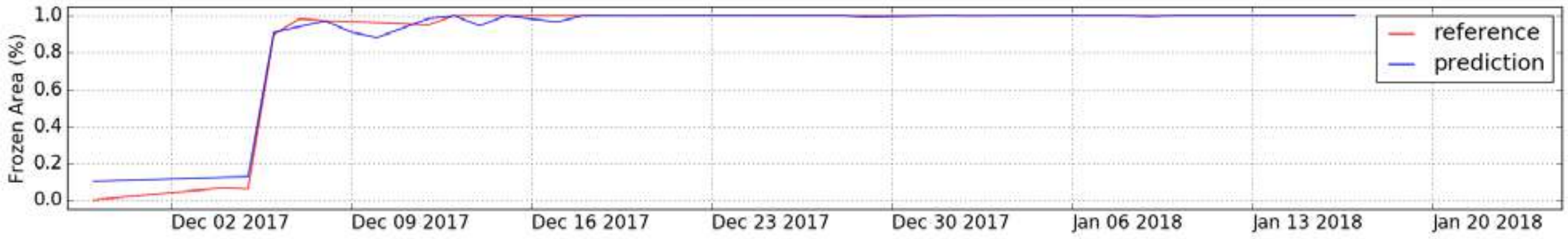

**Fig. 72.** Estimation of frozen area of Cam24, trained on Period1+3, tested on Period3.

### 7.5.7. Training on Easy-To-Label images only

Training deep neural networks in general requires a very large amount of training data, which often has to be manually labelled. This is a labour-intensive task, in particular labelling images with multiple class instances for which multiple complex polygons have to be specified. For the problem at hand labelling of images, which contain one class only (e.g. if the lake is completely covered with snow) is straightforward since the image can be annotated as a whole. Furthermore, this reduces labelling errors introduced by the operator and could be even automated if the dates on which the lake was completely covered by ice or snow or not frozen at all are known. In this test, we evaluate the performance of networks solely trained on data, which is easy to label, or in other words contain a single class only. Therefore, we remove all mixed-class images from the training set of Period1. An exception is the snow images, which contain the clutter class. We have to include these images because too much data would be removed elsewise. However, clutter pixels amount only for about 2% of all pixels and are not relevant in a practical application. For Period1, we train models for Cam21 and Cam24 on the reduced test sets for which multi-class samples are removed. Then, we predict on the complete test set of Period1 including multi-class images. Compared to the model trained on all samples, we observe a slight decrease in overall accuracy (4.8 % for Cam21 and 2.5 % for Cam24). Generally, predictions for mixed-class images poses errors at water-ice class boundaries or ice regions. As a result, also daily estimations of the frozen lake area are expected to be slightly more imprecise than no mixed. Fig. 73 and Fig. 74, clarify that this holds true in particular for the freezing period. However, the performance drop for daily predictions is less due to the multitemporal filtering. For Cam24 results are better than for Cam21.

**Table 39.** Results for Cam21, training on easy (one class) examples.

| | Recall | Precision | IoU |
|---|---|---|---|
| Water | 90.8% | 82.5% | 76.1% |
| Ice | 87.2% | 92.5% | 81.5% |
| Snow | 88.9% | 95.6% | 85.4% |
| Clutter | 94.1% | 49.1% | 47.6% |
| Overall Acc. | 88.9% | | |
| Mean IoU | 72.6% | | |

**Table 40.** Results for Cam24, training on easy (one class) examples.

| | Recall | Precision | IoU |
|---|---|---|---|
| Water | 95.1% | 91.0% | 86.9% |
| Ice | 93.4% | 96.3% | 90.2% |
| Snow | 97.9% | 96.9% | 94.9% |
| Clutter | 68.4% | 85.4% | 61.2% |
| Overall Acc. | 95.2% | | |
| Mean IoU | 83.3% | | |

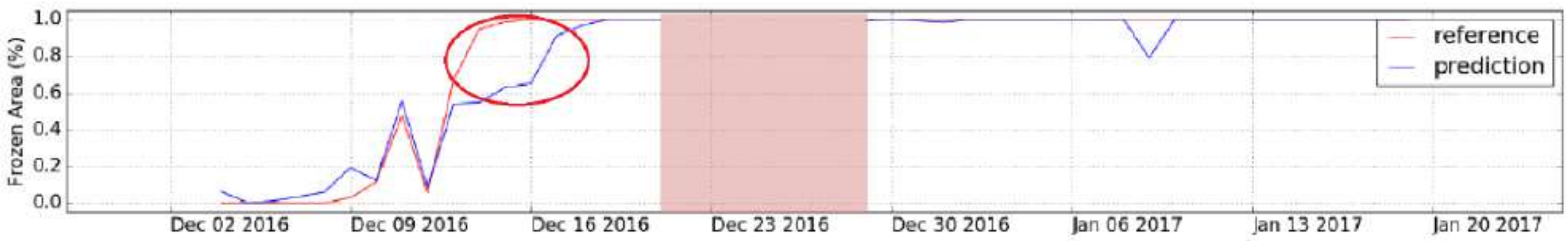

**Fig. 73.** Estimation of frozen area of Cam21, trained on easy examples of Period1, tested on complete test set of Period1.

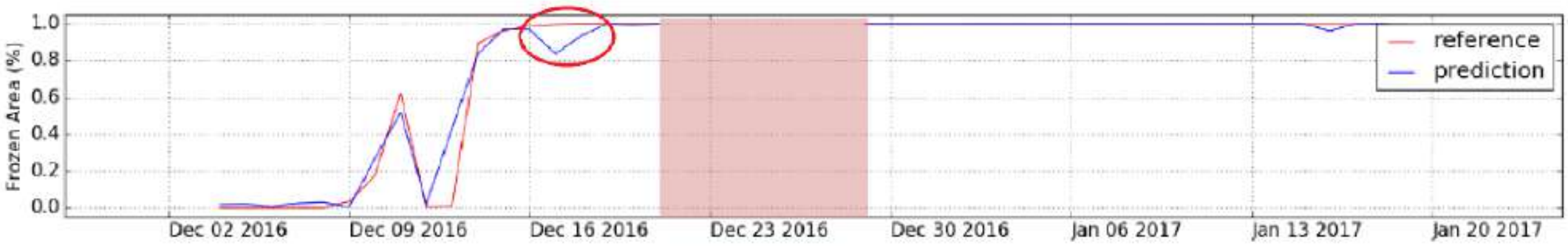

**Fig. 74.** Estimation of frozen area of Cam24, trained on easy examples of Period1, tested on complete test set of Period1.

### 7.5.8. Temporal processing

We trained and tested models of 3D- and LSTM- FC DenseNet on Cam24 of Period1. Since the temporal networks require image sequences instead of randomly sampling images for training, validation and testing, we randomly sample non-overlapping 7-day image sequences from Period1. As a baseline we consider the standard Tiramisu network for single frame processing (Fig. 76) and the fusion of three consecutive single frame predictions via simple majority vote (median of 3 temporally connected pixels) (Fig. 77), while Fig. 75 shows the manually collected ground truth. In all visualization, white represents snow, light-grey represents ice, dark-grey represents water and black represents clutter. Whereas the 3D FC DenseNet was trained from scratch, we used a pre-trained encoder for the LSTM network. Moreover, we fixed the weights of the encoder during training and only optimized parameters of LSTM blocks and decoder. As shown in Table 41, overall accuracy and mean IoU of the single-frame baseline amount to 95.3% and 85.7%, respectively. For the 3-frame median baseline results are only slightly better. These numbers however do not clearly represent the two effects we observed. For temporal static scenes, segmentation maps get slightly more robust. For temporal dynamic scenes, temporal smoothing degrades results. Latter effect is visualized in Fig. 77, which shows the obtained segmentation maps for six consecutive images. Finally, we test, if complex freezing dynamics can be learned by our 3D FC DenseNet and LSTM networks. For the 3D FC DenseNet and the LSTM networks, overall accuracy and mean IoU are slightly worse than the 3-frame median baseline. Figs. 78 and 79 show the obtained semantic segmentations for the most interesting freezing period. In general, semantic segmentations are qualitatively not on par with single frame predictions. For the 3D FC DenseNet, it seems that some sort of temporal smoothing is learned and obtained segmentations are a mix of the 3 involved frames. For LSTM some erroneous ice patches are introduced. We note that training data for the freezing period is strongly limited and might not be enough to train models capturing temporal dynamics. Moreover, many more examples of temporal static scenes are used within the training process. For such temporal consecutive samples the scene and patterns remain almost identical and most reliable results could be obtained by temporal smoothing. We assume that this is learned by the model, so the observed smoothing effects for temporal dynamic scenes are somehow comprehensible.

**Table 41.** Results for single frame prediction.

| | Recall | Precision | IoU |
|---|---|---|---|
| Water | 92.1% | 92.7% | 85.7% |
| Ice | 97.6% | 95.0% | 92.8% |
| Snow | 97.5% | 96.9% | 94.5% |
| Clutter | 75.8% | 89.9% | 69.9% |
| Overall Acc. | 95.3% | | |
| Mean IoU | 85.7% | | |

**Table 42.** Results for majority vote for of 3 subsequent frames.

| | Recall | Precision | IoU |
|---|---|---|---|
| Water | 94.5% | 94.5% | 87.3% |
| Ice | 97,3% | 95.3% | 92.9% |
| Snow | 99.0% | 97.1% | 96.2% |
| Clutter | 72.8% | 93.2% | 69.2% |
| Overall Acc. | 96.0% | | |
| Mean IoU | 86.4% | | |

**Table 43.** Results for 3D FC DenseNet.

| | Recall | Precision | IoU |
|---|---|---|---|
| Water | 93.4% | 88.5% | 83.3% |
| Ice | 93.1% | 93.7% | 87.6% |
| Snow | 97.4% | 97.3% | 94.8% |
| Clutter | 69.9% | 84.6% | 62.1% |
| Overall Acc. | 94.2% | | |
| Mean IoU | 81.9% | | |

**Table 44.** Results for LSTM FC DenseNet.

| | Recall | Precision | IoU |
|---|---|---|---|
| Water | 94.6% | 91.1% | 86.6% |
| Ice | 93.3% | 97.4% | 91.1% |
| Snow | 99.5% | 94.2% | 93.8% |
| Clutter | 63.9% | 95.8% | 62.2% |
| Overall Acc. | 94.7% | | |
| Mean IoU | 83.4% | | |

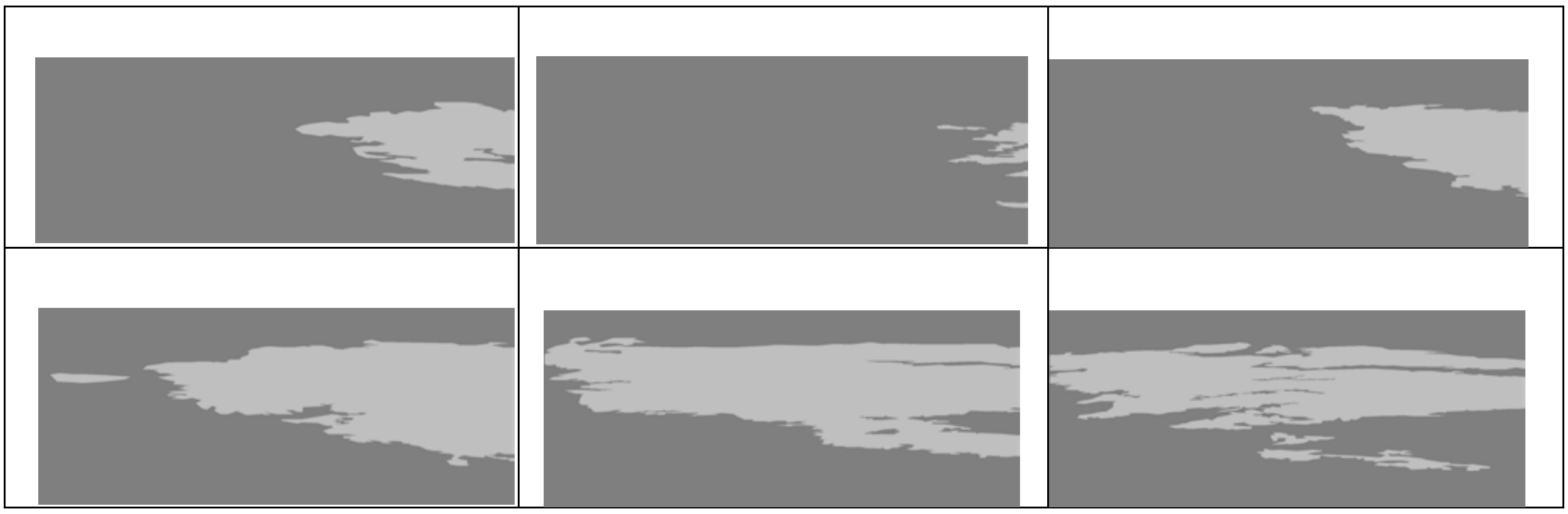

**Fig. 75.** Ground truth label maps of six consecutive images during the freezing period. From left to right, top to bottom: t=1 to t=6.

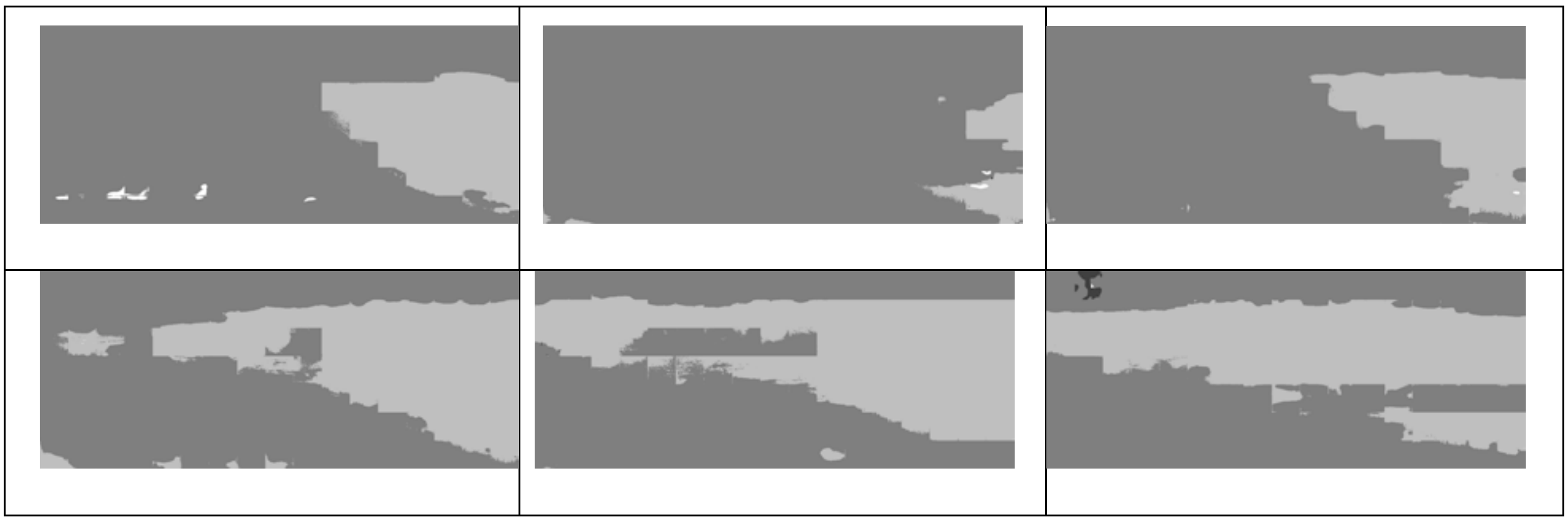

**Fig. 76.** Segmentation maps derived by single frame FC DenseNet. From left to right, top to bottom: t=1 to t=6.

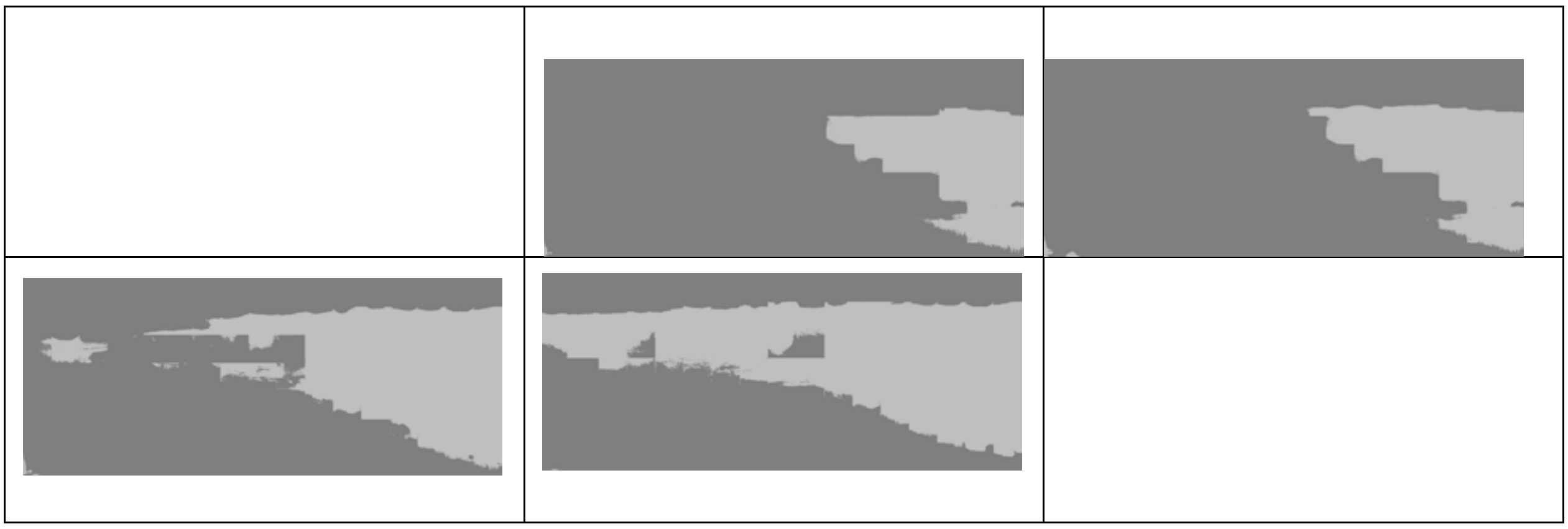

**Fig. 77.** Fused single frame FC DenseNet segmentation maps, median of 3 consecutive frames t-1, t, t+1. (see Section 7.5.8). From left to right, top to bottom: t=2 to t=5. Single frame segmentations for t=1 and t=6 were used to process the label maps t=2and t=5.

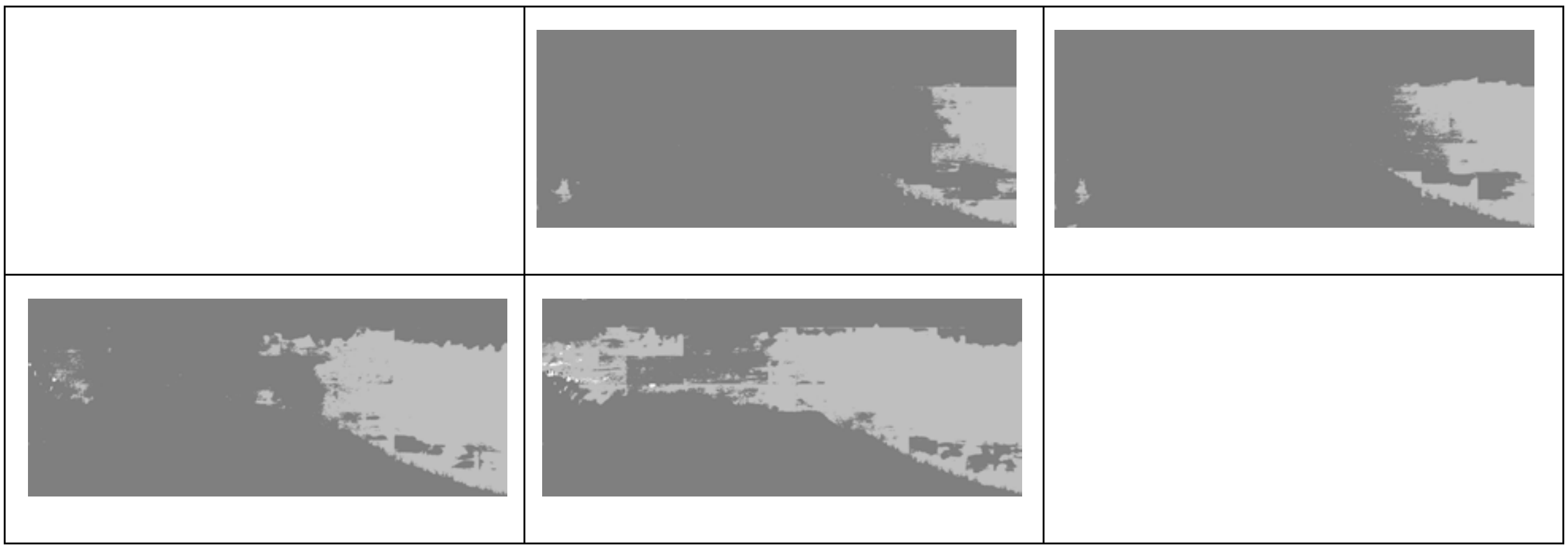

**Fig. 78.** Semantic segmentations generated by 3D FC DenseNet (see Section 7.4.2). From left to right, top to bottom: t=2 to t=5. The final segmentation is for t=i is based on the 3-tuple of RGB images at t=i-1, t=i and t=i+1. RGB images at t=1 and t=6 were used to process the label maps at t=2 and t=5.

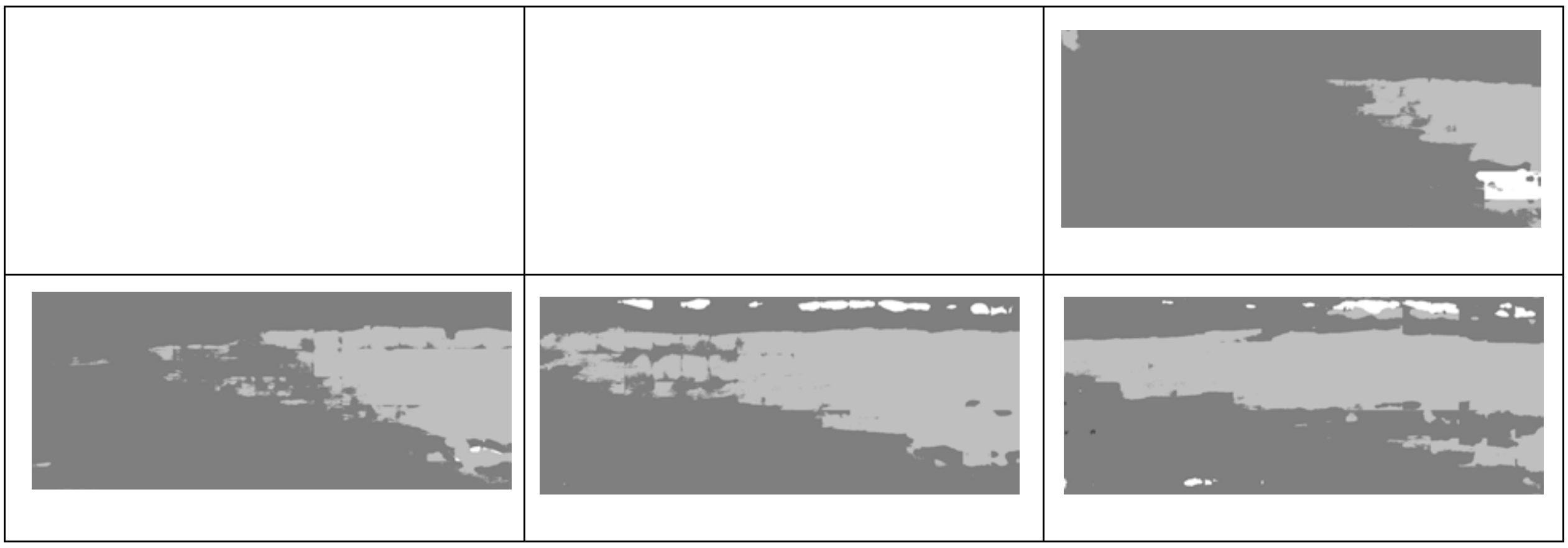

**Fig. 79.** Semantic segmentations generated by LSTM FC DenseNet (see Section 7.4.2). From left to right, top to bottom: t=3 to t=6. The final segmentation is for t=i is based on the 3-tuple of RGB images at t=i, t=i-1 and t=i-2. RGB images at t=1 and t=2 were used to process the label maps at t=3.

### 7.6. Summary

In general, we found a good distribution of publically accessible Webcam data of the defined lakes. However, at two of the six lakes (Lake Aegeri and Lake Sihl) only a limited number of Webcams are available such that only a restricted area of the water body is observed. Manually installing Webcams could solve this issue. Although the two Webcams used for the tests in this report were not affected, we found this to be an issue for some of the other Webcams. Despite the limited image quality, we obtained promising results for the implemented semantic segmentation of the tested Webcam images. Within our tests on data captured in winter 16/17 for the main classes water, ice and snow recall and precision values over 85% were obtained for high- and low-resolution Webcam data. In particular a simple median based technique to fuse predictions of multiple images captured on the same day, gave good estimates of the percentage of frozen water body. Throughout all tests, semantic segmentations of high-resolution Webcam data showed better quality than the low-resolution one. For example, models for the high-resolution Webcam generalized better for data of the winter 2017/18, despite large variations of appearance (different ice patterns). More precisely, the overall accuracy decreases by 13.1 % for Cam21 and 6.4 % for Cam24. Furthermore, it was possible to train models on images of both winters loosing only marginal accuracy. For such models, daily predictions for the Period1 seem to deliver even better results than models trained on Period1 only. This indicates that well-performing, deeper networks featuring larger number of parameters can be trained for multiple winters. Such models should generalize better to new, unseen data. In contrast, generalization capabilities across the two selected cameras was poor, presumably due to large differences in scale and appearance. Models trained on samples from high- and low-resolution Webcams together resulted in better predictions, although, compared to using samples from a single camera only, lower performance was achieved. In future work we want to experiment with data augmentation and scaling of samples to lessen this problem. Our tests indicated that excluding samples capturing mixed-states of the water body (for which manual annotation is labour intensive) for training leads to a slight decrease in prediction performance. Samples of one-class images seem not to fully compensate for the removed mixed-class samples: increased error rates were observed in particular in the freezing period. This holds also true after fusing semantic segmentations to form daily predictions of ice coverage. For the high-resolution Webcam however, ground truth is rather well reproduced. Exploitation of temporal cues for the Webcams was tested using two different network architectures and median-based filtering of consecutive frames. We found that, for latter approach static scene parts segmentations were slightly more robust. To this end, for dynamic scene parts, no improvement compared to single-frame segmentation was achieved. The two implemented 3DConv and LSTM approaches were not able to learn complex dynamic patterns. Although some kind of smoothing behaviour was learned, the models were slightly outperformed by the simple median-based approach. However, the overall accuracy is still larger than 94% for all evaluated temporal models. The availability of more trainings data could improve results. In future work we also want to experiment with balancing in the time domain (static vs. dynamic). Overall, for Webcam data featuring sufficient spatial resolution, we see a great potential for lake ice monitoring.

# 8. FINAL EVALUATION, COMPARISON AND INTEGRATION OF METHODS

## 8.1 Introduction and summary

Here, our aim is to make a proposal for lake ice monitoring mainly in Switzerland and secondarily for continental/global use.

We first make a summary of used data and methods and describe some major parameters, in order to have a compact overview. Note that data of meteorological stations "close" to the target lakes were not directly used, visual observations from lake shores did not exist, and information from Internet sources (e.g. e-media) was scarce, thus not mentioned in Table 45.

**Table 45**. Parameters of used data.

| **Parameter** | **MODIS** | **VIIRS** | **Webcam** | **In-situ** |
|---|---|---|---|---|
| Temporal resolution | 1-2 / d | 1-2 / d | Down to every 10 min, at least every h | Down to every 10 min, at least every h |
| Spatial resolution | 250-1000m | 371x387m (I bands), 742x776m (M-bands) | Very variable from ca. 4mm to 4m (for 10m to 10km distance); useful distance to lake for automated processing ca. 1km (without zoom, and reasonable viewing angle) | At least 2 loggers close to the surface (dz = 0.5 m). Ideally ~10 sensors with a spacing increasing with depth |
| Spectral resolution | 36 bands, 0.41-14.24 μm, only 12 useful bands used, s. www.eoportal.org | 22 bands, 0.41-12.01 μm, only five I-bands used, s. www.eoportal.org | RGB | NA |
| Radiometric resolution | 12 bits | 12 bits (14-bit ADC) | 8-bit, many compression artefacts | NA |
| T-Resolution | NA | 0.1 K at 300K | NA | 0.01 K |
| Costs | free | free | free | Mooring: min 2K CHF/ 5 y and 1-2 monitoring persons twice a year |
| Availability | Very good | Very good | Very good (for some lakes the camera distribution was not that good) | Need to be manually deployed and monitored |
| Security of availability | Very good | Very good | Good, very few gaps, but no guarantee | Good to very good |
| Future availability | No continuity | Continuity guaranteed for several years | Sensors expected to increase in number, stability of availability not guaranteed | On demand |
| Cloud mask problems | Slight | Slight | NA | NA |

[1] The GSD is computed as explained in Section 7.2.

Table 46 gives a summary of characteristics of the processing methods.

**Table 46.** Characteristics of used processing methods.

| Parameter | MODIS | VIIRS (ETHZ) | VIIRS (UniBe) | Webcam (auto) | Webcam (manual) | In-situ |
|---|---|---|---|---|---|---|
| Automation | High, whereby for training we only use the completely frozen or completely non-frozen days, which do not need pixelwise labelling (as with CNNs). | High, training as for MODIS | High; no training needed | High, except training. Training of CNNs requires more labelled samples than traditional ML, moreover pixel-wise segmentations are necessary. | none | Moderate, more work needed |
| Computer time (speed) on common PCs | Order of few min | Order of few min | Order of few minutes for ice retrieval. Atmospheric correction need some minutes. | Computer intensive, needs GPUs | Slow, 1-2 d for one Webcam and season | Very little |
| Near-Real-Time Response | Y | Y | Y | Possibly (excl. training) but not easy | N | possible but more expensive (cable-linked monitoring with data transmission, ~10k CHF /mooring) |
| Need of training | Y | Y | N | Y | N but ex-perience required | N |
| If above Y, time amount for training | Medium (but can be potentially reduced) | Medium (but can be potentially reduced) | | Large (but can be potentially reduced or eliminated) | | |
| Pre-processing needed | Y, largely automated | Y, largely automated | Y, largely automated | Y, largely manual | NA | NA |
| Accuracy (in no. of days (d) for ice-on/off) | Best case: 1 d (ice-off of Sils, ice-on of Sils, St. Moritz)<br><br>Worst case: 4 d (ice-off of Sihl)<br><br>Main problem, mixture of water and thin ice (also little training data with ice). Ice- | Best case: 1 d (ice-off of Sils, ice-on of Sils, Silvaplana)<br><br>Worst case: 3 d (ice-off of Silvaplana)<br><br>Note: | ± 3 d (winter 2016/17) | Due to data gap only ice-on available. Low-resolution. camera: ±1 d, High-resolution camera: ±0 d | | ± 3 d |
| | on/off errors increase, when cloudy days close to these dates | Accuracy of St. Moritz ignored for this analysis as it was tested using mixed pixels | | | | |
| Cloud problems | Severe | Severe | Severe | Much less than satellites (< 10%) | Much less than satellites (< 10%) | None |

## 8.2 Monitoring methods used in the project

### 8.2.1 Optical satellites of MODIS/VIIRS type

Such sensors are definitely useful for lake ice monitoring (especially for global coverage) and the results achieved show a high accuracy level. However, there are several problems, with the first being by far the major one:

- Clouds, especially during lake icing, freezing and melting
- Cloud mask errors
- Coarse GSD (lakes of up to 2 $km^2$ problematic to monitor)
- Mixed pixels and for small lakes larger (e.g. of up to 1 pixel) absolute co-registration errors can cause misclassification of pixels
- If processing needs training, little cloud-free data with (black) ice.
- Mixture of thin ice and water; little training data for only ice.

MODIS data are expected to be discontinued but VIIRS operation is guaranteed over a longer future period (JPSS-1/NOAA-20 until 2024; JPSS-2 with same suite of sensors will be launched in 2021 with designed life time of 7 years; JPSS-3 and -4 are in the planning phase). Especially, the higher spatial resolution I-bands are useful.

### 8.2.2 Webcams

- Generally good ice detection ability with much higher spatial resolution compared to satellites
- No control of position, orientation, lake area coverage, image artefacts, generally poor image quality, some data delivery gaps
- Limited number of publically available Webcams for some lakes
- Much less cloud problems than satellite images
- An extensive number of labelled samples are needed to train CNN models. However, tests regarding the generalization capabilities by integrating training data from both spatial resolution cameras gave promising results.
- Most difficult parts were with floating ice (thin, transparent ice; also its movement caused problems in the multitemporal processing, while there were few training data for these cases).
- Generally, results for the high-resolution Webcam were better than the low-resolution one.
- Could be used also for other environmental purposes (especially new better Webcams, see Section 9.3)
- Especially suitable for small lakes (ca. up to 2 $km^2$), which cannot be monitored by VIIRS-type sensors.
- If processing needs training, less data exist with (black) ice than water and snow.

### 8.2.3 In-situ measurements

Their usage was definitely useful and complemented the other monitoring modes. They also relate to lake properties other than icing, providing a physical interpretation using various models. It is believed that a correct view of the process with only one mooring in the pelagic zone is possible, although we obviously miss the dynamics in the near-shore (and in the bay). This method will work better in round-shaped lakes with regular bathymetry. For lakes such as Lake Sils (multi-basin shape) or Lake Sihl (irregular bathymetry), more mooring may be needed. In any case, in-situ observations (including nearby meteorological observations) offer the possibility to understand and finally correctly model and predict the mechanisms associated with ice formation and melting.

**Summarising** for all monitoring methods (apart in-situ) black ice is a problem, although it could not be quantified to what extent within this project. An overall problem is the lack of extensive and reliable reference data. We believe that this problem cannot be solved in future by manual observations, but by improvement of each monitoring method (providing also reliability of the results) and weighted integration of the various monitoring results.

## 8.3 Relation of monitoring data to lake parameters

The analysis below refers to Swiss lakes, where high altitude and mountains are very strongly correlated. This is not the case of large plateau areas. In addition, Swiss frozen lakes have a rather small area. For global monitoring and much larger lakes, in higher latitudes and often in flat terrain, satellite observations are the most efficient and generally the only option.

Lake parameters/attributes not treated in Table 47:

- lakes used as reservoir or having a dam: there is apparently no influence for in-situ measurements.
- lakes with inflows/outflows have more complicated dynamics and spatial variability due to the river inflow. The area close to a river inflow is expected to freeze later/melt earlier than the rest of the lake. Similar but lesser effects at river outflows.

**Table 47.** Influence (No/Yes) of various lake parameters on monitoring data.

| **Monitoring data** | **Area** | **Average Depth** | **Altitude** | **Surrounding topography** (flat/hilly, mountainous) |
|---|---|---|---|---|
| Medium-resolution optical sensors | Need at least a few clean pixels (area > 2 $km^2$ for VIIRS) | No | At higher altitude more clouds expected and more shadows | At mountainous areas more clouds expected and more shadows |
| Webcams | Max usable distance camera to lake about 1km (automated processing, no zoom, reasonable viewing angle) | No | At high altitude slightly more clouds, less buildings, larger viewing angle, more variable illumination conditions and shadows (for diurnal multitemporal processing) | At mountains slightly more clouds, less buildings, larger viewing angle, more variable illumination conditions and shadows (for diurnal multitemporal processing) |
| In-situ | More moorings will be needed for multibasin lakes (e.g. lake Lucerne) | Yes. Deeper lakes typically freeze | In higher altitude modelling more complicated (require local weather information) | At mountains modelling more complicated (require local weather information) |
| | | later due to larger heat amount stored internally | | |
| **Other monitoring data (not used in this project)** | | | | |
| Sentinel-1 | No (for lakes > ca. 0.05 km$^2$, 103 CH lakes > 0.3 km$^2$ [1]) | No | At high altitude processing more difficult, more wind | At mountains processing more difficult, more wind |
| Sentinel-2 | No (for lakes > ca. 0.05 km$^2$, 103 CH lakes > 0.3 km$^2$ [1]) | No | At higher altitude more clouds expected and more shadows | At mountains more clouds expected and more shadows |
| Better Webcams (Pan-Tilt-Zoom (PTZ), better image quality, possibly more pixels) | With PTZ larger area coverage and at longer distances | No | At high altitude slightly more clouds, less buildings (less problem with Pan and Tilt), larger viewing angle (even better with tilt), more variable illumination conditions and shadows (for diurnal multitemporal processing) | At mountains slightly more clouds, less buildings (less problem with Pan and Tilt), larger viewing angle (even better with tilt), more variable illumination conditions and shadows (for diurnal multitemporal processing) |
| Local mini meteo stations (related mostly to in-situ) | Influence small for freezing lakes in CH (almost never very large) | No | With high altitude, more needed due to greater and local weather variations | With mountains, more needed due to greater and local weather variations |
| UAVs | The larger the area the worse, up to no daily coverage possible with one medium price UAV | No | More cumbersome installation if not local, less flying time and area covered | More cumbersome installation if not local, less flying time and area covered |
| Crowd-sourcing/ Internet-social media images | The larger the area the worse, usually only area close to shore visible | No | Mountain/sky outlines can be used for georeferencing | Mountain/sky outlines can be used for georeferencing |

[1] There are about 215 mountain lakes (height > 800m) in CH, with area > 0.04 km$^2$, of whom 30 with area > 1 km$^2$.

## 8.4 Results of monitoring methods

Detailed results are listed above in the sections for each monitoring method. An important point is the ice-on/off dates detected by each method. They are listed in Appendix 3. Still more detailed results for each individual day are listed in Appendix 6.

From the tables of Appendix 3, one can notice partly significant differences (exceeding the +/- 2 days GCOS requirement) for some of the ice-on/off dates. This is partly due to weaknesses of the monitoring methods, already mentioned above. In some cases, there were also gaps in the acquired data, especially due to clouds. Clouds at or close to the "true" ice-on/off dates can lead to significant errors with the satellite images. An additional source of minor errors may lie in the wrong interpretation by each researcher of the ice-on/off definitions. Additionally, at Lake St. Moritz with its small area, satellite results are expected to be less reliable (and are the less accurate results, while for the larger lakes Sihl and less Sils, the results are better). A precise statement on which method is more accurate is not possible, as the main source of ground truth (visual interpretation of Webcams) had also its own uncertainty. However, when multiple methods deliver the same or very similar dates, this is an indication that these results are most probably correct. The tables of Appendix 3 list three methods for the processing of the in-situ data. EAWAG recommends the first method based on analysing the correlation between two closely located temperature sensors. The method based on the fluctuations of temperature (wavelet analysis) may be too complicated (requires expert user). The first method also gives the best results compared to the ground truth. In the following analysis of the results of Appendix 3 with in-situ we mean the first method. The Webcams give good results compared to the ground truth and in-situ and the results consistent between Webcams 21 and 24. However, this refers only to one lake, so it cannot be generalized. In-situ results are quite good (only for ice-on of Sihl and ice-off of Silvaplana the error was more than +/- 2 days). In-situ results are more stable than satellite observations, as they are not influenced by clouds, cloud mask errors etc. The ground truth could be improved by using better quality Webcams with PTZ capabilities. The two VIIRS methods give results within +/- 2 days, except for ice-off of lakes St. Moritz and Sihl. Differences between the two can be also due to different ice-on/off definition usages. E.g. in Appendix 6 UniBe gives results without probabilities, while ETHZ uses and list probabilities.

As a final conclusion, the integration of different input data and processing methods can significantly increase the precision and reliability of the estimation of ice-on/off dates. However, for a correct combination of the (usually different) results of each method, it is suggested that its method also provides a reliability measure for the estimation.

### 8.5 Suggestions for integration of monitoring methods

We propose to consider the following monitoring data sources (see also Section 9 for new sensors and processing):

1. Optical sensors of VIIRS-type (major problems clouds and small lakes)
2. Active microwave sensors (especially Sentinel-1). Available methods need careful adaptation for Swiss application (e.g. combining ascending and descending orbits; adaptation of algorithms; validation)
3. New, better Webcams (very high temporal resolution, can reduce severely cloud problems, suited for small lakes, can monitor other environmental parameters, processing needs improvement to reduce labelling and training)
4. If needed, usage of Sentinel-2, to cover cloud-free lakes, not monitored by 1-3.
5. In-situ measurements. We suggest coordination with other Federal Agencies, especially FOEN. We suggest continuing the lake ice monitoring allowing longer time series for better understanding the dynamics of the ice covered period (e.g. cooling, freezing, warming, melting). Such dataset will also provide the required information for a precise validation of the ice module in mechanistic hydrodynamic models (e.g. simstrat.ewag.ch). In relation to this monitoring method, usage of better Webcams close-by and new, cheap, local meteorological stations could be of advantage.

For continental/global usage, the methods 1, 2 and 4 are clearly preferable, stressing the importance of satellite observations for large area coverage. Clearly, for focused, small area campaigns and provision of reference data, other monitoring methods like 3 and 5 or UAVs and other airborne platforms could be employed. In relation to international activities, we would like to mention some information on ESA's CCI+ project as listed below.

In the frame of ESA's CCI+ call for new projects also the ECV "Lakes" was included. The consortium addressed the following variables: lake water level, lake water extent, lake surface water temperature, lake ice and lake water reflectance. The variable lake ice will be mainly rely on daily Level 3 Lake Ice Cover (LIC) at 500 m and 1 km resolution derived from MODIS (Aqua/Terra; from 2000) and VIIRS (2012-onward) (Duguay et al., 2015b; Kang and Duguay, 2016). The general approach to be developed in Lake CCI+ will be designed in such a way that new satellite data from Sentinel-3 (OLCI/SLSTR) can be integrated for generation of a 300-m product (2016-onward). Prototype Sentinel-3 and synergy Sentinel-3/Sentinel-1 (SAR; EW mode) ice cover products are envisaged for a small subset of lakes, to be compared to that derived from MODIS/VIIRS for an overlapping period. H2O Geomatics (https://www.h2ogeomatics.com/about) has recently developed an algorithm for the retrieval of lake ice cover from dual-polarization RADARSAT-2 data, which could be integrated into the system and tested on Sentinel-1A/B data (e.g. 2017-2018) for a few lakes, in order to assess in a preliminary manner a multi-frequency approach for the generation of a daily 300-m LIC product (GCOS requirements of spatial and temporal resolutions). More details about the quality of retrieval using RADARSAT-2 data are described in Murfitt et al. (2018).

## 9. OUTLOOK

This section deals with sensors/platforms and processing not investigated in this project or extensions of them. They could facilitate an improved estimation of lake ice. The comments below should be treated with caution, as we did not treat them within this project. The caution refers mainly to operational applicability of UAVs and crowdsourcing/Internet-social media images.

### 9.1 Usage of microwave data (mainly Sentinel-1)

The problem of clouds with optical satellite sensors can only be solved using microwave sensors. In particular, we suggest usage of Sentinel-1, which are freely available for all. The two identical satellites provide a very good temporal resolution (checking the respective archive for one year and a rectangle surrounding Switzerland, an average of 1.3 images / 2d were acquired, though not evenly timely distributed. In the simplest case, methods could be derived to detect lake ice using only the amplitude (ETHZ has done very preliminary investigations on that). On top of that, polarimetric and interferometric (e.g. coherence information) data processing can provide additional information.

### 9.2 Usage of higher resolution optical satellites (mainly Sentinel-2)

Optical sensors can provide an easier interpretation than microwave ones and better than low-resolution optical sensors like MODIS and VIIRS. Sentinel-2 are particularly attractive, as they carry newest sensors with 13 spectral bands (covering VNIR and SWIR) and GSD down to 10m, have a revisit period of 5 days at equator and are freely available for all. Together with Sentinel-1 data they could provide almost daily coverage of most lakes. However, it has to be mentioned that cloud masking is not excellent due to missing thermal band on the Multispectral Imager (MSI).

### 9.3 Better Webcams

As mentioned in the report, Webcams have several advantages, but also disadvantages. For the latter, we suggest the usage of better image quality Webcams, with Pan-Tilt-Zoom (PTZ) capabilities (examples of mainly pan cameras (not common PTZ ones) at https://lejdastaz.roundshot.com/ and https://sils.roundshot.com/). Positioning should be at carefully selected places, especially at buildings with a more secure public control. Thus, with careful positioning, a larger lake area could be covered within the image format, with shorter distance to the lake, and larger intersection angle between the optical axis and the lake surface. Furthermore, with pan and tilt a larger lake area can be covered, while zoom allows to better discriminate between difficult cases (e.g. black ice vs. water), while also allowing investigations on the influence of spatial resolution on the results and allowing generation of richer training datasets, thus reducing time for manual labelling and training. As mentioned in Table 45 (spatial resolution of Webcams) the useful distance between standard Webcams and lake for reliable automated interpretation is about 1 km. With zoom capabilities this range can be extended, however caution should be paid to stitching the multiple overlapping images covering a lake. PTZ cameras can allow also monitoring of other important environmental parameters, like snow cover, land cover, cloud coverage (if they can point towards the sky) etc.

### 9.4 In-situ and related issues

One-dimensional mechanistic models could be used as integrator of information (e.g. remotely sensed observations, Webcams data and in-situ measurements) for lake ice monitoring and forecasting. While such models are relatively easy to setup for lowland lakes (see https://simstrat.eawag.ch/), the quality of the results becomes problematic for alpine lakes (e.g. lakes subject to ice formation) due to the lack of closely located meteorological information. Furthermore, in such systems local atmospheric conditions often strongly differ from the synoptic conditions (e.g. change in altitude, mountains inducing local wind, clouds etc.). We therefore recommend expanding this study by defining pilot sites where in-situ lake temperature, meteorological data (see Section 9.5) and new better Webcams would be integrated. Such pilot sites could be defined in collaboration with the Federal Office of Environment (FOEN), specifically for the high frequency (hourly) in-situ lake temperature observations to optimize resources. Finally, to minimise exploitation costs of such pilot sites, we suggest investigating the potential of developing a strategy similar to the high-altitude (>1800m) lake Sentinel monitoring system at Haute-Savoie (France) (http://www.lacs-sentinelles.org/en) based on synergistic collaborations between scientists, agencies and stakeholders (note: the aim here is not monitoring lake ice but other lake parameters).

## 9.5 New, cheap and compact meteorological stations

We believe it would be an advantage to have compact, cheap meteorological stations close to the lakes, possibly next to new and better Webcams as mentioned above. Such stations could be also used for weather forecast and should include at least measurement of air T, wind speed/direction and solar radiation. Their cost should be about 4-5K CHF.

## 9.6 UAVs

ETHZ has done some experiments in Lake Sils with RGB and thermal cameras. Though the data have not been fully processed, it is believed that snow, ice and water can be separated. However, there are problems for use of this technology, at least in the short-term future. There are restrictions regarding flight regulations (which are expected to be newly determined in CH). For UAVs of medium price range (ca. 2K CHF) flight time is too little (especially for low temperatures and high altitude). There is a lot of manual work and time investment and the degree of automation for data acquisition is low. Combination of images in a block, by using automatically determined tie points, can be difficult in absence of texture (especially fresh snow), though if direct georeferencing is accurate enough, this should not be a problem. Thus, we currently see the usage of UAVs only for specific lake related tasks (e.g. detection of methane holes in St. Moritz lake), and validations, e.g. in specific field campaigns. If in the future, flight permissions allow (especially regarding where to fly and how high and without visual contact), technology (at a reasonable price) allows much longer flying times, and usage (including flying back to a self-charging station) can be automated, then small and medium size (e.g. up to 10 $km^2$) lakes could be observed. An example of automated UAV usage with fewer restrictions in Switzerland is the activities of the company Meteomatics for detailed weather prediction (http://www.meteomatics.com/display/RESEARCH/Meteodrones). However, even in this case, one should compare advantages and disadvantages with other monitoring methods.

## 9.7 Crowd sourcing/Internet-social media images

There is no doubt that there is a great number of images available in the Internet, generally, but also in specific image collections (e.g. Flickr) or in the social media, with their number increasing very rapidly. With the advance of smartphones with cameras continuously improving and the habit of selfies, many personal images with background a lake (frozen or not) are becoming more and more available. There have also been investigations on how to georeference such images, if they are not geo-tagged by usage of GNSS in the phone. However, there are some difficulties in exploiting this data. Images are taken totally randomly and it is not easy to find where the lake is. Georeferencing is not always available, and even if yes it is sometimes not correct and it gives coordinates but not orientation of the optical axis. Data exist more for touristy lakes, not necessarily those that freeze. In addition, the field of view and lake coverage varies greatly. Thus, compared to other monitoring methods, we currently do not suggest using this data as a high priority.

## APPENDIX 1. LIST OF WEBCAMS

**Details of all downloaded Webcam data (2016-2017)**

| CamID | Start Date | End Date | Gap dates | Lake | Acquisition Frequency | Camera Type | Image Dimensions (pixels) (W x H) |
|---|---|---|---|---|---|---|---|
| Cam1 | 1.12.16 | 12.6.17 | Gap1, Gap2 | Greifen | 5 min | Fixed | 1920 x 1080 |
| Cam2 | 4.12.16 | 12.6.17 | Gap1, Gap2 | Greifen | 10 min | Fixed | 1920 x 1080 |
| Cam3 | 4.12.16 | 12.6.17 | 8.3.17 - 17.3.17: showing same image<br>Gap1, Gap2 | Greifen | 10 min | Fixed | 1920 x 1080 |
| Cam4 | 4.12.16 | 12.6.17 | Gap1, Gap2 | Sils | 15 min | Fixed | 640 x 480 |
| Cam5 | 4.12.16 | 12.6.17 | Gap1, Gap2 | Sils | 30 min | Fixed | 640 x 480 |
| Cam6 | 4.12.16 | 12.6.17 | Gap1, Gap2 | Silvaplana | 30 min | Fixed | 640 x 480 |
| Cam7 | 4.12.16 | 12.6.17 | Gap1, Gap2 | Aegeri | 10 min | Fixed | 640 x 240 |
| Cam8 | 28.1.17 | 12.6.17 | Gap2 | Pfaeffiker | 60 min | Fixed | 640 x 480 |
| Cam9 | 15.12.16 | 20.6.17 | 7.4.17 - 27.4.17: caused by script | Sihl | 60 min | Fixed | 420 x 344 |
| Cam10 | 15.12.16 | 20.6.17 | 7.4.17 - 27.4.17: caused by script | Sihl | 60 min | Fixed | 400 x 224 |
| Cam11 | 4.12.16 | 12.6.17 | Gap1, Gap2 | Sihl | 10 min | Rotating | 704 x 576 |
| Cam12 | 4.12.16 | 12.6.17 | Gap1, Gap2 | Sihl | 5 min | Rotating | 704 x 576 |
| Cam13 | 4.12.16 | 9.1.17 | 23.12.17 - 28.12.16: caused by script<br>9.1.17 – end?: same image | Silvaplana | 10 pics/day | Fixed | 485 x 272 |
| Cam14 | 15.12.16 | 20.6.17 | 7.4.17 - 27.4.17 () | Silvaplana | 60min | Fixed | 640 x 480 |
| Cam15 | 4.12.16 | 12.6.17 | Gap1, Gap2 | Silvaplana | 60min | Fixed | 760 x 428 |
| Cam16 | 7.12.16 | 12.6.17 | Gap1, Gap2 | Silvaplana | 6-10 pics/day | Fixed | 1280 x 720 |
| Cam17 | 7.12.16 | 12.6.17 | Gap1, Gap2 | Silvaplana | 60 min | Fixed | 400 x 224 |
| Cam18 | 4.12.16 | 12.6.17 | Gap1, Gap2 | Silvaplana | 7 pics/day | Fixed | |
| Cam19 | 4.12.16 | 5.6.17 | Gap1, Gap2 | Silvaplana | 70 min | Fixed | 640 x 360 |
| Cam20 | 4.12.16 | 12.6.17 | Gap1, Gap2 | St. Moritz | 30 min | Fixed | 640 x 480 |
| Cam21 | 4.12.16 | 5.6.17 | Gap1, Gap2 | St. Moritz | 60 min | Fixed | 1920 x 1080 |
| Cam22 | 16.12.16 | 20.6.17 | 7.4.17 - 27.4.17; caused by script | St. Moritz | 60 min | Fixed | 400 x 224 |
| Cam23 | 4.12.16 | 12.6.17 | Gap1, Gap2 | St. Moritz | 70 min | Fixed | 640 x 360 |
| Cam24 | 4.12.16 | 12.6.17 | Gap1, Gap2 | St. Moritz | 60 min | Fixed | 1920 x 1080 |
| Cam25 | 16.12.16 | 20.6.17 | 7.4.17 - 27.4.17: caused by script | St. Moritz | 60 min | Fixed | 640 x 480 |
| Cam26 | 14.2.17 | 19.4.17 | nil | Sihl (EAWAG) | 3 hours | Fixed | 1280 x 720 |
| Cam27 | 16.12.15 | 13.4.16 | nil | Sihl (EAWAG) | 3 hours | Fixed | 1280 x 720 |
| **Gap1**: 17.2.17, 22.2.17 - 25.2.17, 28.2.17, 18.3.17 - 25.4.17, 24.5.17, 30.5.17, 6.6.17 caused by script | | | | | | | |
| **Gap2:** 24.1.17 - 25.1.17, 14.1.17 caused by script | | | | | | | |

**Details of all downloaded Webcam data (2017-2018)**

| CamID | Start Date | End Date | Gap dates | Lake | Acquisition Frequency | Camera Type | Image Dimensions (pixels) (W x H) |
|---|---|---|---|---|---|---|---|
| Cam1 | 1.12.17 | on-going | nil | Greifen | 5 min | Fixed | 1920 x 1080 |
| Cam2 | 1.12.17 | on-going | nil | Greifen | 10 min | Fixed | 1920 x 1080 |
| Cam3 | 1.12.17 | on-going | nil | Greifen | 10 min | Fixed | 1280 x 960 |
| Cam4 | 1.12.17 | on-going | 15.12.17 till 31.12.17: no upload by provider | Sils | 15 min | Fixed | 640 x 480 |
| Cam5 | 1.12.17 | on-going | nil | Sils | 30 min | Fixed | 640 x 480 |
| Cam6 | 1.12.17 | on-going | nil | Silvaplana | 30 min | Fixed | 640 x 480 |
| Cam7 | 1.12.17 | on-going | nil | Aegeri | 10 min | Fixed | 640 x 240 |
| Cam8 | 2.12.17 | on-going | nil | Pfaeffiker | 60 min | Fixed | 640 x 480 |
| Cam9 | 4.12.17 | on-going | nil | Sihl | 60 min | Fixed | 420 x 343 |
| Cam10 | nil | nil | Showing same image, practically useless | Sihl | 60 min | Fixed | 400 x 224 |
| Cam11 | 2.12.17 | on-going | nil | Sihl | 10 min | Rotating | 704 x 576 |
| Cam12 | 2.12.17 | on-going | nil | Sihl | 5 min | Rotating | 704 x 576 |
| Cam13 | nil | nil | discontinued | Silvaplana | 10 pics/day | Fixed | - |
| Cam14 | nil | nil | Showing same image, practically useless | Silvaplana | 60min | Fixed | 400 x 224 |
| Cam15 | 1.12.17 | on-going | 20.1.18 till 23.1.18: no upload by provider | Silvaplana | 60min | Fixed | 760 x 428 |
| Cam16 | 1.12.17 | on-going | 23.12.17 till 1.1.18: no upload by provider | Silvaplana | 6-10 pics/day | Fixed | 1280 x 720 |
| Cam17 | 2.12.17 | on-going | nil | Silvaplana | 60 min | Fixed | 318 x 220 |
| Cam18 | nil | nil | discontinued | Silvaplana | 7 pics/day | Fixed | - |
| Cam19 | 2.12.17 | on-going | nil | Silvaplana | 70 min | Fixed | 640 x 360 |
| Cam20 | 1.12.17 | on-going | nil | St. Moritz | 30 min | Fixed | 640 x 480 |
| Cam21 | 1.12.17 | on-going | nil | St. Moritz | 60 min | Fixed | 1920 x 1080 |
| Cam22 | 4.12.17 | 18.12.17 | Showing same image, practically useless | St. Moritz | 60 min | Fixed | 400 x 224 |
| Cam23 | 1.12.17 | on-going | nil | St. Moritz | 70 min | Fixed | 640 x 360 |
| Cam24 | 1.12.17 | on-going | nil | St. Moritz | 60 min | Fixed | 1920 x 1080 |
| Cam25 | nil | nil | Showing same image, practically useless | St. Moritz | 60 min | Fixed | 400 x 224 |
| Cam26 | | | no data acquired | Sihl (EAWAG) | 3 hours | Fixed | |
| Cam27 | | | no data acquired | Sihl (EAWAG) | 3 hours | Fixed | |

There are three new cameras with data covering many years (called roundshot):

- One 360 degree at Segl-Maria that includes cameras 5 and 6 or maybe is the same camera.
- One 360 degree (El Paradiso, called Cam28) which covers Champfèr and a small part of Silvaplana (not ideal but useful)
- One 360 degree covering a very small part of Aegeri and other lakes (not useful)

The link to the Website below shows all roundshot Webcams in Switzerland: http://www.roundshot.com/xml_1/internet/en/application/d170/f172.cfm

**Distribution of Webcams and lake coverage** (with zoom in the figure, the camera numbers listed above can be seen)

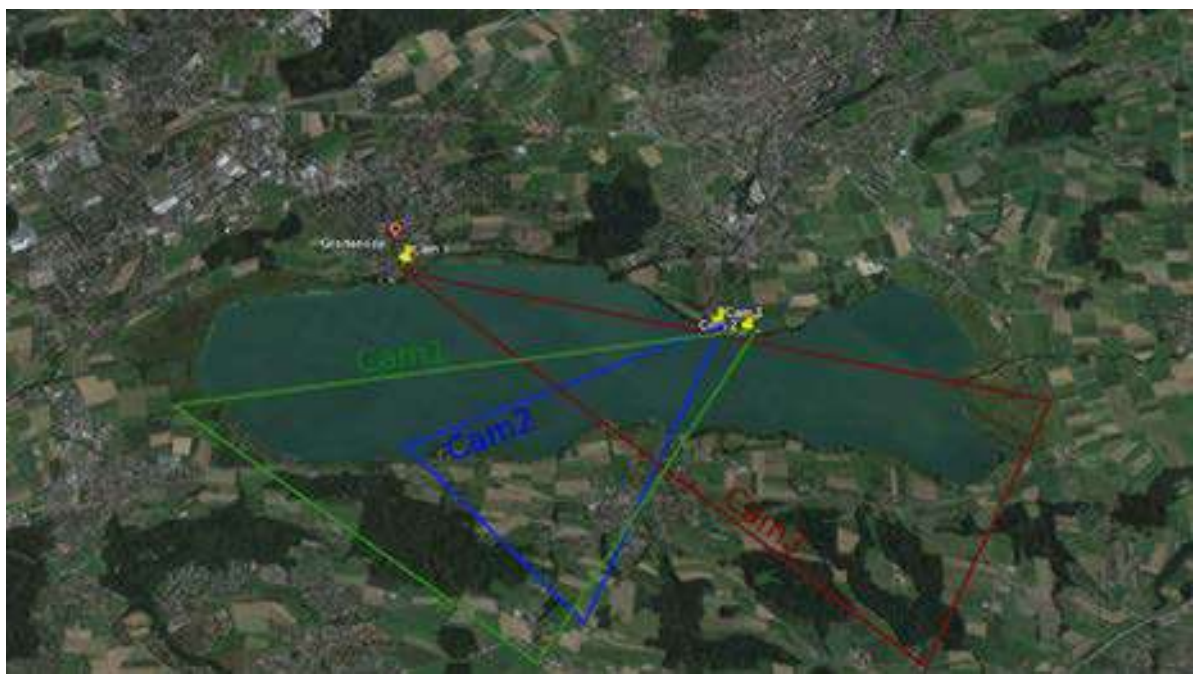

Lake Greifen

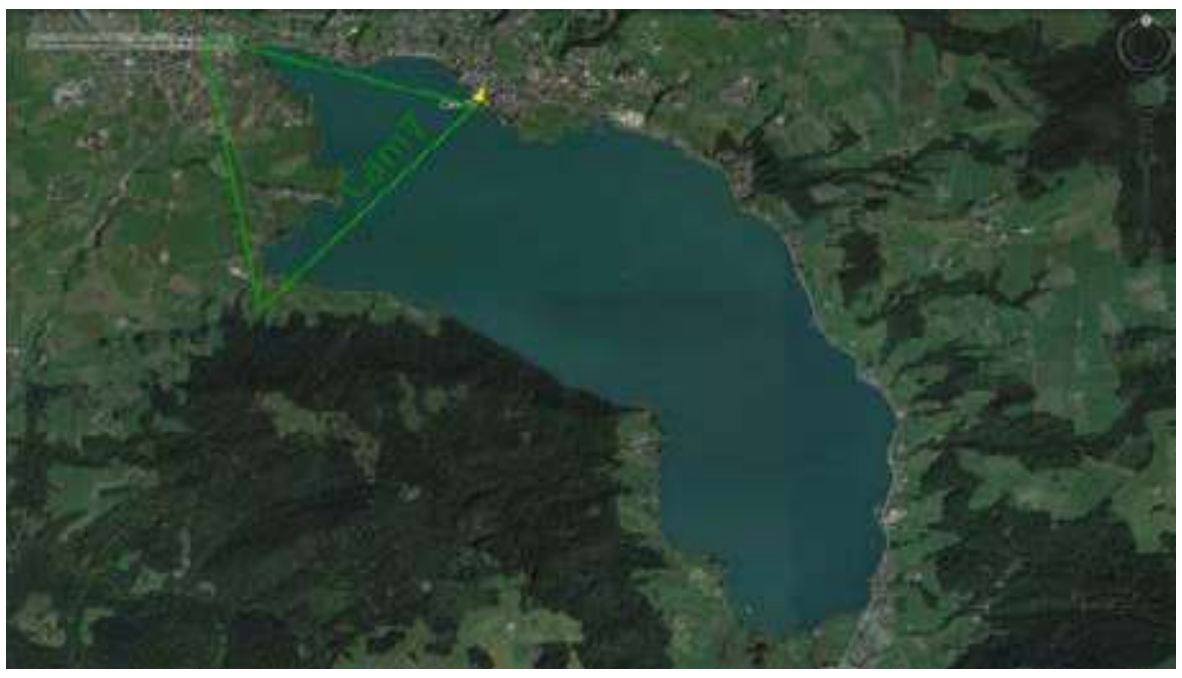

Lake Aegeri

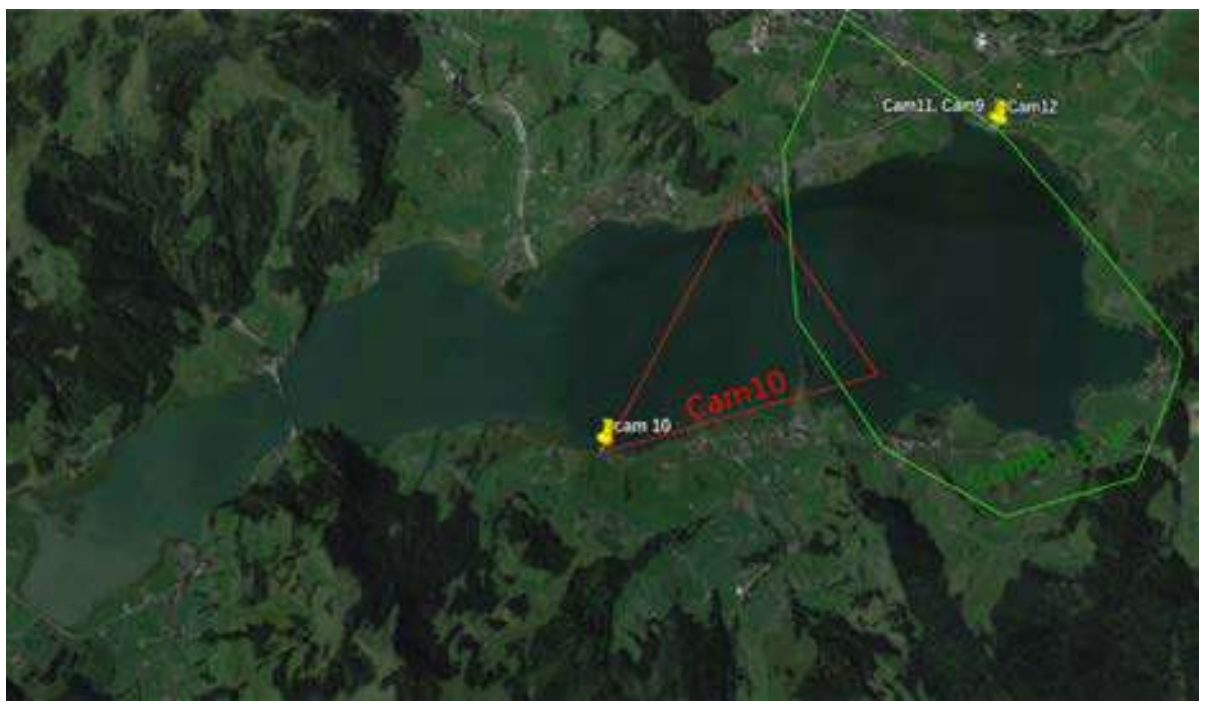

Lake Sihl

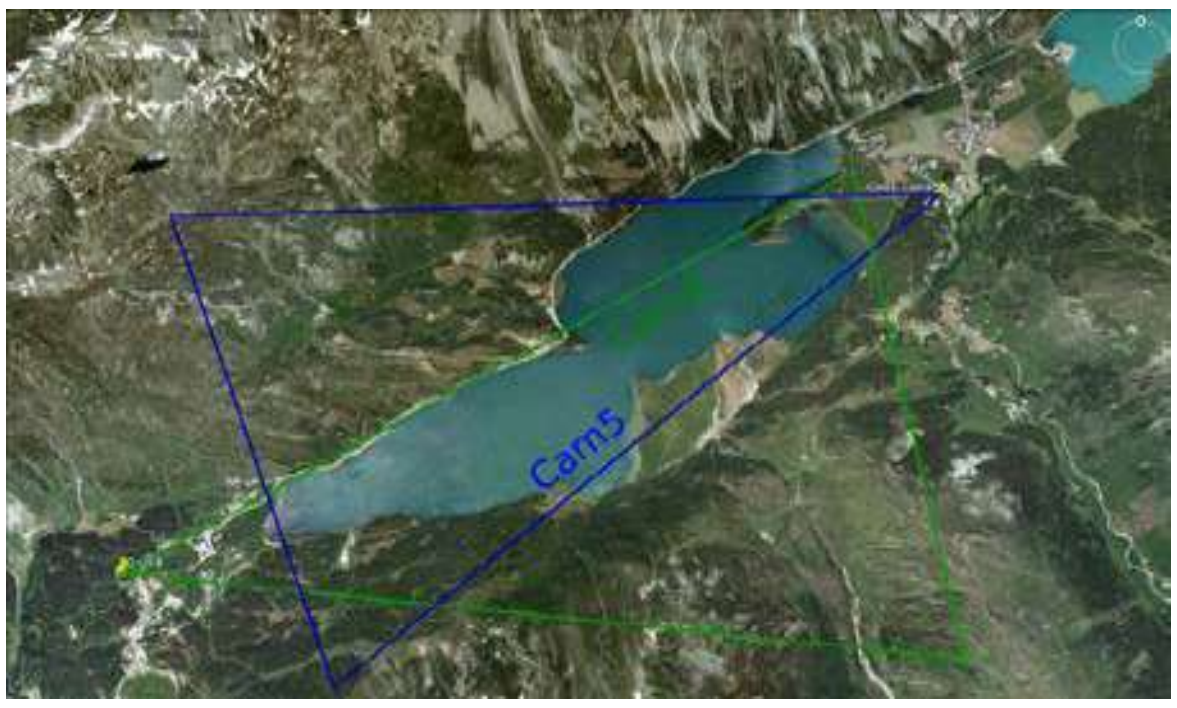

Lake Sils

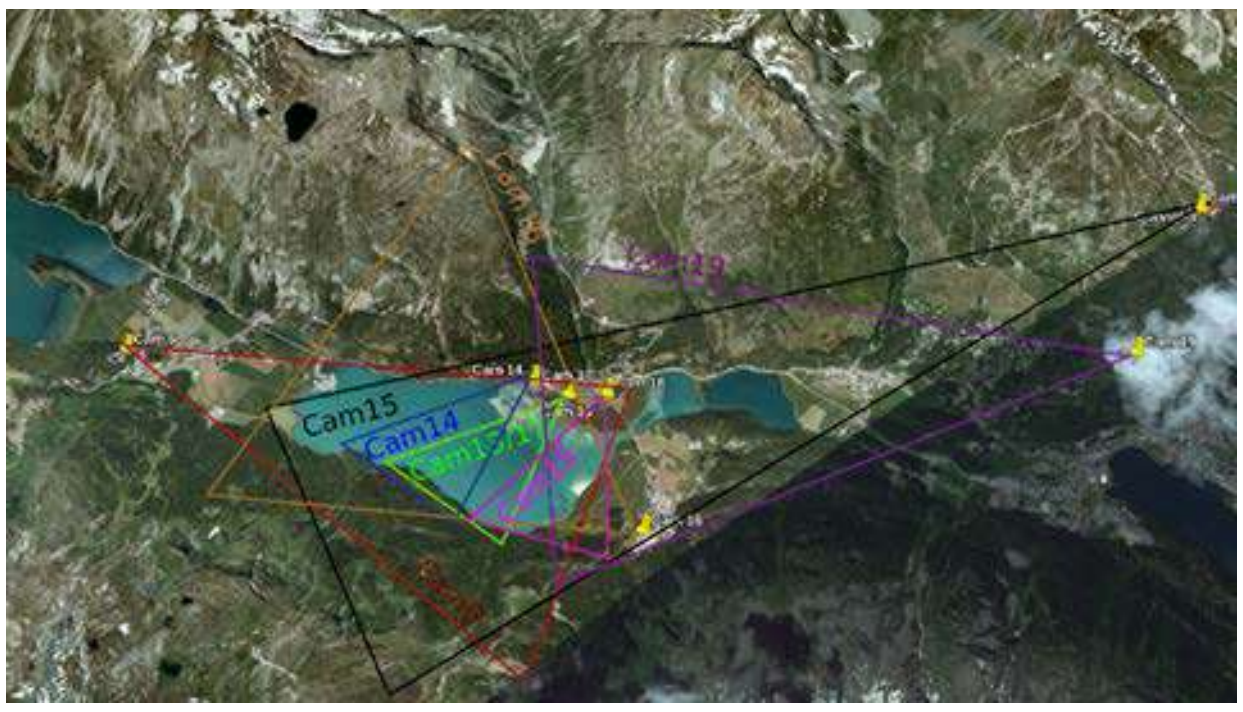

Lake Silvaplana

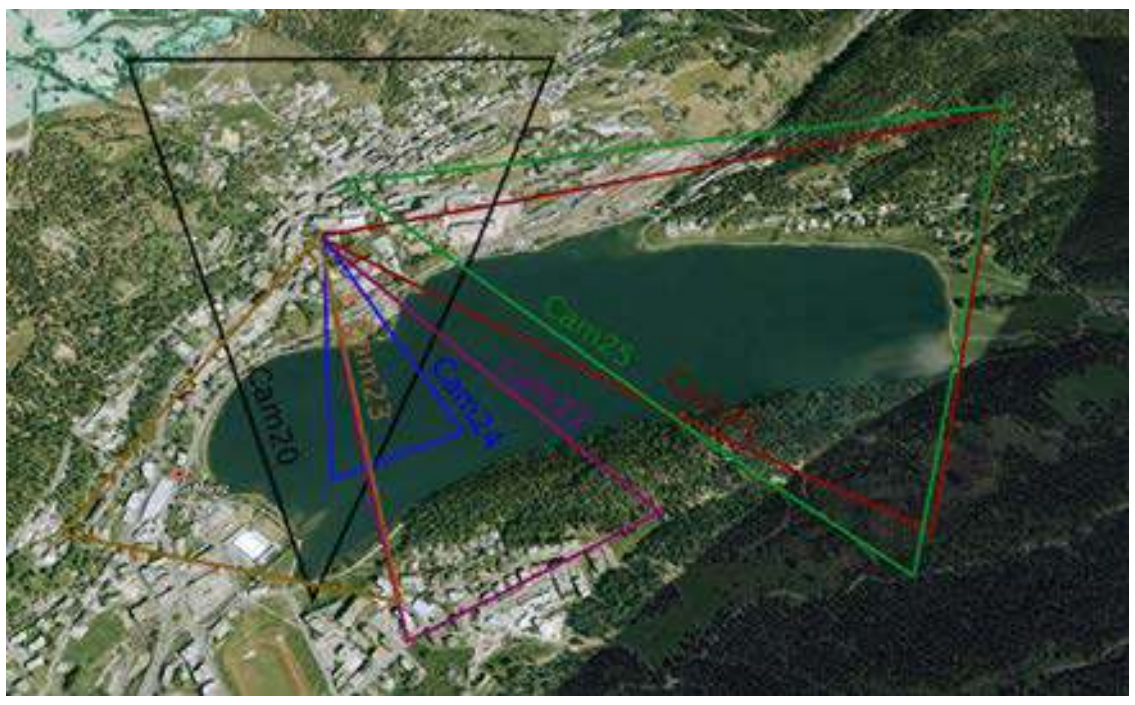

Lake St. Moritz

**Fig. 80.** Webcam distribution and lake coverage for the six target lakes.

## APPENDIX 2. DETAILED MEASUREMENTS OF EAWAG

In-situ installation. All loggers measured at following conditions:

- Temperature at 10 positions from 0.05 to 10 m: 0.05 m ; 0.12 m ; 0.25 m ; 0.5 m ; 0.9 m ; 1.3 m ; 1.9 m ; 2.7 m; 4.6 m; 10 m
- Pressure at 10 m (except Lake Silvaplana on May 2016 at 4.6 m).

**Year 2016 (4 target lakes)**
Lake: Lake Silvaplana
Installation: 7.1.2016
Removal: 6.7.2016

Lake: Aegeri
Installation: 19.1.2016
Removal: 20.4.2016

Lake: Sihl
Installation: 19.1.2016
Removal: 13.4.2016

Lake: Greifen
Installation: 6.1.2016
Removal: 18.4.2016

**Year 2017 (6 target lakes)**
Lake: Lake St. Moritz
Installation: 26.10.2016
Removal: 3.7.2017

Lake: Lake Silvaplana
Installation: 25.10.2016
Removal: 3.7.2017

Lake: Sils
Installation: 26.10.2016
Removal: May 2017

Lake: Aegeri
Installation: 16.11.2016
Removal: 18.5.2017

Lake: Sihl
Installation: 16.11.2016
Removal: 18.5.2017

Lake: Greifen
Installation: 8.11.2016
Removal: 5.5.2017

**Examples of moorings for 4 lakes in winter 2016-17**

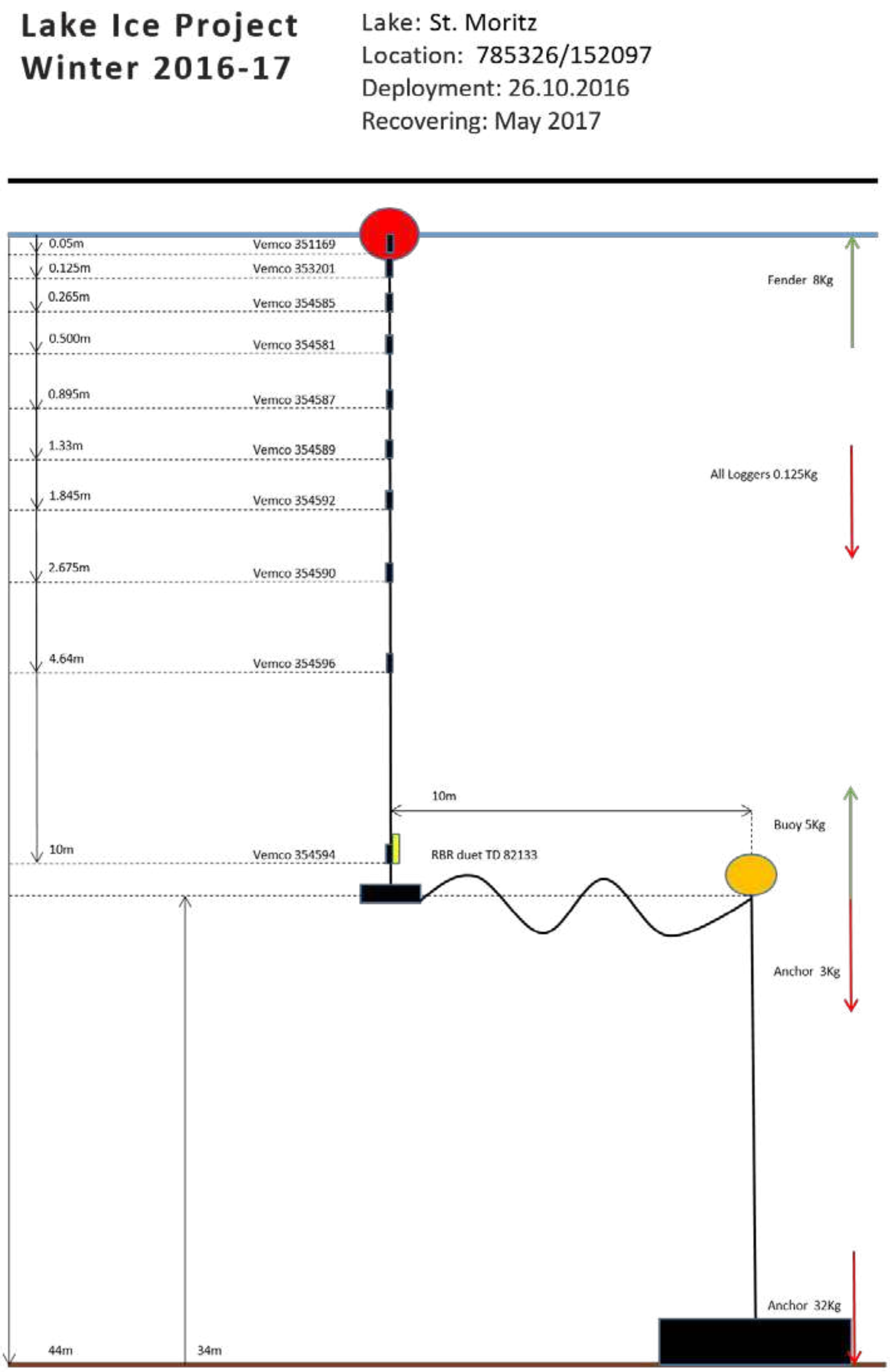

**Lake Ice Project**
**Winter 2016-17**

Lake: Silvaplana
Location: 781243/147298
Deployment: 25.10.2016
Recovering: May 2017

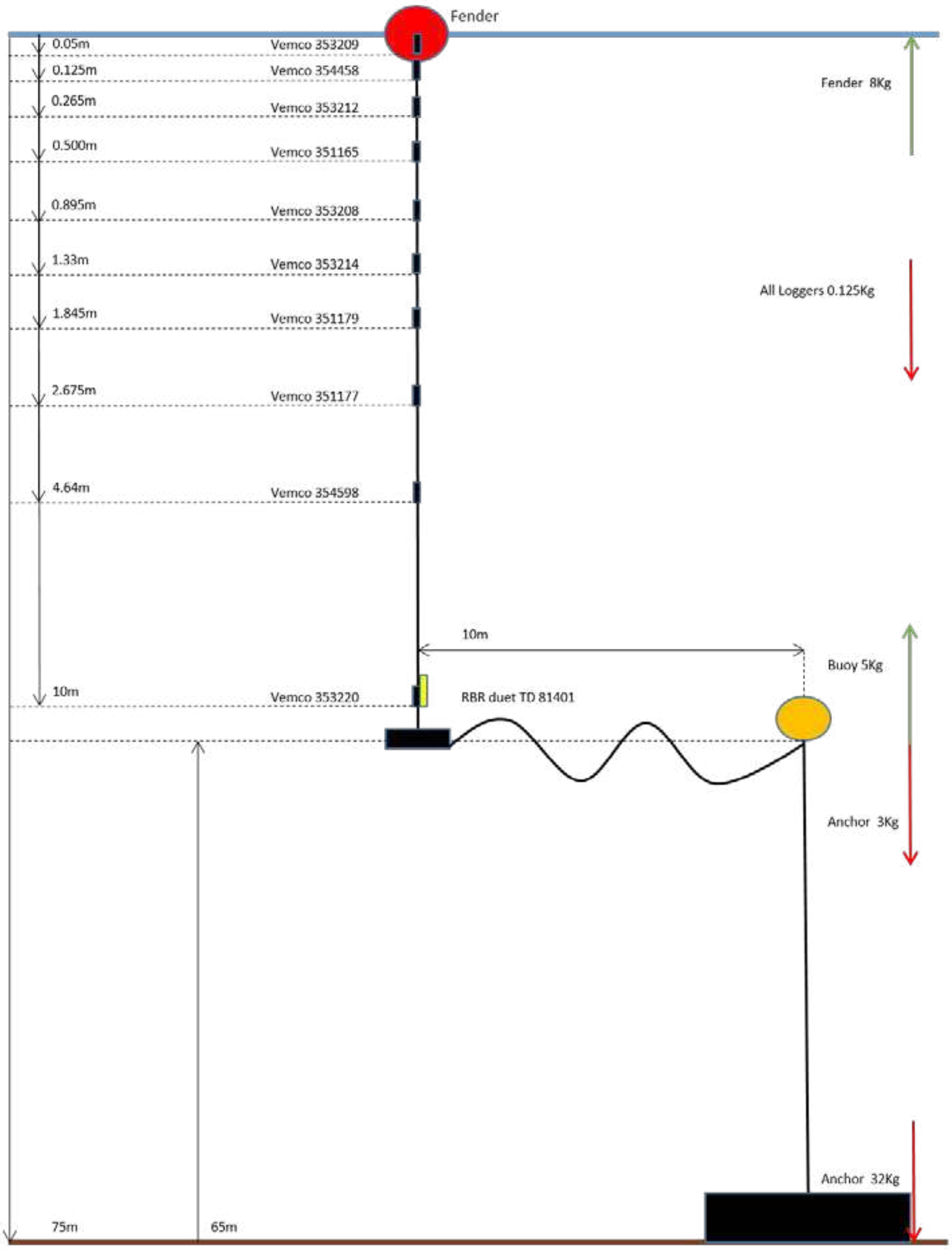

**Lake Ice Project**
**Winter 2016-17**

Lake: Sils
Location: 776384/143999
Deployment: 26.10.2016
Recovering: May 2017

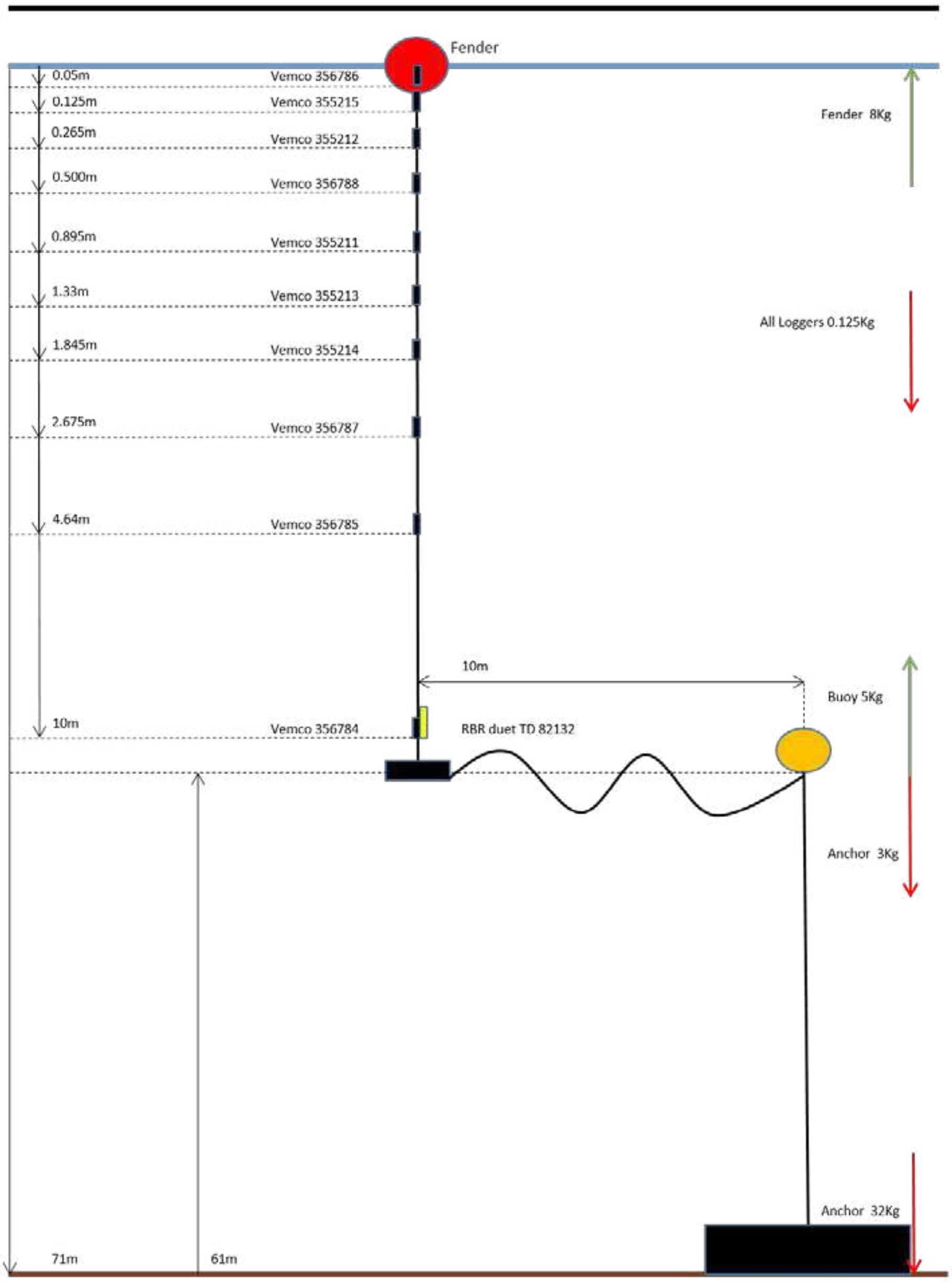

**Lake Ice Project**
**Winter 2016-17**

Lake: Sihlsee
Location: 701877/222818 ADCP
Location: 701949/222816 Loggers
Deployment: 16.11.2016
Recovering: May 2017

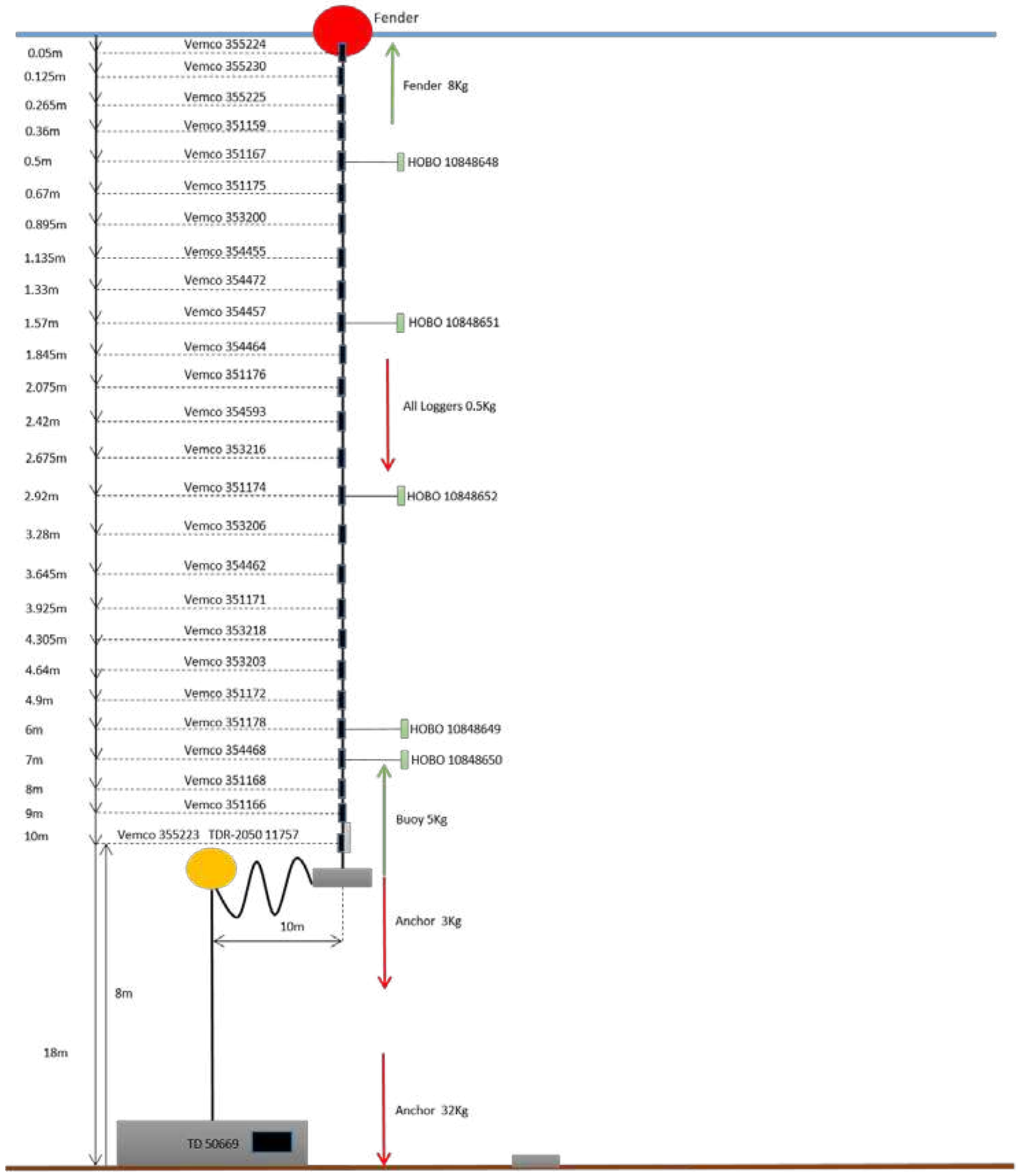

## APPENDIX 3. SUMMARY OF ESTIMATED ICE-ON/OFF DATES

For confidence we use: L (low), M (medium), H (high)

Aegeri lake did not freeze in winter 2016-17.

Sentinel-2A data exist only every 10 or 20 days (if no clouds), so, ice-on/off cannot be decided.

For MODIS and VIIRS (ETHZ) the results shown below refer to the best version using SVM (with RBF kernel) and all channels as input feature vector.

### Lake Sils

| Method | Ice-on date (s) / confidence | Ice-off date (s) / confidence | Remarks |
|---|---|---|---|
| MODIS | 6.1.17 | 12.4.17 | No data available on 5.1.17. There is no MODIS data available on 7.1.17. However, the algorithm detects the next available day (9.1.17) as frozen. Hence, 6.1.17 is chosen as the ice-on date.<br>Note 10.4.17 when 90.9% of the lake is frozen (very close to the threshold) |
| VIIRS (ETHZ) | 6.1.17 | 7.4.17 | No VIIRS data available on 5.1.17. There is no VIIRS data available on 7.1.17. However, the algorithm detects the next available day (9.1.17) as frozen. Hence, 6.1.17 is chosen as the ice-on date. |
| VIIRS (UniBe) | 6.1.2017 / M | 7.4.2017 / M | Before 6.1. cloudy days |
| Webcam | Not processed | Not processed | |
| In-situ (T-based method) | 31.12.16 | 10.4.17 | |
| In-situ (Pressure-based method) | 31.12.16 | 10-12.4.17 | |
| In-situ (Dynamic-based method) | 28.12.16 | 8.4.17 | |
| Visual Webcam * | 2.1.17,5.1.17 / M, L | 8.4.17, 11.4.17 / M,L | Dates used 1.10.16-10.5.17 |

* Visual interpretation of Webcam images. See explanations in Section 2.3.

### Lake Greifen

| Method | Ice-on date (s) / confidence | Ice-off date (s) / confidence | Remarks |
|---|---|---|---|
| MODIS | | | |
| VIIRS (ETHZ) | | | |
| VIIRS (UniBe) | | | |
| Webcam | | | |
| In-situ (T-based method) | 27.1.17 / | 30.1.17 / | |
| Visual Webcam | 29.1.17 / L | 1.2.17 / L | Dates used 26.1.17-3.2.17 |

**Lake Silvaplana**

| Method | Ice-on date (s) / confidence | Ice-off date (s) / confidence | Remarks |
|---|---|---|---|
| MODIS | 1.1.17, 15.1.17 | 8.4.17 | |
| VIIRS (ETHZ) | 11.1.17 | 8.4.17 | |
| VIIRS (UniBe) | Not processed | Not processed | |
| Webcam | Not processed | Not processed | |
| In-situ (T-based method) | 14.1.17 | 14.4.17 | |
| In-situ (Pressure-based method) | 14.1.17 | 15.4.17 | |
| In-situ (Dynamic-based method) | 12.1.17 | 13.4.17 | |
| Visual Webcam | 12.1.17 / M | 11.4.17 / L | Dates used 1.10.16-30.4.17 |

**Lake St. Moritz**

| Method | Ice-on date (s) / confidence | Ice-off date (s) / confidence | Remarks |
|---|---|---|---|
| MODIS | 18.12.16, 3.12.16 | 9.4.17 | Confidence of 3.12.16 as ice-on is very low as it was frozen (3 out of 4 pixels) for 3 consecutive days and then non-frozen for 12 days |
| VIIRS (ETHZ) | 6.1.17, 22.12.16 | 26.3.17 | The confidence of the ice-on/off dates is very low, as these were judged based on the 11 mixed pixels. |
| VIIRS (UniBe) | 21.12.2016 / M | 30.3.2017 /M | Only two pixels located in the lake |
| Webcam | 16.12.16 (low-res Cam21)<br>15.12.16 (high-res Cam24) | No data 18.3-26.4.17 | |
| In-situ (T-based method) | 17.12.16 | 05-8.4.17 | |
| In-situ (Pressure-based method) | 15.12.16 | 8.4.17 | |
| In-situ (Dynamic-based method) | | 5.4.17 | |
| Visual Webcam | 15-17.12.16 / M to L | 30.3-6.4.17 / L | Dates used 4.12.16-30.4.17<br>Data missing for end of March, beginning of April, so no exact ice-off can be determined |

**Lake Sihl**

| Method | Ice-on date (s) / confidence | Ice-off date (s) / confidence | Remarks |
|---|---|---|---|
| MODIS | 3.1.17 | 10.3.17 | There is no MODIS data available on 4.1.17. However, the algorithm detects the next available day (6.1.17) as frozen. Hence, 3.1.17 is chosen as the ice-on date.<br><br>No data was available on 6.3.17 till 9.3.17 |
| VIIRS (ETHZ) | 3.1.17 | 12.3.17 | There is no VIIRS data available on 4.1.17. However, the algorithm detects the next available day (6.1.17) as frozen. Hence, 3.1.17 is chosen as the ice-on date.<br><br>No data was available on 2.1.17<br><br>Note that 10.3 and 11.3 had 90.6 % and 90.9% of frozen pixels (very close to the threshold) |
| VIIRS (UniBe) | 3.1.2017 / H | 15.3.2017 (afternoon) / M | Ice-off: also 14.3. is without ice but the 15$^{th}$ in the morning shows ice cover |
| Webcam | Not processed | Not processed | |
| In-situ (T-based method) | 28-29.12.16 | 16.3.17 | |
| In-situ (Pressure-based method) | | 16.3.17 | |
| In-situ (Dynamic-based method) | 27.12.16 | 16.3.17 | |
| Visual Webcam | 1.1.17 / H | 14.3.17,15.3.17 / H to M, M to L | Dates used 4.12.16-30.4.17 |

## APPENDIX 4. GROUND TRUTH (VISUAL INTERPRETATION OF WEBCAMS)

For full winter 2016-17, one label per day is assigned for the state of each lake from the set below.

Explanation of lake status:
The % below refers to the visible area of the lake, potentially excluding the far part of the lake, which you generally cannot judge, except in case of snow. Clearly, these % are a matter of personal judgement and only approximate.

- ***s*** = snow, when snow is on lake ice, lake frozen to ca. 90-100%
- ***i*** = ice, frozen lake to ca. 90-100%
- ***w*** = water, when lake has ca. 90-100% water
- ***ms*** = more snow, ca. 60-90%, but a small part water
- ***mi*** = more ice, ca. 60-90%, but a small part water
- ***mw*** = more water, ca. 60-90%, but a small part frozen
- ***c*** = clouds or fog covering all lake
- ***u*** = unclear ; when you cannot judge the lake state
- ***n*** = no Webcam data available

In some cases below, c and u may be mixed. Usually, when all lake was cloud covered we used u. If the lake state changes during a day (e.g. when freezing with thin ice or melting), we tried to find the average daily state of the lake, or used unclear. In some unclear cases a double status was used, e.g. mc/c, meaning between more cloudy / cloudy with the first status being usually dominant.

Explanation of cloud status (we included it for lakes Sils and Silvaplana since requested by one project partner, but due to time limitations, it was not included for all lakes):
s (sunny, practically cloudless), ms (more sunny), mc (more clouds), c (totally cloudy)

The data below has 4 parts:

1st: date in format D(D).M(M) The year is clear from the filename.
2nd: lake status
3rd: cloud status.
4th: optionally various comments. When after 2C or 3C a yes exists, it means evaluation of main camera (Segl-Maria) are verified by the camera 2C and/or 3C. The term snow in the comments refers to snow on land at the front part of the camera.

### Lake: Sihl
Operator: Mathias Rothermel
Extent of evaluation: 1.10.2016 to 30.4.2017

ice-on: 1.1
ice-off: 14.3 or less 15.3
start of freezing: 29.12? , many previous days had no data!
end of melting: 18.3? , many days after had missing data

Cameras: Cam12 (this camera is rotating, so the area imaged is not always the same) and during the gap period of Cam12 from 18.3.2017 to 25.4.2017 the EAWAG cam 26 (with very good overview of Lake Sihl from the south) were used.
No cloud observations made.

————————————————

1.10 n
2.10 n
3.10 n
4.10 n
5.10 n
6.10 n
7.10 n
8.10 n
9.10 n
10.10 n
11.10 n
12.10 n
13.10 n
14.10 n
15.10 n
16.10 n
17.10 n
18.10 n
19.10 n
20.10 n
21.10 n
22.10 n
23.10 n
24.10 n
25.10 n
26.10 n
27.10 n
28.10 n
29.10 n
30.10 n
31.10 n
1.11 n
2.11 n
3.11 n
4.11 n
5.11 n
6.11 n
7.11 n
8.11 n
9.11 n
10.11 n
11.11 n
12.11 n
13.11 n
14.11 n
15.11 n
16.11 n
17.11 n
18.11 n

19.11 n
20.11 n
21.11 n
22.11 n
23.11 n
24.11 n
25.11 n
26.11 n
27.11 n
28.11 n
29.11 n
30.11 n
1.12 n
2.12 n
3.12 n
4.12 w
5.12 w
6.12 w
7.12 w
8.12 w
9.12 w
10.12 w
11.12 w
12.12 w
13.12 w
14.12 w
15.12 w
16.12 w
17.12 mw
18.12 w
19.12 w
20.12 w
21.12 w
22.12 mw
23.12 n
24.12 n
25.12 n
26.12 n
27.12 n
28.12 n
29.12 mw
30.12 mi
31.12 mi
1.1 i very probable ice-on
2.1 i
3.1 s
4.1 s
5.1 s
6.1 s
7.1 s
8.1 s
9.1 s

10.1 s
11.1 s
12.1 mi
13.1 mi
14.1 n
15.1 s
16.1 s
17.1 s
18.1 s
19.1 s
20.1 s
21.1 s
22.1 s
23.1 s
24.1 n
25.1 n
26.1 s
27.1 s
28.1 s
29.1 s
30.1 s
31.1 mi
1.2 mi
2.2 mi
3.2 mi
4.2 i
5.2 i
6.2 s
7.2 s
8.2 s
9.2 s
10.2 s
11.2 s
12.2 s
13.2 s
14.2 s
15.2 s
16.2 s
17.2 n
18.2 s
19.2 s
20.2 s
21.2 ms
22.2 n
23.2 n
24.2 n
25.2 n
26.2 s
27.2 s
28.2 n
1.3 s
2.3 i

3.3 i
4.3 i
5.3 i
6.3 i
7.3 i
8.3 s
9.3 s
10.3 i
11.3 i
12.3 i
13.3 i
14.3 mi/i ice-off? or less 15.3?
15.3 mi/i
16.3 mi or rather mw?
17.3 mw
18.3 w (from here to 25.4 data is not available, likely w, from EAWAG cam26)
19.3 w
20.3 w
21.3 w
22.3 w
23.3 w
24.3 w
25.3 w
26.3 w
27.3 w
28.3 w
29.3 w
30.3 w
31.3 w
1.4 w
2.4 w
3.4 w
4.4 w
5.4 w
6.4 w
7.4 w
8.4 w
9.4 w
10.4 w
11.4 w
12.4 w
13.4 w
14.4 w
15.4 w
16.4 w
17.4 w
18.4 w
19.4 w
20.4 w
21.4 w
22.4 w
23.4 w

24.4 w
25.4 w
26.4 w
27.4 w
28.4 w
29.4 w
30.4 w

**Lake: Sils**
Operator: Manos Baltsavias
Extent of evaluation: 1.10.2016 to 10.5.17 and less detailed to 30.6.17.

ice-on, rather 2.1 or less 5.1? (checked by all 3 cameras)
ice-off, 8.4 or 11.4? (checked by Segl-Maria)
start of freezing: 7.12 (but previously starting from 16.10 also days with i, mi, mw; and after 7.12 days with w)
end of melting: very difficult as some ice up to 30.6, but first melting end is on 14.4 or even more on 17.4

Cameras:
Roundshot camera Segl-Maria: best quality, panoramic camera, covers all period, has only one image at 12:00 till and 28.1.17. This camera at 12:00 was the main source of evaluation. Images at other day times were also used.
Cam 4 (called 2C below) acquired from 4.12.16 multiple images / day but is very far and of poor quality.
Cam 5 (called 3C below, a subpart of Segl-Maria) acquired from 4.12.16 multiple images / day, but no good quality.
Cameras 2C and 3C have been used very little in the evaluation, and used especially for the critical periods of ice-on. During ice-off, there was a data acquisition gap.

A first remark. There were very few days when the lake was totally covered by very low clouds and fog. An important advantage compared to satellite images.

Clouds change during the day. Below we used almost always the same time at 12:00 from the Segl-Maria camera. It would be ideal to use an average time between the MODIS and VIIRS overpasses (around 11.30), however for Segl-Maria for many months we had only the 12:00 image. For cloud estimation not only the sky but also the sunshine on the lake surface was checked.

Note that before ice-on, lake started freezing in more than one occasion but then warmed up again. Similarly, after melting end, there were cases of thin ice and then melting on several occasions.

Images till 30.4.2017 checked 2-3 times, during independent periods, with images during ice-on/off checked even more than 3 times.

——————————————————

1.10 w c
2.10 w mc/c
3.10 w s
4.10 w s
5.10 w s strange like ice in the middle
6.10 w s/ms
7.10 w ms
8.10 w ms

9.10 w c
10.10 w mc first snow at lake border and then back at 5.11
11.10 w c no snow
12.10 w s white reflections in front why? and dark shadows in the middle center?
13.10 w c no snow
14.10 u c fog and low clouds
15.10 u c probably water, interesting image (reflections of clouds and objects on lake without sun)
16.10 mi/u s some ice
17.10 w c
18.10 mi mc a bit unclear
19.10 mi mc
20.10 mi s
21.10 mw ms
22.10 w ms
23.10 w c
24.10 w c
25.10 w c
26.10 w c
27.10 w s very little snow even in the mountains
28.10 u ms probably both water and ice
29.10 mw s back may be frozen
30.10 mi s not very clear
31.10 mi s
1.11 w s little ice
2.11 w c ice at back?
3.11 w s
4.11 c c
5.11 w c most area cloud covered, snow
6.11 i c at front maybe water
7.11 u ms probably top part frozen
8.11 w s
9.11 w c
10.11 w c back is unclear
11.11 w c
12.11 w s
13.11 mi c
14.11 w s
15.11 mi s
16.11 mw c
17.11 mw/w c
18.11 c c
19.11 w c evaluation unclear
20.11 w c
21.11 w c
22.11 w c snow starts diminishing and after 26.11 very little
23.11 w c
24.11 mi c
25.11 w c
26.11 w c
27.11 w s
28.11 w s

29.11 w s seems a bit ice, strange shadows at front, camera artefacts? see sudden vertical edge at front/right middle
30.11 w s (why different reflections between 29 and 30.11?)
1.12 u/mi c looks mostly ice
2.12 mw s
3.12 w c
4.12 w ms first time 2nd camera
5.12 w s
6.12 w s why on 6.12 no reflections compared to 5.12 and 7.12 although all sunny?
7.12 w s little thin ice
8.12 mw s ice at back
9.12 u mc
10.12 mi s
11.12 mi/u c
12.12 w s
13.12 mi s
14.12 u s
15.12 w s
16.12 mw s
17.12 mi s
18.12 mi s
19.12 w c what is the white part in front and on the right?
20.12 w c
21.12 mi mc thin ice about 50%
22.12 mw/mi s thin ice about 50%
23.12 mi s thin ice about 50%
24.12 u c seems like 50/50 water/ice
25.12 w c
26.12 u s
27.12 mi s quite some water
28.12 w s
29.12 mw s
30.12 mw/w s unclear in the back whether ice
31.12 mi s quite some water
1.1 mi s more thin ice but also water, much more than 2.1 and 3.1, camera 2C rather yes, 3C yes
2.1 i mc very little water at front, camera 2C rather yes, 3C yes
3.1 i s very little water at front, no snow, what is the black spot at front on 3.1.17 15:30, camera 2C rather yes, 3C yes
4.1 i s no snow, rapid change to 5.1, 11 deg T drop, camera 2C rather yes, 3C yes
5.1 s c little water at front and middle, snow, camera 2C s (12:45), 3C yes
6.1 s s definitely some water, more than 5.1, camera 2C ms/s, 3C yes and even more water
7.1 s/i mc as 6.1, camera 3C yes, more ice (11:30) than Segl-Maria
8.1 s c seems first day with total ice, also from camera 2C, 3C yes
9.1 s s camera 3C yes
10.1 s c
11.1 s s
12.1 s s some water (or thin/black ice?) at front and center left
13.1 s c
14.1 s c
15.1 s s
16.1 s s
17.1 s s

18.1 s s
19.1 s s
20.1 s s
21.1 s s
22.1 s s
23.1 s ms
24.1 s s
25.1 s s bit water at outflow
26.1 s s
27.1 s s
28.1 s c
29.1 s s
30.1 s c
31.1 s c low clouds
1.2 s c very low clouds
2.2 s c low clouds
3.2 u c all clouds
4.2 s c very low clouds
5.2 s c
6.2 s c
7.2 s c
8.2 s c
9.2 s ms
10.2 s c very low clouds
11.2 s ms
12.2 s mc
13.2 s s
14.2 s s
15.2 s s
16.2 s s
17.2 s mc
18.2 s s
19.2 s s
20.2 s ms
21.2 s c/mc
22.2 s s
23.2 s c
24.2 s c
25.2 s s
26.2 s c
27.2 s c
28.2 u c all clouds
1.3 s s
2.3 s mc see some ice or rather cloud shadows?
3.3 s c
4.3 s c
5.3 s c
6.3 s c
7.3 s c
8.3 s mc
9.3 s c
10.3 s s

11.3 s s
12.3 s mc
13.3 s s
14.3 s s
15.3 s ms/mc
16.3 s s snow starts to melt on ice, partly saturated image
17.3 s s snow starts to melt on ice, partly saturated image
18.3 s/i mc snow melting on ice
19.3 i/s c
20.3 i/s s
21.3 i/s c low clouds, start melting? snow melts on ground at front
22.3 u c all clouds
23.3 i/s c
24.3 s/i c
25.3 s/i s outflow starts melting
26.3 i/s s outflow with water
27.3 s/i s
28.3 s/i s
29.3 i/s s
30.3 i/s ms
31.3 i/s s
1.4 i/s mc almost no snow on land front
2.4 i/s c
3.4 i/s s
4.4 i/s c more frozen than 3.4
5.4 i/s mc a little water at front right
6.4 i/s ms more ice than 5.4
7.4 i/s ms a bit more water than 6.4
8.4 i/s s more water than 7.4 at front and at outflow but great majority of lake frozen, start of melting?
9.4 i/s s very little more water than 8.4 but less at front
10.4 i ms similar to 9.4 but at centre white strips why?
11.4 i/s c as 10.4, much more water than 10.4 at front and outflow, clear start of melting
12.4 mi/s s big change to 11.4, much more water than 11.4 but majority frozen, surely melting, rapid change to 13.4
13.4 mw s front almost all water, except maybe centre left and background
14.4 w s back unclear
15.4 w ms some thin ice?
16.4 w c very little thin ice?
17.4 w mc
18.4 w c
19.4 w mc
20.4 w s
21.4 mw/w s seems to have quite some thin ice see 7:00 but not at 9:00
22.4 mw/w s also some thin ice at 8:00 and at back?
23.4 w s
24.4 w c at 9:00am, at 12:00 all clouds
25.4 w c used 18:00, many low clouds
26.4 w c used 18:00 image, too many low clouds, lots of snow on land, check T
27.4 w c used 15:00, low clouds, artefacts at 17:00 and 18:00
28.4 w c low clouds
29.4 w/mw s some or more? thin ice at 14:00 but disappears at 15:00
30.4 w ms snow on land

1.5 w c low clouds
2.5 w mc
3.5 w c strange colours 9:00 = thin ice? rather not
4.5 w c
5.5 w s
6.5 w c snow
7.5 w c check 8:00 thin ice? rather not, medium snow on land
8.5 w c unclear at 12:00 but probably all water, very little snow
9.5 w s no snow
10.5 w ms

Evaluation below was not rechecked and surely includes some errors
11.5 probably all water
12.5 unclear, maybe still ice
13.5 still ice
14.5 probably still ice
15.5 still ice
16.5 seems still ice
17.5 seems still ice
18.5 unclear, maybe ice
19.5 unclear
20.5 ice
21.5 probably all water
22.5 unclear, still ice?
23.5 still ice
24.5 still ice
25.5 still ice
26.5 still ice
27.5 still some ice in morning
28.5 still ice
29.5 still ice
30.5 still ice
31.5 still ice
1.6 unclear, maybe ice
2.6 maybe all water
3.6 still ice
4.6 still ice
5.6 sure ice
6.6 seems also ice
7.6 unclear, probably only water
8.6 still ice
9.6 ice
10.6 ice
11.6 ice
12.6 ice
13.6 ice
14.6 ice
15.6 ice
16.6 ice
17.6 ice
18.6 ice
19.6 ice

20.6 ice
21.6 ice
22.6 ice
23.6 ice
24.6 ice
25.6 unclear, maybe ice
26.6 sure ice
27.6 sure ice
28.6 ice
29.6 ice
30.6 ice

**Lake: Silvaplana**
Operator: Manu Tom
Extent of evaluation: 1.10.2016 to 30.4.2017

ice-on: 11-13.1, more probable 12.1
ice-off: 9-13.4, more probable 11.4
start of freezing: 3-4.12? or more 10.12
end of melting: 18.4 probably

Cameras: Mainly used the camera El Paradiso (called Cam28). In case of doubt, other Webcams were used. Since El Paradiso is located far away from the lake, it is very difficult to judge the state of the lake in general, Hence more than one image during a day were used to infer the state of the lake. There are some days in which the state is unclear. In those cases, a probable state is also proposed. It is often confusing to judge whether it is ice or shadow of the mountains nearby. It is relatively less tough to judge around 12:30-13:30 when the sun comes approximately above the lake. Snow is easy to judge. Ice-water confusion exist a lot with this camera. Most of the other cameras capturing Silvaplana are either providing low quality data (compression artefacts) or very far away (high GSD) or very close (covering very limited lake area). In addition, the panoramic camera Segl-Maria (good resolution but rather far) was used.

A second operator re-examined images during freezing and melting and added some comments in ( ).

There were very few days when the lake was totally covered by very low clouds and fog.

—————————————————

1.10 w c/mc
2.10 u mc/c probably water
3.10 w ms/s
4.10 w s/ms
5.10 w s
6.10 w mc/ms
7.10 w c/mc
8.10 w mc/ms
9.10 w c fog and low clouds
10.10 w mc first snow at lake border
11.10 w ms/mc
12.10 w s/ms
13.10 w c snow in mountains nearby
14.10 w c fog and low clouds

15.10 u c probably water
16.10 w s
17.10 u c probably more water
18.10 u c/mc probably water
19.10 u c/mc probably more water
20.10 u s/ms probably more water
21.10 u mc/ms probably water
22.10 w ms/s
23.10 u c probably water
24.10 w c foggy
25.10 c c foggy
26.10 w c low clouds and fog
27.10 w ms/s
28.10 w s
29.10 mw s
30.10 w s
31.10 w s reflections due to sun makes it a bit confusing
1.11 w s
2.11 u c probably water
3.11 w s
4.11 c c
5.11 c c fog covered
6.11 c c fog covered
7.11 mw mc/ms
8.11 mw s
9.11 u c probably water
10.11 mw mc
11.11 u c probably water
12.11 mw s
13.11 w c shadow of mountain nearby is confusing
14.11 w s
15.11 mw ms
16.11 mw mc
17.11 c c
18.11 c c probably more water
19.11 c c
20.11 w c
21.11 u c probably water (difficult to judge due to fog)
22.11 w c
23.11 w c
24.11 w c
25.11 mw c probably some ice in the south-west part
26.11 u c unclear due to fog, probably more water
27.11 u ms/mc probably more water
28.11 w s
29.11 u s slightly foggy above the lake making it difficult to judge (probably more water)
30.11 w s
1.12 w/mw mc
2.12 w/mw s
3.12 u mc probably more ice
4.12 mi ms
5.12 mw s

6.12 mw s
7.12 mw s
8.12 mw s some floating thin ice
9.12 u mc shadow of mountains making it unclear to judge (probably more water)
10.12 mi s
11.12 mi c
12.12 u ms unclear to judge but there is definitely some ice how much is unclear (probably more water)? Cloud changed very frequently during the day
13.12 mi ms Champfèr is fully frozen
14.12 mw/mi s
15.12 mi/mw s
16.12 mw s some floating ice
17.12 mw s, unclear
18.12 mi s, unclear
19.12 mw c, unclear
20.12 mi c, unclear
21.12 mw/mi s, unclear
22.12 mw s thin ice (16:00 more thin i, but seems some w at back)
23.12 mw s thin ice unclear if all thin i or some w
24.12 mi c (16:00 thin i or w? seems more i, but quite some water)
25.12 u/mi c probably more ice (16:00 thin i or w?)
26.12 mi ms (16:00 thin i or w?)
27.12 mi s (12:00 u, like 28.12, 12:00 thin ice mostly, little w front)
28.12 mw/mi s (12:00 all thin ice? 16:00 seems all thin i)
29.12 mw s (12:00 thin i, some w back, 16:00 most thin i, some w)
30.12 mw/w s (12:00 thin ice?, some w back, 16:00 most thin i, some w (less than 29.12)
31.12 mi/w s (16:00 seems all thin i)
1.1 mi/i s (12:00 some water at front, 16:00 mi, quite some w)
2.1 i/mi ms (12:00 more ice than 1.1, little water, 16:10 mi, quite some water)
3.1 i s (12:00 mi, water (more than 1.1) at back, 16:00 thin ice, quite some water)
4.1 i ms (12:00 like 1.1, 15:00 thin ice, quite some water, more than 3.1)
5.1 u c low clouds, probably ice with snow over it in some parts (12:00, mi, w? 16:00 seems quite some water, like 4.1)
6.1 i/s s probably no water at all, mostly ice and snow above it at some parts (14:00 lots of water, 15:00 like 5.1 more w)
7.1 i/s mc mostly a mix of ice and snow, a little water too (15:00 very little water, much thin ice)
8.1 i/s c (lots of water or thin i?, 14:00 and 15:00)
9.1 i/s s (15:00 quite some water)
10.1 i/s c, maybe a mixture of all 3 classes (only one image, some little water, quite some thin ice forming where on 11.1 frozen)
11.1 s/i s (14:00 little water in NE, maybe ice-on or on 12-13.1, 12.1 more safe)
12.1 i/s s (seems almost all ice, black ice? or also water? seems all ice, see 13:00-14:00)
13.1 ms c rest area either ice or water (15:00, mostly frozen, still some (areawise) water in NE, but not seen in Segl-Maria)
14.1 s c ice in some parts (very little water 16:00, bit more than 15.1 and 16.1)
15.1 s ms ice in some parts (12:00 and 14:00 practically all frozen)
16.1 s s (12:00 practically all frozen)
17.1 s s (as 16.1, a bit water in NE)
18.1 s s (11:10, surely all frozen)
19.1 s s (16:10 all frozen)
20.1 s s
21.1 s s

22.1 s s
23.1 s s
24.1 s s
25.1 s s
26.1 s s
27.1 s s
28.1 s c
29.1 s s
30.1 s c
31.1 s c very cloudy in El Paradiso. Snow easily visible in other Webcams
1.2 s c very cloudy
2.2 s c low clouds
3.2 s c low clouds
4.2 s c
5.2 s c
6.2 s c low clouds
7.2 s ms
8.2 s c
9.2 s c
10.2 s c low clouds
11.2 s ms
12.2 s mc
13.2 s s
14.2 s s
15.2 s s
16.2 s s
17.2 s mc
18.2 s s
19.2 s s
20.2 s s
21.2 s mc
22.2 s s
23.2 s s
24.2 s c
25.2 s s
26.2 s mc
27.2 s s
28.2 s c very low clouds, judged from Cam14
1.3 s mc
2.3 s ms
3.3 s ms/mc
4.3 s c
5.3 s c
6.3 s c
7.3 s c/mc
8.3 s s
9.3 s c very low clouds
10.3 s s
11.3 s s
12.3 s ms/s
13.3 s s
14.3 s s

15.3 s ms/mc
16.3 s s
17.3 s s snow depth decreased
18.3 s/ms c snow melting on ice
19.3 s/i c very thin snow layer on ice
20.3 s/i s
21.3 s/i c low clouds
22.3 s/i c all clouds, judged from image at 18:00
23.3 s/i c
24.3 s/i c
25.3 s/i s
26.3 i/s s
27.3 i/s s
28.3 s/i s
29.3 i/s s
30.3 i/s s
31.3 i/s s (very little water)
1.4 i/s mc (16:20, practically all frozen)
2.4 i/s mc (16:20, practically all frozen)
3.4 i/s s (16:20, practically all frozen)
4.4 i c snow almost disappeared (16:20, practically all frozen)
5.4 i ms (16:20, practically all frozen)
6.4 i/s ms sunny in the evening (16:20, all frozen)
7.4 i/s ms (16:20, practically all frozen)
8.4 i/s s (16:20, mostly frozen, little water?)
9.4 i s snow almost disappeared (16:20, mostly frozen, little water?)
10.4 i s/ms (10:50 etc. almost all ice, water at outflow of Inn)
11.4 i c (mostly ice, little water in evening) ice-off? (though process started on about 9.4 and continued to 13.4)
12.4 i/s s (as 11.4, water at outflow of Inn, more water than 11.4)
13.4 i ms (mostly ice, quite some water, more than 12.4)
14.4 mi s (quite some water, much more than 13.4)
15.4 mi/mw ms (mostly water)
16.4 mw/mi c/mc
17.4 mw/mi c/mc
18.4 w c
19.4 w ms
20.4 w s
21.4 w s
22.4 w s
23.4 w s
24.4 u c unclear due to very low clouds (probably water)
25.4 c c low clouds
26.4 u c low clouds, judged the state from Cam19 but still a bit unclear (probably water)
27.4 w c low clouds, judged the state from Cam17
28.4 u c low clouds, probably water
29.4 w/mw s
30.4 mw ms

**Lake: St. Moritz**
Operator: Mathias Rothermel
Extent of evaluation: 1.10.2016 to 30.4.2017

ice-on: 15-17.12
ice-off: difficult to judge due to lack of images, maybe between 30.3 (little water at inflow of Inn, assume majority is frozen) and 6.4 (which has quite some water)
start of freezing: 9.12
start of melting: 13.4 but many previous days with no data

Cameras used: Cam21. Secondary camera used is Cam23, and even less Cam22.

No cloud observations made.

———————————————————

1.10 n
2.10 n
3.10 n
4.10 n
5.10 n
6.10 n
7.10 n
8.10 n
9.10 n
10.10 n
11.10 n
12.10 n
13.10 n
14.10 n
15.10 n
16.10 n
17.10 n
18.10 n
19.10 n
20.10 n
21.10 n
22.10 n
23.10 n
24.10 n
25.10 n
26.10 n
27.10 n
28.10 n
29.10 n
30.10 n
31.10 n
1.11 n
2.11 n
3.11 n
4.11 n
5.11 n

6.11 n
7.11 n
8.11 n
9.11 n
10.11 n
11.11 n
12.11 n
13.11 n
14.11 n
15.11 n
16.11 n
17.11 n
18.11 n
19.11 n
20.11 n
21.11 n
22.11 n
23.11 n
24.11 n
25.11 n
26.11 n
27.11 n
28.11 n
29.11 n
30.11 n
1.12 n
2.12 n
3.12 n
4.12 w
5.12 w
6.12 w
7.12 w
8.12 w
9.12 mw
10.12 mw
11.12 mw
12.12 mw
13.12 mi
14.12 mi
15.12 i ice-on? small inflow to the east till about 31.12 and outflow at west, at the latest ice-on on 17.12. Ice-off can't be determined due to gap of Webcam images, except some sporadic ones. Based on these sporadic Webcam images ice-off was on or started on 6.4.2017, while water at the outflow of Inn had a similar pattern though larger sequentially on 16.3, 23.3 and 30.3. And this tongue of water though smaller in width existed even on 2.2.2017!. So, my guess for ice-off is between 30.3 and 6.4.2017.
16.12 i
17.12 i
18.12 i
19.12 s
20.12 s
21.12 s
22.12 s
23.12 n

24.12 n
25.12 n
26.12 n
27.12 n
28.12 n
29.12 i
30.12 i
31.12 i
1.1 i
2.1 i
3.1 i
4.1 i
5.1 s
6.1 s
7.1 s
8.1 s
9.1 s
10.1 s
11.1 s
12.1 s
13.1 s
14.1 n
15.1 s
16.1 s
17.1 s
18.1 s
19.1 s
20.1 s
21.1 s
22.1 s
23.1 s
24.1 n
25.1 n
26.1 s
27.1 s
28.1 s
29.1 s
30.1 s
31.1 s
1.2 s
2.2 s
3.2 s
4.2 s
5.2 s
6.2 s
7.2 s
8.2 s
9.2 s
10.2 s
11.2 s
12.2 s
13.2 s

14.2 s
15.2 s
16.2 s
17.2 n
18.2 s
19.2 s
20.2 s
21.2 s
22.2 n
23.2 n
24.2 n
25.2 n
26.2 s
27.2 s
28.2 n
1.3 s
2.3 s
3.3 s
4.3 s
5.3 s
6.3 s
7.3 s
8.3 s
9.3 s
10.3 s
11.3 s
12.3 s
13.3 s
14.3 s
15.3 s
16.3 s
17.3 s
18.3 n
19.3 n
20.3 n
21.3 n
22.3 n
23.3 i some water
24.3 n
25.3 n
26.3 n
27.3 n
28.3 n
29.3 n
30.3 i more water than 23.03
31.3 n
1.4 n
2.4 n
3.4 n
4.4 n
5.4 n
6.4 i more water than 30.03

7.4 n
8.4 n
9.4 n
10.4 n
11.4 n
12.4 n
13.4 w
14.4 n
15.4 n
16.4 n
17.4 n
18.4 n
19.4 n
20.4 w
21.4 n
22.4 n
23.4 n
24.4 n
25.4 n
26.4 w
27.4 w
28.4 w
29.4 w
30.4 w

**Lake: Greifen**
Operator: Manos Baltsavias
Extent of evaluation: only days close to the ice-on/off dates of EAWAG were checked (26.1.2017 (in one case 23.1) to 3.2.2017)

ice-on: considerable ice/practically no snow (majority i compared to w) from 27.1 to max 1.2. Peak of ice 28.1 or rather 29.1.
ice-off: too close to ice-on to decide.
Difficult to say whether there was a real ice-on/off

All 3 cameras used, the last one was a zoom-in of the first one.

Cam 2

23.1 foggy, water, maybe some thin ice from afternoon at the back
24-25.1 no data
26.1 at front water, some thin ice at back especially in the morning, mostly water with some thin ice
27.1 quite some ice at 17:14, less ice before but even before more ice, mostly ice at 15:24 and 14:54 (but water at back and boat moved), even more ice at 14:54 and 13:54, even more ice in the morning, unclear if some water till about 10:00, back unclear

28.1 7:50 great majority ice, getting later more water but majority still ice, back seems ice

29.1 8:30 mostly ice, little water, later more water (also boats) but still great majority ice

30.1 8:00 very little water, 9:12 same, 10:00 bit more water, 11:00 even more water, back rather ice, and towards 17:00 even more water, maybe majority

31.1 foggy in morning, 8:00-10:00 mostly ice, both ice and water, seems more ice, from 11:00 more water, 12:00 more ice? 13:30 and 14:40 ice and back unclear, and 14:00 almost all ice?, then foggy

1.2 foggy, seems more ice, back unclear, 9:00 ice at front, then water, 10:00, 11:00 more ice, 12:00 more ice even at back, from 15:00 and after more water relatively, but majority ice

2.2 very foggy, water at front, later seems all water
3.2 water

Cam 3 (from opposite lake side compared to previous camera, at NE of lake)

26.1 water, back unclear
27.1 ice, back (especially right) unclear, warming and maybe water in later afternoon
28.1 9:00 seems ice overall, back right unclear
29.1 all ice apart from unclear back right, especially in evening
30.1 8:10-9:00 ice, back right unclear, 9:40 more w back right, to 11:30 more w, then seems increasingly w (but also black i?) and after 17:00 more i.
31.1 8:00-9:00 ice, after 10:30 some w back right?, after, all seems ice with increasing fog.
1.2 8:00-9:00 ice (little w), 10:46, 11:16 more ice but quite some w at back, 13:47 more ice than before, back unclear, 15:00 more ice than before but majority seems frozen, then more w till 16:46, 17:26 i comes back
2.2 foggy, 8:00 ice, 9:35 mostly ice, 10:15 mostly water at front, but also ice, 11:00 at front w, rest fog, 12:10 w, 13:00-17:00 w.
3.2 w, back left maybe ice in the morning?

Cam 1 (like Cam 2, but focussed on a smaller area)
Checked camera a bit less detailed

26.1 mw, back thin ice? mainly in morning
27.1 7:30-10:00, almost all i, 10:30-15:00 boats at back, most i, increasingly more w till 17:00
28.1 little w, boats, maybe more w in evening
29.1 quite little w, and almost constant (seems more i in general for the whole period)
30.1 begins like previous day but from noon more w, with increasing w towards evening though majority probably still i
31.1 8:00 fog, i, with more w slowly but after 13:00 more fog and i, maybe at back right and left w
1.2 8:09 fog, front w, 9:40 mostly ice, 10:00 more w, w increasing during the day but majority i
2.2 much fog especially in morning, when visible water
3.2 w

## APPENDIX 5. GROUD TRUTH (VISUAL INTERPRETATION OF SENTINEL-2A DATA)

Data processed from Sentinel-2A interpretation, using mainly the visible bands but for the freezing and melting periods all bands used. Note, that for some dates there were no data in the archive (images only every 20 days instead of 10).

Explanation of lake status:
(Clearly these % are a matter of personal judgement and only approximate)

s = snow, when snow is on lake ice, lake frozen to ca. 90-100%
i = ice, frozen lake to ca. 90-100%
w = water, when lake has ca. 90-100% water
ms = more snow, ca. 60-90%, but a small part water
mi = more ice, ca. 60-90%, but a small part water
mw = more water, ca. 60-90%, but a small part frozen
c = clouds or fog covering all lake
u = unclear, when you cannot judge the lake state

### Lake: Sihl

Operator: Manu Tom
Extent of evaluation: 29.9.2016 to 10.5.2017

————————————————

29.09 w
2.10 c
9.10 c
12.10 w a bit cloudy (judged as water from Band 8)
19.10 c
22.10 w a bit cloudy (judged as water from Band 8)
29.10 w
1.11 w
8.11 w
11.11 c
18.11 c
21.11 c
28.11 c
1.12 c Band 8 gives hint that it might be water but cannot judge
8.12 w some parts of the lake started freezing (south end and some border parts)
11.12 c/w probably w since 8.12 and 18.12 shows consistent state (water)
18.12 w some parts of the lake started freezing (south end and some border parts)
21.12 w snow appears on the frozen parts. Also, a bit more frozen than 18.12
28.12 mw thin ice starts forming near the frozen south end extending more towards the centre of the lake. Also, ice appearing at more borders of the lake
31.12 i fully frozen with ice only (see True Colour Image TCI)
7.01 s thin clouds too
10.01 c probably snow covered
27.01 s
30.01 c probably snow
6.02 c probably snow

9.02 c probably snow
16.02 s some part of south end probably ice or water (cannot judge exact state)
19.02 s some part of south end probably ice or water (cannot judge exact state)
26.02 s some part of south end probably ice or water (cannot judge exact state)
1.03 c/s south end not cloudy, looks almost the same as 26.2 so probably the same state (snow)
8.03 s/ms thin clouds but visible in Band 8. North part almost fully snow covered. South part has less snow. The non-snowy part of the lake could be frozen (probably) or water but cannot judge exactly
11.03 s/ms North part almost fully snow covered. South part has less snow. The non-snowy part of the lake could be frozen (probably) or water but cannot judge exactly
18.03 c
21.03 c
28.03 w some parts of south of lake frozen (concluded from Band 8)
31.03 w some parts of south of lake frozen, some parts at lake border also frozen
7.04 w some parts of south of lake frozen, some parts at lake border also frozen
10.04 w some parts of south of lake frozen (a bit less than 7.04), some parts at lake border also frozen
17.04 c probably water
20.04 c north part not cloudy (can be judged as water from Band 8), south of lake completely covered by clouds
27.04 c
30.04 w but some part of the south of lake covered by snow
7.05 c
10.05 w but some part (very little) of the south of lake covered by ice/snow

**Lake: Sils**
Operator: Manu Tom
Extent of evaluation: 29.9.2016 to 7.5.2017

———————————————

29.09 w
9.10 c
19.10 c
29.10 w
8.11 w
18.11 c
28.11 w
8.12 w
18.12 w some ice at southwest part of lake but <15%, hence judged as w
28.12 w some ice in southwest part of lake but <15%, hence judged as w
7.01 u
27.01 s
6.02 c
16.02 s
26.02 s thin clouds
8.03 s
18.03 c
28.03 i with thin snow layer on top
7.04 i
17.04 w north part of the lake cloudy
27.04 c
7.05 c

**Lake: Silvaplana**
Operator: Manu Tom
Extent of evaluation: 29.9.2016 to 7.5.2017

————————————————

29.09 w
9.10 c
19.10 u probably water
29.10 w
8.11 w
18.11 c
28.11 w
8.12 w
18.12 mw mainly water and a bit of ice at the south part of lake
28.12 mw/mi ice or water ? cannot judge exactly
7.01 u some parts look like water
27.01 s
6.02 c
16.02 s
26.02 s thin clouds
8.03 s
18.03 c
28.03 i with thin snow layer on top
7.04 i
17.04 c
27.04 c
7.05 c

**Lake: St. Moritz**
Operator: Manu Tom
Extent of evaluation: 29.9.2016 to 7.5.2017

————————————————

29.09 w
9.10 c
19.10 u probably water
29.10 w
8.11 w/i (difficult to judge, probably water)
18.11 c
28.11 w/i (difficult to judge, probably water)
8.12 w/i (difficult to judge, probably water)
18.12 u northwest part definitely ice, rest of the lake cannot judge, between ice and water
28.12 u northwest part definitely ice, rest of the lake cannot judge, between ice and water
7.01 c
27.01 s
6.02 c

16.02 s
26.02 s thin clouds
8.03 s
18.03 c
28.03 i with thin snow layer on top
7.04 mi
17.04 c
27.04 c
7.05 c

## APPENDIX 6. SUMMARY OF DAILY RESULTS FOR ALL MONITORING DATA/METHODS AND LAKES ST. MORITZ, SILVAPLANA, SILS AND SIHL IN THE WINTER 2016-2017

For the data listed in the tables below:

1. In green background are the dates of ice-on/off that each monitoring data/method detected. One of these dates maybe not highlighted with green due to missing data. With red, the dates of the ground truth (visual interpretation of Webcams) are highlighted. In case there is a second possible date with lower probability, it is highlighted in yellow (one should also note the reliability of the ground truth as listed in Appendix 3).
2. nd means NO DATA, either because of clouds and for Webcams also fog and serious image quality defects (lens dirt, defocussing etc.) or because the data were not provided/acquired.
3. The approximate lake area covered by each Webcam is shown in Fig. 80. The number of clean pixels per lake for the satellite images for ETHZ are given in Table 6. The cloud-free clean pixels form 100% of the pixels that could be processed.
4. % FP is the percentage of frozen pixels based on the 100% of the pixels as defined in 3. above.
5. VIIRS1 = ETHZ approach ; VIIRS2 = UniBe approach
6. For in-situ data, the following codes are used: State: 1= frozen, 2= probably frozen, 3=probably non-frozen, 4= non-frozen. Whereby below mostly 1 and 4 are used, because of difficulties to clearly define the freeze-up and break-up periods.
7. For Webcams multiple images per day were processed. They vary from day to day and among the two cameras. Because some images were excluded (see 2. above). The images per day were selected manually between 8.00am the earliest and 16:30 the latest. The acquisition frequency for both cameras was 1h. On average, there were ca. 7 images processed per day. Cam21 is the low-resolution camera, Cam24 the high-resolution one.
8. ETHZ was processing for satellite images one lake only if at least 30% of clean pixels in an image were cloud-free. UniBe did not use any such threshold.
9. For both MODIS and VIIRS1, MTA was performed with moving mean filter and window length 3.
10. VIIRS1 on Lake St. Moritz was tested on mixed pixels while the SVM model was trained using clean pixels of other lakes. Hence, for this special case, the results are not 100% trustworthy.
11. For all ETHZ approaches (MODIS and VIIRS1), the lake is considered frozen (label Y) when % FP was >= 90). An exception is Lake St. Moritz (MODIS) where only four clean pixels exist (and even less cloud-free). In this special case, the lake is considered frozen when % FP was >=75. (the threshold of 90% was reduced to 75%).
12. For MODIS and VIIRS1 in the very rare case of multiple acquisitions (max 3) on a single date, results of all acquisitions have been combined (averaged) and a single result is generated (details in Section 3.2.3) and listed below.
13. For MODIS data, the results of the best split (among split 1 and split 2) are displayed. The splits are explained in Section 3.3.4.
14. For MODIS and VIIRS (ETHZ) the results shown below refer to the best version using SVM (with RBF kernel) and all channels as input feature vector.
15. For Lake St. Moritz and VIIRS UniBe reports only max 2 clean pixels, ETHZ none. This difference can be due to slightly different generalization of the OpenStreetMap lake outlines (though a comparison by ETHZ of original and generalized outlines showed minimal differences).
16. For VIIRS, UniBe provides in its results mostly only Y or N without any quantitative value, ETHZ for both VIIRS and MODIS lists the SVM regression score ("probability").
17. UniBe sometimes lists after Y or N a number. This is the number of pixels (cloud-free clean) processed.

**Lake St. Moritz**

| Date | MODIS<br><br>Frozen (Y/N)<br>% FP | VIIRS1<br><br>Frozen (Y/N)<br>% FP | VIIRS2<br><br>Frozen (Y/N)<br>% FP<br>max 2 clean pixels | WCam Cam21<br><br>Frozen (Y/N)<br>% FP | WCam Cam24<br><br>Frozen (Y/N)<br>% FP | In-situ (Temperature based)<br><br>Frozen (Y/N)<br>State (1 to 4) | In-situ (Pressure based)<br><br>Frozen (Y/N)<br>State (1 to 4) | In-situ (Dynamic based)<br><br>Frozen (Y/N)<br>State (1 to 4) |
|---|---|---|---|---|---|---|---|---|
| 1.10.16 | nd | nd | nd | nd | nd | 4 | 4 | 4 |
| 2.10.16 | nd | (N)0 | (N) | nd | nd | 4 | 4 | 4 |
| 3.10.16 | (N)0 | (N)0 | (N) | nd | nd | 4 | 4 | 4 |
| 4.10.16 | (N)0 | (N)0 | (N) | nd | nd | 4 | 4 | 4 |
| 5.10.16 | (N)0 | (N)0 | (N) | nd | nd | 4 | 4 | 4 |
| 6.10.16 | (N)0 | (N)0 | (N) | nd | nd | 4 | 4 | 4 |
| 7.10.16 | (N)0 | (N)0 | (N) | nd | nd | 4 | 4 | 4 |
| 8.10.16 | nd | (N)0 | (N) | nd | nd | 4 | 4 | 4 |
| 9.10.16 | nd | nd | nd | nd | nd | 4 | 4 | 4 |
| 10.10.16 | nd | nd | nd | nd | nd | 4 | 4 | 4 |
| 11.10.16 | nd | nd | nd | nd | nd | 4 | 4 | 4 |
| 12.10.16 | (N)0 | (N)0 | (N) | nd | nd | 4 | 4 | 4 |
| 13.10.16 | nd | nd | nd | nd | nd | 4 | 4 | 4 |
| 14.10.16 | nd | nd | nd | nd | nd | 4 | 4 | 4 |
| 15.10.16 | nd | nd | nd | nd | nd | 4 | 4 | 4 |
| 16.10.16 | (N)0 | (N)0 | (N) | nd | nd | 4 | 4 | 4 |
| 17.10.16 | nd | nd | nd | nd | nd | 4 | 4 | 4 |
| 18.10.16 | (N)0 | (N)0 | (N) | nd | nd | 4 | 4 | 4 |
| 19.10.16 | nd | nd | nd | nd | nd | 4 | 4 | 4 |
| 20.10.16 | (N)0 | (N)0 | (N) | nd | nd | 4 | 4 | 4 |
| 21.10.16 | nd | nd | nd | nd | nd | 4 | 4 | 4 |
| 22.10.16 | (N)0 | (N)0 | (N) | nd | nd | 4 | 4 | 4 |
| 23.10.16 | nd | nd | nd | nd | nd | 4 | 4 | 4 |
| 24.10.16 | nd | nd | nd | nd | nd | 4 | 4 | 4 |
| 25.10.16 | nd | nd | nd | nd | nd | 4 | 4 | 4 |
| 26.10.16 | nd | nd | nd | nd | nd | 4 | 4 | 4 |
| 27.10.16 | (N)0 | (N)0 | (N) | nd | nd | 4 | 4 | 4 |
| 28.10.16 | (N)0 | nd | nd | nd | nd | 4 | 4 | 4 |
| 29.10.16 | (N)0 | (N)0 | (N) | nd | nd | 4 | 4 | 4 |
| 30.10.16 | (N)0 | (N)0 | (N) | nd | nd | 4 | 4 | 4 |
| 31.10.16 | nd | (N)0 | (N) | nd | nd | 4 | 4 | 4 |
| 1.11.16 | (N)0 | (N)0 | (N) | nd | nd | 4 | 4 | 4 |
| 2.11.16 | nd | nd | nd | nd | nd | 4 | 4 | 4 |
| 3.11.16 | (N)0 | (N)0 | (N) | nd | nd | 4 | 4 | 4 |
| 4.11.16 | nd | nd | nd | nd | nd | 4 | 4 | 4 |
| 5.11.16 | nd | nd | nd | nd | nd | 4 | 4 | 4 |
| 6.11.16 | nd | nd | nd | nd | nd | 4 | 4 | 4 |
| 7.11.16 | nd | (N)63.6 | (N) | nd | nd | 4 | 4 | 4 |
| 8.11.16 | (Y)75 | (N)81.8 | (Y)1 | nd | nd | 4 | 4 | 4 |
| 9.11.16 | (N)33.3 | (Y)100 | (Y)2 | nd | nd | 4 | 4 | 4 |
| 10.11.16 | (Y)75 | nd | nd | nd | nd | 4 | 4 | 4 |
| 11.11.16 | nd | nd | nd | nd | nd | 4 | 4 | 4 |
| 12.11.16 | (N)0 | (N)63.6 | (N) | nd | nd | 4 | 4 | 4 |
| 13.11.16 | nd | nd | nd | nd | nd | 4 | 4 | 4 |
| 14.11.16 | (N)25 | (N)81.8 | (N) | nd | nd | 4 | 4 | 4 |
| 15.11.16 | (N)25 | (N)81.8 | (Y)2 | nd | nd | 4 | 4 | 4 |
| 16.11.16 | nd | nd | nd | nd | nd | 4 | 4 | 4 |
| 17.11.16 | nd | (N)36.4 | (N) | nd | nd | 4 | 4 | 4 |
| 18.11.16 | nd | nd | nd | nd | nd | 4 | 4 | 4 |
| 19.11.16 | nd | (N)16.7 | (N) | nd | nd | 4 | 4 | 4 |
| 20.11.16 | nd | nd | nd | nd | nd | 4 | 4 | 4 |
| 21.11.16 | nd | nd | nd | nd | nd | 4 | 4 | 4 |
| 22.11.16 | nd | nd | nd | nd | nd | 4 | 4 | 4 |
| 23.11.16 | nd | nd | nd | nd | nd | 4 | 4 | 4 |
| 24.11.16 | nd | nd | nd | nd | nd | 4 | 4 | 4 |
| 25.11.16 | nd | nd | nd | nd | nd | 4 | 4 | 4 |
| 26.11.16 | nd | (N)0 | (N) | nd | nd | 4 | 4 | 4 |
| 27.11.16 | (N)0 | (N)0 | (N) | nd | nd | 4 | 4 | 4 |
| 28.11.16 | (N)25 | (N)9.1 | (N) | nd | nd | 4 | 4 | 4 |
| 29.11.16 | (N)25 | (N)9.1 | (N) | nd | nd | 4 | 4 | 4 |
| 30.11.16 | (N)25 | (N)36.4 | (N) | nd | nd | 4 | 4 | 4 |
| 1.12.16 | (N)0 | (N)0 | nd | nd | nd | 4 | 4 | 4 |
| 2.12.16 | (N)0 | (N)9.1 | (N) | nd | nd | 4 | 4 | 4 |
| 3.12.16 | (Y)75 | (N)9.1 | (N) | nd | nd | 4 | 4 | 4 |
| 4.12.16 | (Y)75 | (N)9.1 | (N) | 0.01 | 0.00 | 4 | 4 | 4 |
| 5.12.16 | (Y)75 | (N)36.4 | (N) | 0.07 | 0.00 | 4 | 4 | 4 |
| 6.12.16 | (N)50 | (N)18.2 | (N) | 0.17 | 0.22 | 4 | 4 | 4 |
| 7.12.16 | (N)50 | (N)18.2 | (N) | 0.01 | 0.12 | 4 | 4 | 4 |
| 8.12.16 | (N)0 | (N)18.2 | (N) | 0.12 | 0.14 | 4 | 4 | 4 |
| 9.12.16 | (N)0 | nd | nd | 0.09 | 0.11 | 4 | 4 | 4 |
| 10.12.16 | (N)0 | (N)0 | (N) | 0.28 | 0.27 | 4 | 4 | 4 |
| 11.12.16 | nd | nd | nd | 0.66 | 0.55 | 4 | 4 | 4 |
| 12.12.16 | (N)0 | (N)9.1 | (N) | 0.43 | 0.29 | 4 | 4 | 4 |
| 13.12.16 | (N)0 | (N)9.1 | (N) | 0.71 | nd | 4 | 4 | 4 |
| 14.12.16 | (N)0 | (N)18.2 | (N) | 0.81 | 0.57 | 4 | 4 | 4 |
| 15.12.16 | (N)25 | (N)18.2 | (N) | 0.79 | 0.99 | 4 | 1 | 4 |
| 16.12.16 | (N)50 | (N)63.6 | (N) | 0.90 | 1.00 | 4 | 1 | 4 |
| 17.12.16 | (N)50 | (N)54.5 | (N) | 0.98 | 0.98 | 1 | 1 | 3 |
| 18.12.16 | (Y)100 | (N)63.6 | (N) | 0.95 | 1.00 | 1 | 1 | 1 |
| 19.12.16 | nd | nd | nd | 1.00 | 1.00 | 1 | 1 | 1 |
| 20.12.16 | nd | nd | nd | 1.00 | 1.00 | 1 | 1 | 1 |
| 21.12.16 | (Y)100 | (N)81.8 | (Y)2 | 1.00 | 1.00 | 1 | 1 | 1 |
| 22.12.16 | (Y)100 | (Y)100 | (Y)2 | 0.97 | 1.00 | 1 | 1 | 1 |
| 23.12.16 | (Y)75 | (Y)100 | (Y)2 | nd | nd | 1 | 1 | 1 |
| 24.12.16 | nd | nd | nd | nd | nd | 1 | 1 | 1 |
| 25.12.16 | nd | nd | nd | nd | nd | 1 | 1 | 1 |
| 26.12.16 | (Y)75 | (N)45.5 | (N) | nd | nd | 1 | 1 | 1 |
| 27.12.16 | (N)50 | (N)45.5 | (N) | nd | nd | 1 | 1 | 1 |
| 28.12.16 | (Y)75 | (N)54.5 | (N) | nd | nd | 1 | 1 | 1 |
| 29.12.16 | (Y)75 | (N)37.5 | nd | 1.00 | 1.00 | 1 | 1 | 1 |
| 30.12.16 | (Y)75 | (N)54.5 | (N) | 1.00 | 1.00 | 1 | 1 | 1 |
| 31.12.16 | (Y)75 | (N)81.8 | (N) | 1.00 | 1.00 | 1 | 1 | 1 |
| 1.1.17 | (Y)75 | (N)81.8 | (N) | 1.00 | 1.00 | 1 | 1 | 1 |
| 2.1.17 | (Y)100 | nd | nd | 0.99 | 1.00 | 1 | 1 | 1 |
| 3.1.17 | (Y)100 | (N)72.7 | (N) | 1.00 | 1.00 | 1 | 1 | 1 |
| 4.1.17 | nd | (N)50 | (N) | 0.98 | 1.00 | 1 | 1 | 1 |
| 5.1.17 | nd | nd | nd | nd | nd | 1 | 1 | 1 |
| 6.1.17 | (Y)100 | (Y)100 | (Y)2 | 1.00 | 1.00 | 1 | 1 | 1 |
| 7.1.17 | nd | nd | nd | 1.00 | 1.00 | 1 | 1 | 1 |
| 8.1.17 | nd | nd | nd | 1.00 | 1.00 | 1 | 1 | 1 |
| 9.1.17 | (Y)100 | (Y)100 | (Y)2 | 1.00 | 1.00 | 1 | 1 | 1 |
| 10.1.17 | nd | (Y)100 | (Y)2 | 1.00 | 1.00 | 1 | 1 | 1 |
| 11.1.17 | (Y)100 | (Y)100 | (Y)2 | 1.00 | 1.00 | 1 | 1 | 1 |
| 12.1.17 | (Y)100 | (Y)100 | (Y)2 | 0.92 | 1.00 | 1 | 1 | 1 |
| 13.1.17 | nd | nd | nd | nd | 1.00 | 1 | 1 | 1 |
| 14.1.17 | nd | nd | nd | nd | nd | 1 | 1 | 1 |
| 15.1.17 | nd | nd | nd | 1.00 | 1.00 | 1 | 1 | 1 |
| 16.1.17 | (Y)100 | (Y)100 | nd | 1.00 | 1.00 | 1 | 1 | 1 |
| 17.1.17 | (Y)100 | (Y)100 | (Y)2 | 1.00 | 1.00 | 1 | 1 | 1 |
| 18.1.17 | (Y)100 | (Y)100 | (Y)2 | 1.00 | 1.00 | 1 | 1 | 1 |
| 19.1.17 | (Y)100 | (Y)100 | (Y)2 | 1.00 | 1.00 | 1 | 1 | 1 |
| 20.1.17 | (Y)100 | (Y)100 | (Y)2 | 1.00 | 1.00 | 1 | 1 | 1 |
| 21.1.17 | (Y)100 | (Y)100 | (Y)2 | nd | 1.00 | 1 | 1 | 1 |
| 22.1.17 | (Y)100 | (Y)100 | (Y)2 | 1.00 | nd | 1 | 1 | 1 |
| 23.1.17 | (Y)100 | (Y)100 | (Y)2 | 1.00 | 1.0 | 1 | 1 | 1 |
| 24.1.17 | (Y)100 | (Y)100 | (Y)2 | 1.00 | nd | 1 | 1 | 1 |
| 25.1.17 | (Y)100 | (Y)100 | (Y)2 | 1.00 | nd | 1 | 1 | 1 |
| 26.1.17 | (Y)100 | (Y)100 | (Y)2 | 1.00 | 1.00 | 1 | 1 | 1 |
| 27.1.17 | (Y)100 | (Y)100 | (Y)2 | 1.00 | 1.00 | 1 | 1 | 1 |
| 28.1.17 | nd | nd | nd | 1.00 | 1.00 | 1 | 1 | 1 |
| 29.1.17 | (Y)100 | (Y)100 | (Y)2 | 1.00 | 1.00 | 1 | 1 | 1 |
| 30.1.17 | (Y)100 | (Y)100 | (Y)1 | 1.00 | 1.00 | 1 | 1 | 1 |
| 31.1.17 | nd | nd | nd | nd | nd | 1 | 1 | 1 |
| 1.2.17 | nd | nd | nd | 1.00 | 1.00 | 1 | 1 | 1 |
| 2.2.17 | nd | nd | nd | 1.00 | 1.00 | 1 | 1 | 1 |
| 3.2.17 | nd | nd | nd | nd | 1.00 | 1 | 1 | 1 |
| 4.2.17 | nd | (Y)100 | nd | 1.00 | nd | 1 | 1 | 1 |
| 5.2.17 | nd | nd | nd | 1.00 | 1.00 | 1 | 1 | 1 |
| 6.2.17 | nd | nd | nd | 1.00 | 1.00 | 1 | 1 | 1 |
| 7.2.17 | nd | nd | nd | 1.00 | 1.00 | 1 | 1 | 1 |
| 8.2.17 | nd | nd | nd | 1.00 | 1.00 | 1 | 1 | 1 |
| 9.2.17 | (Y)100 | (Y)100 | (Y)2 | nd | 1.00 | 1 | 1 | 1 |
| 10.2.17 | nd | nd | nd | 1.00 | 1.00 | 1 | 1 | 1 |
| 11.2.17 | (Y)100 | (Y)100 | (Y)2 | 1.00 | 1.00 | 1 | 1 | 1 |
| 12.2.17 | nd | (Y)100 | (Y)2 | 1.00 | 1.00 | 1 | 1 | 1 |
| 13.2.17 | (Y)100 | (Y)100 | (Y)2 | 1.00 | 1.00 | 1 | 1 | 1 |
| 14.2.17 | (Y)100 | (Y)100 | (Y)2 | 1.00 | 1.00 | 1 | 1 | 1 |
| 15.2.17 | (Y)100 | (Y)100 | (Y)2 | 1.00 | 1.00 | 1 | 1 | 1 |
| 16.2.17 | (Y)100 | (Y)100 | (Y)2 | 1.00 | 1.00 | 1 | 1 | 1 |
| 17.2.17 | nd | (Y)100 | (Y)2 | nd | nd | 1 | 1 | 1 |
| 18.2.17 | (Y)100 | (Y)100 | (Y)2 | 1.00 | 1.00 | 1 | 1 | 1 |
| 19.2.17 | (Y)100 | (Y)100 | (Y)2 | 1.00 | 1.00 | 1 | 1 | 1 |
| 20.2.17 | (Y)100 | (Y)100 | (Y)2 | 1.00 | 1.00 | 1 | 1 | 1 |
| 21.2.17 | nd | (Y)100 | (Y)2 | 1.00 | 1.00 | 1 | 1 | 1 |
| 22.2.17 | nd | (Y)100 | (N) | nd | nd | 1 | 1 | 1 |
| 23.2.17 | nd | nd | nd | nd | nd | 1 | 1 | 1 |
| 24.2.17 | nd | nd | nd | nd | nd | 1 | 1 | 1 |
| 25.2.17 | (Y)100 | (Y)100 | (Y)2 | nd | nd | 1 | 1 | 1 |
| 26.2.17 | nd | (Y)100 | (Y)2 | 1.00 | 1.00 | 1 | 1 | 1 |
| 27.2.17 | (Y)100 | (Y)100 | (Y)2 | 1.00 | 1.00 | 1 | 1 | 1 |
| 28.2.17 | nd | nd | nd | nd | nd | 1 | 1 | 1 |
| 1.3.17 | (Y)100 | (Y)100 | (Y)2 | 1.00 | 1.00 | 1 | 1 | 1 |
| 2.3.17 | nd | nd | nd | 1.00 | 1.00 | 1 | 1 | 1 |
| 3.3.17 | nd | nd | nd | 1.00 | 1.00 | 1 | 1 | 1 |
| 4.3.17 | nd | nd | nd | nd | 1.00 | 1 | 1 | 1 |
| 5.3.17 | nd | (Y)100 | nd | 0.99 | 1.00 | 1 | 1 | 1 |
| 6.3.17 | nd | nd | nd | 1.00 | 1.00 | 1 | 1 | 1 |
| 7.3.17 | nd | (Y)100 | (Y)2 | 1.00 | 1.00 | 1 | 1 | 1 |
| 8.3.17 | (Y)100 | (Y)100 | (Y)2 | 1.00 | 1.00 | 1 | 1 | 1 |
| 9.3.17 | nd | nd | nd | 1.00 | 1.00 | 1 | 1 | 1 |
| 10.3.17 | (Y)100 | (Y)100 | (Y)2 | 1.00 | 1.00 | 1 | 1 | 1 |
| 11.3.17 | (Y)100 | (Y)100 | (Y)1 | 1.00 | 1.00 | 1 | 1 | 1 |
| 12.3.17 | (Y)100 | (Y)100 | (Y)2 | 1.00 | 1.00 | 1 | 1 | 1 |
| 13.3.17 | (Y)100 | (Y)100 | (Y)2 | 1.00 | 1.00 | 1 | 1 | 1 |
| 14.3.17 | (Y)100 | (Y)100 | (N) | 1.00 | 1.00 | 1 | 1 | 1 |
| 15.3.17 | nd | (Y)100 | nd | 1.00 | 1.00 | 1 | 1 | 1 |
| 16.3.17 | (Y)100 | (Y)100 | (N) | 1.00 | 1.00 | 1 | 1 | 1 |
| 17.3.17 | (Y)100 | (Y)100 | (Y)2 | 1.00 | 1.00 | 1 | 1 | 1 |
| 18.3.17 | nd | (Y)100 | (Y)1 | nd | nd | 1 | 1 | 1 |
| 19.3.17 | nd | nd | nd | nd | nd | 1 | 1 | 1 |
| 20.3.17 | (Y)100 | (Y)100 | (Y)2 | nd | nd | 1 | 1 | 1 |
| 21.3.17 | nd | nd | nd | nd | nd | 1 | 1 | 1 |
| 22.3.17 | nd | (Y)100 | nd | nd | nd | 1 | 1 | 1 |
| 23.3.17 | nd | nd | nd | nd | nd | 1 | 1 | 1 |
| 24.3.17 | nd | nd | nd | nd | nd | 1 | 1 | 1 |
| 25.3.17 | (Y)100 | (Y)90.9 | (Y)2 | nd | nd | 1 | 1 | 1 |
| 26.3.17 | (Y)100 | (N)81.8 | (N) | nd | nd | 1 | 1 | 1 |
| 27.3.17 | (Y)100 | (N)63.6 | (N) | nd | nd | 1 | 1 | 1 |
| 28.3.17 | (Y)100 | (N)27.3 | (Y)2 | nd | nd | 1 | 1 | 1 |
| 29.3.17 | (Y)100 | (N)18.2 | (Y)1 | nd | nd | 1 | 1 | 1 |
| 30.3.17 | (Y)100 | (N)0 | (N) | nd | nd | 1 | 1 | 1 |
| 31.3.17 | (Y)100 | (N)0 | (N) | nd | nd | 1 | 1 | 1 |
| 1.4.17 | nd | (N)0 | (N) | nd | nd | 1 | 1 | 1 |
| 2.4.17 | (Y)100 | nd | nd | nd | nd | 1 | 1 | 1 |
| 3.4.17 | (Y)100 | (N)0 | (N) | nd | nd | 1 | 1 | 1 |
| 4.4.17 | nd | nd | nd | nd | nd | 1 | 1 | 1 |
| 5.4.17 | nd | nd | nd | nd | nd | 2 | 1 | 3 |
| 6.4.17 | nd | (N)18.2 | (N) | nd | nd | 2 | 1 | 4 |
| 7.4.17 | (Y)100 | (N)9.1 | (N) | nd | nd | 2 | 1 | 4 |
| 8.4.17 | (Y)100 | (N)27.3 | (N) | nd | nd | 4 | 4 | 4 |
| 9.4.17 | (N)25 | (N)20 | nd | nd | nd | 4 | 4 | 4 |
| 10.4.17 | (N)25 | (N)25 | nd | nd | nd | 4 | 4 | 4 |
| 11.4.17 | nd | (N)27.3 | (N) | nd | nd | 4 | 4 | 4 |
| 12.4.17 | (N)50 | (N)0 | (N) | nd | nd | 4 | 4 | 4 |
| 13.4.17 | (N)50 | (N)0 | (N) | nd | nd | 4 | 4 | 4 |
| 14.4.17 | nd | (N)0 | (N) | nd | nd | 4 | 4 | 4 |
| 15.4.17 | nd | nd | nd | nd | nd | 4 | 4 | 4 |
| 16.4.17 | nd | nd | nd | nd | nd | 4 | 4 | 4 |
| 17.4.17 | nd | nd | nd | nd | nd | 4 | 4 | 4 |
| 18.4.17 | nd | nd | nd | nd | nd | 4 | 4 | 4 |
| 19.4.17 | nd | nd | nd | nd | nd | 4 | 4 | 4 |
| 20.4.17 | (N)0 | (N)0 | (N) | nd | nd | 4 | 4 | 4 |
| 21.4.17 | (N)0 | (N)0 | (N) | nd | nd | 4 | 4 | 4 |
| 22.4.17 | (N)0 | (N)0 | (N) | nd | nd | 4 | 4 | 4 |
| 23.4.17 | (Y)100 | (N)0 | (N) | nd | | 4 | 4 | 4 |
| 24.4.17 | nd | (N)0 | (N) | nd | | 4 | 4 | 4 |
| 25.4.17 | nd | nd | nd | nd | | 4 | 4 | 4 |
| 26.4.17 | nd | nd | nd | nd | | 4 | 4 | 4 |
| 27.4.17 | nd | nd | nd | 0.00 | | 4 | 4 | 4 |
| 28.4.17 | nd | (Y)100 | nd | 0.00 | | 4 | 4 | 4 |
| 29.4.17 | (N)0 | (Y)90.9 | (N) | 0.00 | 0.07 | 4 | 4 | 4 |
| 30.4.17 | (N)0 | (N)63.6 | (N) | 0.01 | 0.07 | 4 | 4 | 4 |

•

**Lake Silvaplana**

| Date | MODIS<br><br>Frozen (Y/N)<br>% FP | VIIRS1<br><br>Frozen (Y/N)<br>% FP | VIIRS2<br><br>Frozen (Y/N)<br>% FP | In-situ<br>(Temperature based)<br><br>Frozen (Y/N)<br>State (1 to 4) | In-situ<br>(Pressure based)<br><br>Frozen (Y/N)<br>State (1 to 4) | In-situ<br>(Dynamic based)<br><br>Frozen (Y/N)<br>State (1 to 4) |
|---|---|---|---|---|---|---|
| 1.10.16 | nd | nd | | 4 | 4 | 4 |
| 2.10.16 | (N)0 | (N)0 | | 4 | 4 | 4 |
| 3.10.16 | (N)0 | (N)0 | | 4 | 4 | 4 |
| 4.10.16 | (N)0 | (N)0 | | 4 | 4 | 4 |
| 5.10.16 | (N)0 | (N)0 | | 4 | 4 | 4 |
| 6.10.16 | (N)0 | (N)0 | | 4 | 4 | 4 |
| 7.10.16 | (N)0 | (N)0 | | 4 | 4 | 4 |
| 8.10.16 | (N)0 | (N)0 | | 4 | 4 | 4 |
| 9.10.16 | nd | nd | | 4 | 4 | 4 |
| 10.10.16 | nd | nd | | 4 | 4 | 4 |
| 11.10.16 | nd | nd | | 4 | 4 | 4 |
| 12.10.16 | (N)0 | (N)0 | | 4 | 4 | 4 |
| 13.10.16 | nd | nd | | 4 | 4 | 4 |
| 14.10.16 | nd | nd | | 4 | 4 | 4 |
| 15.10.16 | nd | nd | | 4 | 4 | 4 |
| 16.10.16 | (N)0 | (N)0 | | 4 | 4 | 4 |
| 17.10.16 | nd | nd | | 4 | 4 | 4 |
| 18.10.16 | nd | nd | | 4 | 4 | 4 |
| 19.10.16 | nd | nd | | 4 | 4 | 4 |
| 20.10.16 | (N)0 | (N)0 | | 4 | 4 | 4 |
| 21.10.16 | nd | nd | | 4 | 4 | 4 |
| 22.10.16 | (N)0 | nd | | 4 | 4 | 4 |
| 23.10.16 | nd | nd | | 4 | 4 | 4 |
| 24.10.16 | nd | nd | | 4 | 4 | 4 |
| 25.10.16 | nd | nd | | 4 | 4 | 4 |
| 26.10.16 | nd | nd | | 4 | 4 | 4 |
| 27.10.16 | (N)0 | (N)0 | | 4 | 4 | 4 |
| 28.10.16 | (N)0 | (N)0 | | 4 | 4 | 4 |
| 29.10.16 | (N)0 | (N)0 | | 4 | 4 | 4 |
| 30.10.16 | (N)0 | (N)0 | | 4 | 4 | 4 |
| 31.10.16 | (N)0 | (N)0 | | 4 | 4 | 4 |
| 1.11.16 | (N)0 | (N)0 | | 4 | 4 | 4 |
| 2.11.16 | nd | nd | | 4 | 4 | 4 |
| 3.11.16 | (N)0 | (N)0 | | 4 | 4 | 4 |
| 4.11.16 | nd | nd | | 4 | 4 | 4 |
| 5.11.16 | nd | nd | | 4 | 4 | 4 |
| 6.11.16 | nd | nd | | 4 | 4 | 4 |
| 7.11.16 | (N)9.5 | (N)11.1 | | 4 | 4 | 4 |
| 8.11.16 | (N)9.5 | (N)66.7 | | 4 | 4 | 4 |
| 9.11.16 | nd | (N)77.8 | | 4 | 4 | 4 |
| 10.11.16 | (N)0 | (Y)100 | | 4 | 4 | 4 |
| 11.11.16 | nd | nd | | 4 | 4 | 4 |
| 12.11.16 | (N)0 | (N)0 | | 4 | 4 | 4 |
| 13.11.16 | nd | nd | | 4 | 4 | 4 |
| 14.11.16 | (N)4.8 | (N)33.3 | | 4 | 4 | 4 |
| 15.11.16 | (N)4.8 | (N)50 | | 4 | 4 | 4 |
| 16.11.16 | nd | nd | | 4 | 4 | 4 |
| 17.11.16 | (N)0 | (Y)100 | | 4 | 4 | 4 |
| 18.11.16 | nd | nd | | 4 | 4 | 4 |
| 19.11.16 | nd | nd | | 4 | 4 | 4 |
| 20.11.16 | nd | nd | | 4 | 4 | 4 |
| 21.11.16 | nd | nd | | 4 | 4 | 4 |
| 22.11.16 | nd | nd | | 4 | 4 | 4 |
| 23.11.16 | nd | nd | | 4 | 4 | 4 |
| 24.11.16 | nd | nd | | 4 | 4 | 4 |
| 25.11.16 | nd | nd | | 4 | 4 | 4 |
| 26.11.16 | nd | (N)0 | | 4 | 4 | 4 |
| 27.11.16 | (N)0 | (N)0 | | 4 | 4 | 4 |
| 28.11.16 | (N)14.3 | (N)0 | | 4 | 4 | 4 |
| 29.11.16 | (N)23.8 | (N)0 | | 4 | 4 | 4 |
| 30.11.16 | (N)0 | (N)0 | | 4 | 4 | 4 |
| 1.12.16 | (N)0 | nd | | 4 | 4 | 4 |
| 2.12.16 | (N)0 | (N)0 | | 4 | 4 | 4 |
| 3.12.16 | (N)10.5 | (N)0 | | 4 | 4 | 4 |
| 4.12.16 | (N)18.2 | (N)0 | | 4 | 4 | 4 |
| 5.12.16 | (N)47.6 | (N)0 | | 4 | 4 | 4 |
| 6.12.16 | (N)28.6 | (N)0 | | 4 | 4 | 4 |
| 7.12.16 | (N)5.3 | (N)0 | | 4 | 4 | 4 |
| 8.12.16 | (N)0 | (N)0 | | 4 | 4 | 4 |
| 9.12.16 | (N)0 | nd | | 4 | 4 | 4 |
| 10.12.16 | (N)0 | (N)0 | | 4 | 4 | 4 |
| 11.12.16 | (N)12.5 | nd | | 4 | 4 | 4 |
| 12.12.16 | (N)66.7 | (N)0 | | 4 | 4 | 4 |
| 13.12.16 | (N)28.6 | (N)0 | | 4 | 4 | 4 |
| 14.12.16 | (N)57.1 | (N)0 | | 4 | 4 | 4 |
| 15.12.16 | (N)38.1 | (N)0 | | 4 | 4 | 4 |
| 16.12.16 | (N)57.1 | (N)0 | | 4 | 4 | 4 |
| 17.12.16 | (N)57.1 | (N)0 | | 4 | 4 | 4 |
| 18.12.16 | (N)85.7 | (N)0 | | 4 | 4 | 4 |
| 19.12.16 | nd | (N)0 | | 4 | 4 | 4 |
| 20.12.16 | nd | nd | | 4 | 4 | 4 |
| 21.12.16 | (Y)90.5 | (N)11.1 | | 4 | 4 | 4 |
| 22.12.16 | (N)33.3 | (N)0 | | 4 | 4 | 4 |
| 23.12.16 | (N)5.3 | (N)0 | | 4 | 4 | 4 |
| 24.12.16 | nd | nd | | 4 | 4 | 4 |
| 25.12.16 | nd | nd | | 4 | 4 | 4 |
| 26.12.16 | (N)52.4 | (N)11.1 | | 4 | 4 | 4 |
| 27.12.16 | (N)42.9 | (N)0 | | 4 | 4 | 4 |
| 28.12.16 | (N)81 | (N)0 | | 4 | 4 | 4 |
| 29.12.16 | (N)42.9 | (N)0 | | 4 | 4 | 4 |
| 30.12.16 | (N)42.9 | (N)0 | | 4 | 4 | 4 |
| 31.12.16 | (N)42.9 | (N)0 | | 4 | 4 | 4 |
| 1.1.17 | (Y)90.5 | (N)33.3 | | 2 | 4 | 4 |
| 2.1.17 | (Y)100 | nd | | 2 | 4 | 4 |
| 3.1.17 | (Y)100 | (N)50 | | 4 | 4 | 4 |
| 4.1.17 | nd | nd | | 4 | 4 | 4 |
| 5.1.17 | nd | nd | | 4 | 4 | 4 |
| 6.1.17 | (Y)100 | (Y)100 | | 4 | 4 | 4 |
| 7.1.17 | nd | nd | | 4 | 4 | 4 |
| 8.1.17 | nd | nd | | 4 | 4 | 4 |
| 9.1.17 | (Y)100 | (N)66.7 | | 4 | 4 | 4 |
| 10.1.17 | nd | nd | | 4 | 4 | 4 |
| 11.1.17 | (N)71.4 | (Y)100 | | 4 | 4 | 3 |
| 12.1.17 | (N)71.4 | (Y)100 | | 4 | 4 | 2 |
| 13.1.17 | nd | nd | | 4 | 4 | 1 |
| 14.1.17 | nd | nd | | 1 | 1 | 1 |
| 15.1.17 | (Y)100 | (Y)100 | | 1 | 1 | 1 |
| 16.1.17 | (Y)100 | (Y)100 | | 1 | 1 | 1 |
| 17.1.17 | (Y)100 | (Y)100 | | 1 | 1 | 1 |
| 18.1.17 | (Y)100 | (Y)100 | | 1 | 1 | 1 |
| 19.1.17 | (Y)100 | (Y)100 | | 1 | 1 | 1 |
| 20.1.17 | (Y)100 | (Y)100 | | 1 | 1 | 1 |
| 21.1.17 | (Y)100 | (Y)100 | | 1 | 1 | 1 |
| 22.1.17 | (Y)100 | (Y)100 | | 1 | 1 | 1 |
| 23.1.17 | (Y)100 | (Y)100 | | 1 | 1 | 1 |
| 24.1.17 | (Y)100 | (Y)100 | | 1 | 1 | 1 |
| 25.1.17 | (Y)100 | (Y)100 | | 1 | 1 | 1 |
| 26.1.17 | (Y)100 | (Y)100 | | 1 | 1 | 1 |
| 27.1.17 | (Y)100 | (Y)100 | | 1 | 1 | 1 |
| 28.1.17 | nd | nd | | 1 | 1 | 1 |
| 29.1.17 | (Y)100 | (Y)100 | | 1 | 1 | 1 |
| 30.1.17 | (Y)100 | nd | | 1 | 1 | 1 |
| 31.1.17 | nd | nd | | 1 | 1 | 1 |
| 1.2.17 | nd | nd | | 1 | 1 | 1 |
| 2.2.17 | nd | nd | | 1 | 1 | 1 |
| 3.2.17 | nd | (Y)100 | | 1 | 1 | 1 |
| 4.2.17 | nd | nd | | 1 | 1 | 1 |
| 5.2.17 | nd | (Y)100 | | 1 | 1 | 1 |
| 6.2.17 | nd | nd | | 1 | 1 | 1 |
| 7.2.17 | nd | nd | | 1 | 1 | 1 |
| 8.2.17 | nd | nd | | 1 | 1 | 1 |
| 9.2.17 | (Y)100 | nd | | 1 | 1 | 1 |
| 10.2.17 | (Y)100 | nd | | 1 | 1 | 1 |
| 11.2.17 | (Y)100 | (Y)100 | | 1 | 1 | 1 |
| 12.2.17 | nd | (Y)100 | | 1 | 1 | 1 |
| 13.2.17 | (Y)100 | (Y)100 | | 1 | 1 | 1 |
| 14.2.17 | (Y)100 | (Y)100 | | 1 | 1 | 1 |
| 15.2.17 | (Y)100 | (Y)100 | | 1 | 1 | 1 |
| 16.2.17 | (Y)100 | (Y)100 | | 1 | 1 | 1 |
| 17.2.17 | nd | nd | | 1 | 1 | 1 |
| 18.2.17 | (Y)100 | (Y)100 | | 1 | 1 | 1 |
| 19.2.17 | (Y)100 | (Y)100 | | 1 | 1 | 1 |
| 20.2.17 | (Y)100 | (Y)100 | | 1 | 1 | 1 |
| 21.2.17 | nd | nd | | 1 | 1 | 1 |
| 22.2.17 | (Y)100 | (Y)100 | | 1 | 1 | 1 |
| 23.2.17 | nd | nd | | 1 | 1 | 1 |
| 24.2.17 | nd | nd | | 1 | 1 | 1 |
| 25.2.17 | (Y)100 | (Y)100 | | 1 | 1 | 1 |
| 26.2.17 | nd | nd | | 1 | 1 | 1 |
| 27.2.17 | (Y)100 | (Y)100 | | 1 | 1 | 1 |
| 28.2.17 | nd | nd | | 1 | 1 | 1 |
| 1.3.17 | (Y)100 | (Y)100 | | 1 | 1 | 1 |
| 2.3.17 | nd | (Y)100 | | 1 | 1 | 1 |
| 3.3.17 | nd | nd | | 1 | 1 | 1 |
| 4.3.17 | (Y)100 | nd | | 1 | 1 | 1 |
| 5.3.17 | nd | (Y)100 | | 1 | 1 | 1 |
| 6.3.17 | nd | nd | | 1 | 1 | 1 |
| 7.3.17 | nd | nd | | 1 | 1 | 1 |
| 8.3.17 | (Y)100 | (Y)100 | | 1 | 1 | 1 |
| 9.3.17 | nd | nd | | 1 | 1 | 1 |
| 10.3.17 | (Y)100 | (Y)100 | | 1 | 1 | 1 |
| 11.3.17 | (Y)100 | (Y)100 | | 1 | 1 | 1 |
| 12.3.17 | (Y)100 | (Y)100 | | 1 | 1 | 1 |
| 13.3.17 | (Y)100 | (Y)100 | | 1 | 1 | 1 |
| 14.3.17 | (Y)100 | (Y)100 | | 1 | 1 | 1 |
| 15.3.17 | nd | (Y)100 | | 1 | 1 | 1 |
| 16.3.17 | (Y)100 | (Y)100 | | 1 | 1 | 1 |
| 17.3.17 | (Y)100 | (Y)100 | | 1 | 1 | 1 |
| 18.3.17 | nd | (Y)100 | | 1 | 1 | 1 |
| 19.3.17 | nd | nd | | 1 | 1 | 1 |
| 20.3.17 | (Y)100 | (Y)100 | | 1 | 1 | 1 |
| 21.3.17 | nd | nd | | 1 | 1 | 1 |
| 22.3.17 | nd | nd | | 1 | 1 | 1 |
| 23.3.17 | nd | nd | | 1 | 1 | 1 |
| 24.3.17 | nd | nd | | 1 | 1 | 1 |
| 25.3.17 | (Y)100 | (Y)100 | | 1 | 1 | 1 |
| 26.3.17 | (Y)100 | (Y)100 | | 1 | 1 | 1 |
| 27.3.17 | (Y)100 | (Y)100 | | 1 | 1 | 1 |
| 28.3.17 | (Y)100 | (Y)100 | | 1 | 1 | 1 |
| 29.3.17 | (Y)100 | (Y)100 | | 1 | 1 | 1 |
| 30.3.17 | (Y)100 | (Y)100 | | 1 | 1 | 1 |
| 31.3.17 | (Y)100 | (Y)100 | | 1 | 1 | 1 |
| 1.4.17 | nd | (Y)100 | | 1 | 1 | 1 |
| 2.4.17 | (Y)100 | nd | | 1 | 1 | 1 |
| 3.4.17 | (Y)100 | (Y)100 | | 1 | 1 | 1 |
| 4.4.17 | nd | nd | | 1 | 1 | 1 |
| 5.4.17 | (Y)94.1 | (Y)100 | | 1 | 1 | 1 |
| 6.4.17 | nd | (Y)100 | | 1 | 1 | 1 |
| 7.4.17 | (Y)100 | (Y)100 | | 1 | 1 | 1 |
| 8.4.17 | (N)85.7 | (N)77.8 | | 1 | 1 | 1 |
| 9.4.17 | (N)42.9 | (N)88.9 | | 1 | 1 | 1 |
| 10.4.17 | (N)23.8 | (N)88.9 | | 1 | 1 | 1 |
| 11.4.17 | nd | (N)88.9 | | 1 | 1 | 1 |
| 12.4.17 | (N)14.3 | nd | | 1 | 1 | 1 |
| 13.4.17 | (N)0 | (N)88.9 | | 1 | 1 | 4 |
| 14.4.17 | (N)0 | (N)88.9 | | 4 | 1 | 4 |
| 15.4.17 | nd | nd | | 4 | 4 | 4 |
| 16.4.17 | nd | nd | | 4 | 4 | 4 |
| 17.4.17 | nd | nd | | 4 | 4 | 4 |
| 18.4.17 | nd | nd | | 4 | 4 | 4 |
| 19.4.17 | nd | (N)0 | | 4 | 4 | 4 |
| 20.4.17 | (N)42.9 | (N)0 | | 4 | 4 | 4 |
| 21.4.17 | (N)0 | (N)0 | | 4 | 4 | 4 |
| 22.4.17 | (N)0 | (N)0 | | 4 | 4 | 4 |
| 23.4.17 | (N)0 | (N)0 | | 4 | 4 | 4 |
| 24.4.17 | nd | (N)0 | | 4 | 4 | 4 |
| 25.4.17 | (N)63.2 | (N)14.3 | | 4 | 4 | 4 |
| 26.4.17 | nd | nd | | 4 | 4 | 4 |
| 27.4.17 | nd | nd | | 4 | 4 | 4 |
| 28.4.17 | nd | nd | | 4 | 4 | 4 |
| 29.4.17 | (N)0 | (N)0 | | 4 | 4 | 4 |
| 30.4.17 | (N)0 | (N)0 | | 4 | 4 | 4 |

-

**Lake Sils**

| Date | MODIS<br><br>Frozen (Y/N)<br>% FP | VIIRS1<br><br>Frozen (Y/N)<br>% FP | VIIRS2<br><br>Frozen (Y/N)<br>% FP<br>0-9 (max) = no. of clean pixels | In-situ (Temperature based)<br><br>Frozen (Y/N)<br>State (1 to 4) | In-situ (Pressure based)<br><br>Frozen (Y/N)<br>State (1 to 4) | In-situ (Dynamic based)<br><br>Frozen (Y/N)<br>State (1 to 4) |
|---|---|---|---|---|---|---|
| 1.10.16 | nd | nd | nd | 4 | 4 | 4 |
| 2.10.16 | (N)0 | nd | nd | 4 | 4 | 4 |
| 3.10.16 | (N)0 | (N)0 | (N)0 | 4 | 4 | 4 |
| 4.10.16 | (N)0 | (N)0 | (N)0 | 4 | 4 | 4 |
| 5.10.16 | (N)0 | (N)0 | (N)0 | 4 | 4 | 4 |
| 6.10.16 | (N)0 | nd | nd | 4 | 4 | 4 |
| 7.10.16 | (N)0 | (N)0 | (N)0 | 4 | 4 | 4 |
| 8.10.16 | (N)0 | (N)0 | (N)0 | 4 | 4 | 4 |
| 9.10.16 | nd | nd | nd | 4 | 4 | 4 |
| 10.10.16 | nd | (N)12.5 | (N)0 | 4 | 4 | 4 |
| 11.10.16 | nd | nd | nd | 4 | 4 | 4 |
| 12.10.16 | (N)0 | (N)0 | (N)0 | 4 | 4 | 4 |
| 13.10.16 | nd | nd | nd | 4 | 4 | 4 |
| 14.10.16 | nd | nd | nd | 4 | 4 | 4 |
| 15.10.16 | nd | nd | nd | 4 | 4 | 4 |
| 16.10.16 | (N)0 | (N)0 | (N)0 | 4 | 4 | 4 |
| 17.10.16 | nd | nd | (N)0 | 4 | 4 | 4 |
| 18.10.16 | nd | (N)0 | (N)0 | 4 | 4 | 4 |
| 19.10.16 | nd | nd | nd | 4 | 4 | 4 |
| 20.10.16 | (N)0 | (N)0 | (N)0 | 4 | 4 | 4 |
| 21.10.16 | (N)0 | nd | (N)30 | 4 | 4 | 4 |
| 22.10.16 | nd | nd | nd | 4 | 4 | 4 |
| 23.10.16 | nd | nd | nd | 4 | 4 | 4 |
| 24.10.16 | nd | nd | nd | 4 | 4 | 4 |
| 25.10.16 | nd | nd | nd | 4 | 4 | 4 |
| 26.10.16 | nd | nd | nd | 4 | 4 | 4 |
| 27.10.16 | (N)0 | (N)0 | (N)0 | 4 | 4 | 4 |
| 28.10.16 | (N)0 | (N)0 | (N)0 | 4 | 4 | 4 |
| 29.10.16 | (N)0 | (N)0 | (N)0 | 4 | 4 | 4 |
| 30.10.16 | (N)0 | (N)0 | (N)0 | 4 | 4 | 4 |
| 31.10.16 | (N)0 | (N)0 | (N)0 | 4 | 4 | 4 |
| 1.11.16 | (N)0 | (N)0 | (N)0 | 4 | 4 | 4 |
| 2.11.16 | nd | nd | nd | 4 | 4 | 4 |
| 3.11.16 | (N)0 | (N)0 | (N)0 | 4 | 4 | 4 |
| 4.11.16 | nd | nd | nd | 4 | 4 | 4 |
| 5.11.16 | nd | nd | nd | 4 | 4 | 4 |
| 6.11.16 | nd | nd | nd | 4 | 4 | 4 |
| 7.11.16 | (N)13.3 | (N)9.1 | (N)2 | 4 | 4 | 4 |
| 8.11.16 | (N)12.1 | (N)9.1 | (N)1 | 4 | 4 | 4 |
| 9.11.16 | nd | nd | nd | 4 | 4 | 4 |
| 10.11.16 | (Y)100 | nd | nd | 4 | 4 | 4 |
| 11.11.16 | nd | nd | nd | 4 | 4 | 4 |
| 12.11.16 | (N)3 | (N)0 | (N)2 | 4 | 4 | 4 |
| 13.11.16 | nd | nd | nd | 4 | 4 | 4 |
| 14.11.16 | (N)3 | (N)10 | (N)1 | 4 | 4 | 4 |
| 15.11.16 | (N)3 | (N)0 | (N)0 | 4 | 4 | 4 |
| 16.11.16 | nd | nd | nd | 4 | 4 | 4 |
| 17.11.16 | nd | nd | nd | 4 | 4 | 4 |
| 18.11.16 | nd | nd | nd | 4 | 4 | 4 |
| 19.11.16 | nd | (N)30 | (N)8 | 4 | 4 | 4 |
| 20.11.16 | nd | nd | nd | 4 | 4 | 4 |
| 21.11.16 | nd | nd | nd | 4 | 4 | 4 |
| 22.11.16 | nd | nd | nd | 4 | 4 | 4 |
| 23.11.16 | nd | nd | nd | 4 | 4 | 4 |
| 24.11.16 | nd | nd | nd | 4 | 4 | 4 |
| 25.11.16 | nd | nd | nd | 4 | 4 | 4 |
| 26.11.16 | nd | nd | nd | 4 | 4 | 4 |
| 27.11.16 | (N)3 | (N)0 | (N)0 | 4 | 4 | 4 |
| 28.11.16 | (N)6.1 | (N)0 | (N)0 | 4 | 4 | 4 |
| 29.11.16 | (N)0 | (N)0 | (N)0 | 4 | 4 | 4 |
| 30.11.16 | (N)0 | (N)0 | (N)1 | 4 | 4 | 4 |
| 1.12.16 | (N)0 | (N)0 | (N)0 | 4 | 4 | 4 |
| 2.12.16 | (N)0 | (N)0 | (N)0 | 4 | 4 | 4 |
| 3.12.16 | nd | (N)0 | (N)0 | 4 | 4 | 4 |
| 4.12.16 | nd | (N)0 | (N)0 | 4 | 4 | 4 |
| 5.12.16 | (N)13.3 | (N)0 | (N)0 | 4 | 4 | 4 |
| 6.12.16 | (N)0 | (N)0 | (N)2 | 4 | 4 | 4 |
| 7.12.16 | (N)0 | (N)0 | (N)0 | 4 | 4 | 4 |
| 8.12.16 | (N)0 | (N)0 | (N)0 | 4 | 4 | 4 |
| 9.12.16 | (N)0 | nd | nd | 4 | 4 | 4 |
| 10.12.16 | (N)0 | (N)0 | (N)0 | 4 | 4 | 4 |
| 11.12.16 | nd | nd | nd | 4 | 4 | 4 |
| 12.12.16 | (N)3 | (N)0 | (N)0 | 4 | 4 | 4 |
| 13.12.16 | (N)0 | (N)0 | (N)0 | 4 | 4 | 4 |
| 14.12.16 | (N)6.3 | (N)0 | (N)1 | 4 | 4 | 4 |
| 15.12.16 | (N)3 | (N)0 | (N)1 | 4 | 4 | 4 |
| 16.12.16 | (N)6.1 | (N)0 | (N)0 | 4 | 4 | 4 |
| 17.12.16 | (N)3.2 | (N)0 | (N)0 | 4 | 4 | 4 |
| 18.12.16 | (N)3 | (N)0 | (N)0 | 4 | 4 | 4 |
| 19.12.16 | nd | (N)0 | nd | 4 | 4 | 4 |
| 20.12.16 | nd | nd | nd | 4 | 4 | 4 |
| 21.12.16 | (N)12.1 | (N)0 | (N)1 | 4 | 4 | 4 |
| 22.12.16 | (N)0 | (N)0 | (N)3 | 4 | 4 | 4 |
| 23.12.16 | (N)0 | (N)0 | (N)0 | 4 | 4 | 4 |
| 24.12.16 | nd | nd | nd | 4 | 4 | 4 |
| 25.12.16 | nd | nd | nd | 4 | 4 | 4 |
| 26.12.16 | (N)12.1 | (N)0 | (N)1 | 4 | 4 | 4 |
| 27.12.16 | (N)9.4 | (N)0 | (N)1 | 4 | 4 | 4 |
| 28.12.16 | (N)15.2 | (N)0 | (N)0 | 4 | 4 | 1 |
| 29.12.16 | (N)3 | (N)0 | (N)0 | 4 | 4 | 1 |
| 30.12.16 | (N)0 | (N)0 | (N)0 | 4 | 4 | 1 |
| 31.12.16 | (N)3 | (N)0 | (N)1 | 1 | 1 | 1 |
| 1.1.17 | (N)15.2 | (N)0 | (N)1 | 1 | 1 | 1 |
| 2.1.17 | (N)6.1 | nd | nd | 1 | 1 | 1 |
| 3.1.17 | nd | nd | nd | 1 | 1 | 1 |
| 4.1.17 | (N)16.7 | (N)0 | (N)0 | 1 | 1 | 1 |
| 5.1.17 | nd | nd | nd | 1 | 1 | 1 |
| 6.1.17 | (Y)93.9 | (Y)100 | (Y)9 | 1 | 1 | 1 |
| 7.1.17 | nd | nd | nd | 1 | 1 | 1 |
| 8.1.17 | nd | nd | nd | 1 | 1 | 1 |
| 9.1.17 | (Y)100 | (Y)100 | (Y)9 | 1 | 1 | 1 |
| 10.1.17 | (Y)100 | nd | nd | 1 | 1 | 1 |
| 11.1.17 | (Y)93.9 | (Y)100 | (Y)8 | 1 | 1 | 1 |
| 12.1.17 | (Y)93.9 | (Y)100 | (Y)8 | 1 | 1 | 1 |
| 13.1.17 | nd | nd | nd | 1 | 1 | 1 |
| 14.1.17 | nd | nd | nd | 1 | 1 | 1 |
| 15.1.17 | (Y)100 | (Y)100 | (Y)9 | 1 | 1 | 1 |
| 16.1.17 | (Y)100 | (Y)100 | (Y)9 | 1 | 1 | 1 |
| 17.1.17 | (Y)100 | (Y)100 | (Y)9 | 1 | 1 | 1 |
| 18.1.17 | (Y)100 | (Y)100 | (Y)9 | 1 | 1 | 1 |
| 19.1.17 | (Y)100 | (Y)100 | (Y)9 | 1 | 1 | 1 |
| 20.1.17 | (Y)100 | (Y)100 | (Y)9 | 1 | 1 | 1 |
| 21.1.17 | (Y)100 | (Y)100 | (Y)9 | 1 | 1 | 1 |
| 22.1.17 | (Y)100 | (Y)100 | (Y)9 | 1 | 1 | 1 |
| 23.1.17 | (Y)100 | (Y)100 | (Y)9 | 1 | 1 | 1 |
| 24.1.17 | (Y)100 | (Y)100 | (Y)9 | 1 | 1 | 1 |
| 25.1.17 | (Y)100 | (Y)100 | (Y)9 | 1 | 1 | 1 |
| 26.1.17 | (Y)100 | (Y)100 | (Y)9 | 1 | 1 | 1 |
| 27.1.17 | (Y)100 | (Y)100 | (Y)9 | 1 | 1 | 1 |
| 28.1.17 | nd | nd | nd | 1 | 1 | 1 |
| 29.1.17 | (Y)100 | (Y)100 | (Y)9 | 1 | 1 | 1 |
| 30.1.17 | (Y)100 | (Y)100 | (Y)9 | 1 | 1 | 1 |
| 31.1.17 | nd | nd | nd | 1 | 1 | 1 |
| 1.2.17 | nd | nd | nd | 1 | 1 | 1 |
| 2.2.17 | nd | nd | nd | 1 | 1 | 1 |
| 3.2.17 | nd | (Y)100 | (Y)9 | 1 | 1 | 1 |
| 4.2.17 | nd | nd | nd | 1 | 1 | 1 |
| 5.2.17 | nd | (N)87.5 | (N)5 | 1 | 1 | 1 |
| 6.2.17 | nd | nd | nd | 1 | 1 | 1 |
| 7.2.17 | nd | nd | (N)1 | 1 | 1 | 1 |
| 8.2.17 | nd | nd | nd | 1 | 1 | 1 |
| 9.2.17 | nd | nd | nd | 1 | 1 | 1 |
| 10.2.17 | nd | nd | nd | 1 | 1 | 1 |
| 11.2.17 | nd | (Y)100 | (N)4 | 1 | 1 | 1 |
| 12.2.17 | nd | (Y)100 | (Y)8 | 1 | 1 | 1 |
| 13.2.17 | (Y)100 | (Y)100 | (Y)9 | 1 | 1 | 1 |
| 14.2.17 | (Y)100 | (Y)100 | (Y)9 | 1 | 1 | 1 |
| 15.2.17 | (Y)100 | (Y)100 | (Y)9 | 1 | 1 | 1 |
| 16.2.17 | (Y)100 | (Y)100 | (Y)9 | 1 | 1 | 1 |
| 17.2.17 | nd | nd | nd | 1 | 1 | 1 |
| 18.2.17 | (Y)100 | (Y)100 | (Y)9 | 1 | 1 | 1 |
| 19.2.17 | (Y)100 | (Y)100 | (Y)9 | 1 | 1 | 1 |
| 20.2.17 | (Y)100 | (Y)100 | (Y)9 | 1 | 1 | 1 |
| 21.2.17 | nd | nd | nd | 1 | 1 | 1 |
| 22.2.17 | (Y)100 | (Y)100 | (Y)7 | 1 | 1 | 1 |
| 23.2.17 | nd | nd | nd | 1 | 1 | 1 |
| 24.2.17 | nd | nd | nd | 1 | 1 | 1 |
| 25.2.17 | (Y)100 | (Y)100 | (Y)9 | 1 | 1 | 1 |
| 26.2.17 | nd | nd | nd | 1 | 1 | 1 |
| 27.2.17 | (Y)100 | (Y)100 | (Y)9 | 1 | 1 | 1 |
| 28.2.17 | nd | nd | nd | 1 | 1 | 1 |
| 1.3.17 | (Y)100 | (Y)100 | (Y)9 | 1 | 1 | 1 |
| 2.3.17 | nd | (Y)100 | (Y)9 | 1 | 1 | 1 |
| 3.3.17 | nd | nd | nd | 1 | 1 | 1 |
| 4.3.17 | nd | nd | nd | 1 | 1 | 1 |
| 5.3.17 | nd | (Y)100 | (N)3 | 1 | 1 | 1 |
| 6.3.17 | nd | nd | nd | 1 | 1 | 1 |
| 7.3.17 | nd | nd | (N)1 | 1 | 1 | 1 |
| 8.3.17 | (N)63.6 | (Y)100 | (Y)9 | 1 | 1 | 1 |
| 9.3.17 | nd | nd | nd | 1 | 1 | 1 |
| 10.3.17 | (Y)100 | (Y)100 | (Y)9 | 1 | 1 | 1 |
| 11.3.17 | (Y)100 | (Y)100 | (Y)9 | 1 | 1 | 1 |
| 12.3.17 | (Y)100 | (Y)100 | (Y)9 | 1 | 1 | 1 |
| 13.3.17 | (Y)100 | (Y)100 | (Y)9 | 1 | 1 | 1 |
| 14.3.17 | (Y)100 | (Y)100 | (Y)8 | 1 | 1 | 1 |
| 15.3.17 | nd | (Y)100 | (Y)9 | 1 | 1 | 1 |
| 16.3.17 | (Y)100 | (Y)100 | (Y)8 | 1 | 1 | 1 |
| 17.3.17 | (Y)100 | (Y)100 | (N)5 | 1 | 1 | 1 |
| 18.3.17 | nd | nd | (N)1 | 1 | 1 | 1 |
| 19.3.17 | nd | nd | nd | 1 | 1 | 1 |
| 20.3.17 | (Y)100 | (Y)100 | (Y)7 | 1 | 1 | 1 |
| 21.3.17 | nd | nd | nd | 1 | 1 | 1 |
| 22.3.17 | nd | nd | nd | 1 | 1 | 1 |
| 23.3.17 | nd | nd | nd | 1 | 1 | 1 |
| 24.3.17 | nd | nd | nd | 1 | 1 | 1 |
| 25.3.17 | (Y)100 | (Y)100 | (Y)8 | 1 | 1 | 1 |
| 26.3.17 | (Y)100 | (Y)100 | (Y)9 | 1 | 1 | 1 |
| 27.3.17 | (Y)100 | (Y)100 | (Y)7 | 1 | 1 | 1 |
| 28.3.17 | (Y)100 | (Y)100 | (Y)7 | 1 | 1 | 1 |
| 29.3.17 | (Y)100 | (Y)100 | (Y)9 | 1 | 1 | 1 |
| 30.3.17 | nd | (Y)100 | (Y)9 | 1 | 1 | 1 |
| 31.3.17 | (Y)100 | (Y)100 | (Y)8 | 1 | 1 | 1 |
| 1.4.17 | nd | (Y)100 | (N)6 | 1 | 1 | 1 |
| 2.4.17 | nd | nd | nd | 1 | 1 | 1 |
| 3.4.17 | (Y)100 | (Y)100 | (Y)7 | 1 | 1 | 1 |
| 4.4.17 | nd | nd | nd | 1 | 1 | 1 |
| 5.4.17 | (Y)100 | (Y)90.9 | (Y)9 | 1 | 1 | 1 |
| 6.4.17 | nd | (Y)90.9 | (Y)8 | 1 | 1 | 1 |
| 7.4.17 | (Y)100 | (N)81.8 | (N)5 | 1 | 1 | 1 |
| 8.4.17 | (Y)90.9 | (N)20 | (N)1 | 1 | 1 | 4 |
| 9.4.17 | (Y)97 | (N)18.2 | (N) | 1 | 1 | 4 |
| 10.4.17 | (Y)90.9 | (N)18.2 | (N) | 4 | 2 | 4 |
| 11.4.17 | nd | (N)27.3 | (N)4 | 4 | 3 | 4 |
| 12.4.17 | (N)73.3 | nd | nd | 4 | 4 | 4 |
| 13.4.17 | (N)78.8 | (N)0 | (N) | 4 | 4 | 4 |
| 14.4.17 | (N)45.5 | (N)0 | (N) | 4 | 4 | 4 |
| 15.4.17 | nd | nd | nd | 4 | 4 | 4 |
| 16.4.17 | nd | nd | nd | 4 | 4 | 4 |
| 17.4.17 | nd | (N)0 | (N) | 4 | 4 | 4 |
| 18.4.17 | nd | (N)0 | (N) | 4 | 4 | 4 |
| 19.4.17 | (N)0 | nd | nd | 4 | 4 | 4 |
| 20.4.17 | (N)0 | (N)0 | (N) | 4 | 4 | 4 |
| 21.4.17 | (N)0 | (N)0 | (N) | 4 | 4 | 4 |
| 22.4.17 | (N)0 | (N)0 | (N) | 4 | 4 | 4 |
| 23.4.17 | (N)0 | (N)0 | (N) | 4 | 4 | 4 |
| 24.4.17 | nd | (N)0 | (N) | 4 | 4 | 4 |
| 25.4.17 | nd | nd | nd | 4 | 4 | 4 |
| 26.4.17 | nd | nd | nd | 4 | 4 | 4 |
| 27.4.17 | nd | nd | nd | 4 | 4 | 4 |
| 28.4.17 | (N)0 | nd | nd | 4 | 4 | 4 |
| 29.4.17 | (N)3 | (N)0 | (N) | 4 | 4 | 4 |
| 30.4.17 | (N)3 | (N)0 | (N) | 4 | 4 | 4 |

•

**Lake Sihl**

| Date | MODIS<br><br>Frozen (Y/N) % FP | VIIRS1<br><br>Frozen (Y/N) % FP | VIIRS2<br><br>Frozen (Y/N) % FP | In-situ (Temperature based)<br><br>Frozen (Y/N) State (1 to 4) | In-situ (Pressure based)<br><br>Frozen (Y/N) State (1 to 4) | In-situ (Dynamic based)<br><br>Frozen (Y/N) State (1 to 4) |
|---|---|---|---|---|---|---|
| 1.10.16 | (N)84.3 | nd | nd | 4 | 4 | 4 |
| 2.10.16 | nd | nd | nd | 4 | 4 | 4 |
| 3.10.16 | (N)0 | (N)0 | (N) | 4 | 4 | 4 |
| 4.10.16 | (N)0 | (N)0 | (N) | 4 | 4 | 4 |
| 5.10.16 | nd | nd | nd | 4 | 4 | 4 |
| 6.10.16 | (N)0 | (N)0 | (N) | 4 | 4 | 4 |
| 7.10.16 | (N)0 | (N)0 | (N) | 4 | 4 | 4 |
| 8.10.16 | nd | nd | nd | 4 | 4 | 4 |
| 9.10.16 | nd | nd | nd | 4 | 4 | 4 |
| 10.10.16 | (N)0 | nd | nd | 4 | 4 | 4 |
| 11.10.16 | (N)0 | (N)3.7 | (N) | 4 | 4 | 4 |
| 12.10.16 | (N)0.9 | (N)0 | (N) | 4 | 4 | 4 |
| 13.10.16 | nd | nd | nd | 4 | 4 | 4 |
| 14.10.16 | nd | nd | nd | 4 | 4 | 4 |
| 15.10.16 | nd | nd | nd | 4 | 4 | 4 |
| 16.10.16 | (N)0 | (N)0 | (N) | 4 | 4 | 4 |
| 17.10.16 | nd | nd | nd | 4 | 4 | 4 |
| 18.10.16 | nd | nd | nd | 4 | 4 | 4 |
| 19.10.16 | nd | nd | nd | 4 | 4 | 4 |
| 20.10.16 | (N)0 | nd | nd | 4 | 4 | 4 |
| 21.10.16 | nd | nd | nd | 4 | 4 | 4 |
| 22.10.16 | (N)0 | (N)0 | (N) | 4 | 4 | 4 |
| 23.10.16 | nd | nd | nd | 4 | 4 | 4 |
| 24.10.16 | nd | nd | nd | 4 | 4 | 4 |
| 25.10.16 | nd | nd | nd | 4 | 4 | 4 |
| 26.10.16 | nd | nd | nd | 4 | 4 | 4 |
| 27.10.16 | nd | nd | (N) | 4 | 4 | 4 |
| 28.10.16 | (N)0 | (N)0 | (N) | 4 | 4 | 4 |
| 29.10.16 | (N)0 | (N)0 | (N) | 4 | 4 | 4 |
| 30.10.16 | (N)0 | (N)0 | (N) | 4 | 4 | 4 |
| 31.10.16 | nd | nd | nd | 4 | 4 | 4 |
| 1.11.16 | (N)0 | (N)0 | (N) | 4 | 4 | 4 |
| 2.11.16 | nd | nd | nd | 4 | 4 | 4 |
| 3.11.16 | (N)0 | (N)3.2 | (N) | 4 | 4 | 4 |
| 4.11.16 | (N)0 | (N)2.3 | (N) | 4 | 4 | 4 |
| 5.11.16 | nd | nd | nd | 4 | 4 | 4 |
| 6.11.16 | nd | nd | nd | 4 | 4 | 4 |
| 7.11.16 | nd | nd | nd | 4 | 4 | 4 |
| 8.11.16 | nd | nd | nd | 4 | 4 | 4 |
| 9.11.16 | nd | nd | nd | 4 | 4 | 4 |
| 10.11.16 | nd | nd | nd | 4 | 4 | 4 |
| 11.11.16 | nd | nd | nd | 4 | 4 | 4 |
| 12.11.16 | (N)3.5 | (N)11.4 | (N) | 4 | 4 | 4 |
| 13.11.16 | nd | nd | nd | 4 | 4 | 4 |
| 14.11.16 | nd | nd | nd | 4 | 4 | 4 |
| 15.11.16 | nd | (N)0 | (N) | 4 | 4 | 4 |
| 16.11.16 | nd | nd | nd | 4 | 4 | 4 |
| 17.11.16 | nd | nd | nd | 4 | 4 | 4 |
| 18.11.16 | nd | nd | nd | 4 | 4 | 4 |
| 19.11.16 | nd | nd | nd | 4 | 4 | 4 |
| 20.11.16 | nd | nd | nd | 4 | 4 | 4 |
| 21.11.16 | nd | nd | nd | 4 | 4 | 4 |
| 22.11.16 | nd | nd | nd | 4 | 4 | 4 |
| 23.11.16 | nd | nd | nd | 4 | 4 | 4 |
| 24.11.16 | nd | nd | nd | 4 | 4 | 4 |
| 25.11.16 | nd | nd | nd | 4 | 4 | 4 |
| 26.11.16 | nd | nd | nd | 4 | 4 | 4 |
| 27.11.16 | nd | nd | nd | 4 | 4 | 4 |
| 28.11.16 | nd | nd | nd | 4 | 4 | 4 |
| 29.11.16 | nd | nd | nd | 4 | 4 | 4 |
| 30.11.16 | (N)0 | (N)0 | (N) | 4 | 4 | 4 |
| 1.12.16 | (N)0 | (N)0 | (N) | 4 | 4 | 4 |
| 2.12.16 | (N)0 | (N)0 | (N) | 4 | 4 | 4 |
| 3.12.16 | (N)0 | (N)0 | (N) | 4 | 4 | 4 |
| 4.12.16 | (N)0 | (N)0 | (N) | 4 | 4 | 4 |
| 5.12.16 | (N)0 | (N)0 | (N) | 4 | 4 | 4 |
| 6.12.16 | (N)0 | (N)0 | (N) | 4 | 4 | 4 |
| 7.12.16 | (N)0 | (N)0 | (N) | 4 | 4 | 4 |
| 8.12.16 | (N)0 | (N)0 | (N) | 4 | 4 | 4 |
| 9.12.16 | nd | (N)0 | (N) | 4 | 4 | 4 |
| 10.12.16 | (N)0 | (N)0 | (N) | 4 | 4 | 4 |
| 11.12.16 | (N)0 | nd | nd | 4 | 4 | 4 |
| 12.12.16 | (N)0 | (N)0 | (N) | 4 | 4 | 4 |
| 13.12.16 | (N)0 | nd | nd | 4 | 4 | 4 |
| 14.12.16 | (N)0 | (N)0 | (N) | 4 | 4 | 4 |
| 15.12.16 | (N)0 | (N)0 | (N) | 4 | 4 | 4 |
| 16.12.16 | (N)0 | (N)0 | (N) | 4 | 4 | 4 |
| 17.12.16 | (N)0 | (N)0 | (N) | 4 | 4 | 4 |
| 18.12.16 | (N)0 | nd | nd | 4 | 4 | 4 |
| 19.12.16 | (N)1.8 | (N)0 | (N) | 4 | 4 | 4 |
| 20.12.16 | nd | nd | nd | 4 | 4 | 4 |
| 21.12.16 | (N)0 | (N)0 | (N) | 4 | 4 | 4 |
| 22.12.16 | (N)0 | (N)0 | (N) | 4 | 4 | 4 |
| 23.12.16 | (N)0 | (N)0 | (N) | 4 | 4 | 4 |
| 24.12.16 | (N)0 | nd | nd | 4 | 4 | 4 |
| 25.12.16 | nd | nd | nd | 4 | 4 | 4 |
| 26.12.16 | (N)0 | (N)0 | (N) | 4 | 4 | 4 |
| 27.12.16 | nd | nd | nd | 4 | 4 | 1 |
| 28.12.16 | (N)69.6 | (N)2.2 | (N) | 2 | 4 | 1 |
| 29.12.16 | (N)83.5 | (N)4.4 | (N) | 1 | 4 | 1 |
| 30.12.16 | (N)67 | (N)34.4 | (N) | 1 | 4 | 1 |
| 31.12.16 | (N)0 | (N)65.5 | (N) | 1 | 1 | 1 |
| 1.1.17 | (N)0 | (N)71.1 | (N) | 1 | 1 | 1 |
| 2.1.17 | (N)84.1 | nd | (N) | 1 | 1 | 1 |
| 3.1.17 | (Y)100 | (Y)100 | (Y) | 1 | 1 | 1 |
| 4.1.17 | nd | nd | nd | 1 | 1 | 1 |
| 5.1.17 | nd | nd | nd | 1 | 1 | 1 |
| 6.1.17 | (Y)100 | (Y)100 | (Y) | 1 | 1 | 1 |
| 7.1.17 | nd | nd | nd | 1 | 1 | 1 |
| 8.1.17 | nd | nd | nd | 1 | 1 | 1 |
| 9.1.17 | nd | nd | nd | 1 | 1 | 1 |
| 10.1.17 | nd | nd | nd | 1 | 1 | 1 |
| 11.1.17 | nd | nd | nd | 1 | 1 | 1 |
| 12.1.17 | nd | nd | nd | 1 | 1 | 1 |
| 13.1.17 | (Y)91.3 | nd | nd | 1 | 1 | 1 |
| 14.1.17 | nd | nd | nd | 1 | 1 | 1 |
| 15.1.17 | nd | nd | nd | 1 | 1 | 1 |
| 16.1.17 | nd | nd | nd | 1 | 1 | 1 |
| 17.1.17 | nd | nd | nd | 1 | 1 | 1 |
| 18.1.17 | (Y)100 | (Y)100 | (Y) | 1 | 1 | 1 |
| 19.1.17 | (Y)100 | (Y)100 | (Y) | 1 | 1 | 1 |
| 20.1.17 | (Y)100 | (Y)100 | (Y) | 1 | 1 | 1 |
| 21.1.17 | (Y)100 | (Y)100 | (Y) | 1 | 1 | 1 |
| 22.1.17 | (Y)100 | (Y)100 | (Y) | 1 | 1 | 1 |
| 23.1.17 | (Y)100 | (Y)100 | (Y) | 1 | 1 | 1 |
| 24.1.17 | nd | nd | nd | 1 | 1 | 1 |
| 25.1.17 | nd | nd | (Y)3 | 1 | 1 | 1 |
| 26.1.17 | nd | (Y)100 | (Y) | 1 | 1 | 1 |
| 27.1.17 | (Y)100 | (Y)100 | (Y) | 1 | 1 | 1 |
| 28.1.17 | (Y)100 | (Y)100 | (Y) | 1 | 1 | 1 |
| 29.1.17 | (Y)100 | (Y)100 | (Y) | 1 | 1 | 1 |
| 30.1.17 | nd | nd | nd | 1 | 1 | 1 |
| 31.1.17 | nd | nd | nd | 1 | 1 | 1 |
| 1.2.17 | nd | (Y)100 | (Y) | 1 | 1 | 1 |
| 2.2.17 | nd | (Y)100 | (Y) | 1 | 1 | 1 |
| 3.2.17 | nd | (Y)100 | (N) | 1 | 1 | 1 |
| 4.2.17 | nd | nd | (Y) | 1 | 1 | 1 |
| 5.2.17 | nd | nd | nd | 1 | 1 | 1 |
| 6.2.17 | nd | (Y)100 | (Y) | 1 | 1 | 1 |
| 7.2.17 | nd | nd | nd | 1 | 1 | 1 |
| 8.2.17 | nd | nd | nd | 1 | 1 | 1 |
| 9.2.17 | nd | nd | nd | 1 | 1 | 1 |
| 10.2.17 | (Y)100 | nd | nd | 1 | 1 | 1 |
| 11.2.17 | (Y)100 | (Y)100 | (Y) | 1 | 1 | 1 |
| 12.2.17 | nd | (Y)100 | (Y) | 1 | 1 | 1 |
| 13.2.17 | nd | nd | nd | 1 | 1 | 1 |
| 14.2.17 | (Y)100 | (Y)100 | (Y) | 1 | 1 | 1 |
| 15.2.17 | (Y)100 | (Y)100 | (Y) | 1 | 1 | 1 |
| 16.2.17 | (Y)100 | (Y)100 | (Y) | 1 | 1 | 1 |
| 17.2.17 | nd | nd | nd | 1 | 1 | 1 |
| 18.2.17 | (Y)100 | (Y)100 | (Y) | 1 | 1 | 1 |
| 19.2.17 | (Y)100 | (Y)100 | (Y) | 1 | 1 | 1 |
| 20.2.17 | nd | nd | nd | 1 | 1 | 1 |
| 21.2.17 | nd | nd | nd | 1 | 1 | 1 |
| 22.2.17 | nd | (Y)100 | (Y) | 1 | 1 | 1 |
| 23.2.17 | nd | nd | nd | 1 | 1 | 1 |
| 24.2.17 | nd | nd | nd | 1 | 1 | 1 |
| 25.2.17 | (Y)97.4 | (Y)100 | (Y) | 1 | 1 | 1 |
| 26.2.17 | nd | (Y)100 | (Y) | 1 | 1 | 1 |
| 27.2.17 | (Y)100 | (Y)100 | (Y) | 1 | 1 | 1 |
| 28.2.17 | nd | nd | (Y) | 1 | 1 | 1 |
| 1.3.17 | nd | nd | nd | 1 | 1 | 1 |
| 2.3.17 | nd | nd | nd | 1 | 1 | 1 |
| 3.3.17 | (Y)100 | (Y)100 | (Y) | 1 | 1 | 1 |
| 4.3.17 | nd | nd | nd | 1 | 1 | 1 |
| 5.3.17 | (Y)99.1 | (Y)95.3 | (Y) | 1 | 1 | 1 |
| 6.3.17 | nd | nd | nd | 1 | 1 | 1 |
| 7.3.17 | nd | nd | nd | 1 | 1 | 1 |
| 8.3.17 | nd | nd | nd | 1 | 1 | 1 |
| 9.3.17 | nd | nd | nd | 1 | 1 | 1 |
| 10.3.17 | (N)88.7 | (Y)90.6 | (N)* | 1 | 1 | 1 |
| 11.3.17 | (N)87.1 | (Y)90.9 | (N)* | 1 | 1 | 1 |
| 12.3.17 | (N)83.5 | (N)82.2 | (N)* | 1 | 1 | 1 |
| 13.3.17 | (N)72.2 | (N)60.5 | (N)* | 1 | 1 | 1 |
| 14.3.17 | (N)69.6 | (N)60 | (N)* | 1 | 1 | 1 |
| 15.3.17 | (N)60.9 | (N)48.9 | (N) | 1 | 1 | 1 |
| 16.3.17 | (N)22.6 | (N)33.3 | (N) | 4 | 4 | 4 |
| 17.3.17 | (N)7.8 | (N)15.6 | (N) | 4 | 4 | 4 |
| 18.3.17 | nd | nd | nd | 4 | 4 | 4 |
| 19.3.17 | nd | nd | nd | 4 | 4 | 4 |
| 20.3.17 | (N)0 | (N)0 | (N) | 4 | 4 | 4 |
| 21.3.17 | nd | nd | nd | 4 | 4 | 4 |
| 22.3.17 | nd | nd | nd | 4 | 4 | 4 |
| 23.3.17 | (N)0 | (N)0 | (N) | 4 | 4 | 4 |
| 24.3.17 | nd | nd | nd | 4 | 4 | 4 |
| 25.3.17 | (N)8.7 | (N)0 | (N) | 4 | 4 | 4 |
| 26.3.17 | nd | nd | nd | 4 | 4 | 4 |
| 27.3.17 | (N)0 | (N)0 | (N) | 4 | 4 | 4 |
| 28.3.17 | (N)0 | (N)0 | (N) | 4 | 4 | 4 |
| 29.3.17 | (N)0 | (N)0 | (N) | 4 | 4 | 4 |
| 30.3.17 | (N)0 | (N)0 | (N) | 4 | 4 | 4 |
| 31.3.17 | (N)0 | nd | (N) | 4 | 4 | 4 |
| 1.4.17 | (N)0 | (N)0 | (N) | 4 | 4 | 4 |
| 2.4.17 | nd | nd | nd | 4 | 4 | 4 |
| 3.4.17 | (N)0 | nd | (N) | 4 | 4 | 4 |
| 4.4.17 | nd | nd | nd | 4 | 4 | 4 |
| 5.4.17 | nd | nd | nd | 4 | 4 | 4 |
| 6.4.17 | (N)0 | (N)0 | (N) | 4 | 4 | 4 |
| 7.4.17 | (N)0 | (N)0 | (N) | 4 | 4 | 4 |
| 8.4.17 | (N)0 | (N)0 | (N) | 4 | 4 | 4 |
| 9.4.17 | (N)0 | (N)0 | (N) | 4 | 4 | 4 |
| 10.4.17 | (N)0 | (N)0 | (N) | 4 | 4 | 4 |
| 11.4.17 | nd | nd | nd | 4 | 4 | 4 |
| 12.4.17 | (N)0 | (N)0 | (N) | 4 | 4 | 4 |
| 13.4.17 | (N)0 | (N)0 | (N) | 4 | 4 | 4 |
| 14.4.17 | nd | nd | nd | 4 | 4 | 4 |
| 15.4.17 | nd | nd | nd | 4 | 4 | 4 |
| 16.4.17 | nd | nd | nd | 4 | 4 | 4 |
| 17.4.17 | nd | nd | nd | 4 | 4 | 4 |
| 18.4.17 | nd | nd | nd | 4 | 4 | 4 |
| 19.4.17 | (N)31.4 | nd | nd | 4 | 4 | 4 |
| 20.4.17 | nd | nd | nd | 4 | 4 | 4 |
| 21.4.17 | (N)16.2 | (N)0 | (N) | 4 | 4 | 4 |
| 22.4.17 | (N)11.3 | (N)0 | (N) | 4 | 4 | 4 |
| 23.4.17 | (N)0 | (N)0 | (N) | 4 | 4 | 4 |
| 24.4.17 | nd | (N)0 | (N) | 4 | 4 | 4 |
| 25.4.17 | nd | nd | nd | 4 | 4 | 4 |
| 26.4.17 | nd | nd | nd | 4 | 4 | 4 |
| 27.4.17 | nd | nd | nd | 4 | 4 | 4 |
| 28.4.17 | nd | nd | nd | 4 | 4 | 4 |
| 29.4.17 | (N)40 | (N)0 | (N) | 4 | 4 | 4 |
| 30.4.17 | (N)35.5 | (N)0 | (N) | 4 | 4 | 4 |

* Approx. 50% of the lake is ice covered.